\pdfoutput = 1
\documentclass[11pt]{article}
\usepackage{enumerate}
\usepackage[OT1]{fontenc}
\usepackage[usenames]{color}
\usepackage{smile}
\usepackage[colorlinks,
linkcolor=red,
anchorcolor=blue,
citecolor=blue
]{hyperref}
\usepackage{mathrsfs}
\usepackage{fullpage}
\usepackage{hyperref}
\usepackage[protrusion=true, expansion=true]{microtype}
\usepackage{float}
\usepackage{subfigure}
\usepackage{cancel}

\usepackage{amsfonts,amsmath,amssymb,amsthm,url,xspace,mathtools,authblk}
\usepackage{tikz}
\usepackage{verbatim}
\usetikzlibrary{arrows,shapes}
\usepackage[bottom]{footmisc}
\usepackage{enumitem}
\usepackage{lscape,diagbox}
\usepackage{caption}
\usepackage{algorithmicx,algpseudocode}
\usepackage{math}
\usepackage{color}
\usepackage{multirow}

\ifx\counterwithout\undefined\usepackage{chngcntr}\fi
\counterwithout{equation}{section}

\allowdisplaybreaks[1]
\mathtoolsset{showonlyrefs=true}

\usepackage{xargs}
\usepackage[colorinlistoftodos,prependcaption,textsize=tiny]{todonotes}
\newcommandx{\unsure}[2][1=]{\todo[linecolor=red,backgroundcolor=red!25,bordercolor=red,#1]{#2}}
\newcommandx{\change}[2][1=]{\todo[linecolor=blue,backgroundcolor=blue!25,bordercolor=blue,#1]{#2}}
\newcommandx{\info}[2][1=]{\todo[linecolor=OliveGreen,backgroundcolor=OliveGreen!25,bordercolor=OliveGreen,#1]{#2}}
\newcommandx{\improvement}[2][1=]{\todo[linecolor=Plum,backgroundcolor=Plum!25,bordercolor=Plum,#1]{#2}}

\usepackage{alphalph}
\begin{document}

\title{Online Covariance Matrix Estimation in Sketched Newton Methods}

\author[1]{Wei Kuang}
\author[1,2]{Mihai Anitescu}
\author[3]{Sen Na}

\affil[1]{Department of Statistics, The University of Chicago}
\affil[2]{Mathematics and Computer Science Division, Argonne National Laboratory}
\affil[3]{School of Industrial and Systems Engineering, Georgia Institute of Technology}

\date{}

\maketitle

\begin{abstract}

Given the ubiquity of streaming data, online algorithms have been widely used for parameter estimation, with second-order methods particularly standing out for their efficiency and robustness. In this paper, we study an online sketched Newton method that leverages a randomized sketching technique to perform an approximate Newton step in each iteration, thereby eliminating the computational bottleneck of second-order methods.
While existing studies have established the asymptotic normality of sketched Newton methods, a consistent estimator of the limiting covariance matrix remains an open problem. We propose a fully online covariance matrix estimator that is constructed entirely from the Newton iterates and requires no matrix factorization. Compared to covariance estimators for first-order online methods, our estimator for second-order methods is \textit{batch-free}. 
We establish the consistency and convergence rate of our estimator, and coupled with asymptotic normality results, we can then perform online statistical inference for the model parameters based on sketched Newton methods. We also discuss the extension of our estimator to constrained problems, and demonstrate its superior performance on regression problems as well as benchmark problems in the CUTEst set.

\end{abstract}

\section{Introduction}\label{sec:1}

We consider the following stochastic optimization problem:
\begin{equation}\label{prob}
\min_{\bx\in\mathbb{R}^d} F(\bx)=\mE_{\P}[f(\bx;\xi)],
\end{equation}
where $F: \mathbb{R}^d \rightarrow \mathbb{R}$ is a stochastic, strongly convex objective function, $f(\cdot;\xi)$ is its noisy observation, and $\xi \sim \P$ is a random variable.
Problems of form \eqref{prob} appear in various decision-making applications in statistics and data science, including online recommendation \citep{Li2010contextual}, precision medicine \citep{Kosorok2019Precision}, energy control \citep{Wallace2005Applications}, portfolio allocation \citep{Fan2012Vast}, and e-commerce  \citep{Chen2022Statistical}.
In these applications, \eqref{prob} is often interpreted as a model parameter estimation problem, where $\bx$ denotes the model parameter and $\xi$ denotes a random data sample. The true model parameter $\bx^{\star} = \argmin_{\bx\in\mathbb{R}^d} F(\bx)$ is the minimizer of the expected population loss $F$.

The classic offline approach to solving \eqref{prob} is \textit{sample average approximation} or $M$-estimation, which generates $t$ i.i.d. samples $\xi_1,\dots,\xi_t\hskip-1.5pt\sim\hskip-1.5pt\mathcal{P}$ and approximates the population loss $F$ by the empirical loss:
\begin{equation}\label{nequ:1}
\hat{\bx}_t = \argmin_{\bx\in \mR^d} \cbr{\hat{F}_t(\bx) \coloneqq \frac{1}{t}\sum_{i=1}^{t} f(\bx; \xi_i) }.
\end{equation} 
The statistical properties, e.g., $\sqrt{t}$-consistency and asymptotic normality, of $M$-estimators $\hat{\bx}_t$ are well-known in the literature \citep{Vaart1998Asymptotic, Hastie2009Elements}, and numerous deterministic optimization methods can be applied to solve Problem \eqref{nequ:1}, such as gradient descent and Newton's method \citep{Boyd2004Convex}. However, deterministic methods are not appealing for large datasets due to their significant computation and memory costs. In contrast, online methods via \textit{stochastic approximation} have recently attracted much attention. 
These methods efficiently process each sample once received and then discard, making them well-suited for modern streaming data. Thus, it is particularly critical to quantify the uncertainty of online methods and leverage the methods to perform \textit{online statistical inference} for model parameters.

One of the most fundamental online methods is stochastic gradient descent (SGD) \citep{Robbins1951Stochastic, Kiefer1952Stochastic}, which takes the form
\begin{equation}\label{equ:sgd}
\bx_{t+1} = \bx_t - \alpha_t \nabla f(\bx_t; \xi_t),\quad t\geq 1.
\end{equation}
There exists a long sequence of literature that quantifies the uncertainty of SGD and its many variants. Early works established almost sure convergence and asymptotic normality results of SGD in restricted settings \citep{Sacks1958Asymptotic, Fabian1968Asymptotic, Robbins1971convergence, Fabian1973Asymptotically, Ljung1977Analysis, Ermoliev1983Stochastic, Lai2003Stochastic}. Later on, \citet{Ruppert1988Efficient, Polyak1992Acceleration} proposed averaging SGD iterates as $\bar{\bx}_t =\sum_{i=1}^{t}\bx_i/t$ and established generic asymptotic normality results for $\bar{\bx}_t$. This seminal asymptotic study has then been generalized to other gradient-based methods, including implicit SGD \citep{Toulis2014Statistical, Toulis2017Asymptotic}, constant-stepsize SGD \citep{Li2018Statistical, Mou2020Linear}, moment-adjusted SGD \citep{Liang2019Statistical}, momentum-accelerated SGD \citep{Tang2023Acceleration}, and projected SGD \citep{Duchi2021Asymptotic, Davis2024Asymptotic}. Additionally, studies under non-i.i.d. settings have also been reported \citep{Chen2020Statisticala, Liu2023Online, Li2023statistical}.$\;$

With the asymptotic normality result for the averaged iterate $\bar{\bx}_t$ (see \eqref{exp:Omegastar} for the definition of $\tOmega$):\;\;
\begin{equation}\label{nequ:2}
\sqrt{t}(\barx_t - \tx) \stackrel{d}{\longrightarrow} \mN(0, \tOmega),
\end{equation}
estimating the limiting covariance matrix $\tOmega$ is the crucial next step to perform online statistical inference. 
We note that some inferential procedures may bypass the need for this estimation, such as bootstrapping \citep{Fang2018Online, Liu2023Statistical, Zhong2023Online, Lam2023Resampling} and random scaling \citep{Li2021Statistical, Lee2022Fast}. These procedures may offer certain benefits under favorable settings, but they also suffer from some practical and statistical limitations. For example, online bootstrap methods can readily adapt to non-normal limiting distributions; however, they typically require running multiple resampled trajectories, resulting in substantial computational overhead. Random scaling methods construct pivotal statistics by exploiting martingale or self-normalization structures, yielding limiting distributions that are parameter-free. Nevertheless, such methods are primarily suited for marginal inference, do not naturally apply to joint confidence regions, and are \textit{statistically conservative}. In particular, self-normalized confidence intervals, while asymptotically valid, are generally wider than those based on asymptotic normality, the latter being \textit{asymptotically minimax optimal} in the sense of H\'ajek and Le Cam \citep{Hajek1972Local, LeCam1972Limits}. In contrast, estimating the asymptotic covariance not only enables the construction of asymptotically optimal confidence regions, supports hypothesis testing for linear and nonlinear functionals via delta method, but also provides valuable information for downstream algorithm design, tuning, and diagnostic purposes (e.g., acceleration) that may go beyond inference itself.

With these motivations, many works have indeed focused on \eqref{nequ:2} and proposed different online covariance matrix estimators. In particular, \citet{Chen2020Statistical} proposed two estimators: a plug-in estimator and a batch-means estimator. Compared to the plug-in estimator, which averages the estimated objective Hessians and then computes its inverse, resulting in significant computational costs, the batch-means estimator is obtained simply through the SGD iterates. \citet{Chen2020Statistical} investigated the choice of batch sizes given a fixed total sample size, while \citet{Zhu2021Online} refined that estimator by not fixing the total sample size in advance. The two aforementioned works utilized increasing batch sizes, which has been relaxed to equal batch sizes recently \citep{Zhu2021Constructing, Singh2023Utility}. Combining the asymptotic normality with covariance estimation, we can then construct online confidence intervals for model parameters $\tx$ based on averaged SGD iterates.$\hskip1.3cm$

From the optimization aspect, SGD is computationally efficient with a cheap per-iteration time complexity of $O(d)$; however, when it comes to online statistical inference, SGD still requires updating proper covariance (or random scaling) matrices at each step, leading to $O(d^2)$ time and memory complexity. This more costly inference task provides an opportunity to utilize higher-order methods, such as Newton methods. 
In addition, SGD is known to be sensitive to stepsize tuning, noise heterogeneity, and ill-conditioning of the objective. It tends to perform poorly when the Hessian has eigenvalues on widely different scales. For example, even in problems with dimension as low as $d=20$, one can observe a clear undercoverage when using SGD for online inference (see \cite{Zhu2021Online} and Section \ref{sec:5}). As a reliable alternative class of methods,
stochastic Newton methods address these limitations and often enjoy improved and more robust performance by preconditioning the gradient direction with an (approximate) inverse Hessian \citep{Byrd2016Stochastic, Kovalev2019Stochastic, Bercu2020Efficient}. Although Newton methods rely on second-order Hessian information that may be difficult to access in some cases, such information can be approximated using gradients (via quasi-Newton schemes \citep{Dennis1977Quasi}) or even function values (via finite differences \citep{Spall1998overview, Na2025Derivative}). 
In fact, second-order methods have indeed been widely implemented for parameters estimation of generalized linear models \cite[Sections 4.8 and 6.6]{Dunn2018Generalized}.
The online updating scheme takes the form:
\begin{equation}\label{nequ:3}
\bx_{t+1} = \bx_t + \alpha_t \Delta\bx_t \quad\quad \text{ with }\quad\quad B_t\Delta\bx_t = - \nabla f(\bx_t; \xi_t),
\end{equation}
where $B_t\approx \nabla^2 F(\bx_t)$ is an estimate of the objective Hessian. 
A growing body of literature focuses on performing online statistical inference based on \eqref{nequ:3}. \citet{Leluc2023Asymptotic} considered $B_t$ as a general preconditioning matrix and established the asymptotic normality for the \textit{last} iterate $\bx_t$ assuming the convergence of $B_t$. The authors showed that $\bx_t$ achieves asymptotic efficiency (i.e., minimal covariance) when $B_t \rightarrow \nabla^2 F(\tx)$, corresponding to online Newton methods. 
\citet{Bercu2020Efficient} developed an online Newton method for logistic regression and established similar asymptotic normality for $\bx_t$. \citet{Cenac2020efficient, Boyer2022asymptotic} expanded that approach to more general regression problems and investigated statistical inference on weighted Newton iterates $\bar{\bx}_t = \sum_{i=1}^{t} \bx_i/t$. The above studies revolved around regression problems where the estimated Hessian $B_t$ can be expressed as an average of rank-one matrices, allowing its inverse $B_t^{-1}$ to be updated by the Sherman-Morrison formula \citep{Sherman1950Adjustment}. However, computing the inverse of a general Hessian matrix can be computationally demanding, with an $O(d^3)$ time complexity.

To address the above computational bottleneck, \citet{Na2025Statistical} introduced an \textit{online sketched Newton method} that leverages randomized sketching techniques to approximately solve the Newton system \eqref{nequ:3}, without requiring the approximation error to vanish. Specifically, the time complexity of solving the Newton system can be reduced to $O(\tau\cdot\texttt{nnz}(S)d)$, where $S\in\mR^{d\times q}$ is the sketching matrix with $q\ll d$; $\tau$ denotes the number of sketching steps; and $\texttt{nnz}(S)$ denotes the number of nonzero entries of $S$. For instance, when $S$ is a sparse sketching vector, one can have $\tau = O(d)$ and lead to the time complexity of $O(d^2)$. 
\citet{Na2025Statistical} quantified the uncertainty of both sampling and sketching and established the asymptotic normality for the last iterate $\bx_t$ of the sketched Newton method (see \eqref{equ:lyap} for the definition of $\tXi$):
\begin{equation}\label{nequ:4}
1/\sqrt{\alpha_t}\cdot(\bx_t-\tx)\stackrel{d}{\longrightarrow}\mN(0, \tXi),
\end{equation}
where the limiting covariance $\tXi\neq \tOmega$ depends on the underlying sketching distribution in a complex manner. Due to the challenges of estimating the sketching components in $\tXi$, the authors proposed a plug-in estimator for $\tOmega$ instead. That estimator raises two major concerns. First, the plug-in estimator is generally not asymptotically consistent, although the bias $\tOmega-\tXi$ is controlled by the approximation error of solving the Newton system. It is only consistent when solving the Newton system exactly (so that the approximation error is zero). This bias may significantly compromise the performance of online statistical inference. Second, the plug-in estimator involves the inversion of the estimated Hessian, leading to an $O(d^3)$ time complexity that contradicts the spirit of using sketching solvers.

Motivated by the limitations of plug-in estimators and the success of batch-means estimators in first-order methods, we propose a novel \textit{weighted sample covariance} estimator for $\tXi$. Our estimator is constructed entirely from the sketched Newton iterates with varying weights, and does not involve any matrix inversion, making it computationally efficient. Additionally, our estimator has a simple recursive form, aligning well with the online nature of the method. Unlike batch-means estimators in first-order methods, our estimator is \textit{batch-free}.
We establish the consistency and convergence rate of our estimator, and coupled with the asymptotic normality in \eqref{nequ:4}, we can then construct asymptotically valid confidence intervals for the true model parameters $\bx^{\star}$ based on the Newton iterates $\{\bx_t\}$.
The challenge in our analysis lies in quantifying multiple sources of randomness (sampling, sketching, and adaptive stepsize introduced later); all of them affect the asymptotic behavior of online Newton methods. 
We emphasize that our analysis naturally holds for degenerate designs where the Newton systems are exactly solved and/or the stepsizes are deterministic. To our knowledge, the proposed estimator is the first online construction of a consistent limiting covariance matrix estimator for online second-order methods. We demonstrate its superior empirical performance through extensive experiments on regression problems and benchmark problems from the CUTEst test set.

As a side note, we should mention that this paper focuses on online inference based on the \textit{last} Newton iterate. There is a recent follow-up work \cite{Du2025Online}, after the present paper, focusing on online inference based on the \textit{averaged} Newton iterate $\bar{\bx}_t = \sum_{i=1}^{t} \bx_i/t$. The authors established the asymptotic normality of $\barx_t$ with a covariance $\bar{\Xi}^\star$ different from $\tXi$ in \eqref{nequ:4}, and developed a random scaling inference procedure. While both inference procedures perform competitively and improve upon first-order methods (\cite[Section 5]{Du2025Online}), our last-iterate inference is still of great values due to the following three reasons.$\hskip3cm$

\vskip2pt
\hskip-0.2cm \textbf{(a)} 
While iterate averaging is known to improve statistical efficiency for first-order methods, this effect is less pronounced for second-order methods.
In particular, for exact Newton methods, both the last and averaged iterates are equally, asymptotically minimax optimal ($\tOmega = \tXi = \bar{\Xi}^\star$).
For sketched Newton methods, where computational-statistical trade-offs arise from the sketching solver, both the last and averaged iterates are suboptimal, although the averaging does offer certain efficiency gain ($\tOmega\preceq \bar{\Xi}^\star\preceq \tXi$).
That said, this gain is mild since both suboptimality gaps, $\bar{\Xi}^\star-\tOmega$ and $\tXi-\tOmega$, are explicitly controlled by the sketching approximation error, decaying exponentially fast with the number of sketching steps. $\quad\;$

\vskip2pt
\hskip-0.2cm \textbf{(b)} 
As discussed after \eqref{nequ:2}, compared with covariance estimation, random scaling inference methods may suffer from intrinsic statistical limitations. Such methods are suboptimal even for exact Newton methods (while covariance estimation inference methods are optimal), and are primarily applied for marginal inference \cite[Theorem 4.2]{Du2025Online}, rather than for constructing joint confidence regions or nonlinear hypothesis testing via delta method.

\vskip2pt
\hskip-0.2cm \textbf{(c)} 
More fundamentally, estimating the limiting covariance itself is a problem of independent interest. Our analysis offers a new perspective on understanding the benefits of leveraging second-order information in statistical inference. We note that the Hessian information not only improves the stationarity property of the last Newton iterate compared to the last SGD iterate (optimal v.s. suboptimal), but also enables a batch-free covariance estimator that has no intrusive tuning parameters and a provably faster convergence rate compared to the batch-means covariance estimator for SGD ($O_p(1/\sqrt{t\alpha_t})$ v.s. $O_p(1/\sqrt[4]{t\alpha_t})$).

To better situate the present work within existing literature, we summarize related inference procedures based on first- and second-order methods in Table \ref{tab:1}.

\begin{table}[t!]
{\centering
\resizebox{\linewidth}{!}{
\begin{tabular}{|c|c|c|c|c|c|c|c|c|}
\hline
\multirow{2}{*}{\begin{tabular}[x]{@{}c@{}}Online \\ Algorithm\end{tabular}} & \multirow{2}{*}{Iterate} & \multicolumn{7}{c|}{Inference Procedure and Property} \rule{0pt}{2ex}\\
\cline{3-9}
& & \multicolumn{2}{c|}{Method} & Rate & Efficiency & Param-free & Computation & Memory \rule{0pt}{2ex} \\
\hline
\multirow{4}{*}{\begin{tabular}[x]{@{}c@{}}\\[-5pt] SGD\end{tabular}} & Last & \multicolumn{2}{c|}{?} & ? & Suboptimal & ? & ? & ? \rule{0pt}{2ex}\\
\cline{2-9}
& \multirow{3}{*}{Ave} & \multirow{2}{*}{\begin{tabular}[x]{@{}c@{}}Cov \\ Est\end{tabular}} & Plug-in \citep{Chen2020Statistical} & $O_p(\sqrt{\alpha_t})$ & \multirow{2}{*}{Optimal} & Yes & $O(d^3)$ & \multirow{3}{*}{$O(d^2)$}  \rule{0pt}{2.3ex}\\
\cline{4-5}\cline{7-8}
& &  & Batch-Means \citep{Zhu2021Online} & $O_p(1/\sqrt[4]{t\alpha_t})$ & & No & \multirow{2}{*}{$O(d^2)$} &  \rule{0pt}{2.3ex}\\
\cline{3-7}
& & \multicolumn{2}{c|}{Random Scaling \citep{Lee2022Fast}} & $-$ & Suboptimal & Yes &  &  \rule{0pt}{2.3ex}\\
\hline

\multirow{5}{*}{\begin{tabular}[x]{@{}c@{}}\\ Stochastic \\ Newton \\ (exact)\end{tabular}} & \multirow{2}{*}{\begin{tabular}[x]{@{}c@{}}\\[-8pt] Last \end{tabular}} & \multirow{2}{*}{\begin{tabular}[x]{@{}c@{}}Cov \\ Est\end{tabular}} & Plug-in \citep{Na2025Statistical} & $O_p(\sqrt{\alpha_t})$ & \multirow{2}{*}{\begin{tabular}[x]{@{}c@{}}\\[-8pt] {\blue Optimal}\end{tabular}} & \multirow{2}{*}{\begin{tabular}[x]{@{}c@{}}\\[-8pt] {\blue Yes}\end{tabular}} & $O(d^3)$ &\multirow{5}{*}{\begin{tabular}[x]{@{}c@{}}\\[-5pt] $O(d^2)$\end{tabular}}  \rule{0pt}{2.3ex}\\
\cline{4-5}\cline{8-8}
& & & {\blue Batch-Free (ours)} & {\blue $O_p(1/\sqrt{t\alpha_t})$} & & & {\blue $O(d^2)$} &  \rule{0pt}{2.3ex}\\
\cline{2-8}
& \multirow{3}{*}{\begin{tabular}[x]{@{}c@{}}\\[-8pt] Ave\end{tabular}} & \multirow{2}{*}{\begin{tabular}[x]{@{}c@{}}Cov \\ Est\end{tabular}} & Plug-in \citep{Na2025Statistical} & $O_p(\sqrt{\alpha_t})$ & \multirow{2}{*}{\begin{tabular}[x]{@{}c@{}}\\[-8pt] {\blue Optimal}\end{tabular}} & \multirow{3}{*}{\begin{tabular}[x]{@{}c@{}}\\[-8pt] {\blue Yes}\end{tabular}} & $O(d^3)$ &  \rule{0pt}{2.3ex}\\
\cline{4-5}\cline{8-8}
& & & {\blue Batch-Free (ours)} & {\blue $O_p(1/\sqrt{t\alpha_t})$} & & & \multirow{2}{*}{\begin{tabular}[x]{@{}c@{}}\\[-9pt] {\blue $O(d^2)$}\end{tabular}} &  \rule{0pt}{2.3ex}\\
\cline{3-6}
& & \multicolumn{2}{c|}{Random Scaling \citep{Du2025Online}} & $-$ & Suboptimal & &  &  \rule{0pt}{2.3ex}\\
\hline

\multirow{5}{*}{\begin{tabular}[x]{@{}c@{}}\\ Stochastic \\ Newton \\ (sketched)\end{tabular}} & \multirow{2}{*}{\begin{tabular}[x]{@{}c@{}}\\[-8pt] Last \end{tabular}} & \multirow{2}{*}{\begin{tabular}[x]{@{}c@{}}Cov \\ Est\end{tabular}} & Plug-in \citep{Na2025Statistical} & $O_p(\sqrt{\alpha_t})+\text{Bias}$ & \multirow{2}{*}{\begin{tabular}[x]{@{}c@{}}\\[-8pt] {\blue Suboptimal}\end{tabular}} & \multirow{2}{*}{\begin{tabular}[x]{@{}c@{}}\\[-8pt] {\blue Yes}\end{tabular}} & $O(d^3)$ &\multirow{5}{*}{\begin{tabular}[x]{@{}c@{}}\\[-5pt] $O(d^2)$\end{tabular}}  \rule{0pt}{2.3ex}\\
\cline{4-5}\cline{8-8}
& & & {\blue Batch-Free (ours)} & {\blue $O_p(1/\sqrt{t\alpha_t})$} & & & {\blue $O(d^2)$} &  \rule{0pt}{2.3ex}\\
\cline{2-8}
& \multirow{3}{*}{\begin{tabular}[x]{@{}c@{}}\\[-8pt] Ave\end{tabular}} & \multirow{2}{*}{\begin{tabular}[x]{@{}c@{}}Cov \\ Est\end{tabular}} & Plug-in? & ? &  \multirow{3}{*}{\begin{tabular}[x]{@{}c@{}}\\[-8pt] Suboptimal\end{tabular}} & Yes & $O(d^3)$ &  \rule{0pt}{2.3ex}\\
\cline{4-5}\cline{7-8}
& & & Batch-Free/Means? & ? & & Yes/No? & \multirow{2}{*}{\begin{tabular}[x]{@{}c@{}}\\[-9pt] $O(d^2)$\end{tabular}} &  \rule{0pt}{2.3ex}\\
\cline{3-5}\cline{7-7}
& & \multicolumn{2}{c|}{Random Scaling \citep{Du2025Online}} & $-$ &  & Yes & &  \rule{0pt}{2.3ex}\\
\hline
\end{tabular}}
\vskip2pt
\caption{\textit{Summary of existing inference results. Here, $t$ is the iteration index and $\alpha_t$ is the stepsize. The symbol ``?'' denotes cases where results are missing in existing literature, either because the setting is of limited interest (the last SGD iterate) or because the setting remains open (covariance estimation of the averaged sketched Newton iterate). The symbol ``$-$'' denotes inapplicable cases, in particular for random scaling inference procedures, for which non-asymptotic rates are not available in existing literature. 
We explain that our batch-free covariance estimator applies directly to the averaged exact Newton method, since both the last and averaged exact Newton iterates share the same (optimal) limiting covariance matrix.}}
\label{tab:1} \vspace*{-0.37cm}}
\end{table}

\vskip0.1cm
\noindent \textbf{Structure of the paper:} 
We introduce the online sketched Newton method in Section \ref{sec:2}, and present assumptions and some preliminary theoretical results in Section \ref{sec:3}. In Section \ref{sec:4}, we introduce the weighted sample covariance matrix estimator and present its theoretical guarantees. The numerical experiments are provided in Section \ref{sec:5}, followed by conclusions and future work in Section \ref{sec:6}.

\vskip0.1cm

\noindent \textbf{Notation:} 
Throughout the paper, we use $\|\cdot\|$ to denote the $\ell_2$ norm for vectors and the spectral norm for matrices, $\|\cdot\|_F$ to denote the Frobenius norm for matrices, and $\text{Tr}(\cdot)$ to denote the trace of a matrix. We use $O(\cdot)$ and $o(\cdot)$ to denote the big and small $O$ notation in the usual sense. In particular, for two positive sequences $\{a_t, b_t\}$, $a_t=O(b_t)$ (also denoted as $a_t\lesssim b_t$) if $a_t\leq cb_t$ for a positive constant $c$ and all large enough $t$. Analogously, $a_t=o(b_t)$ if $a_t/b_t\rightarrow 0$ as $t\rightarrow\infty$. For two scalars $a$ and $b$, $a\wedge b=\min(a, b)$ and $a\vee b = \max(a,b)$. We let $I$ denote the identity matrix, $\boldsymbol{0}$ denote zero vector or matrix, $\boldsymbol{e}_i$ denote the vector with $i$-th entry being 1 and 0 otherwise, and $\boldsymbol{1}$ denote all-ones vector; their dimensions are clear from the context. For a sequence of compatible matrices $\{A_i\}$, $\prod_{k=i}^jA_k = A_j A_{j-1}\cdots A_i$ if $j\geq i$ and $I$ if $j<i$. For a matrix $A$, $\lambda_{\min}(A)$ ($\lambda_{\max}(A)$) denotes the smallest (largest) eigenvalue of $A$. We also let $F_t = F(\bx_t)$ and $F^\star = F(\bx^\star)$ (similar for $\nabla F_t, \nabla^2F_t$, etc.), and let $\mathbf{1}_{\{\cdot\}}$ denote the indicator function.

\section{Online Sketched Newton Method}\label{sec:2}

\vspace{-0.1cm}

At a high level, the online sketched Newton method takes the following update scheme:$\hskip2cm$
\begin{equation}\label{nequ:7}
\bx_{t+1} = \bx_t + \baralpha_t \barDelta\bx_t,
\end{equation}
where $\barDelta\bx_t$ \textit{approximately} solves the Newton system $B_t\Delta\bx_t = - \nabla f(\bx_t; \xi_t)$ via the sketching solver (see \eqref{nequ:8}) and $\baralpha_t$ is an \textit{adaptive}, \textit{potentially random} stepsize (see \eqref{nequ:6}). 

More precisely, given the current iterate $\bx_t$, we randomly generate a sample $\xi_t\sim \P$ and obtain the gradient and Hessian estimates:
\begin{equation*}
\bar{g}_t = \nabla f(\bx_t;\xi_t)\hskip1.3cm \text{ and }\hskip1.3cm \bar{H}_t = \nabla^2 f(\bx_t;\xi_t).
\end{equation*}
Then, we define $B_t$ to be the Hessian average using samples $\{\xi_i\}_{i=0}^{t-1}$, expressed as \vskip-0.1cm
\begin{equation}\label{nequ:5}
B_t = \frac{1}{t}\sum_{i=0}^{t-1} \barH_i \hskip2cm \stackrel{\text{online update}}{\Longrightarrow} \hskip0.5cm B_t = \frac{t-1}{t}B_{t-1} + \frac{1}{t}\barH_{t-1}.
\end{equation}
In this paper, we use $\bar{(\cdot)}$ to denote a random quantity that depends on the current sample $\xi_t$. Note that the estimate $\barH_t$ is only used in the $(t+1)$-th iteration; thus $B_t$ is deterministic conditional on $\bx_t$ (this is why we do not use the notation $\bar{B}_t$). The Hessian average is widely used in Newton methods to accelerate the convergence rate \citep{Na2022Hessian}. In certain problems, $B_t$ can be expressed as the sum of rank-1 matrices, allowing its inverse to be updated online in a manner similar to \eqref{nequ:5} \citep{Bercu2020Efficient, Cenac2020efficient, Boyer2022asymptotic, Leluc2023Asymptotic}. However, solving the Newton system $B_t\Delta\bx_t = -\barg_t$ for a generic stochastic function can be expensive.

We now employ the sketching solver to approximately solve $B_t\Delta\bx_t = -\bar{g}_t$. At each inner iteration $j$, we generate a sketching matrix/vector $S_{t,j}\in\mathbb{R}^{d\times q}\sim S$ for some $q\geq 1$ and solve the subproblem:$\hskip1cm$
\begin{equation}\label{sec2:equ1}
\Delta\bx_{t,j+1} = \argmin_{\Delta\bx}\|\Delta\bx-\Delta\bx_{t,j}\|^2,\quad\quad\text{ s.t. }\quad \;\; S_{t,j}^TB_t\Delta\bx = -S_{t,j}^T\bar{g}_t.
\end{equation}
In particular, we only aim to solve the sketched Newton system $S_{t,j}^TB_t\Delta\bx = -S_{t,j}^T\bar{g}_t$ at the $j$-th inner iteration, and we prefer the solution that is as close as possible to the current solution approximation $\Delta\bx_{t,j}$. The closed-form recursion of \eqref{sec2:equ1} is  ($\Delta\bx_{t,0}=\0$):
\begin{equation}\label{sec2:equ2}
\Delta\bx_{t,j+1} = \Delta\bx_{t,j} - B_tS_{t,j}(S_{t,j}^TB_t^2S_{t,j})^{\dagger}S_{t,j}^T(B_t\Delta\bx_{t,j}+\bar{g}_t),
\end{equation}
where $(\cdot)^{\dagger}$ denotes the Moore-Penrose pseudoinverse. When we employ a sketching vector ($q=1$), the pseudoinverse reduces to the reciprocal, meaning that solving the Newton system is \textit{matrix-free} --- no matrix factorization is needed. 
We briefly discuss various trade-offs of choosing dense/sparse sketching matrices/vectors in Remark \ref{rem:2}, while we refer to \cite{Strohmer2008Randomized, Woodruff2014Computational, Gower2015Randomized, Derezinski2024Sharp, Na2025Statistical} for detailed quantitative analyses under proper conditions on the Hessians' sparsity patterns and their eigenvalue decay structures.$\hskip 2cm$

For a deterministic integer $\tau$, we let
\begin{equation}\label{nequ:8}
\barDelta\bx_t = \Delta\bx_{t, \tau},
\end{equation}
and then we update the iterate $\bx_t$ as in \eqref{nequ:7} with a potentially random stepsize $\baralpha_t$ satisfying
\begin{equation}\label{nequ:6}
\beta_{t}\leq \baralpha_t\leq \beta_{t}+\chi_t \quad\quad\quad \text{with}\quad\quad \beta_t=\frac{c_{\beta}}{(t+1)^{\beta}} \quad\text{and}\quad \chi_t=\frac{c_{\chi}}{(t+1)^{\chi}}.
\end{equation}
The motivation for using a well-controlled random stepsize is to enhance the adaptivity of the method without compromising the asymptotic normality guarantee. Particularly, different directions may prefer different stepsizes, so that $\baralpha_t$ depends on $\barDelta\bx_t$ and is random. \cite{Berahas2021Sequential, Berahas2023Stochastic, Curtis2024Stochastic} have proposed various adaptive stepsize selection schemes for Newton methods on constrained problems that precisely satisfy the condition in \eqref{nequ:6}.

\begin{remark}[Discussions on sketching parameters.]\label{rem:2}

In this remark, we discuss the computation-accuracy trade-offs of the sketching solver \eqref{sec2:equ2} under different choices of sketching parameters. We should mention that the present paper focuses on covariance matrix estimation for sketched Newton methods; refined convergence analyses of sketching solvers under various Hessian conditions and their practical guidelines have been extensively explored in existing literature. See \cite{Strohmer2008Randomized, Gower2015Randomized, Derezinski2024Sharp, Woodruff2014Computational, Na2025Statistical} and references therein.$\quad\;\;$

The quality of the approximate solution $\barDelta\bx_t$ is governed by three interacting quantities: the sketching dimension $q$, the sketching distribution $S$, and the number of sketching steps $\tau$. Suppose the sketching distribution $S$ satisfies $\mE[B_tS(S^TB_t^2S)^{\dagger}S^TB_t \mid \bx_{0:t}]\succeq \gamma_S\cdot I$ for some $\gamma_S>0$ (as assumed in Assumption \ref{ass:4}), it has been shown in aforementioned literature that
\begin{equation*}
\mE[\|\barDelta\bx_t - \Delta\bx_t\|^2\mid \bx_{0:t}] \leq \rho^\tau \|\Delta\bx_t\|^2\quad\quad \text{ with }\quad \rho = 1-\gamma_S.
\end{equation*}
Furthermore, for many common sketching distributions (e.g., Gaussian, Rademacher, Kaczmarz sketches etc.), it is well known that $\gamma_S = O(q/d)$, and that decaying the expected error below a given threshold requires $\tau = O(1/\{\log(1/\rho)\}) = O(1/\gamma_S) = O(d/q)$ sketching steps.

Consequently, for matrix sketches $S \in \mR^{d \times q}$ with $q=O(d)$, the sketching solver exhibits a constant (with respect to $d$) convergence rate $\rho$, resulting in a constant number of sketching steps $\tau=O(1)$. In this case, the dominant computational cost arises from the update \eqref{sec2:equ2}, which requires $O(d^2 q)$ flops in total, or $O(\texttt{nnz}(S)\,d)$ for sparse sketches, where $\texttt{nnz}(S)$ denotes the number of nonzero entries of $S$. Thus, increasing the sketching dimension $q$ improves the convergence rate of the sketching solver but also incurs a higher one-shot computational cost, making matrix sketches attractive when a small number of high-quality sketching steps is desired. In contrast, vector sketches $S \in \mR^d$ (i.e., $q=1$) incur a lower per-step cost of $O(d^2)$ (or $O(d)$ for sparse vectors), but require $\tau = O(d)$ sketching steps to decay the approximation error. 
Overall, the total computational cost of the sketching solver, given by the product of the number of sketching steps and the cost per step, ranges from $O(d^2)$ to $O(d^3)$, with the worst-case complexity matching that of exact Newton solvers.

In our covariance matrix estimation theory, different sketching parameters do not affect the convergence rate of the covariance matrix estimator, which is determined by the stepsize of the Newton method, i.e., the convergence rate to stationarity of the Newton iterates. That said, sketching parameters do affect constant factors in the rate. Our Lemma \ref{sec4:lem1} and Theorem \ref{sec4:thm1} make the dependence on $\gamma_S$ and $\tau$ explicit.$\quad\;\;$
\end{remark}

\section{Assumptions and Asymptotic Normality}\label{sec:3}

In this section, we introduce assumptions and present the asymptotic normality guarantee for sketched Newton methods. Our presentation is adapted from \cite[Theorems 4.8 and 5.6]{Na2025Statistical} by restricting to the unconstrained strongly convex setting and refining some assumptions.
Throughout the paper, we let $\mF_t = \sigma(\{\xi_i\}_{i=0}^t)$, $\forall t\geq 0$ be the filtration of $\sigma$-algebras generated by the sample sequence $\xi_0, \xi_1,\xi_2\ldots$.

\subsection{Assumptions}\label{sec:3.1}

We first impose a Lipschitz continuity condition on the objective Hessian $\nabla^2  F(\bx)$, which is standard in existing literature \citep{Bercu2020Efficient, Cenac2020efficient, Na2025Statistical}.

\vspace{-0.15cm}

\begin{assumption}\label{ass:1}
We assume that $F(\bx)$ is twice continuously differentiable and its Hessian $\nabla^2F(\bx)$ is $\Upsilon_{L}$-Lipschitz continuous. In particular, for any $\bx$ and $\bx^{\prime}$, we have $\|\nabla^2 F(\bx) - \nabla^2 F(\bx^{\prime})\|\leq \Upsilon_{L}\|\bx-\bx^{\prime}\|$.
\end{assumption}

\vspace{-0.15cm}

The next assumption regards the noise in stochastic gradients. We assume that the fourth conditional moment of the gradient noise satisfies a growth condition.

\begin{assumption}\label{ass:2}

We assume the function $f(\bx;\xi)$ is twice continuously differentiable with respect to $\bx$ for any $\xi$, and $\|\nabla f(\bx;\xi)\|$ is uniformly integrable for any $\bx$. This implies $\mathbb{E}[\bar{g}_t\mid \mathcal{F}_{t-1}]=\nabla F_t$. Furthermore, there exist constants $C_{g,1}, C_{g,2}>0$ such that
\begin{equation}\label{ass:2:4th}
\mE[\|\bar{g}_t - \nabla F_t\|^4\mid\mathcal{F}_{t-1}]\leq C_{g,1}\|\bx_t-\bx^\star\|^4 + C_{g,2}, \quad\quad \forall t\geq 0.
\end{equation}
\end{assumption}

Assumption \ref{ass:2} assumes that at each step, the gradient estimate $\barg_t=\nabla f(\bx_t;\xi_t)$ is unbiased and the noise $\barg_t-\nabla F_t$ satisfies a fourth-moment growth condition conditional on $\bx_t$.
In particular, we do not require the noise to have a uniformly bounded fourth moment; instead, we allow the moment to grow as $\bx_t$ moves away from the solution $\tx$. This assumption is implied by requiring (1) the noise at the solution $\nabla f(\tx;\xi)-\nabla F^\star$ has a bounded fourth moment; (2) the noise difference between $\bx_t$ and $\tx$ satisfies a Lipschitz continuity property. The latter condition follows directly from standard smoothness assumptions.

Assumption \ref{ass:2} is standard in existing literature on online inference. In fact, for establishing asymptotic normality alone, the fourth-moment condition in \eqref{ass:2:4th} can be relaxed to a $(2+\epsilon)$-moment condition, as assumed in \cite[Assumptions 5 and 8]{Leluc2023Asymptotic}. 
However, the fourth-moment growth condition is widely imposed for limiting covariance estimation, since estimating covariance relies on higher-order information about the noise compared to merely characterizing the stationary distribution.
See \cite[Assumption 3.2(3)]{Chen2020Statistical} and \cite[Assumption 2(3)]{Zhu2021Online} for precisely the same assumption for covariance estimation of SGD. For nonlinear and nonsmooth problems, even stronger moment conditions have been imposed in the literature. For example, a uniformly bounded fourth-moment condition is assumed in \cite[Assumption 4.2]{Na2025Statistical} for the plug-in covariance estimator of stochastic Newton method, and an eighth-moment growth condition is assumed in \cite[Assumption 3.4(3)]{Jiang2025Online} for the batch-means covariance estimator of projected SGD. With these being said, Assumption \ref{ass:2} has been explicitly verified for linear and logistic regression models with Gaussian design covariates in \cite[Lemma 3.1 and Appendix A]{Chen2020Statistical}. In fact, the same analysis applies to more general design covariates, as long as the covariates have sufficient high-order moments (i.e., they can be heavy-tailed).

The next assumption imposes lower and upper bounds for stochastic Hessians, with a growth condition on the Hessian noise.

\begin{assumption}\label{ass:3}
There exist constants $\Upsilon_H>\gamma_H>0$ such that for any $\xi$ and any $\bx$,
\begin{equation}\label{ass:3:a}
\gamma_H\leq\lambda_{\min}(\nabla^2 f(\bx;\xi))\leq \lambda_{\max}(\nabla^2 f(\bx;\xi))\leq \Upsilon_H,
\end{equation}
which implies $\mathbb{E}[\bar{H}_t\mid \mathcal{F}_{t-1}]=\nabla^2 F_t$. Furthermore, there exist constants $C_{H,1}, C_{H,2}>0$ such that
\begin{equation}\label{ass:3:4th}
\mE\left[\|\bar{H}_t-\nabla^2 F_t\|^4\mid\mF_{t-1}\right]\leq C_{H,1}\|\bx_t-\bx^\star\|^4 + C_{H,2}, \quad\quad \forall t\geq 0.
\end{equation}

\end{assumption}

The condition \eqref{ass:3:a} is widely used in the literature on stochastic second-order methods \citep{Byrd2016Stochastic, Berahas2016Multi, Moritz2016Linearly}. By the averaging structure of $B_t$, we know \eqref{ass:3:a} implies
\begin{equation}\label{ass:3:c2}
\gamma_H\leq\lambda_{\min}(B_t)\leq\lambda_{\max}(B_t)\leq\Upsilon_H.
\end{equation}
As discussed, \cite[Lemma 3.1]{Chen2020Statistical} shows that condition \eqref{ass:3:a} together with the boundedness of $\mE[|\nabla f(\bx^\star;\xi)|^4]$ implies \eqref{ass:2:4th}.
The growth condition on the Hessian noise in \eqref{ass:3:4th} is analogous to that imposed on the gradient noise in \eqref{ass:2:4th}. We require the same order of moments due to the observation that the Hessian preconditions the gradient in Newton methods, transforming the direction from $\barg_t$ to $B_t^{-1}\barg_t$. Moreover, the analysis in \cite[Appendix A]{Chen2020Statistical} shows that condition \eqref{ass:3:4th} holds for linear and logistic regression models with Gaussian designs, and the analysis extends directly to more general designs satisfying bounded moment conditions.

We finally require the following assumption regarding the sketching distribution.

\begin{assumption}\label{ass:4}
For $t\geq 0$, we assume the sketching matrix $S_{t,j}\stackrel{iid}{\sim}S$ satisfies $\mE[B_tS(S^TB_t^2S)^{\dagger}S^TB_t \mid \mF_{t-1}]\succeq \gamma_S I$ and $\mE[\|S\|^2\|S^{\dagger}\|^2]\leq \Upsilon_S$ for some constants $\gamma_S, \Upsilon_S>0$.
\end{assumption}

The above two expectations are taken over the randomness of the sketching matrix $S$. The lower bound of the projection matrix $B_tS(S^TB_t^2S)^{\dagger}S^TB_t$ is commonly required by sketching solvers to ensure convergence \citep{Gower2015Randomized}. We trivially have $\gamma_S\leq 1$. The bounded second moment of the condition number of $S$ is necessary to analyze the difference $\|B_tS(S^TB_t^2S)^{\dagger}S^TB_t-B^{\star}S(S^T(B^{\star})^2S)^{\dagger}S^TB^{\star}\|$ between two projection matrices \citep[Lemma 5.2]{Na2025Statistical}. Under \eqref{ass:3:c2}, both conditions easily hold for various sketching distributions, such as Gaussian sketching $S\sim \mN(\0, \Sigma)$ and Uniform sketching $S\sim \text{Unif}(\{\be_i\}_{i=1}^d)$, where $\be_i$ is the $i$-th canonical basis of $\mR^d$ (called randomized Kaczmarz method \citep{Strohmer2008Randomized}).

\subsection{Almost sure convergence and asymptotic normality}\label{sec3:subsec2}

We review the almost sure convergence and asymptotic normality of the sketched Newton method, which serve as fundamental results for subsequent covariance estimation and statistical inference on $\tx$. As mentioned, our following theoretical presentation is adapted from \cite[Theorems 4.8 and 5.6]{Na2025Statistical}; however, we refine the study of \cite{Na2025Statistical} in two aspects. (1) The growth conditions on the gradient and Hessian noises in Assumptions \ref{ass:2} and \ref{ass:3} are weaker than the uniformly bounded moment conditions in \cite[Assumption 4.2]{Na2025Statistical}. (2) By a sharper analysis, the requirement on the stepsize adaptivity gap $\chi_t$ is relaxed from $\chi>1$ in \cite[(4.4)]{Na2025Statistical} to $\chi>0.5(\beta+1)$. We show that the results \cite[Theorems 4.8 and 5.6]{Na2025Statistical} still hold under these weaker conditions, with proofs deferred to Appendix \ref{pf:sec3:subsec2} for completeness.

\begin{theorem}[Almost sure convergence]\label{sec3:thm1}

\hspace{-0.15cm}Consider the iteration scheme \eqref{nequ:7}. Suppose Assumptions \ref{ass:1} -- \ref{ass:4} hold, the number of sketches satisfies $\tau\geq \log(\gamma_H/4\Upsilon_H)/\log \rho$ with $\rho=1-\gamma_S$, and the stepsize parameters satisfy $\beta\in(0.5,1]$, $\chi>0.5(\beta+1)$, and $c_{\beta}, c_{\chi}>0$. Then, we have $\bx_t\rightarrow \tx$ as $t\rightarrow\infty$ almost surely.

\end{theorem}

To present the normality result, we need to introduce some additional notation. Let $B^\star = \nabla^2F(\tx)$. For $S_1,\ldots,S_\tau\stackrel{iid}{\sim}S$, we define the product of $\tau$ projection matrices as
\begin{equation}\label{exp:Cstar}
\tilde{C}^\star = \prod_{j=1}^\tau(I - B^\star S_j(S_j^T(B^\star)^2S_j)^\dagger S_j^TB^\star)
\end{equation}
and let $C^\star=\mE[\tilde{C}^\star]$. Then, we denote the eigenvalue decomposition of $I-C^\star$ as
\begin{equation}\label{eigendecomp}
I-C^\star = U\Sigma U^T\quad\quad\text{with}\quad\quad \Sigma=\text{diag}(\sigma_1,\dots,\sigma_d).
\end{equation}
We also define
\begin{equation}\label{exp:Omegastar}
\Omega^\star = (B^\star)^{-1}\mE[\nabla f(\bx^\star;\xi)\nabla^T f(\bx^\star;\xi)](B^\star)^{-1}.
\end{equation} 

With the above notation, we have the following normality guarantee for the scheme \eqref{nequ:7}.

\begin{theorem}[Asymptotic normality]\label{sec3:thm2}

\hspace{-0.15cm}Suppose Assumptions \ref{ass:1} -- \ref{ass:4} hold, the number of sketches satisfies $\tau\geq \log(\gamma_H/4\Upsilon_H)/\log \rho$ with $\rho = 1-\gamma_S$, and the stepsize parameters satisfy $\beta\in(0.5,1]$, $\chi>1.5\beta$, and $c_{\beta}>1/\{1.5(1-\rho^\tau)\}$ for $\beta=1$. Then, we have 
\begin{equation}\label{equ:NTclt}
\sqrt{1/\baralpha_t}(\bx_t-\bx^\star)\stackrel{d}{\longrightarrow}\mN(\mathbf{0},\Xi^\star),
\end{equation}
where $\Xi^{\star}$ is the solution to the following Lyapunov equation:
\begin{equation}\label{equ:lyap}
\rbr{\cbr{1 - \frac{\b1_{\{\beta=1\}}}{2c_\beta} }I - C^\star}\Xi^{\star} + \Xi^{\star}\rbr{\cbr{1 - \frac{\b1_{\{\beta=1\}}}{2c_\beta}}I - C^\star} = \mathbb{E}\big[(I-\widetilde{C}^{\star}) \Omega^{\star}(I-\widetilde{C}^{\star})^T\big].
\end{equation}
\end{theorem}

In fact, the limiting covariance $\Xi^{\star}$ has an explicit form as:
\begin{equation}\label{exp:Xistar}
\Xi^\star = U\big(\Theta\circ U^T\mE[(I-\tilde{C}^*)\Omega^\star(I-\tilde{C}^*)^T]U\big)U^T\quad\text{with}\quad [\Theta]_{k,l}= \frac{1}{\sigma_k + \sigma_l - \b1_{\{\beta=1\}}/c_\beta},
\end{equation}
where $\circ$ denotes the matrix Hadamard product. There exists a degenerate case. When the Newton systems are exactly solved ($\tau=\infty$), then $\tilde{C}^\star = C^\star = \0$, $\Sigma=I$, and $\Xi^\star = \Omega^\star/(2- \b1_{\{\beta=1\}}/c_\beta)$. In this case, we have $\Xi^\star = \Omega^\star/2$ for $\beta\in(0.5,1)$ and $\Xi^\star = \Omega^\star$ for $\beta=c_{\beta}=1$. For the latter setup, we know $\Xi^\star = \Omega^\star$ achieves the asymptotic minimax lower bound \citep{Duchi2021Asymptotic}.

\section{Online Covariance Matrix Estimation}\label{sec:4}

In this section, we build upon the results in Section \ref{sec:3} and construct an online estimator for the limiting covariance matrix $\Xi^\star$. With the covariance estimator, we are then able to perform online statistical inference, such as constructing asymptotically valid confidence intervals for model parameters. $\hskip2.5cm$

\subsection{Weighted sample covariance estimator}\label{sec:4.1}

As introduced in Section \ref{sec:1}, existing literature \cite{Leluc2023Asymptotic, Na2025Statistical} has shown that Hessian preconditioning improves the stationarity properties of the last Newton iterate compared to the last SGD iterate. In particular, the last exact Newton iterate, without iterate averaging, can achieve the same statistical optimality as the averaged SGD iterate (both with proper stepsizes). This observation motivates estimating the covariance $\Xi^\star$ of the last Newton iterate by exploiting its normality in \eqref{equ:NTclt}. To this end, we draw inspiration from covariance estimation of a stationary sequence $\{\bx_t\}$ with mean $\tx$, for which $\Cov(\bx_t)$ can be consistently estimated by the sample averaging $\frac{1}{t}\sum_{i=1}^t (\bx_i-\tx)(\bx_i-\tx)^T$. In our study, the iterates $\{\bx_t\}$ are clearly nonstationary; according to the asymptotic normality in \eqref{equ:NTclt} (see also Lemma \ref{sec4:lem1} with $\chi>1.5\beta$, i.e., $\chi_t=o(\beta_t^{1.5})$), we know $\mE[\|\bx_t-\tx\|^2] = O(\beta_t)$. Therefore, we propose reweighting each sample variance $(\bx_i-\tx)(\bx_i-\tx)^T$ by a factor of $O(1/\beta_i)$, and replacing the unknown mean $\tx$ with the averaged iterate $\barx_t=\sum_{i=1}^t \bx_i/t$. This leads to the weighted sample covariance estimator proposed in this work.

Specifically, let $\varphi_t = \beta_t + \chi_t/2$ be the centered stepsize. Our weighted sample covariance estimator is defined as
\begin{equation}\label{exp:Xihat}
\hat{\Xi}_t = \frac{1}{t}\sum_{i=1}^t \frac{1}{\varphi_{i-1}}(\bx_i-\bar{\bx}_t)(\bx_i-\bar{\bx}_t)^T\quad\quad\quad\text{with}\quad\quad\quad \bar{\bx}_t = \frac{1}{t}\sum_{i=1}^t\bx_i.
\end{equation}
This estimator can be rewritten as
\begin{equation*}
\hat{\Xi}_t = W_t - \bv_t\bar{\bx}_t^T - \bar{\bx}_t\bv_t^T + a_t\bar{\bx}_t\bar{\bx}_t^T,
\end{equation*}
where
\begin{equation}\label{nequ:9}
W_t = \frac{1}{t}\sum_{i=1}^t \frac{1}{\varphi_{i-1}}\bx_i\bx_i^T,\quad \quad\quad\bv_t = \frac{1}{t}\sum_{i=1}^t \frac{1}{\varphi_{i-1}}\bx_i,\quad\quad\quad a_t = \frac{1}{t}\sum_{i=1}^t\frac{1}{\varphi_{i-1}}.
\end{equation}
We mention that $W_t, \bv_t, \bar{\bx}_t, a_t$ can all be updated recursively, meaning that $\hat{\Xi}_t$ can be computed in a fully online fashion. The detailed steps are shown in Algorithm \ref{alg:2}.

\begin{algorithm}[t!]
\caption{Construction of Weighted Sample Covariance Estimator}	\label{alg:2}
\begin{algorithmic}[1]
\State\textbf{Input:} initial iterate $\bx_0$, positive sequences $\{\beta_t, \chi_t\}$, an integer $\tau>0$, $B_0 = I$;
\State\textbf{Initialize:} $W_0 = \boldsymbol{0}\in\mathbb{R}^{d\times d}$, $\bv_0 = \bar{\bx}_0 = \boldsymbol{0}\in\mathbb{R}^d$, $a_0 = 0$
\For{$t = 0,1,2,\ldots$}
\State Obtain the sketched Newton iterate $\bx_{t+1}$ and let $\varphi_t = \beta_t+\chi_t/2$;
\State Update the quantities as:
\begin{align*}
W_{t+1} & = \frac{t}{t+1}W_t + \frac{1/\varphi_t}{(t+1)}\;\bx_{t+1}\bx_{t+1}^T,\quad\quad && \bv_{t+1}  = \frac{t}{t+1}\bv_t + \frac{1/\varphi_t}{t+1}\;\bx_{t+1};\\
\bar{\bx}_{t+1} & = \frac{t}{t+1}\bar{\bx}_t + \frac{1}{t+1}\bx_{t+1},\quad\quad && a_{t+1} = \frac{t}{t+1}a_t + \frac{1/\varphi_t}{t+1};
\end{align*}
\EndFor
\State\textbf{Output:} Covariance estimator $\hat{\Xi}_t = W_t - \bv_t\bar{\bx}_t^T - \bar{\bx}_t\bv_t^T + a_t\bar{\bx}_t\bar{\bx}_t^T$.
\end{algorithmic}
\end{algorithm}

We note that the estimator $\hat{\Xi}_t$ is in a similar flavor to batch-means covariance estimators designed for first-order online methods. In particular, \citet{Chen2020Statistical, Zhu2021Online} grouped SGD iterates into multiple batches and estimated the covariance $\tOmega$ in \eqref{nequ:2} by computing the sample covariance  among batches. Here, we clarify three differences. First, our estimator $\hat{\Xi}_t$ targets estimating the limiting covariance of the \textit{last Newton} iterate, while existing literature \cite{Chen2020Statistical, Zhu2021Online} targets estimating the limiting covariance of the \textit{averaged SGD} iterate. We recall the end of Section \ref{sec:1} and the beginning of Section \ref{sec:4.1} (cf. Table \ref{tab:1}) for the justification of inference based on the last Newton iterate. In fact, as suggested by \eqref{equ:lyap}, the two limiting covariance matrices are closely related. In particular, for exact Newton methods, we have $\tXi = \tOmega$ when $\beta_t = 1/(t+1)$ and $\tXi = \tOmega/2$ when $\beta_t = 1/(t+1)^\beta$ for any $\beta\in(0.5,1)$. Thus, both our estimator $\hat{\Xi}_t$ and batch-means estimators in \cite{Chen2020Statistical, Zhu2021Online} can be used to estimate the optimal covariance $\tOmega$. Second, compared to batch-means estimators in \cite{Chen2020Statistical, Zhu2021Online}, our estimator $\hat{\Xi}_t$ is \textit{batch-free}. Specifically, batch-means estimators rely on additional batch size sequences that must satisfy certain conditions and largely affect both theoretical and empirical performance of the estimators. In contrast, we assign proper weights to the iterates based on the stepsizes, eliminating the need to tune any extra parameters beyond those required by the online method itself. Third, as shown in Theorem \ref{sec4:thm1} later, the convergence rate of our estimator $\hat{\Xi}_t$ is provably faster than that of batch-means estimators, improving from $O_p(1/\sqrt[4]{t\beta_t})$ (with an optimal batch size sequence) to $O_p(1/\sqrt{t\beta_t})$. This improved convergence rate provides further evidence on the benefits of leveraging Hessian information (if available) in statistical inference.

We observe that the memory and computational complexities of inference based on sketched Newton methods are comparable to those of first-order methods. The memory complexity is dominated by storing $B_t$ and $W_t$, incurring a cost of $O(d^2)$, independent of the sample size $t$. Furthermore, as discussed in Remark \ref{rem:2}, the computational complexity includes $O(\tau\cdot\texttt{nnz}(S)d)$ flops for computing the sketched Newton direction and $O(d^2)$ flops for updating $\hat{\Xi}_t$, where $\texttt{nnz}(S)$ denotes the number of nonzero entries of $S$. For instance, when $S\sim \text{Unif}(\{\be_i\}_{i=1}^d)$, \citet{Na2025Statistical} showed $\tau = O(d)$, suggesting that the overall computational complexity of sketched Newton inference is $O(d^2)$. This order precisely matches the complexity in \cite{Chen2020Statistical, Zhu2021Online}.

\begin{remark}
In this remark, we explain the reason of using the averaged iterate $\barx_t = \sum_{i=1}^{t}\bx_i/t$ in the definition \eqref{exp:Xihat} instead of the last iterate $\bx_t$, although both converge to $\tx$. The key reason is that $\barx_t$ converges to $\tx$ at a faster rate than $\bx_t$. In particular, when $\chi>1.5\beta$ (i.e., $\chi_t = o(\beta_t^{1.5})$), we have
\begin{equation*}
\mE [\|\barx_t-\tx\|^2] = O(1/t) < O(\beta_t) = \mE [\|\bx_t-\tx\|^2],
\end{equation*}
where the first equality is established in Lemma \ref{appen:A5:lem2} and the last equality in Lemma \ref{sec4:lem1}. Employing an estimator of $\tx$ with a faster convergence rate is crucial for ensuring the consistency of $\hat{\Xi}_t$. At a high level, the estimation error $\hat{\Xi}_t-\tXi$ can be decomposed as (see Appendix \ref{appen:A5} for rigorous derivations):$\hskip1.5cm$
\begin{equation*}
\mE [\|\hat{\Xi}_t-\tXi\|] \;\lesssim\; \mE\!\bigg[\Big\|\frac{1}{t}\sum_{i=1}^t\frac{1}{\varphi_{i-1}}(\bx_i-\bx^\star)(\bx_i-\bx^\star)^T-\Xi^\star\Big\|\bigg] \;+\; \sqrt{\frac{1}{t}\sum_{i=1}^t\frac{1}{\varphi_{i-1}}\mE\!\left[\|\barx_t-\bx^\star\|^2\right]},
\end{equation*}
where the first term characterizes the error of estimating $\tXi$ by weighted sample covariance of the centered iterates, and the second term characterizes the error of estimating $\tx$ by $\barx_t$.
For the second term, it can be shown that $\frac{1}{t}\sum_{i=1}^t \frac{1}{\varphi_{i-1}} = O(1/\beta_t)$. Combined with the rate $\mE[\|\barx_t-\tx\|^2] = O(1/t)$, we know the second term is of order $O(1/\sqrt{t\beta_t})$. In contrast, if $\bx_t$ were used in place of $\barx_t$, the second term would be of constant order $O(1)$, breaking the consistency of the estimator $\hat{\Xi}_t$.
\end{remark}

\begin{remark}\label{rem:1}
We compare $\hat{\Xi}_t$ with the plug-in estimator proposed in \cite{Na2025Statistical}. Due to significant challenges in estimating sketch-related quantities in \eqref{equ:lyap}, \citet{Na2025Statistical} simply neglected all those quantities and estimated $\tOmega/(2-\b1_{\{\beta=1\}}/c_\beta)$ instead. Their plug-in estimator is defined as:
\begin{equation}\label{exp:PI}
\tilde{\Xi}_t = \frac{1}{2 - \b1_{\{\beta=1\}}/c_\beta} \cdot B_t^{-1}\Big(\frac{1}{t}\sum_{i=1}^t\bar{g}_i\bar{g}_i^T\Big)B_t^{-1}.
\end{equation}
Comparing $\tilde{\Xi}_t$ with $\hat{\Xi}_t$, we clearly see that $\tilde{\Xi}_t$ is not matrix-free as it involves the inverse of $B_t$ (i.e., $O(d^3)$ flops), which contradicts the spirit of using sketching solvers. Furthermore, $\tilde{\Xi}_t$ is a biased estimator of $\Xi^\star$, leading to invalid confidence coverage even as $t\rightarrow\infty$.
\end{remark}

\subsection{Convergence rate of the estimator }\label{sec4:subsec1}

To establish the convergence rate of $\hat{\Xi}_t$, we first present a preparation result that provides error bounds for the Newton iterate $\bx_t$ and the averaged Hessian $B_t$. We show that the fourth moments of $\|\bx_t-\tx\|$ and $\|B_t- B^\star\|$ scale as $O(\beta_t^2+\chi_t^4/\beta_t^4)$. 
When $\chi_t \gtrsim \beta_t^{1.5}$, the error $\chi_t^4/\beta_t^4$ incurred by the adaptivity of stepsize dominates. In contrast, when $\chi_t\lesssim \beta_t^{1.5}$, adaptive stepsizes lead only to a higher-order error. A matching error bound (for the iterate $\bx_t$) has been established for SGD methods with $\chi_t=0$ \citep{Chen2020Statistical}. That said, our analysis is more involved due to higher-order methods, sketching components, and randomness in stepsizes. 

\begin{lemma}[Error bounds of $\bx_t$ and $B_t$]\label{sec4:lem1}
Suppose Assumptions \ref{ass:1} -- \ref{ass:4} hold, the number of sketches satisfies $\tau\geq \log(\gamma_H/4\Upsilon_H)/\log \rho$ with $\rho = 1-\gamma_S$, and the stepsize parameters satisfy $\beta\in(0,1)$, $\chi>\beta$, and $c_{\beta}, c_{\chi}>0$. Then, we have
\begin{equation*}
\mE\big[\|\bx_t-\bx^\star\|^4\big]\lesssim \beta_t^2 + \frac{\chi_t^4}{\beta_t^4}\quad\quad\text{ and }\quad\quad \mE\big[\|B_t-B^\star\|^4\big]\lesssim \beta_t^2 + \frac{\chi_t^4}{\beta_t^4}.
\end{equation*}

\end{lemma}

With the above lemma, we show the convergence rate of $\hat{\Xi}_t$ in the following theorem.

\begin{theorem}\label{sec4:thm1}
Under the conditions in Lemma \ref{sec4:lem1}, except for strengthening $\chi>\beta$ to $\chi>1.5\beta$, the covariance estimator $\hat{\Xi}_t$ defined in \eqref{exp:Xihat} satisfies
\begin{equation*}
\mE\sbr{\big\|\hat{\Xi}_t-\Xi^\star\big\|}\lesssim
\begin{dcases}
\frac{1}{{(1-\rho^\tau)^{1.5}}}\bigg(\sqrt{\beta_t} \; +\; \frac{\chi_t}{\beta_t^{1.5}}\bigg), \quad & \text{ for } 0<\beta\leq 0.5,\\
\frac{1}{{(1-\rho^\tau)^{1.5}}}\bigg(\frac{1}{\sqrt{t\beta_t}} \; +\; \frac{\chi_t}{\beta_t^{1.5}}\bigg),\quad & \text{ for } 0.5<\beta< 1,
\end{dcases}
\end{equation*}
where we explicitly track the dependency of the constant factor on $\rho = 1-\gamma_S$.
\end{theorem}

Since $\sqrt{\beta_t} \vee \chi_t/\beta_t^{1.5}\rightarrow 0$ and $t\beta_t\rightarrow \infty$ as $t\rightarrow\infty$ (because $\chi>1.5\beta$), Theorem \ref{sec4:thm1} states that $\hat{\Xi}_t$ is an (asymptotically) consistent estimator of $\Xi^{\star}$. Note that $\chi>1.5\beta$ is already required by the asymptotic normality guarantee (cf. Theorem \ref{sec3:thm2}). 
From the result, we see that if $\beta \in (0,0.5]$, then choosing $\chi > 2\beta$ (i.e., $\chi_t \lesssim \beta_t^2$) makes the adaptivity error term $\chi_t / \beta_t^{1.5}$ of higher order (although this case is less interesting, since existing normality analyses typically require $\beta > 0.5$).
If $\beta \in (0.5,1)$, then choosing $\chi > \beta + 0.5$ (i.e., $\chi_t\lesssim\beta_t/\sqrt{t}$) makes the adaptivity error term $\chi_t / \beta_t^{1.5}$ of higher order.

We next discuss the effect of sketching on the convergence behavior of our covariance estimator $\hat{\Xi}_t$ when asymptotic normality is attained ($\beta>0.5$). First, sketched Newton preserves the same convergence rate as exact Newton and only affects the convergence rate through the constant factor. As shown in \eqref{pf:res}, the dependency of the constant factor on sketching parameters $(\rho, \tau)$ is fully captured by the ratio $1/(1-\rho^\tau)^{1.5}$. 
Second, for a fixed $\rho = 1-\gamma_S$, the constant factor improves as the number of sketching steps $\tau$ increases, indicating that additional sketching iterations reduce the constant factor at the expense of higher computational cost. In the extreme case of $\tau = \infty$, the dependency of the constant factor on $\rho$ (i.e., on different choices of sketching matrices) vanishes, since regardless of the sketching distribution, the sketching solver reduces to the exact Newton solver when $\tau=\infty$. Third, similarly, a larger lower bound $\gamma_S$ (cf. Assumption \ref{ass:4}) implies a smaller $\rho = 1-\gamma_S$, which also yields a better constant factor. In the extreme case of $S = I$, we have $\gamma_S = 1$, and the sketching operator \eqref{sec2:equ2} fully preserves the information in $B_t$. In this case, the dependency of the constant factor on $\tau$ vanishes, since, regardless of the number of sketching steps, the sketching solver again reduces to the exact Newton solver. 
We refer to Remark \ref{rem:2} for the discussion of the relationship between computational cost and the choice of sketching parameters $(\tau,\gamma_S)$.

When we suppress the sketching solver, the limiting covariance is $\tXi = \tOmega/2$ for $\beta\in(0.5,1)$, meaning that $2\hat{\Xi}_t$ is a consistent estimator of $\tOmega$. Notably, this result suggests that we can estimate the optimal covariance matrix $\tOmega$ without grouping the iterates, computing the batch means, and tuning batch size sequences, which significantly differs and simplifies the estimation procedure in first-order {methods}. This advantage is indeed achieved by leveraging Hessian estimates; however, we preserve the computation and memory costs as low as those of first-order methods. We defer a comprehensive discussion of Theorem \ref{sec4:thm1} to Section \ref{sec:4.3}.

Theorem \ref{sec4:thm1} immediately implies the following corollary, which demonstrates the construction of confidence intervals/regions.

\begin{corollary}\label{sec4:cor1}

Let us set the coverage probability as $1-q$ with $q\in (0,1)$. Consider performing the online scheme \eqref{nequ:7} and computing the covariance estimator \eqref{exp:Xihat}. Suppose Assumptions \ref{ass:1} -- \ref{ass:4} hold, the number of sketches satisfies $\tau\geq \log(\gamma_H/4\Upsilon_H)/\log \rho$ with $\rho = 1-\gamma_S$, and the stepsize parameters satisfy $\beta\in(0.5,1)$, $\chi> 1.5\beta$, and $c_{\beta},c_{\chi}>0$. Then, we have 
\begin{equation*}
P\rbr{\tx \in \E_{t,q}} \rightarrow 1 - q\quad\quad \text{ as }\quad t\rightarrow\infty,
\end{equation*}
where $\E_{t,q} = \{\bx\in\mR^d: (\bx-\bx_t)^T\hat{\Xi}_t^{-1}(\bx-\bx_t)/\baralpha_t \leq \chi_{d,1-q}^2\}$. Furthermore, for any direction $\bw\in\mR^d$,
\begin{equation*}
P\rbr{\bw^T\tx \in \sbr{\bw^T\bx_t \pm z_{1-q/2}\sqrt{\baralpha_t\cdot \bw^T\hat{\Xi}_t\bw}}} \rightarrow 1-q \quad\quad \text{ as }\quad t\rightarrow\infty.
\end{equation*}
Here, $\chi_{d,1-q}^2$ is the $(1-q)$-quantile of $\chi_d^2$ distribution, while $z_{1-q/2}$ is the $(1-q/2)$-quantile of standard Gaussian distribution.
\end{corollary}

We would like to emphasize that the above statistical inference procedure is fully online and matrix-free. In particular, $\bx_t$ is updated with online nature; for confidence intervals, $\bw^T\hat{\Xi}_t\bw$ is computed online as introduced in Section \ref{sec:4.1}; for confidence region, $\hat{\Xi}_t^{-1}$ can also be updated online:
\begin{equation*}
\hat{\Xi}_{t+1}^{-1} = \frac{t+1}{t}\hat{\Xi}_{t}^{-1} - \frac{t+1}{t}\hat{\Xi}_{t}^{-1} R_t\rbr{\Pi_t + R_t^T\hat{\Xi}_{t}^{-1}R_t}^{-1} R_t^T\hat{\Xi}_{t}^{-1},
\end{equation*}
where $R_t = \rbr{\bv_t - a_t\barx_t; \barx_t - \barx_{t+1}; \bx_{t+1} - \barx_{t+1}} \in \mR^{d\times 3}$ and $\Pi_t = (a_t, 1, 0; 1, 0, 0;0,0, t\varphi_t) \in\mR^{3\times 3}$. See Appendix \ref{appendix:1} for the derivation of the above recursion.

\subsection{Comparison and generalization of existing studies}\label{sec:4.3}

In this section, we compare our weighted sample covariance estimator $\hat{\Xi}_t$ with other existing covariance estimators for both first- and second-order online methods. A summary of theoretical convergence results of all related estimators has been provided in Table \ref{tab:1}. We also discuss the generalization of our estimator to other methods.

\vskip4pt
\noindent$\bullet$ \textbf{Plug-in estimator of online sketched Newton.}
Recall from Remark \ref{rem:1} that, due to the challenges of estimating sketch-related quantities $\tilde{C}^\star$ and $C^\star$ in \eqref{equ:lyap}, a recent work \cite{Na2025Statistical} simply neglected these quantities and designed a plug-in covariance estimator $\tilde{\Xi}_t$ in \eqref{exp:PI}. In addition to concerns about excessive computation, \cite[Theorem 5.10]{Na2025Statistical} indicated that for $\beta\in(0.5,1)$,
\begin{equation}\label{equ:1}
\|\tilde{\Xi}_t - \tXi\| = O(\sqrt{\beta_t \log(1/\beta_t)})+O((1-\gamma_S)^\tau).
\end{equation}
Here, the second term accounts for the oversight in estimating sketch-related quantities. It decays exponentially with the sketching steps $\tau$ but \textit{does not vanish} for any finite $\tau$. Thus, $\tilde{\Xi}_t$ is not a consistent estimator of $\tXi$. In the degenerate case where the Newton systems are solved exactly ($\tau = \infty$), $\tilde{\Xi}_t$ converges to $\tXi$ at a rate of $O(\sqrt{\beta_t \log(1/\beta_t)})$, which is faster than the $O(1/\sqrt{t\beta_t})$ rate achieved by our estimator $\hat{\Xi}_t$ (since $\beta>0.5\Rightarrow\beta_t=o(1/\sqrt{t})$). In this case, choosing between $\hat{\Xi}_t$ and $\tilde{\Xi}_t$ involves a trade-off between faster convergence and computational efficiency: the plug-in estimator converges faster but requires computing the inverse of the Hessian, leading to an $O(d^3)$ per-step computational cost; in contrast, the weighted sample covariance converges slower but uses only the iterates themselves for the update, leading to an $O(d^2)$ per-step computational cost. It is worth noting that the faster convergence of the plug-in estimator is anticipated (see \cite{Chen2020Statistical} for a comparison of plug-in and batch-means estimators in SGD methods), since its convergence rate is fully tied to that of the iterates. As a comparison, the convergence rate of our sample covariance is additionally confined by the slow decay of correlations among the iterates.

\vskip4pt
\noindent$\bullet$ \textbf{Batch-means estimator of SGD.} 
As introduced in Sections \ref{sec:1} and \ref{sec:4.1}, \citet{Chen2020Statistical, Zhu2021Online} grouped SGD iterates into batches and estimated the limiting covariance $\tOmega$ by the sample covariance among batches (each batch mean is treated as one sample). \citet{Singh2023Utility} further relaxed their conditions from increasing batch sizes to equal batch sizes. Compared to this type of estimators, our estimator $\hat{\Xi}_t$ is batch-free, requiring no additional parameters beyond those of the algorithm itself.
Furthermore, the aforementioned works all showed that the convergence rate of the batch-means estimators is $O(1/\sqrt[4]{t\beta_t})$, which is slower than that of $\hat{\Xi}_t$ in Theorem \ref{sec4:thm1}. Intuitively, the batch-means estimators require a long batch of iterates to obtain a single sample, while our batch-free estimator treats each individual iterate as a single sample, making it more efficient in utilizing (correlated) iterates. 
That being said, we should mention that our estimator $\hat{\Xi}_t$ targets estimating the limiting covariance of the last Newton iterate, which differs from the aforementioned works that target estimating the limiting covariance of the averaged SGD iterate.$\quad\quad$

\vskip4pt
\noindent$\bullet$ \textbf{Generalization to conditioned SGD.} 
We point out that $\hat{\Xi}_t$ can also serve as a consistent covariance estimator for conditioned SGD methods, which follow the update form \eqref{nequ:3} though $B_t$ may not approximate the objective Hessian $\nabla^2F_t$. \citet{Leluc2023Asymptotic} established asymptotic normality for conditioned SGD methods under the assumption of convergence of the conditioning matrix $B_t$. These methods include AdaGrad \citep{Duchi2011Adaptive}, RMSProp \citep{Tieleman2012Lecture}, and quasi-Newton methods \citep{Byrd2016Stochastic} as special cases.  
Notably, Theorem \ref{sec4:thm1} does not require $B_t$ to converge to the true Hessian $\nabla^2 F(\tx)$, making our analysis directly applicable to conditioned SGD methods.

\vskip4pt
\noindent$\bullet$ \textbf{Generalization to sketched Sequential Quadratic Programming.} We consider a constrained stochastic optimization problem:
\begin{equation*}
\min_{\bx\in\mR^d}\; F(\bx) = \mE_\P[f(\bx;\xi)] \quad\;\;\text{s.t.}\quad c(\bx)=\0,
\end{equation*}
where $F:\mR^d\rightarrow\mR$ is a stochastic objective with $f(\cdot;\xi)$ as the noisy observation, and $c:\mR^d\rightarrow \mR^m$ imposes deterministic constraints on the model parameters $\bx$. The constraints $c(\bx)$ are assumed to satisfy standard regularity conditions, including smoothness and linear independence constraint qualification assumptions. We refer the reader to \cite[Assumption 4.1]{Na2025Statistical} and \cite[Example 1]{Davis2024Asymptotic} for details. Such problems appear widely in statistical machine learning, including constrained $M$-estimation and algorithmic fairness. \citet{Na2025Statistical} designed an online sketched Sequential Quadratic Programming (SQP) method for solving the problem. Define $\mL(\bx, \blambda) = F(\bx) + \blambda^Tc(\bx)$ as the Lagrangian function, where $\blambda \in \mR^m$ are the dual variables. The sketched SQP method can be regarded as applying the sketched Newton method to $\mL(\bx, \blambda)$, leading to the update $(\bx_{t+1}, \blambda_{t+1}) = (\bx_t, \blambda_t)+ \baralpha_t(\barDelta \bx_t, \barDelta\blambda_t)$, where $(\barDelta \bx_t, \barDelta \blambda_t)$ is the sketched solution to the primal-dual Newton system:
\begin{equation*}
\begin{pmatrix}
B_t & G_t^T\\
G_t & \0
\end{pmatrix}\begin{pmatrix}
\Delta \bx_t\\
\Delta\blambda_t
\end{pmatrix} = -\begin{pmatrix}
\bar{\nabla}_\bx\mL_t\\
c_t
\end{pmatrix}.
\end{equation*}
Here, analogous to \eqref{nequ:3}, $B_t\approx \nabla_{\bx}^2\mL_t$ is an estimate of the Lagrangian Hessian with respect to $\bx$, $G_t = \nabla c_t\in\mR^{m\times d}$ is the constraints Jacobian, and $\bar{\nabla}_\bx\mL_t = \nabla F(\bx_t;\xi_t) + G_t^T\blambda_t$ is the estimate of the Lagrangian gradient with respect to $\bx$. \citet{Na2025Statistical} established asymptotic normality for the SQP iterate $(\bx_t, \blambda_t)$. We observe that the constraints are not essential in the SQP analysis; therefore, our construction of $\hat{\Xi}_t$ is naturally applied to the covariance estimation of the sketched SQP method. An empirical demonstration of $\hat{\Xi}_t$ for constrained problems is presented in Section \ref{sec5:subsec2}.

\section{Numerical Experiment}\label{sec:5}

In this section, we demonstrate the empirical performance of the weighted sample covariance matrix $\hat{\Xi}_t$ on both regression problems and benchmark CUTEst problems \citep{Gould2014CUTEst}. We compare $\hat{\Xi}_t$ with two other online covariance estimators: the plug-in estimator $\tilde{\Xi}_t$ in \eqref{exp:PI} (based on sketched Newton) and the batch-means estimator $\bar{\Xi}_t$ (based on SGD) \cite[Algorithm 2]{Zhu2021Online}. We evaluate the performance of each estimator by both the (relative) covariance estimation error and the coverage rate of constructed confidence intervals. 
Due to the large number of possible combinations of problem setups, we organize the experimental section as follows. In Sections \ref{sec5:linear} and \ref{sec5:logistic}, we study linear and logistic regression problems across different methods, varying the problem dimension $d$, the covariate design covariance $\Sigma_a$, and the number of sketching steps $\tau$. Throughout these experiments, we apply the randomized Kaczmarz method with $q=1$ (i.e., a single sketching vector) for the sketched Newton methods.
In Section \ref{sec:5.3}, we explore the impact of different sketching configurations on statistical inference, including the sketching distribution (Kaczmarz v.s. Gaussian), the sketching dimension $q$, and the number of sketching steps $\tau$.
In Section \ref{sec:5.4}, we explore the impact of Hessian preconditioning on the asymptotics of the estimates. In particular, we compare exact Newton with (averaged) SGD under different stepsizes, and show how Hessian preconditioning improves statistical efficiency. For simplicity, we use linear regression with different design covariances as an example in Sections \ref{sec:5.3}--\ref{sec:5.4}. 
In Section \ref{sec:5.5}, we examine how the numbers of sketching iterations affects the convergence rate of covariance estimation.
Finally, in Section \ref{sec5:subsec2}, we explore the proposed batch-free covariance estimator in constrained optimization problems.
Due to the space limit, we defer some experimental results to Appendix \ref{appen:exp}. Our code is available at \url{https://github.com/weikuang97/SketchedNT-Inf}.

\subsection{Linear regression}\label{sec5:linear}

We consider the linear regression model $\xi_{b} = \bxi_{a}^T\bx^{\star} + \varepsilon$, where $\xi = (\bxi_a, \xi_b)\in\mR^d\times \mR$ is the feature-response vector and $\varepsilon\sim\mN(0, \sigma^2)$ is the Gaussian noise. For this model, we use the squared loss defined as $f(\bx; \xi) = \frac{1}{2}(\xi_{b}-\bxi_{a}^T\bx)^2$. Similar to existing studies \citep{Chen2020Statistical, Zhu2021Online, Na2025Statistical}, we apply Gaussian features $\bxi_a \sim \mN(0, \Sigma_a)$ with different dimensions and covariance matrices $\Sigma_a$. In particular, we vary $d\in\{20, 40, 60, 100\}$, and for each $d$, we consider three types of covariance matrices. (i) Identity: $\Sigma_a = I$. (ii) Toeplitz: $[\Sigma_a]_{i,j} = r^{|i-j|}$ with $r\in\{0.4, 0.5, 0.6\}$. (iii) Equi-correlation: $[\Sigma_a]_{i,i}=1$ and $[\Sigma_a]_{i,j} = r$ for $i\neq j$, with $r\in\{0.1,0.2,0.3\}$. The true model parameter is set as $\bx^\star = (1/d,\dots,1/d)\in\mR^d$.

\begin{figure}[!t]
\centering     
\subfigure[SGD]{\label{A11}\includegraphics[width=0.32\textwidth]{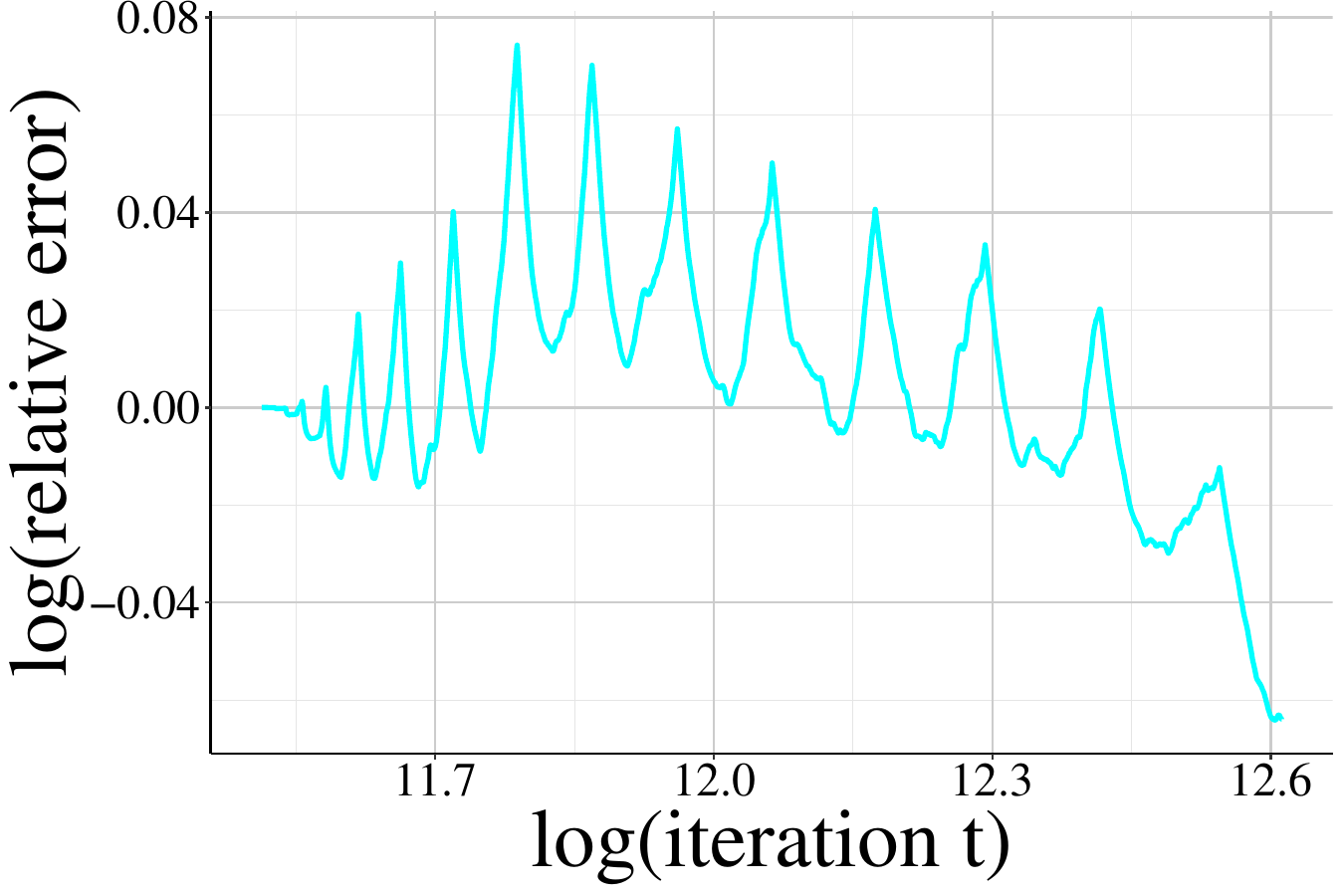}}
\subfigure[exact Newton ($\tau=\infty$)]{\label{A12}\includegraphics[width=0.32\textwidth]{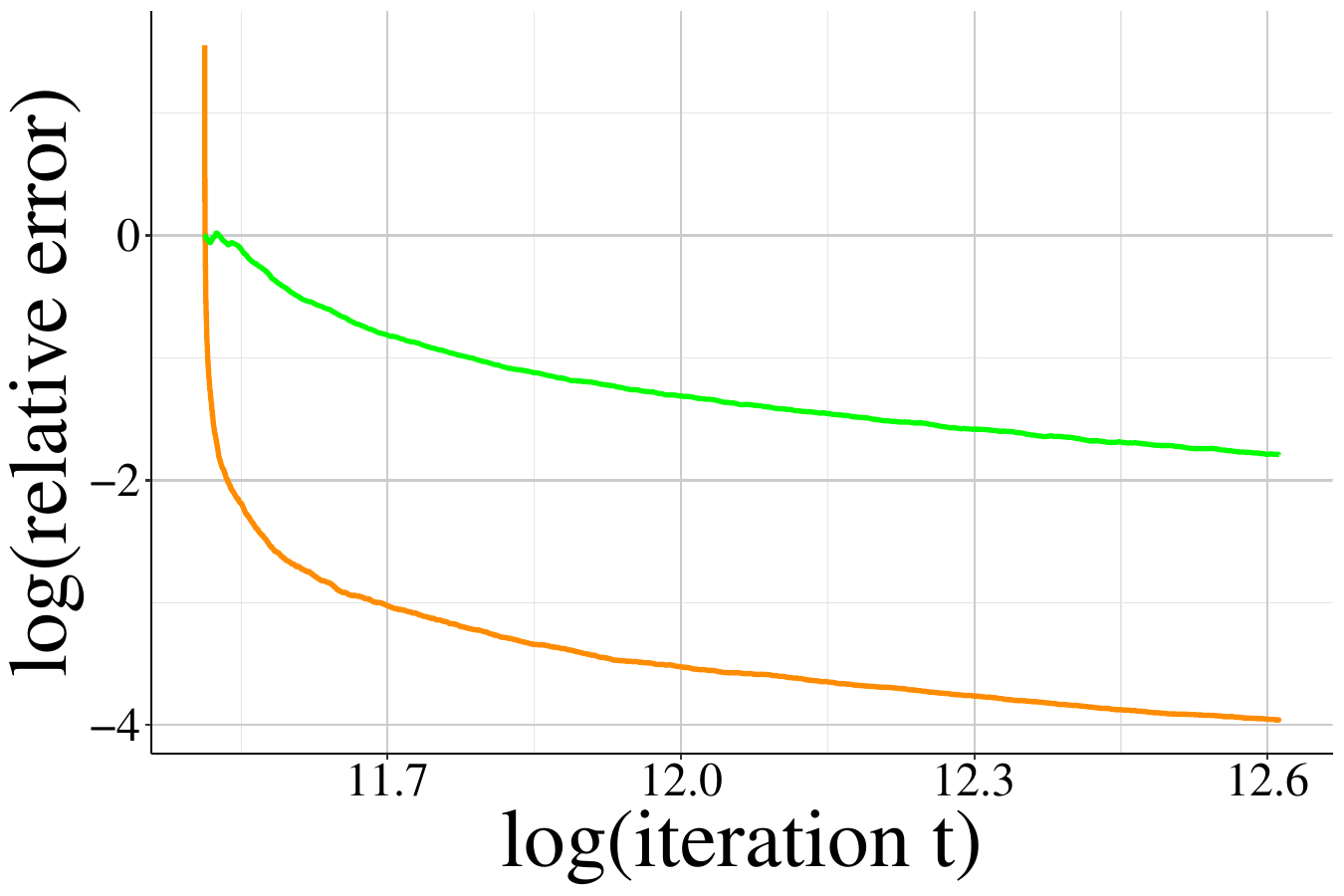}}
\subfigure[sketched Newton ($\tau=2$)]{\label{A13}\includegraphics[width=0.32\textwidth]{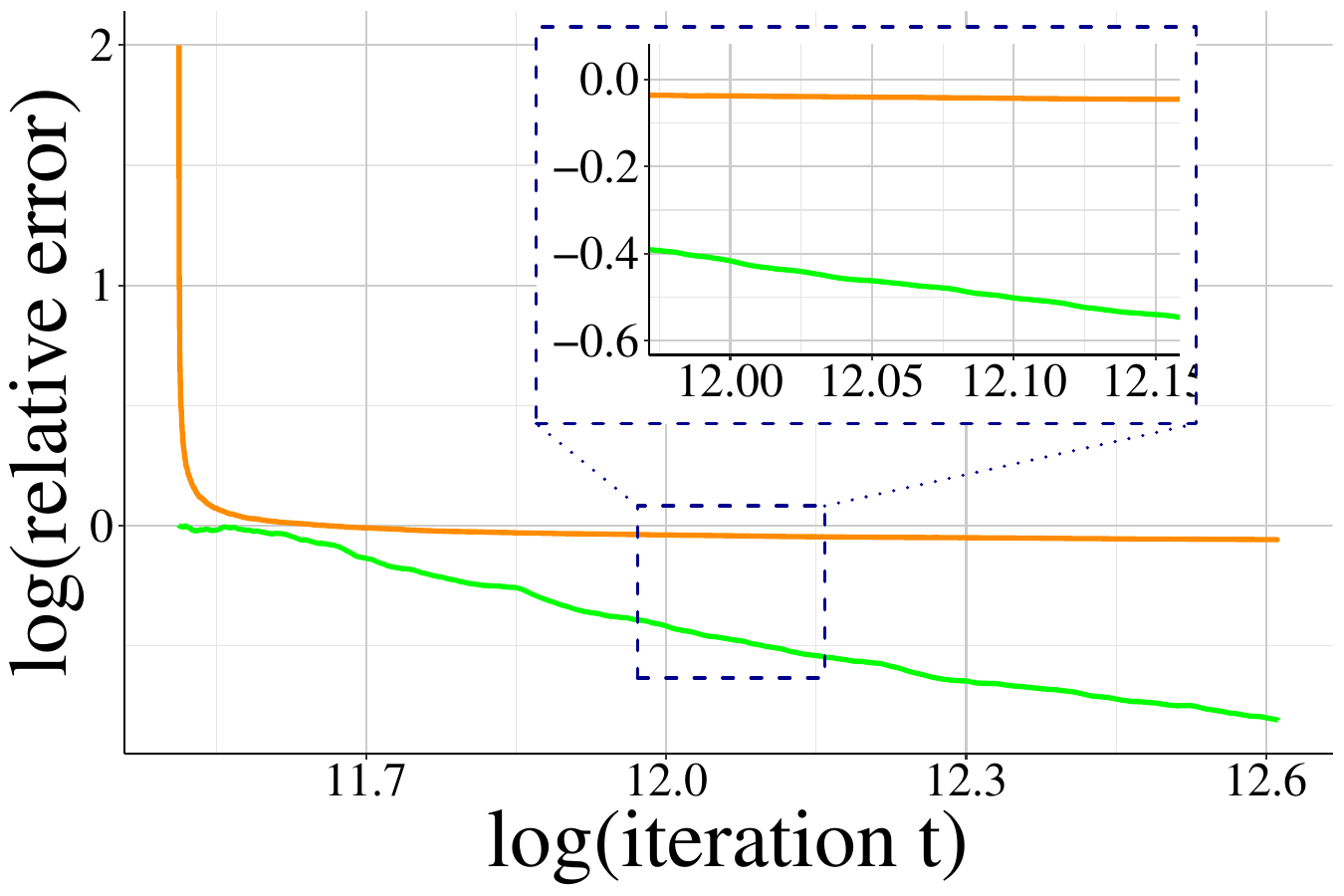}}
\vskip5pt
\centering{Relative covariance estimation error}

\subfigure[SGD]{\label{A21}\includegraphics[width=0.32\textwidth]{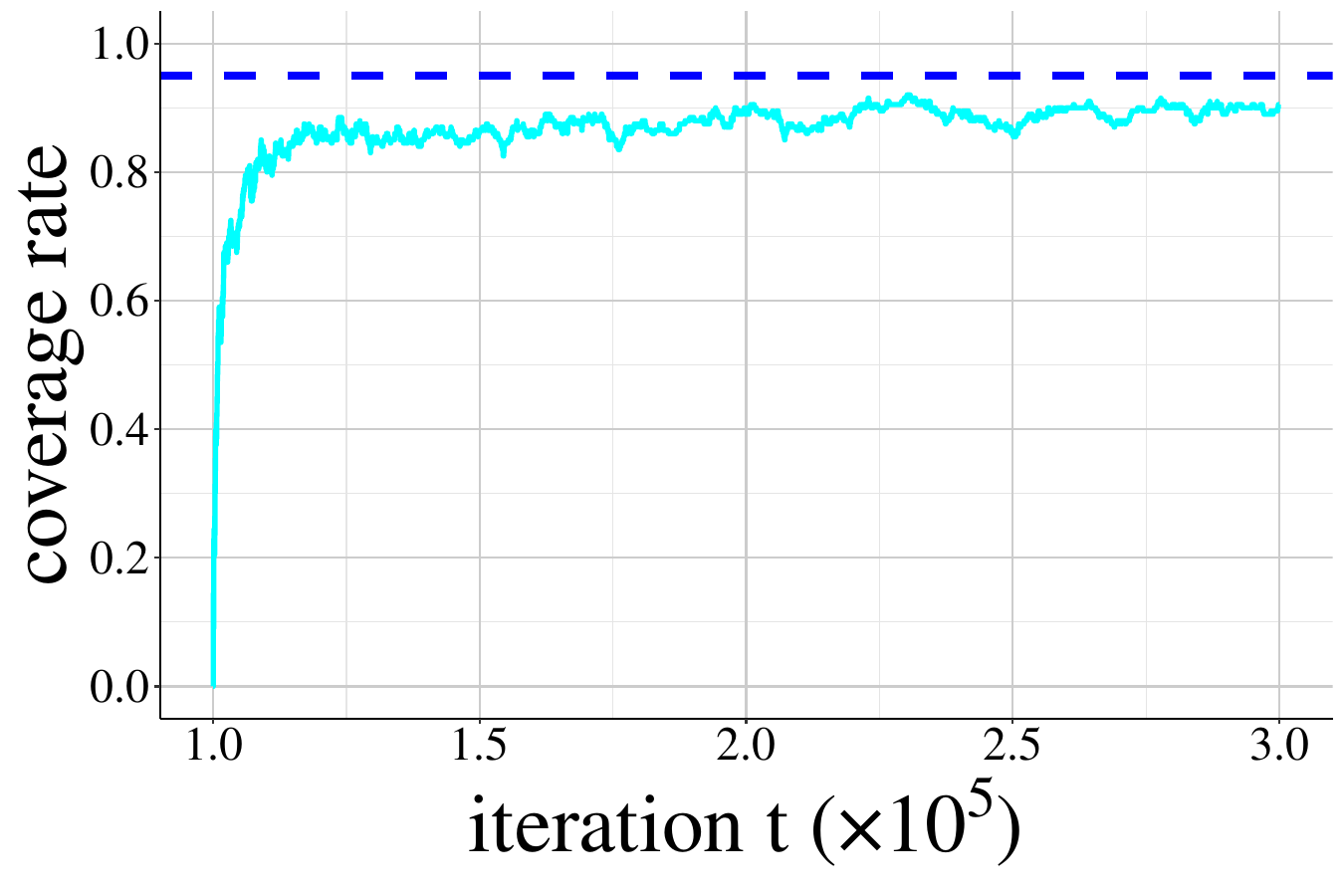}}
\subfigure[exact Newton ($\tau=\infty$)]{\label{A22}\includegraphics[width=0.32\textwidth]{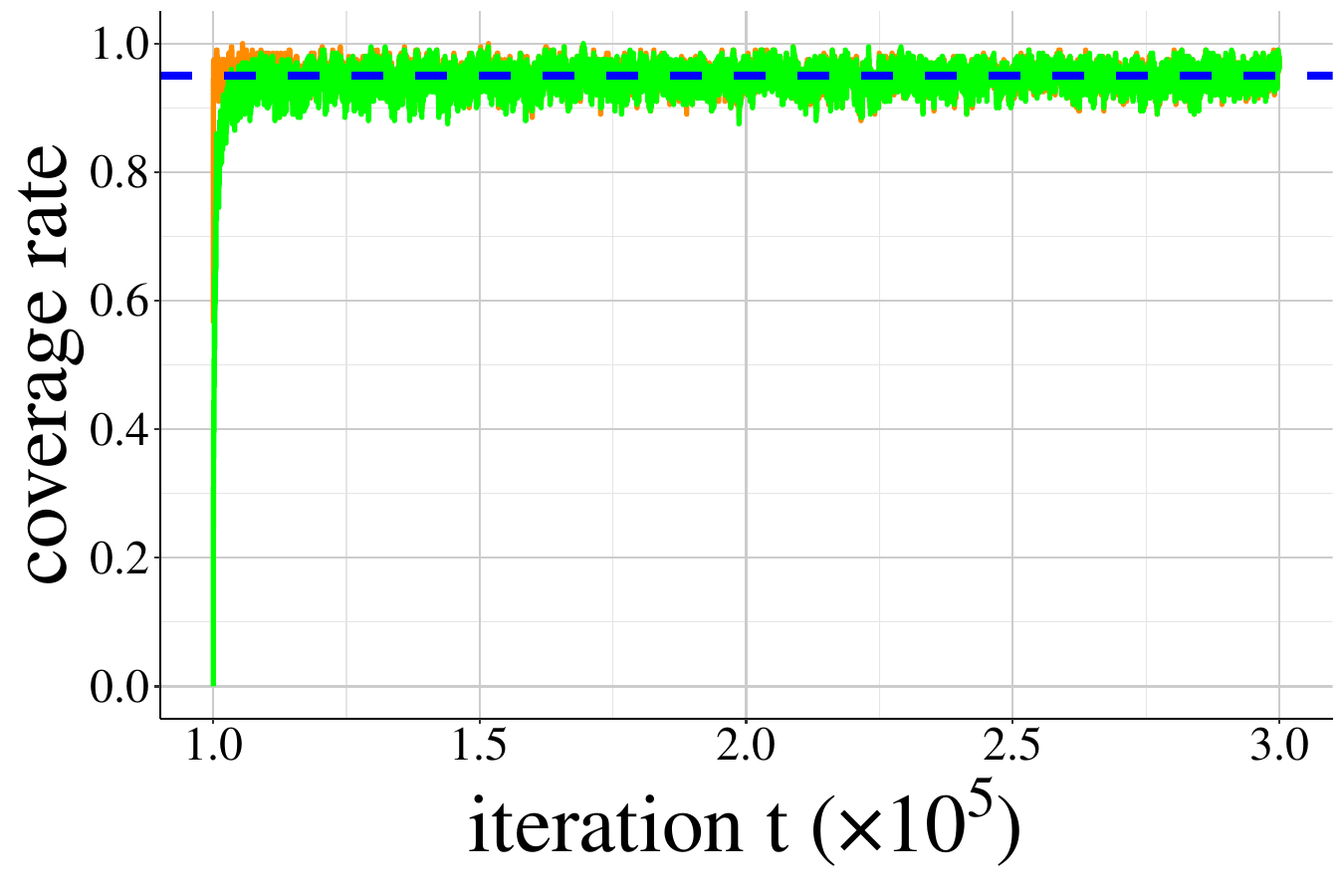}}
\subfigure[sketched Newton ($\tau=2$)]{\label{A23}\includegraphics[width=0.32\textwidth]{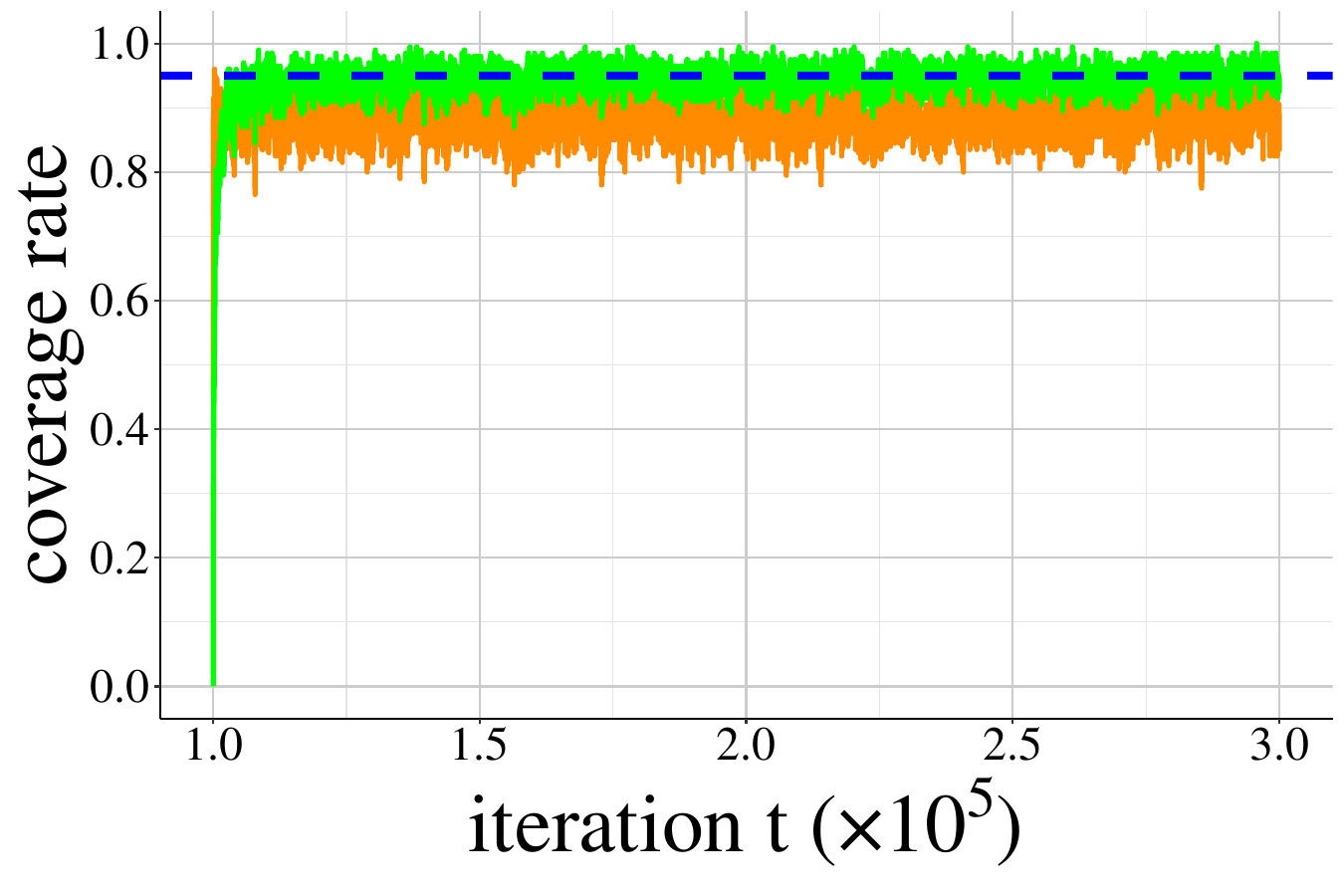}}
\vskip5pt
\centering{Empirical coverage rate of 95\% confidence intervals}	

\subfigure[SGD]{\label{A31}\includegraphics[width=0.32\textwidth]{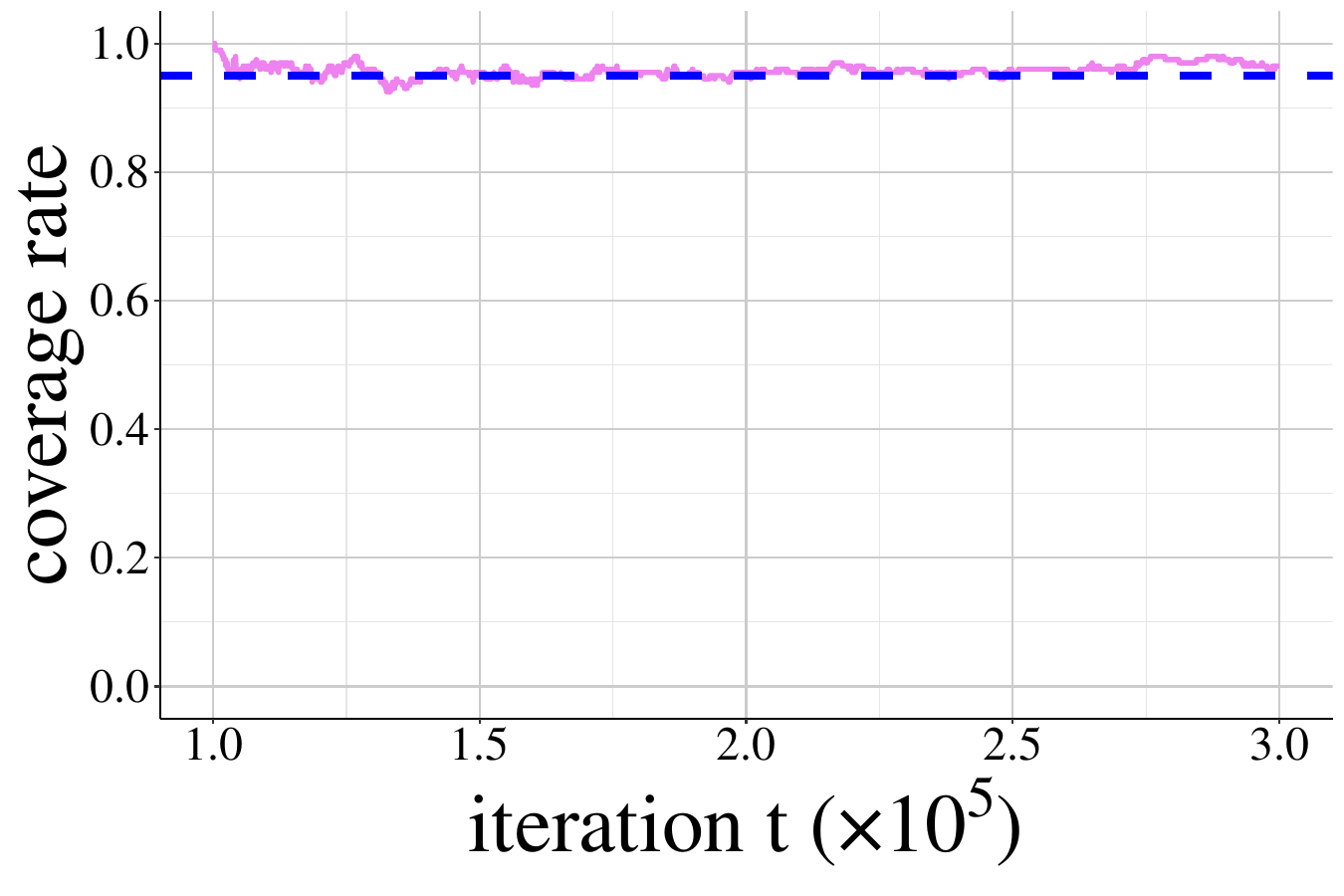}}
\subfigure[exact Newton ($\tau=\infty$)]{\label{A32}\includegraphics[width=0.32\textwidth]{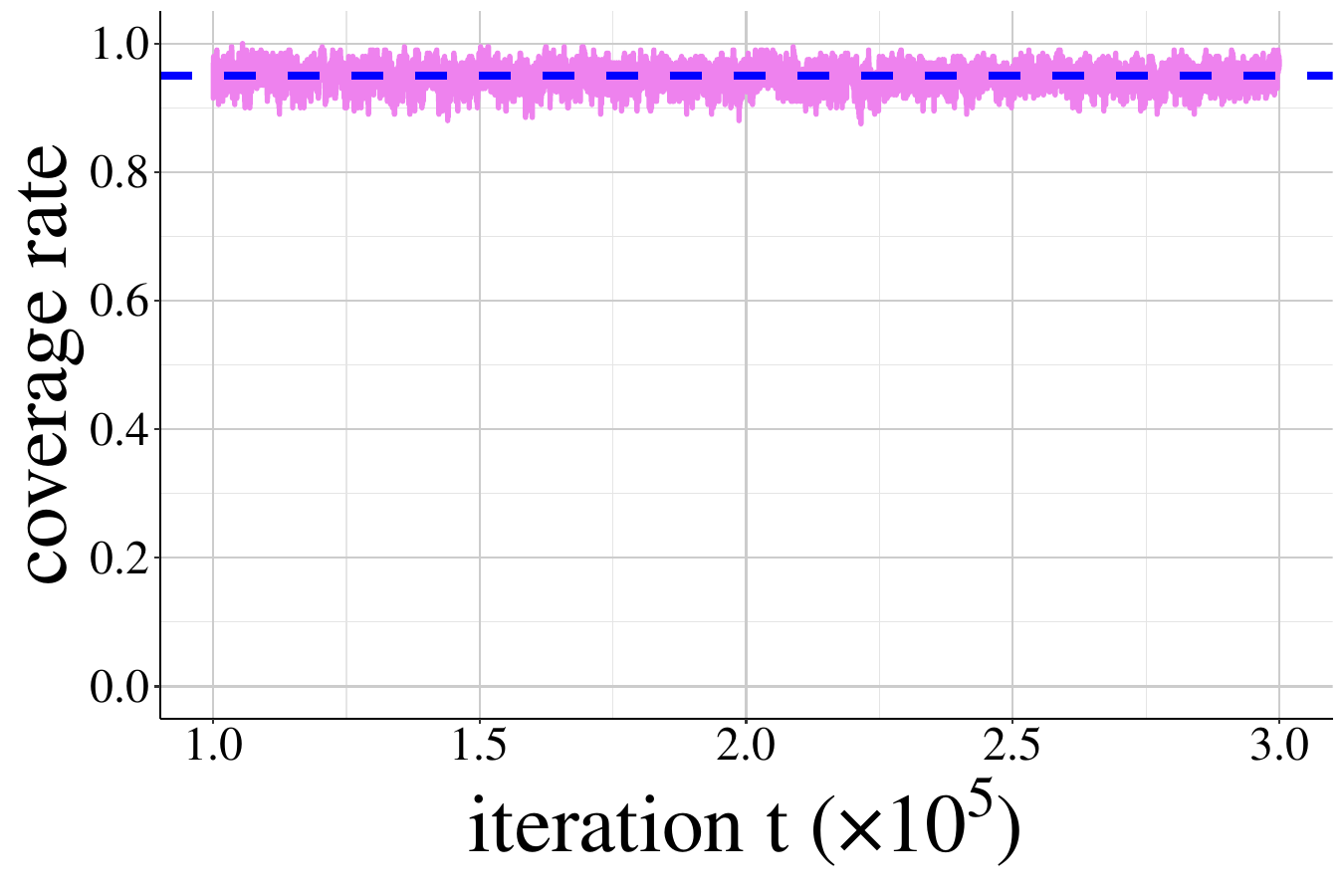}}
\subfigure[sketched Newton ($\tau=2$)]{\label{A33}\includegraphics[width=0.32\textwidth]{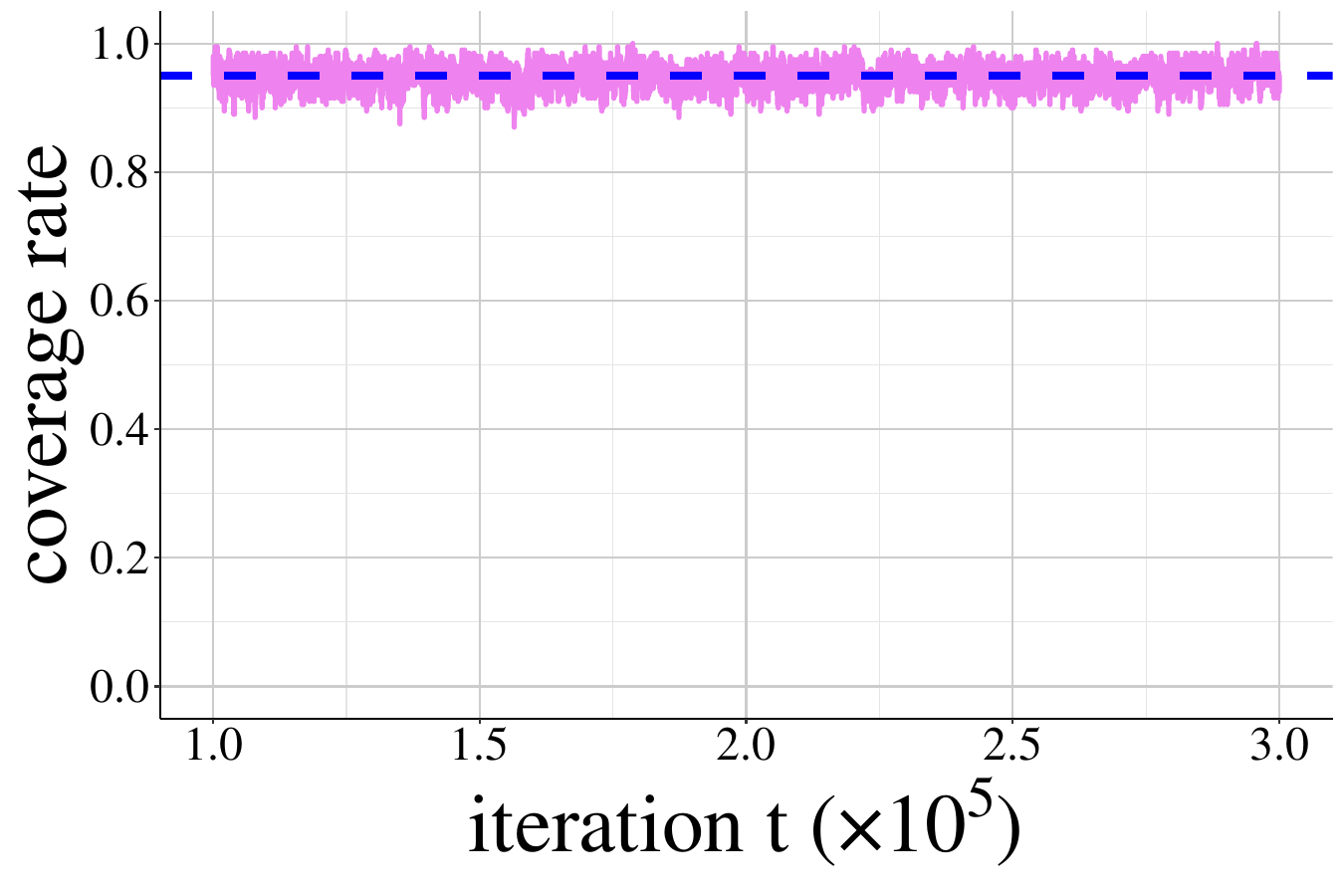}}
\vskip5pt
\centering{Empirical coverage rate of 95\% oracle confidence intervals}	
\vskip5pt
\includegraphics[width=0.95\textwidth]{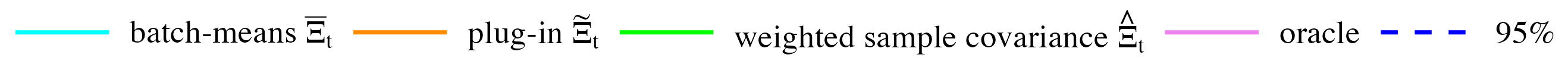}
\caption{\textit{
The averaged trajectories for linear regression problems with $d=5$ and Equi-correlation $\Sigma_a \;(r=0.3)$. From left to right, the columns correspond to SGD, exact Newton method, and sketched Newton method $(\tau = 2)$. For averaged SGD, the limiting covariance $\Xi^\star$ is estimated using the batch-means estimator $\bar{\Xi}_t$. For exact and sketched Newton methods, $\Xi^\star$ is estimated using both the plug-in estimator $\tilde{\Xi}_t$ and the proposed sample covariance $\hat{\Xi}_t$. 
The first row shows the log relative covariance estimation error $(\textit{e.g.,}\; \log(\|\hat{\Xi}_t-\Xi^\star\|/\|\Xi^\star\|))$ v.s $\log t$. The second row shows the coverage rate of the 95\% confidence intervals for $\sum_{i=1}^d\bx^{\star}_i/d$ constructed using corresponding estimators of $\Xi^\star$. The third row shows the coverage rate of the oracle 95\% confidence intervals, where the oracle confidence intervals are constructed using the true covariance $\Xi^\star$. The figures demonstrate the consistency of $\hat{\Xi}_t$ and its superior performance in statistical inference.}}\label{fig:1}	
\end{figure}

For the batch-means estimator $\bar{\Xi}_t$, we adopt the setup in \cite{Zhu2021Online} by setting the stepsize of SGD as $\beta_t = 0.5t ^{-\beta}$ and the batch size as $a_m  =  \lfloor m^{2/(1-\beta)}\rfloor$ (in their notation) with $\beta = 0.505$. For both plug-in estimator $\tilde{\Xi}_t$ and our sample covariance estimator $\hat{\Xi}_t$, we implement sketched Newton methods with varying sketching steps $\tau\in\{10,20,40,\infty\}$. When $\tau=\infty$, the scheme reduces to the standard Newton method. We apply the Kaczmarz method, where the sketching distribution in \eqref{sec2:equ2} is $S\sim \text{Unif}(\{\be_i\}_{i=1}^d)$ (cf. Section \ref{sec:3.1}). We set $\beta_t=t^{-\beta}$ and $\baralpha_t\sim\text{Unif}[\beta_t,\beta_t+\beta_t^2]$. For all estimators, we initialize the method at $\bx_0=\boldsymbol{0}$, run $3\times 10^5$ iterations, and aim to construct 95\% confidence intervals for the averaged parameters $\sum_{i=1}^d\bx^{\star}_i/d$. All the results are averaged over 200 independent runs.\;

We present the averaged trajectories of relative covariance estimation error and the empirical coverage rate of confidence intervals using three covariance estimators in Figure \ref{fig:1}.
From Figures \ref{A12} and \ref{A13}, we observe that $\hat{\Xi}_t$ is a consistent estimator for both exact and sketched Newton methods. Additionally, the tails of the green lines (corresponding to $\hat{\Xi}_t$) in both figures form nearly straight lines, with absolute slope values greater than $(1-0.505)/2$. This behavior aligns with the theoretical upper bound established in Theorem \ref{sec4:thm1}. Although $\tilde{\Xi}_t$ converges faster than $\hat{\Xi}_t$, it is consistent only for exact Newton methods. For sketched Newton method, the estimation error of $\tilde{\Xi}_t$ quickly stabilizes at a positive constant due to the bias introduced by ignoring the sketching effect (cf. \eqref{equ:1}).
From Figure \ref{A11}, we see that $\bar{\Xi}_t$ converges more slowly than $\hat{\Xi}_t$, which is also consistent with the theoretical results in \cite{Zhu2021Online}. The estimation error of $\bar{\Xi}_t$ exhibits oscillations along with the batching process. This occurs because the limited sample size in each newly created batch introduces additional errors. This phenomenon is undesirable, as increasing the sample size does not always lead to a reduction in estimation error. Our batch-free estimator effectively resolves this issue.

In terms of statistical inference, Figures \ref{A31}--\ref{A33} show that the coverage rates of all oracle confidence intervals, constructed using the true limiting covariance $\Xi^\star$ under different iterative algorithms, remain close to the target confidence level of 95\%. This reconfirms the established asymptotic normality of these algorithms and highlights the importance of accurately estimating $\Xi^\star$ for constructing valid confidence intervals. 
From Figure \ref{A23}, confidence intervals based on $\hat{\Xi}_t$ achieve a coverage rate close to 95\%, while those based on $\tilde{\Xi}_t$ exhibit undercoverage due to bias. In Figure \ref{A22}, the coverage rate trajectories of $\hat{\Xi}_t$ and $\tilde{\Xi}_t$ nearly overlap, indicating that although $\hat{\Xi}_t$ converges slower than $\tilde{\Xi}_t$ in exact Newton method, its accuracy is sufficient for constructing reliable confidence intervals. However, updating $\tilde{\Xi}_t$ is significantly more computationally expensive due to the inverse of $B_t$. Regarding $\bar{\Xi}_t$, Figure \ref{A21} shows that its confidence intervals exhibit undercoverage due to slow convergence. 
Overall, across all figures, we observe that the consistency and fast convergence of $\hat{\Xi}_t$ make it a reliable and computationally efficient choice for constructing confidence intervals.

To comprehensively evaluate the performance of the three estimators in statistical inference, we present part of results under various settings in Table \ref{table:1}, while a complete table is provided in Appendix \ref{appen:exp}. The table reports the empirical coverage rate of the confidence intervals and the averaged relative estimation error in the variance of $\b1^T\tx/d = \sum_{i=1}^d \bx^\star_i/d$, expressed as $\b1^T(\hat{\Xi}_t - \tXi)\b1/\b1^T\Xi^\star\b1$, at the last iteration.
From the table, we observe that, overall, the coverage rate of the confidence intervals based on $\hat{\Xi}_t$ remains around 95\% in most cases. In contrast, the coverage rates for $\tilde{\Xi}_t$ in sketched Newton methods and $\bar{\Xi}_t$ in SGD tend to exhibit undercoverage, as previously explained for Figure \ref{fig:1}. It is important to note that the sketching distribution used in our experiments does not introduce bias when $\Sigma_a = I$. When $\tilde{\Xi}_t$ is a consistent estimator (i.e., when $\Sigma_a = I$ or $\tau = \infty$), $\hat{\Xi}_t$ performs competitively compared to $\tilde{\Xi}_t$. However, in other cases, the relative variance estimation error of $\tilde{\Xi}_t$ is significantly larger than that of $\hat{\Xi}_t$ due to bias, leading to differences in statistical inference performance. The influence of the dimension $d$ and the sketching iteration number $\tau$ is more pronounced for the Equi-correlation $\Sigma_a$. 
For instance, when $\tau = 10$, we observe that the coverage rate of $\tilde{\Xi}_t$ decreases as $d$ increases, indicating that higher dimensionality makes the problem more challenging. Conversely, when fixing $d = 100$, the coverage rate for $\tilde{\Xi}_t$ gradually increases as $\tau$ increases from $10$ to $40$. This occurs because increasing $\tau$ reduces the approximation error of the Newton direction, thereby reducing the bias introduced by sketching techniques.
The results in Table \ref{table:1} further demonstrate the superior performance of our proposed estimator $\hat{\Xi}_t$.

\begin{table}[!t]
\centering
\resizebox{\linewidth}{!}{
\begin{tabular}{|c|c|c|c|cccccccc|}	
\hline
\multirow{3}{*}{$\Sigma_a$} & \multirow{3}{*}{d} & \multirow{3}{*}{Criterion} & \multirow{2}{*}{SGD} & \multicolumn{8}{c|}{Sketched Newton Method} \\ \cline{5-12} & & & &
\multicolumn{2}{c|}{$\tau=\infty$} & \multicolumn{2}{c|}{$\tau=10$} & \multicolumn{2}{c|}{$\tau=20$} & \multicolumn{2}{c|}{$\tau=40$} \\ \cline{4-12} & & &
{\footnotesize$\bar{\Xi}_t$} & {\footnotesize$\tilde{\Xi}_t$} & \multicolumn{1}{c|}{{\footnotesize$\hat{\Xi}_t$}} & {\footnotesize$\tilde{\Xi}_t$} & \multicolumn{1}{c|}{{\footnotesize$\hat{\Xi}_t$}} & {\footnotesize$\tilde{\Xi}_t$} & \multicolumn{1}{c|}{{\footnotesize$\hat{\Xi}_t$}} & {\footnotesize$\tilde{\Xi}_t$} & {\footnotesize$\hat{\Xi}_t$} \\ \hline
\multirow{8}{*}{Identity} & \multirow{2}{*}{20} & Cov (\%) &
89.50 & 94.00 & \multicolumn{1}{c|}{94.00} & 93.00 &
\multicolumn{1}{c|}{93.00} & 95.00 & \multicolumn{1}{c|}{95.50} & 97.00 & 96.00 \\
& & Var Err & -0.178 & 0.024 & \multicolumn{1}{c|}{0.008} & 0.025 & \multicolumn{1}{c|}{0.026} & 0.025 & \multicolumn{1}{c|}{0.028} & 0.025 & 0.021 \\ \cline{2-12} 
& \multirow{2}{*}{40} & Cov (\%) & 88.00 &  94.00 &
\multicolumn{1}{c|}{93.50} &  96.00 &  \multicolumn{1}{c|}{97.50} &  96.00 &  \multicolumn{1}{c|}{96.00} &  95.00 &  95.00 \\
& &  Var Err &  -0.145 &  0.049 &  \multicolumn{1}{c|}{0.048} &  0.048 &  \multicolumn{1}{c|}{0.035} &  0.049 &  \multicolumn{1}{c|}{0.036} &  0.049 &  0.044 \\ \cline{2-12} 
&  \multirow{2}{*}{60} &  Cov (\%) &  \textbf{85.50} &  91.50 &  \multicolumn{1}{c|}{91.00} &  \textbf{94.50} &  \multicolumn{1}{c|}{\textbf{94.00}} &  94.00 &  \multicolumn{1}{c|}{94.50} &  94.50 &  94.00 \\
&   &  Var Err &  \textbf{-0.174} &  0.072 &  \multicolumn{1}{c|}{0.070} &  \textbf{0.074} &  \multicolumn{1}{c|}{\textbf{0.035}} &  0.073 &  \multicolumn{1}{c|}{0.044} &  0.073 &  0.058 \\ \cline{2-12} 
&  \multirow{2}{*}{100} &  Cov (\%) &  88.00 &  100.0 &  \multicolumn{1}{c|}{100.0} &  95.50 &  \multicolumn{1}{c|}{95.50} &  94.00 &  \multicolumn{1}{c|}{93.00} &  95.00 &  95.50 \\
&   &  Var Err &  -0.185 &  $\infty$ &  \multicolumn{1}{c|}{$\infty$} &  0.129 &  \multicolumn{1}{c|}{0.096} &  0.128 &  \multicolumn{1}{c|}{0.076} &  0.126 &  0.109 \\ \hline
\multirow{8}{*}{\shortstack{Toeplitz\\$r=0.5$}} &  \multirow{2}{*}{20} &  Cov (\%) &  \textbf{87.00} &  94.50 &  \multicolumn{1}{c|}{94.50} &  \textbf{89.00} &  \multicolumn{1}{c|}{\textbf{96.50}} &  89.00 &  \multicolumn{1}{c|}{94.00} &  90.00 &  93.00 \\
&   &  Var Err &  \textbf{-0.104} &  0.025 &  \multicolumn{1}{c|}{0.026} &  \textbf{-0.339} &  \multicolumn{1}{c|}{\textbf{0.003}} &  -0.283 &  \multicolumn{1}{c|}{0.009} &  -0.208 &  0.018 \\ \cline{2-12} 
&  \multirow{2}{*}{40} &  Cov (\%) &  91.00 &  96.50 &  \multicolumn{1}{c|}{96.50} &  89.50 &  \multicolumn{1}{c|}{94.00} &  85.50 &  \multicolumn{1}{c|}{95.50} &  89.00 &  94.50 \\
&   &  Var Err &  -0.074 &  0.048 &  \multicolumn{1}{c|}{0.040} &  -0.376 &  \multicolumn{1}{c|}{0.016} &  -0.343 &  \multicolumn{1}{c|}{0.022} &  -0.285 &  0.029 \\ \cline{2-12} 
&  \multirow{2}{*}{60} &  Cov (\%) &  86.50 &  94.00 &  \multicolumn{1}{c|}{94.50} &  83.50 &  \multicolumn{1}{c|}{92.50} &  85.50 &  \multicolumn{1}{c|}{93.00} &  84.50 &  94.00 \\
&   &  Var Err &  -0.061 &  0.072 &  \multicolumn{1}{c|}{0.074} &  -0.383 &  \multicolumn{1}{c|}{0.044} &  -0.361 &  \multicolumn{1}{c|}{0.029} &  -0.317 &  0.046 \\ \cline{2-12} 
&  \multirow{2}{*}{100} &  Cov (\%) &  93.50 &  100.0 &  \multicolumn{1}{c|}{100.0} &  90.00 &  \multicolumn{1}{c|}{96.00} &  89.00 &  \multicolumn{1}{c|}{95.00} &  89.50 &  97.00 \\
&   &  Var Err &  -0.083 &  $\infty$ &  \multicolumn{1}{c|}{$\infty$} &  1.156 &  \multicolumn{1}{c|}{2.659} &  -0.069 &  \multicolumn{1}{c|}{0.582} &  -0.335 &  0.067 \\ \hline
\multirow{8}{*}{\shortstack{Equi-corr\\$r=0.2$}} &  \multirow{2}{*}{20} &  Cov (\%) &  92.00 &  93.00 &  \multicolumn{1}{c|}{92.50} &  \textbf{79.00} &  \multicolumn{1}{c|}{\textbf{94.00}} &  83.00 &  \multicolumn{1}{c|}{94.00} &  91.50 &  95.50 \\
&   &  Var Err &  -0.063 &  0.024 &  \multicolumn{1}{c|}{0.023} &  \textbf{-0.538} &  \multicolumn{1}{c|}{\textbf{0.013}} &  -0.468 &  \multicolumn{1}{c|}{0.016} &  -0.334 &  0.012 \\ \cline{2-12} 
&  \multirow{2}{*}{40} &  Cov (\%) &  \textbf{90.50} &  95.50 &  \multicolumn{1}{c|}{94.50} &  \textbf{75.00} &  \multicolumn{1}{c|}{\textbf{96.50}} &  82.50 &  \multicolumn{1}{c|}{96.50} &  80.50 &  94.50 \\
&   &  Var Err &  \textbf{-0.139} &  0.048 &  \multicolumn{1}{c|}{0.040} & \textbf{ -0.654} &  \multicolumn{1}{c|}{\textbf{0.022}} &  -0.630 &  \multicolumn{1}{c|}{0.018} &  -0.580 &  0.024 \\ \cline{2-12} 
&  \multirow{2}{*}{60} &  Cov (\%) &  91.00 &  95.50 &  \multicolumn{1}{c|}{95.50} &  \textbf{72.00} &  \multicolumn{1}{c|}{\textbf{91.50}} &  68.00 &  \multicolumn{1}{c|}{94.50} &  81.50 &  96.50 \\
&   &  Var Err &  -0.015 &  0.072 &  \multicolumn{1}{c|}{0.067} &  \textbf{-0.697} &  \multicolumn{1}{c|}{\textbf{0.019}} &  -0.685 &  \multicolumn{1}{c|}{0.027} &  -0.660 &  0.029 \\ \cline{2-12} 
&  \multirow{2}{*}{100} &  Cov (\%) &  93.50 &  100.0 &  \multicolumn{1}{c|}{100.0} &  \textbf{69.50} &  \multicolumn{1}{c|}{\textbf{96.50}} &  \textbf{68.00} &  \multicolumn{1}{c|}{\textbf{97.50}} &  \textbf{73.00} &  \textbf{97.50} \\
&   &  Var Err &  -0.022 &  $\infty$ &  \multicolumn{1}{c|}{$\infty$} &  \textbf{-0.732} &  \multicolumn{1}{c|}{\textbf{0.030}} &  \textbf{-0.727} &  \multicolumn{1}{c|}{\textbf{0.028}} &  \textbf{-0.718} &
\textbf{0.035} \\ \hline
\end{tabular}}
\caption{\textit{Linear regression: the empirical coverage rate of 95\% confidence intervals $(\textit{Cov})$ and the averaged relative estimation error of the variance $(\textit{Var Err})$ of $\b1^T\bx_t/d$, given by $\b1^T(\hat{\Xi}_t-\Xi^\star)\b1/\b1^T\Xi^\star\b1$. We bold entries to highlight scenarios where $\hat{\Xi}_t$ performs significantly better than others.}}
\label{table:1}
\end{table}

\begin{remark}
We should also mention in this remark that the bias of the plug-in covariance estimator $\tilde{\Xi}_t$ does not necessarily always result in undercoverage; it can also lead to overcoverage that may be less apparent than undercoverage. Moreover, the bias is closely related to the condition number of the Hessian matrix $B_t$ \cite[(4.3)]{Na2025Statistical}. In particular, we observe that the coverage rate of the plug-in estimator $\tilde{\Xi}_t$ in \cite{Na2025Statistical}, although it still exhibits some undercoverage, is close to the nominal 95\% level in many settings (see their Appendix F.2), while its coverage rate drops substantially in our setting (see Table \ref{table:1}).

Although this comparison is not very rigorous, since constrained and unconstrained problems are fundamentally different (the former relies on the Lagrangian function and inference directions must be related to active constraints, while the latter only involves the objective and allows inference in any directions), it is still of interest to explore underlying mechanisms governing the behavior of the plug-in estimator across different problems settings. Note that the bias of $\tilde{\Xi}_t$ is of order $O((1-\gamma_S)^\tau)$ (see \eqref{equ:1}), which depends critically on $\gamma_S$. As shown in \cite[(4.3)]{Na2025Statistical}, $\gamma_S$ is bounded below by the reciprocal of the squared condition number of the Hessian $B_t$, implying that a larger condition number leads to a larger bias gap. The Hessian in our setting indeed has a substantially higher condition number than that considered in \cite{Na2025Statistical}.
For example, in the linear regression model with Toeplitz covariance matrix $\Sigma_a$, when $r = 0.6$ and $d = 60$, our condition number is $15.94$, compared to $3.41$ in their study. A similar gap appears in the Equi-correlation case: when $r = 0.3$ and $d = 60$, our condition number is $26.71$, while theirs is $8.38$. In fact, the authors of \cite{Na2025Statistical} increased the variance of the design covariates from $1$ to $6$ in order to reduce the condition number of the true Hessian (of the Lagrangian function), which is otherwise inflated by a randomly drawn, normally distributed constraint matrix $A$ that is not included in our study.
\end{remark}

\begin{figure}[!t]
\centering   
\subfigure[SGD]{\label{B11}\includegraphics[width=0.32\textwidth]{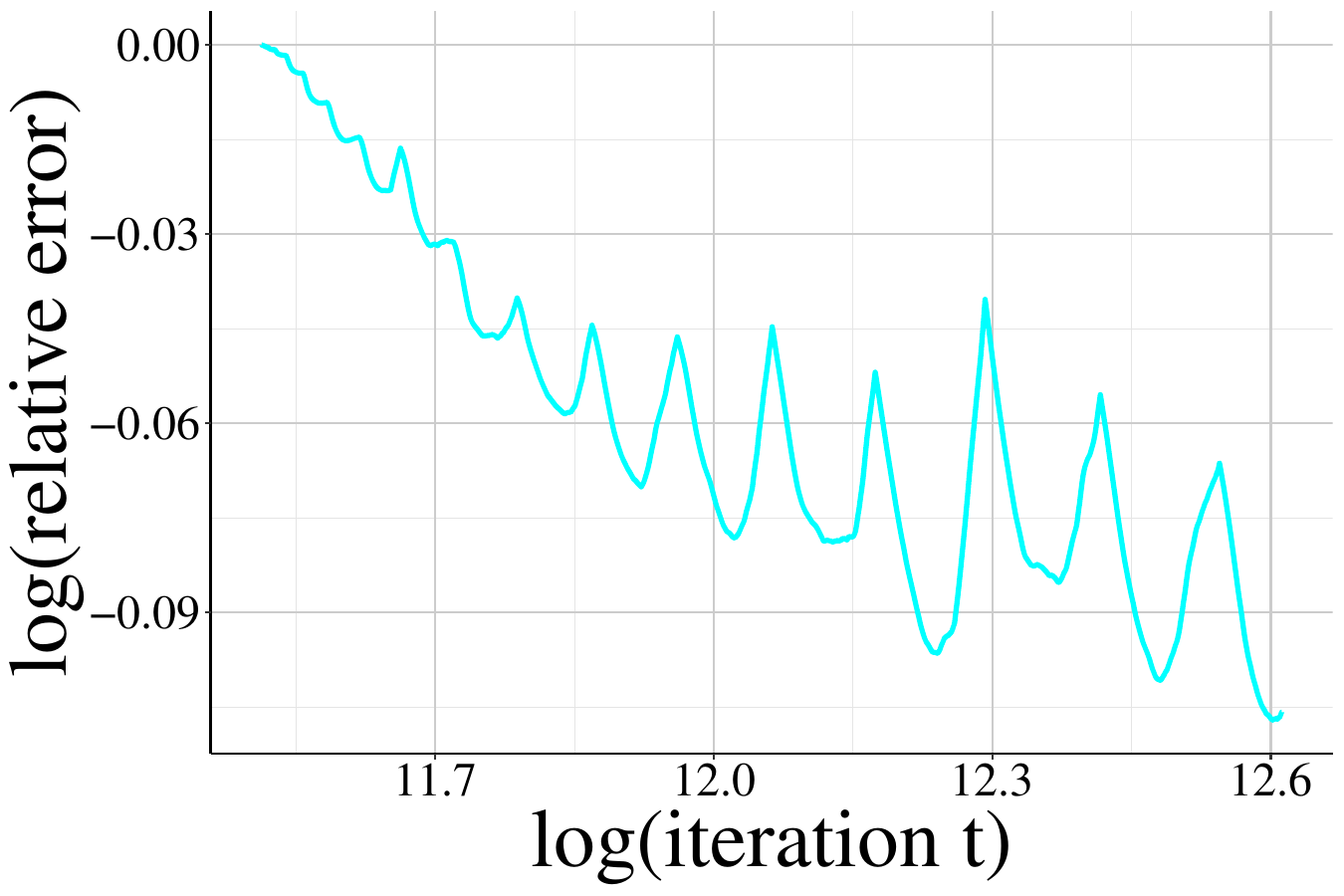}}
\subfigure[exact Newton ($\tau=\infty$)]{\label{B12}\includegraphics[width=0.32\textwidth]{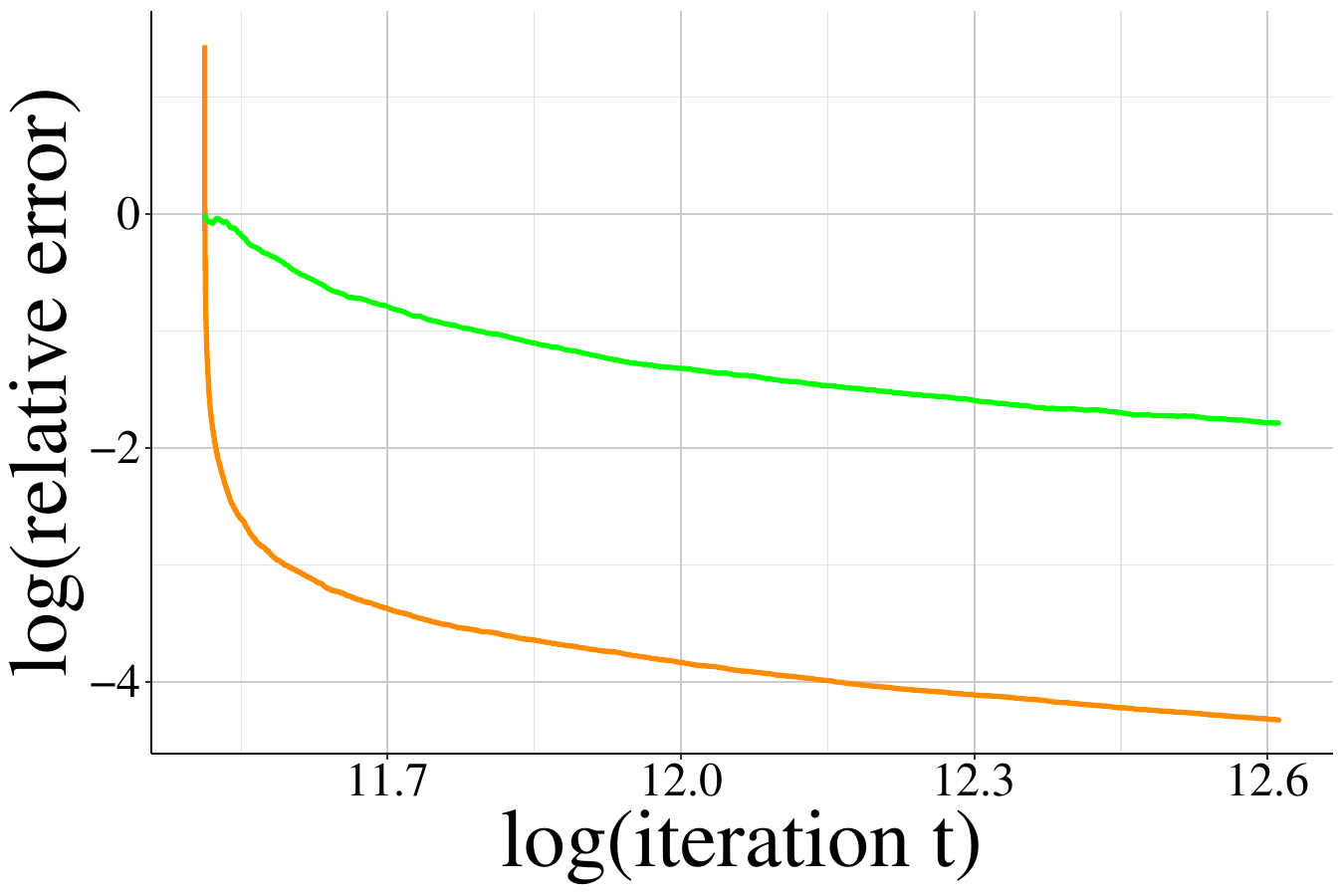}}
\subfigure[sketched Newton ($\tau=2$)]{\label{B13}\includegraphics[width=0.32\textwidth]{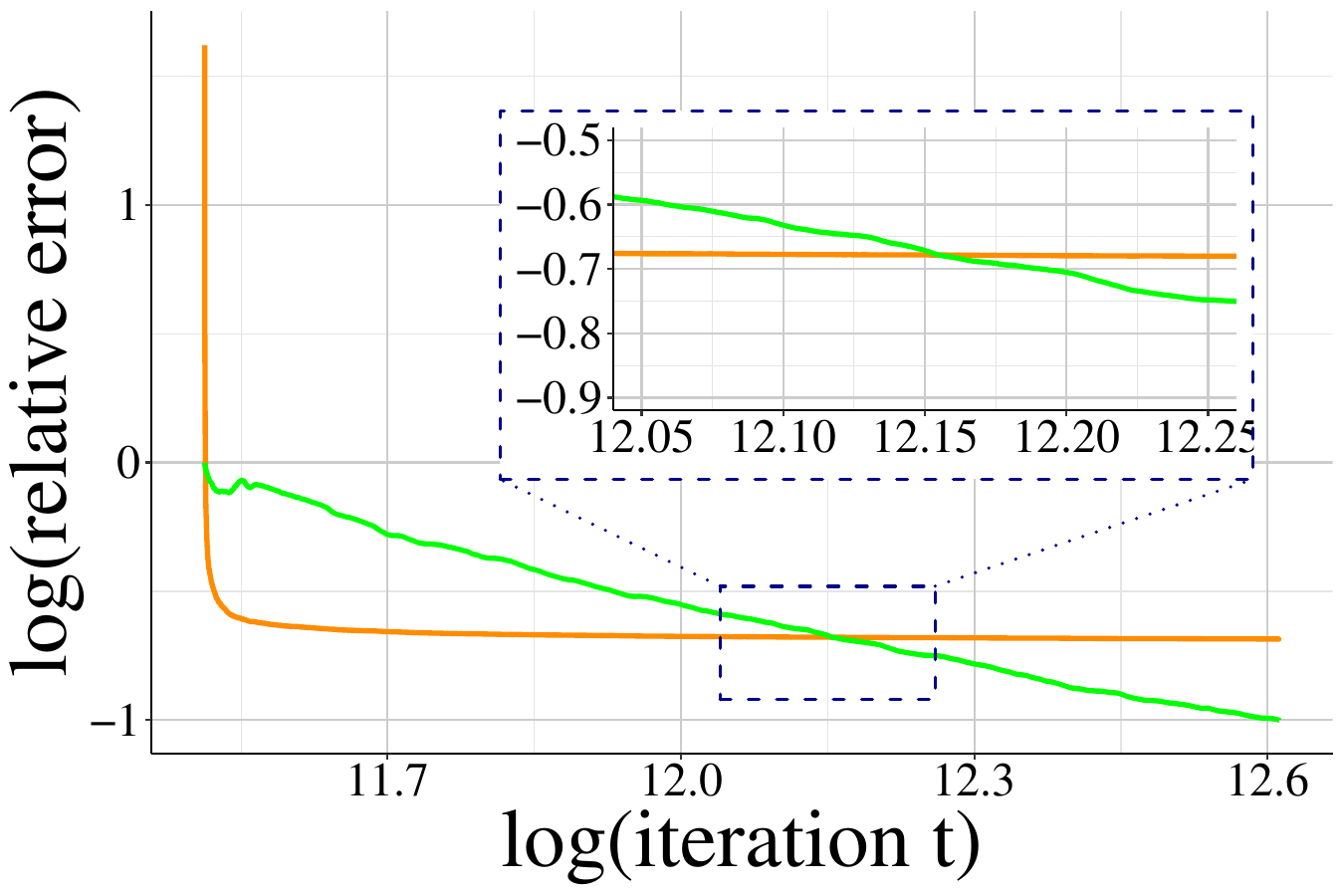}}
\vskip5pt
\centering{Relative covariance estimation error}
	
\subfigure[SGD]{\label{B21}\includegraphics[width=0.32\textwidth]{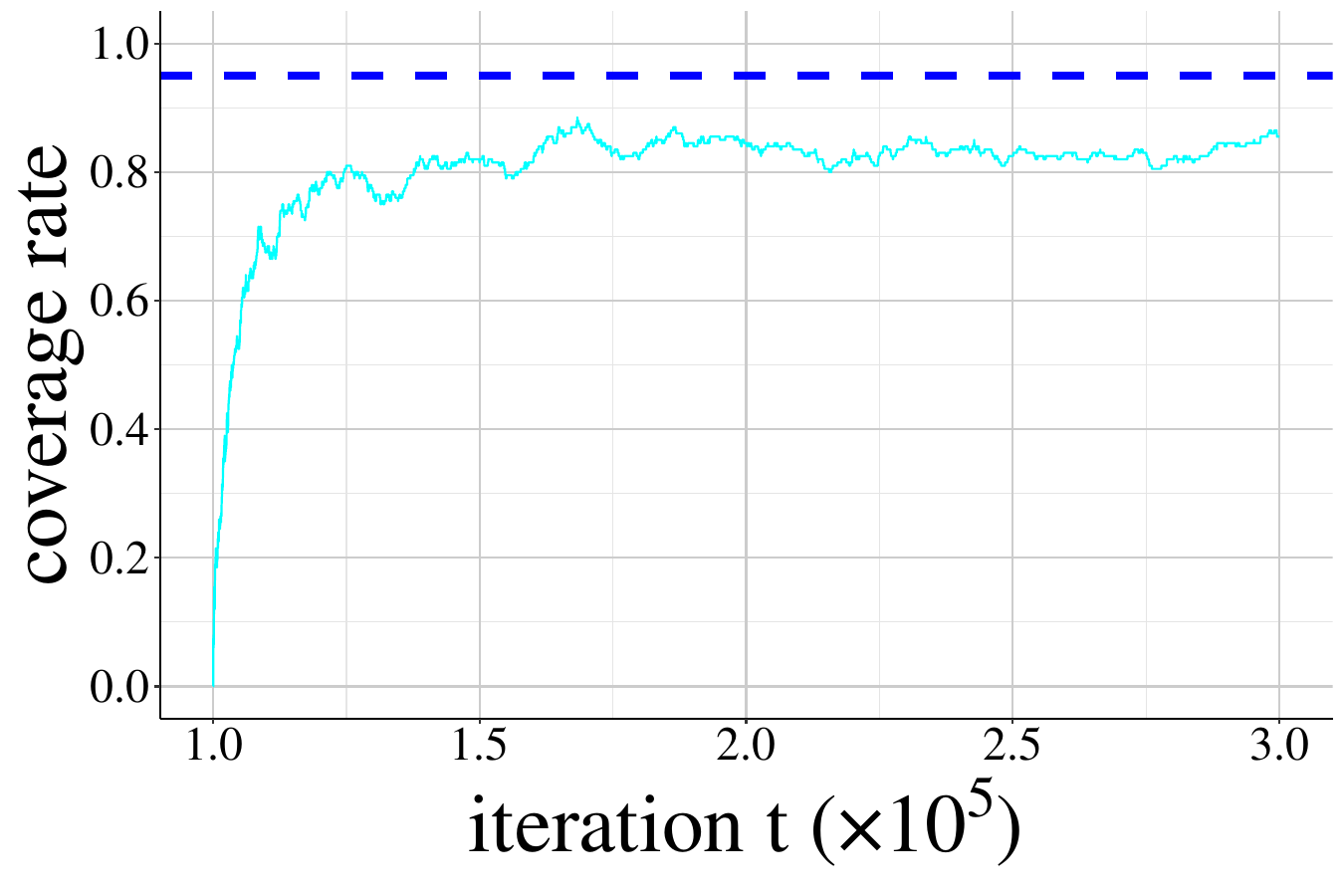}}
\subfigure[exact Newton ($\tau=\infty$)]{\label{B22}\includegraphics[width=0.32\textwidth]{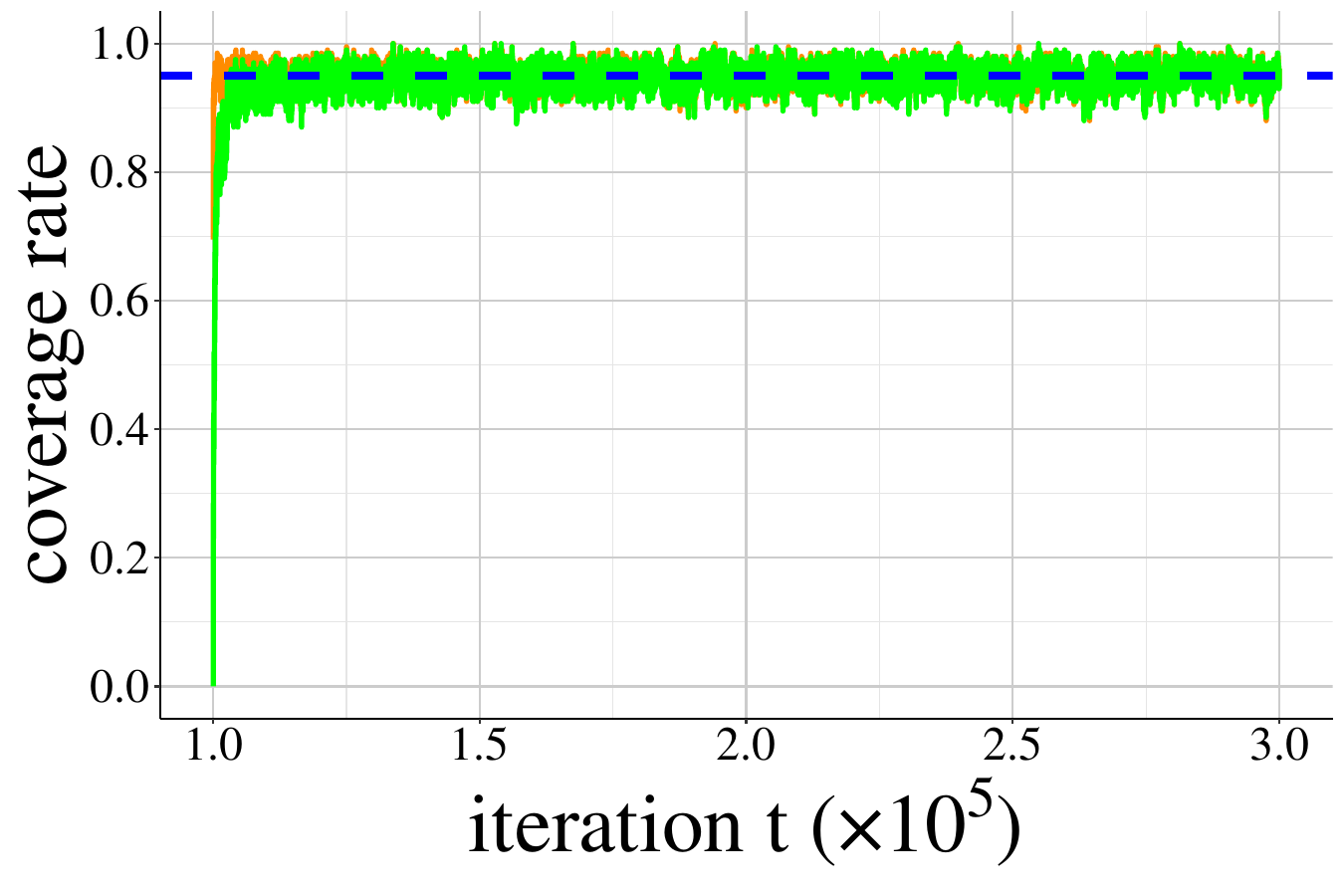}}
\subfigure[sketched Newton ($\tau=2$)]{\label{B23}\includegraphics[width=0.32\textwidth]{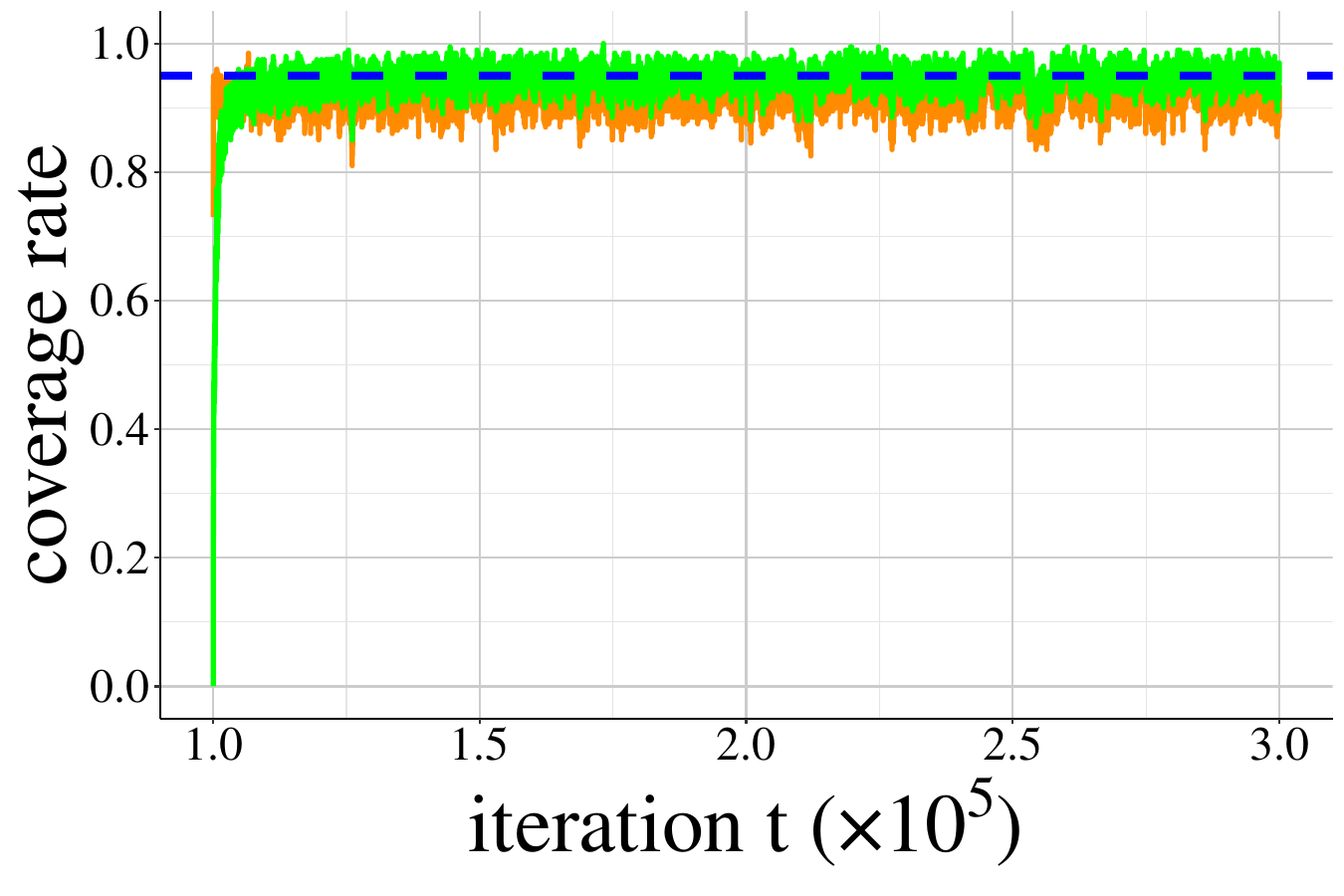}}
\vskip5pt
\centering{Empirical coverage rate of 95\% confidence intervals}

\subfigure[SGD]{\label{B31}\includegraphics[width=0.32\textwidth]{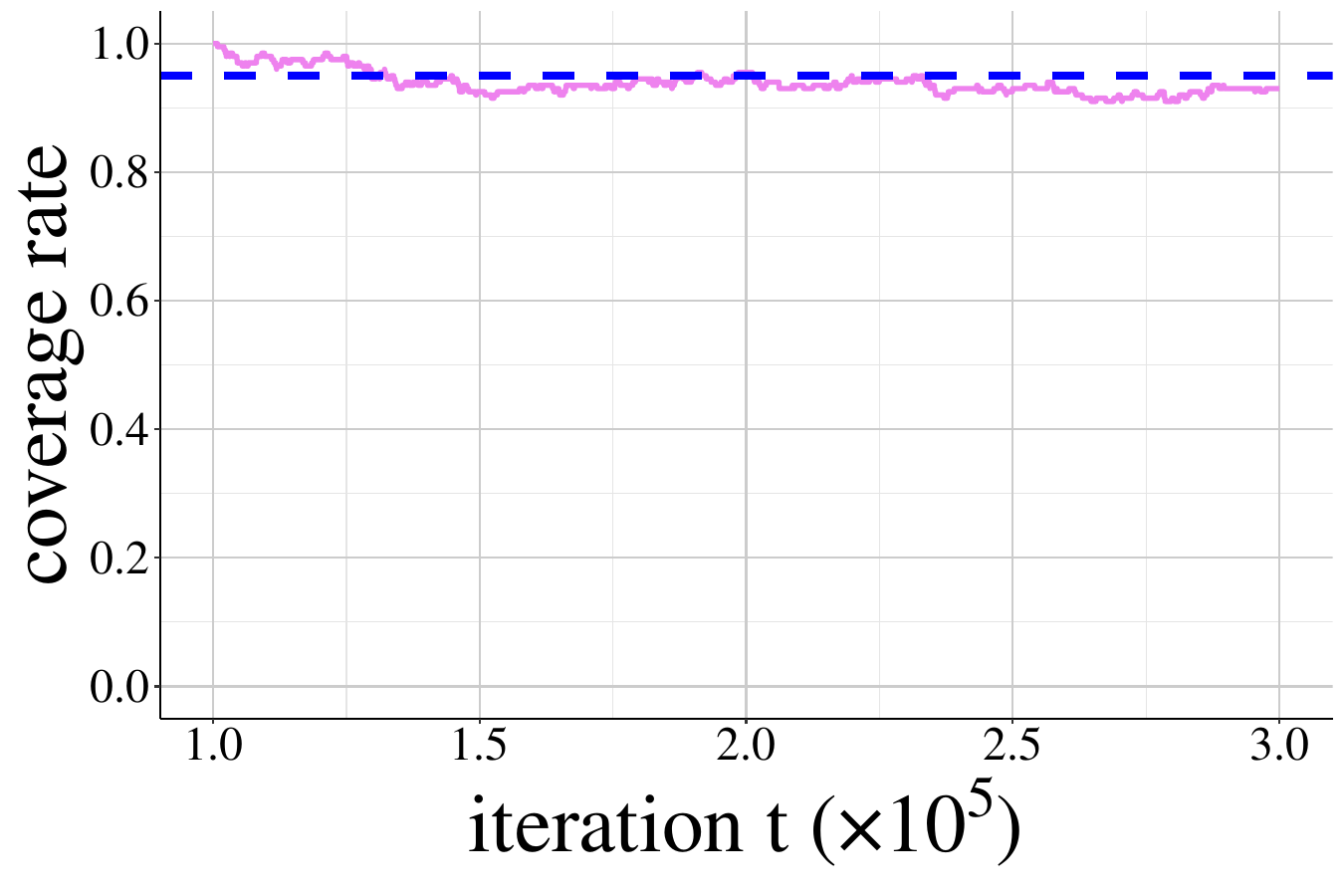}}
\subfigure[exact Newton ($\tau=\infty$)]{\label{B32}\includegraphics[width=0.32\textwidth]{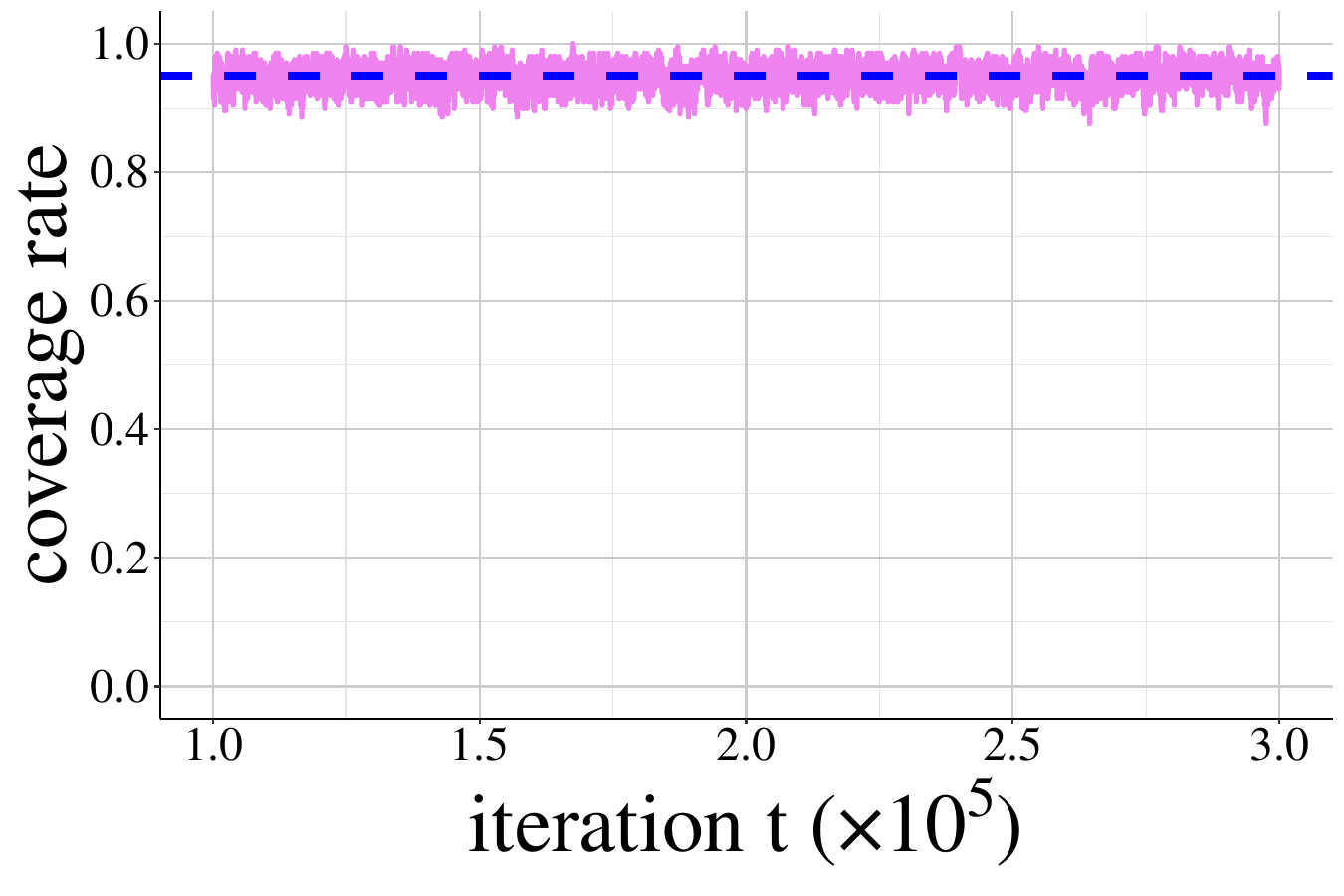}}
\subfigure[sketched Newton ($\tau=2$)]{\label{B33}\includegraphics[width=0.32\textwidth]{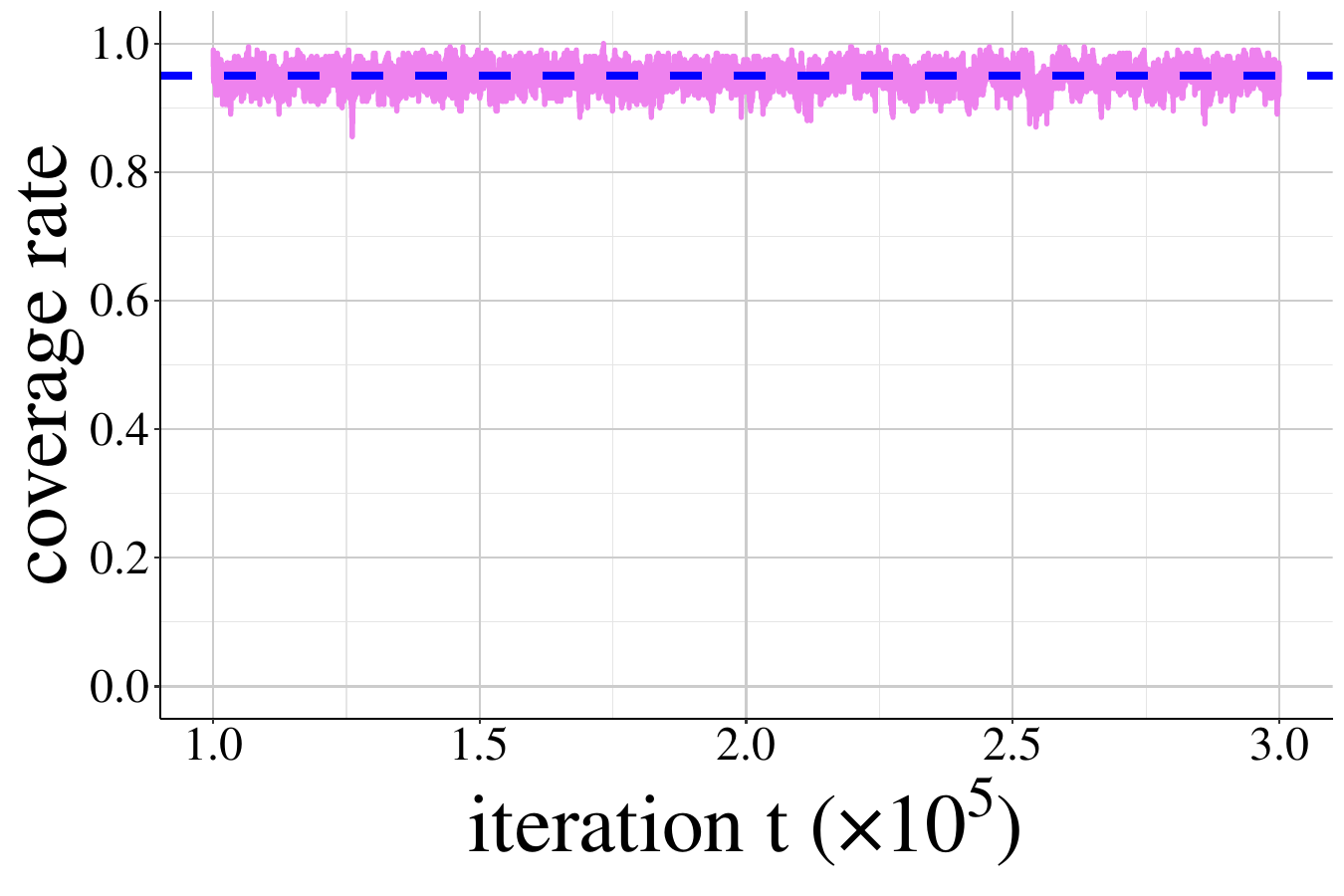}}
\vskip5pt
\centering{Empirical coverage rate of 95\% oracle confidence intervals}
\vskip5pt
\includegraphics[width=0.95\textwidth]{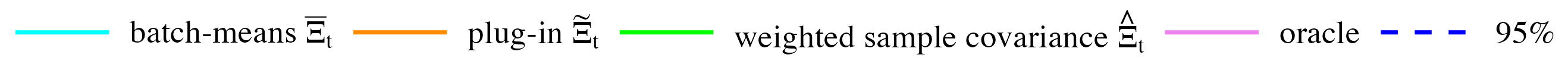}
\caption{\textit{The averaged trajectories for logistic regression problems with $d=5$ and Toeplitz $\Sigma_a$ $(r=0.6)$. See Figure \ref{fig:1} for interpretation.}}\label{fig:2}
\end{figure}

\subsection{Logistic regression}\label{sec5:logistic}

Next we consider the logistic regression model $P(\xi_{b}\mid\bxi_{a})\propto \exp(0.5\xi_{b}\cdot\bxi_{a}^T\bx^{\star})$ with $\xi_{b}\in\{-1,1\}$. For this model, we use the log loss defined as $f(\bx; \xi)=\log\big(1+\exp(-\xi_{b}\cdot\bxi_{a}^T\bx)\big)$. We follow the same experimental setup as in the linear regression model in Section \ref{sec5:linear}. Following Section \ref{sec5:linear}, we summarize part of results for logistic regression in Figure \ref{fig:2} and Table \ref{table:2}; a complete result is provided in Appendix \ref{appen:exp}.

The findings largely align with those observed for linear regression. The only noticeable difference is that $\bar{\Xi}_t$ for the SGD method exhibits worse performance in terms of coverage rate compared to linear regression problems. In contrast, second-order (sketched) Newton methods perform consistently well, with the sample covariance $\hat{\Xi}_t$ excelling in the majority of cases. These results further reconfirm the consistency of $\hat{\Xi}_t$ and illustrate that the confidence intervals constructed using $\hat{\Xi}_t$ are asymptotically valid.

\begin{table}[!t]
\centering
\resizebox{\linewidth}{!}{
\begin{tabular}{|c|c|c|c|cccccccc|}
\hline
\multirow{3}{*}{$\Sigma_a$} &  \multirow{3}{*}{d} &  \multirow{3}{*}{Criterion} &  \multirow{2}{*}{SGD} &  \multicolumn{8}{c|}{Sketched Newton Method} \\ \cline{5-12} 
&   &   &   &  \multicolumn{2}{c|}{$\tau=\infty$} &  \multicolumn{2}{c|}{$\tau=10$} &  \multicolumn{2}{c|}{$\tau=20$} &  \multicolumn{2}{c|}{$\tau=40$} \\ \cline{4-12} 
&   &   &  {\footnotesize$\bar{\Xi}_t$} &  {\footnotesize$\tilde{\Xi}_t$} &  \multicolumn{1}{c|}{{\footnotesize$\hat{\Xi}_t$}} &  {\footnotesize$\tilde{\Xi}_t$} &  \multicolumn{1}{c|}{{\footnotesize$\hat{\Xi}_t$}} &  {\footnotesize$\tilde{\Xi}_t$} &  \multicolumn{1}{c|}{{\footnotesize$\hat{\Xi}_t$}} &  {\footnotesize$\tilde{\Xi}_t$} &  {\footnotesize$\hat{\Xi}_t$} \\ \hline
\multirow{8}{*}{Identity} &  \multirow{2}{*}{20} &  Cov (\%) &  88.00 &  95.00 &  \multicolumn{1}{c|}{95.50} &  93.50 &  \multicolumn{1}{c|}{94.50} &  94.00 &  \multicolumn{1}{c|}{94.00} &  94.00 &  94.00 \\
&   &  Var Err   & -0.262 &  0.040 &  \multicolumn{1}{c|}{0.038} &  0.052 &  \multicolumn{1}{c|}{0.021} &  0.052 &  \multicolumn{1}{c|}{0.036} &  0.048 &  0.031 \\ \cline{2-12} 
&  \multirow{2}{*}{40} &  Cov (\%) &  \textbf{84.50} &  93.00 &  \multicolumn{1}{c|}{93.50} &  \textbf{95.00} &  \multicolumn{1}{c|}{\textbf{94.00}} &  95.50 & \multicolumn{1}{c|}{95.50} &  97.00 &  97.50 \\
&   &  Var Err &  \textbf{-0.307} &  0.084 &  \multicolumn{1}{c|}{0.070} &  \textbf{0.090} &  \multicolumn{1}{c|}{\textbf{0.052}} &  0.090 & \multicolumn{1}{c|}{0.073} &  0.089 &  0.083 \\ \cline{2-12} 
&  \multirow{2}{*}{60} &  Cov (\%) &  89.00 &  92.00 &  \multicolumn{1}{c|}{92.50} &  94.00 &  \multicolumn{1}{c|}{93.50} &  95.00 &  \multicolumn{1}{c|}{95.00} &  92.50 &  92.50 \\
&   &  Var Err &  -0.240 &  0.129 &  \multicolumn{1}{c|}{0.123} &  0.134 &  \multicolumn{1}{c|}{0.102} &  0.134 &  \multicolumn{1}{c|}{0.115} &  0.132 &  0.110 \\ \cline{2-12} 
&  \multirow{2}{*}{100} &  Cov (\%) &  86.50 &  95.50 &  \multicolumn{1}{c|}{96.00} &  97.00 &  \multicolumn{1}{c|}{96.50} &  95.50 &  \multicolumn{1}{c|}{96.00} &  93.50 &  93.00 \\
&   &  Var Err &  -0.220 &  0.229 &  \multicolumn{1}{c|}{0.219} &  0.236 &  \multicolumn{1}{c|}{0.195} &  0.233 &  \multicolumn{1}{c|}{0.200} &  0.231 &  0.221 \\ \hline
\multirow{8}{*}{\shortstack{Toeplitz\\$r=0.5$}} &  \multirow{2}{*}{20} &  Cov (\%) &  88.00 &  98.50 &  \multicolumn{1}{c|}{98.00} &  93.50 &  \multicolumn{1}{c|}{96.00} &  94.00 &  \multicolumn{1}{c|}{95.50} &  93.50 &  94.50 \\
&   &  Var Err &  -0.199 &  0.038 &  \multicolumn{1}{c|}{0.032} &  -0.226 &  \multicolumn{1}{c|}{0.028} &  -0.179 &  \multicolumn{1}{c|}{0.028} &  -0.097 &  0.020 \\ \cline{2-12} 
&  \multirow{2}{*}{40} &  Cov (\%) &  \textbf{85.50} &  96.00 &  \multicolumn{1}{c|}{96.50} &  \textbf{91.50} &  \multicolumn{1}{c|}{\textbf{95.00}} &  90.00 &  \multicolumn{1}{c|}{93.50} &  95.00 &  97.50 \\
&   &  Var Err &  \textbf{-0.190} &  0.079 &  \multicolumn{1}{c|}{0.077} &  \textbf{-0.233} &  \multicolumn{1}{c|}{\textbf{0.063}} &  -0.217 &  \multicolumn{1}{c|}{0.066} &  -0.160 &  0.064 \\ \cline{2-12} 
&  \multirow{2}{*}{60} &  Cov (\%) &  92.00 &  95.50 &  \multicolumn{1}{c|}{94.50} &  92.00 &  \multicolumn{1}{c|}{94.50} &  89.50 &  \multicolumn{1}{c|}{94.00} &  92.50 &  97.00 \\
&   &  Var Err &  -0.170 &  0.122 &  \multicolumn{1}{c|}{0.115} &  -0.215 &  \multicolumn{1}{c|}{0.081} &  -0.208 &  \multicolumn{1}{c|}{0.100} &  -0.174 &  0.094 \\ \cline{2-12} 
&  \multirow{2}{*}{100} &  Cov (\%) &  87.50 &  97.00 &  \multicolumn{1}{c|}{96.00} &  90.50 &  \multicolumn{1}{c|}{93.50} &  92.00 &  \multicolumn{1}{c|}{96.50} &  90.00 &  93.50 \\
&   &  Var Err &  -0.159 &  0.221 &  \multicolumn{1}{c|}{0.215} &  -0.158 &  \multicolumn{1}{c|}{0.163} &  -0.161 &  \multicolumn{1}{c|}{0.166} &  -0.146 &  0.164 \\ \hline
\multirow{8}{*}{\shortstack{Equi-corr\\$r=0.2$}} &  \multirow{2}{*}{20} &  Cov (\%) &  90.00 &  96.00 &  \multicolumn{1}{c|}{96.00} &  88.50 &  \multicolumn{1}{c|}{96.50} &  \textbf{89.00} &  \multicolumn{1}{c|}{\textbf{96.50}} &  92.50 &  96.00 \\
&   &  Var Err &  -0.172 &  0.041 &  \multicolumn{1}{c|}{0.037} &  -0.394 &  \multicolumn{1}{c|}{0.028} &  \textbf{-0.302} &  \multicolumn{1}{c|}{\textbf{0.037}} &  -0.153 &  0.039 \\ \cline{2-12}  
&  \multirow{2}{*}{40} &  Cov (\%) &  \textbf{86.00} &  95.00 &  \multicolumn{1}{c|}{95.00} &  \textbf{78.00} &  \multicolumn{1}{c|}{\textbf{95.00}} &  \textbf{81.00} &  \multicolumn{1}{c|}{\textbf{94.50}} &  \textbf{88.00} &  \textbf{96.50} \\
&   &  Var Err &  \textbf{-0.111} &  0.083 &  \multicolumn{1}{c|}{0.084} &  \textbf{-0.530} &  \multicolumn{1}{c|}{\textbf{0.062}} &  \textbf{-0.490} &  \multicolumn{1}{c|}{\textbf{0.046}} &  \textbf{-0.402} &  \textbf{0.050} \\ \cline{2-12} 
&  \multirow{2}{*}{60} &  Cov (\%) &  80.00 &  94.00 &  \multicolumn{1}{c|}{93.50} &  78.50 &  \multicolumn{1}{c|}{94.00} &  \textbf{80.00} &  \multicolumn{1}{c|}{\textbf{97.00}} &  82.50 &  96.00 \\
&   &  Var Err &  -0.144 &  0.130 &  \multicolumn{1}{c|}{0.110} &  -0.592 &  \multicolumn{1}{c|}{0.076} &  \textbf{-0.569} &  \multicolumn{1}{c|}{\textbf{0.068}} &  -0.518 &  0.072 \\ \cline{2-12} 
&  \multirow{2}{*}{100} &  Cov (\%) &  66.50 &  97.50 &  \multicolumn{1}{c|}{96.00} &  73.50 &  \multicolumn{1}{c|}{96.00} &  \textbf{73.00} &  \multicolumn{1}{c|}{\textbf{96.00}} &  80.00 &  97.00 \\
&   &  Var Err &  -0.108 &  0.234 &  \multicolumn{1}{c|}{0.227} &  -0.647 &  \multicolumn{1}{c|}{0.116} &  \textbf{-0.636} &  \multicolumn{1}{c|}{\textbf{0.115}} &  -0.615 &  0.109 \\ \hline
\end{tabular}}
\caption{\textit{Logistic regression: the empirical coverage rate of 95\% confidence intervals $(\textit{Cov})$ and the averaged relative estimation error of the variance $(\textit{Var Err})$ of $\b1^T\bx_t/d$, given by $\b1^T(\hat{\Xi}_t-\Xi^\star)\b1/\b1^T\Xi^\star\b1$. We bold entries to highlight scenarios where $\hat{\Xi}_t$ performs significantly better than others.}}
\label{table:2}
\end{table}

\subsection{Inference under different sketching configurations}\label{sec:5.3}

\begin{table}[!t]
\centering
\resizebox{\linewidth}{!}{
\begin{tabular}{|c|c|c|ccc|ccc|ccc|}
\hline
\multirow{3}{*}{Sketching} & \multirow{3}{*}{$r$} & \multirow{3}{*}{$q$} 
& \multicolumn{3}{c|}{$\tau = 5$} 
& \multicolumn{3}{c|}{$\tau = 10$} 
& \multicolumn{3}{c|}{$\tau = 20$} \\ \cline{4-12}
& & 
& Err & Cov & Len 
& Err & Cov & Len 
& Err & Cov & Len \\ [-4pt]
& & 
& {\footnotesize (ratio)} & {\footnotesize (\%)} & {\footnotesize $(\times 10^{-2})$} 
& {\footnotesize (ratio)} & {\footnotesize (\%)} & {\footnotesize $(\times 10^{-2})$}
& {\footnotesize (ratio)} & {\footnotesize (\%)} & {\footnotesize $(\times 10^{-2})$} \\ \hline
\multirow{8}{*}{Gaussian}
& \multirow{4}{*}{0}
& 1
& 1.072 & 95.50 & 2.547
& 0.786 & 93.00 & 2.575
& \textit{0.607} & 96.00 & 2.591 \\ \cline{3-12}
& & 5
& 0.546 & 94.00 & 2.594
& 0.491 & 95.50 & 2.583
& \textit{\textbf{0.476}} & 94.00 & 2.588 \\ \cline{3-12}
& & 10
& 0.484 & 95.50 & 2.602
& 0.477 & 93.50 & 2.596
& \textit{\textbf{0.476}} & 94.50 & 2.579 \\ \cline{3-12}
& & 20
& \textbf{0.476} & 95.00 & 2.600
& \textit{\textbf{0.471}} & 95.50 & 2.598
& 0.477 & 94.00 & 2.605 \\ 
\cline{2-12}\\[-10pt]
\cline{2-12}
& \multirow{4}{*}{0.2}
& 1
& 2.546 & 96.50 & 1.814
& 1.804 & 92.00 & 1.738
& \textit{1.309} & 96.50 & 1.589 \\ \cline{3-12}
& & 5
& 1.882 & 94.50 & 1.299
& 1.600 & 97.00 & 1.216
& \textit{1.324} & 92.00 & 1.182 \\ \cline{3-12}
& & 10
& 1.783 & 96.50 & 1.195
& 1.576 & 96.00 & 1.182
& \textit{1.331} & 94.50 & 1.179 \\ \cline{3-12}
& & 20
& \textbf{1.755} & 93.50 & 1.180
& \textbf{1.555} & 96.50 & 1.184
& \textit{\textbf{1.323}} & 94.00 & 1.188 \\ \hline\hline
\multirow{8}{*}{Kaczmarz}
& \multirow{4}{*}{0}
& 1
& 1.075 & 96.00 & 2.560
& 0.786 & 95.00 & 2.578
& \textit{0.596} & 94.00 & 2.593 \\ \cline{3-12}
& & 5
& 0.574 & 95.50 & 2.597
& 0.494 & 96.50 & 2.594
& \textit{\textbf{0.474}} & 95.50 & 2.592 \\ \cline{3-12}
& & 10
& 0.492 & 95.50 & 2.586
& \textit{\textbf{0.474}} & 95.00 & 2.602
& 0.476 & 97.00 & 2.581 \\ \cline{3-12}
& & 20
& \textit{\textbf{0.473}} & 93.50 & 2.594
& 0.477 & 97.00 & 2.581
& 0.482 & 93.00 & 2.594 \\ 
\cline{2-12}\\[-10pt]
\cline{2-12}
& \multirow{4}{*}{0.2}
& 1
& 1.755 & 93.50 & 1.827
& \textbf{1.209} & 98.50 & 1.757
& \textit{\textbf{0.858}} & 93.50 & 1.636 \\ \cline{3-12}
& & 5
& 1.930 & 95.00 & 1.301
& 1.610 & 94.50 & 1.225
& \textit{1.321} & 95.00 & 1.184 \\ \cline{3-12}
& & 10
& 1.788 & 94.50 & 1.201
& 1.563 & 95.50 & 1.185
& \textit{1.309} & 94.50 & 1.184 \\ \cline{3-12}
& & 20
& \textbf{1.751} & 95.00 & 1.180
& 1.568 & 96.00 & 1.187
& \textit{1.334} & 94.50 & 1.183 \\ \hline			
\end{tabular}}
\caption{\textit{Linear regression: the averaged relative estimation error $($Err$)$ $\|\hat{\Xi}_t-\Xi^\star\|/\|\Xi^\star\|$, the empirical coverage rate of 95\% confidence intervals $($Cov$)$, and the averaged confidence interval length $($Len$)$ of $\b1^T\bx^\star/d$, evaluated under different sketching matrices $($Gaussian and Kaczmarz$)$, sketching dimension $q$, and sketching iteration number $\tau$. Fixing the sketching distribution and correlation parameter $r$, for each $\tau$ we bold the smallest error achieved across different values of $q$ (i.e., the smallest entry within each column), and for each $q$ we italicize the smallest error achieved across different values of $\tau$ (i.e., the smallest entry within each row).}
\label{table:5}}
\end{table}

In this section, we examine the empirical behavior of the sample covariance estimator $\hat{\Xi}_t$ under different sketching schemes. The experimental setup largely mirrors that of Section \ref{sec5:linear}. We focus on the linear regression model with an Equi-correlation design covariance matrix $\Sigma_a$, considering two correlation levels: $r = 0$ (Identity) and $r = 0.2$. The problem dimension is fixed at $d = 20$. We construct $95\%$ confidence intervals for $\b1^T\bx^\star/d$ based on $\hat{\Xi}_t$.
We investigate two sketching mechanisms. For Gaussian sketching, the sketching matrix $S = [S_{i,j}]\in\mathbb{R}^{d\times q}$ has entries $S_{i,j}$ drawn i.i.d. from standard normal distribution. For Kaczmarz sketching, the sketch $S = [S_1, \ldots, S_q]$ is formed by selecting $\{S_1,\dots,S_q\}$ uniformly from the canonical basis vectors $\{\be_1,\ldots,\be_d\}$. For both schemes, we vary the sketching dimension $q\in\{1,5,10,20\}$ and the number of sketching steps $\tau\in\{5,10,20\}$. All reported results are averaged over $200$ independent replications.
Table \ref{table:5} summarizes the averaged relative covariance estimation error (Err), the empirical coverage rate of $95\%$ confidence intervals (Cov), and the averaged confidence interval length (Len) under different sketching schemes, sketching dimensions $q$, and numbers of sketching steps $\tau$.

We first examine the results for Gaussian sketching. When $\tau=5$ and $\tau=10$, the relative estimation error decreases monotonically as $q$ increases, indicating that a larger sketching dimension improves the accuracy of covariance estimation. For moderate sketching dimensions ($q=1,5,10$), increasing the number of sketching iterations $\tau$ also reduces the estimation error, suggesting that more sketching iterations lead to more accurate Hessian approximations. 
As comparisons, when $q=20$, the estimation error remains largely unchanged across different values of $\tau$; and similarly, for $\tau=20$ and $q\geq 5$, further increases in $q$ yield only marginal improvements. These observations indicate that increasing either $q$ or $\tau$ enhances estimation accuracy, but once one of them is sufficiently large, further increases in the other yield diminishing returns.
We then turn to the length of the confidence interval. When $r=0$ (i.e., $\Sigma_a=I$), the interval length remains relatively stable across different values of $q$ and $\tau$. However, when $r=0.2$, corresponding to a more ill-conditioned covariance structure, increasing either $q$ or $\tau$ leads to shorter confidence intervals, particularly when $\tau$ is small. 
Overall, the order of the confidence interval length is largely determined by the number of samples (i.e., the number of iterations) used in the estimation procedure. In particular, the order of $10^{-2}$ achieved by our Newton method matches that of the confidence intervals constructed by SGD-based methods in \cite{Chen2020Statistical, Zhu2021Online}. This observation is also consistent with the confidence interval formula in Corollary \ref{sec4:cor1}.

We next consider the results for Kaczmarz sketching. When $r = 0$, the relative estimation error decreases sharply as $q$ increases from $1$ to $5$, while additional gains from larger values of $q$ are more moderate. When $r = 0.2$, increasing the sketching dimension from $q = 5$ to $q = 20$ continues to reduce the estimation error. Across both values of $r$ in the Kaczmarz setting, increasing $\tau$ consistently improves estimation accuracy. Similar to Gaussian sketching, the confidence interval length remains nearly unchanged when $r = 0$, while for $r = 0.2$, larger values of $q$ or $\tau$ lead to shorter confidence intervals. The order of confidence interval length of $10^{-2}$ is retained for different covariance structures.
Across all sketching configurations, the empirical coverage rate remains close to the nominal level of $95\%$, demonstrating that confidence intervals constructed using $\hat{\Xi}_t$ maintain reliable coverage under both Gaussian and Kaczmarz sketching schemes.

\subsection{Improved efficiency via Hessian preconditioning}\label{sec:5.4}

\begin{figure}[!t]
\centering     
\subfigure[ASGD ($\alpha_t = 1/t^{0.505}$) v.s. Newton ($\beta_t=1/t$)]{\label{C11}\includegraphics[width=0.45\textwidth]{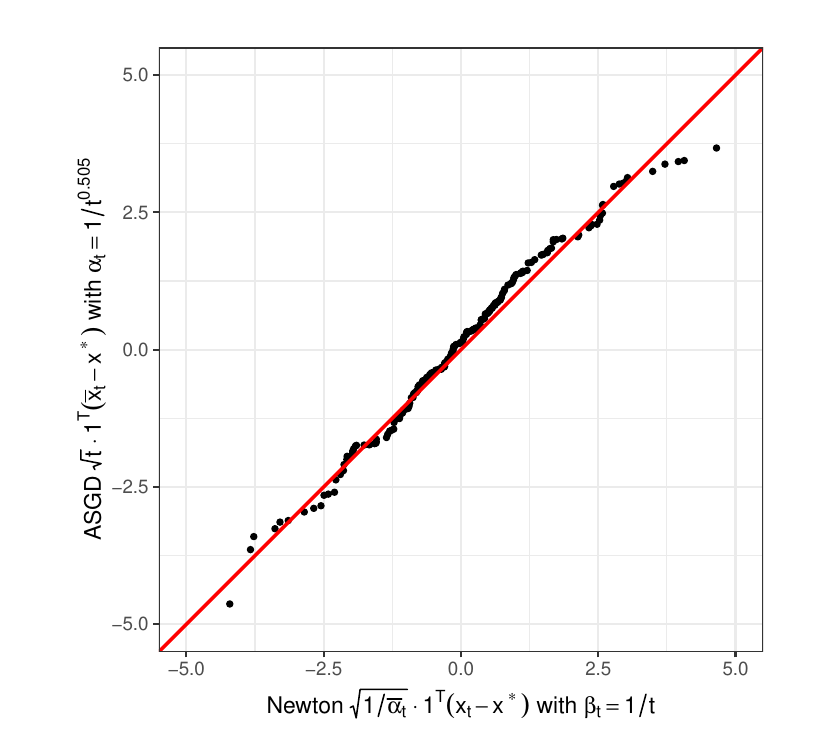}}
\subfigure[Newton ($\beta_t = 1/t^{0.505})$ v.s. Newton ($\beta_t = 1/t$)]{\label{C12}\includegraphics[width=0.45\textwidth]{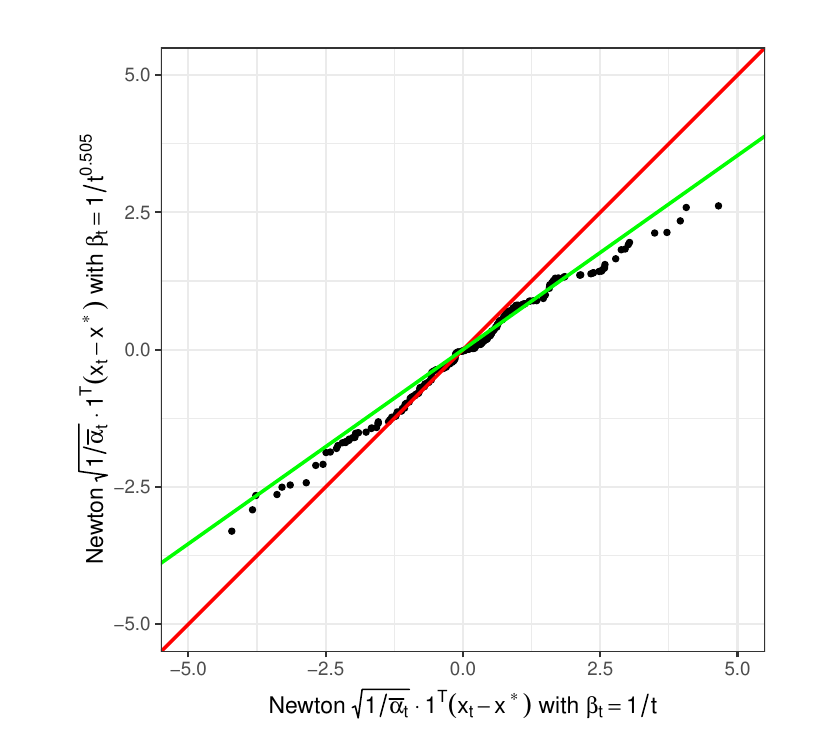}}	
\subfigure[SGD ($\alpha_t = 1/t^{0.505}$) v.s. Newton ($\beta_t = 1/t^{0.505}$)]{\label{C21}\includegraphics[width=0.45\textwidth]{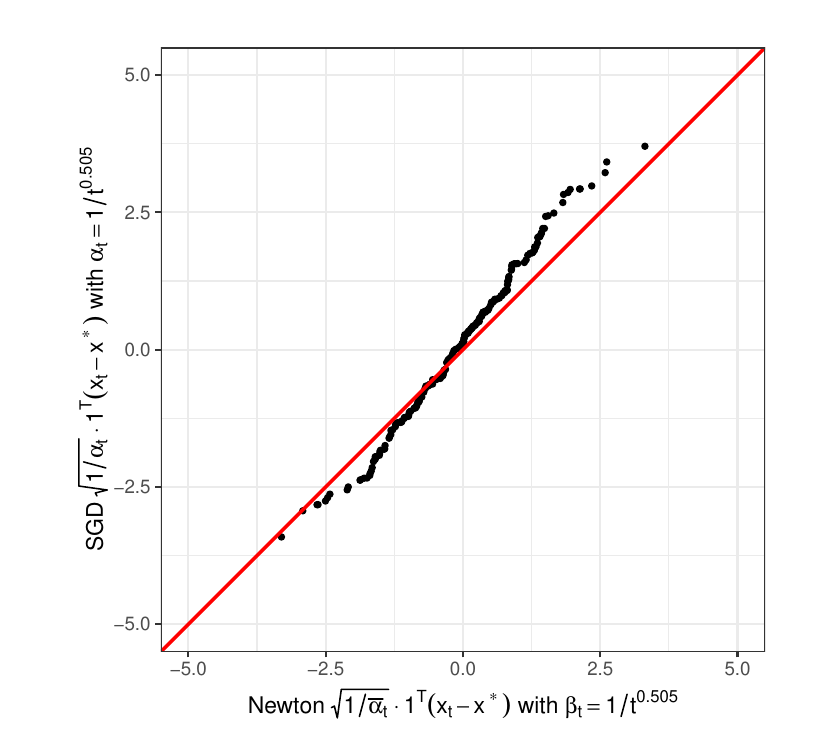}}
\subfigure[SGD ($\alpha_t = 1/t$) v.s. Newton ($\beta_t = 1/t$)]{\label{C22}\includegraphics[width=0.45\textwidth]{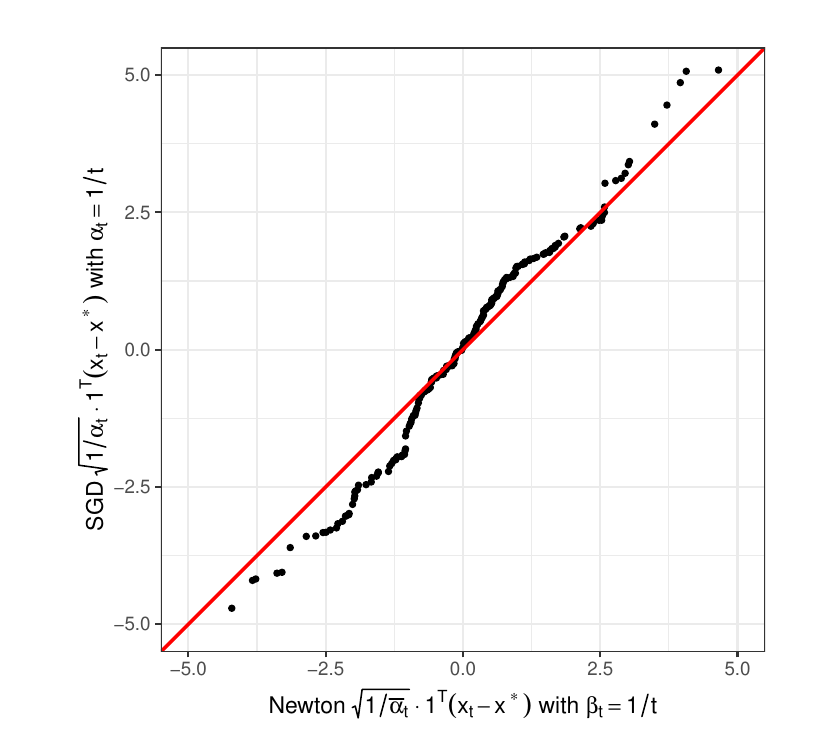}}	
\vskip3pt
\centering{Two-sample Q-Q plot of online methods}	
\vskip3pt
\caption{\textit{
The two-sample Q-Q plots for the linear regression problem with $d=5$ and Equi-correlation covariance $\Sigma_a$ ($r=0.2$). We record the scaled errors $\sqrt{1/\alpha_t}\cdot \mathbf{1}^T(\bx_t-\bx^\star)$, $\sqrt{t}\cdot \mathbf{1}^T(\bar{\bx}_t-\bx^\star)$, and $\sqrt{1/\bar{\alpha}_t}\cdot \mathbf{1}^T(\bx_t-\bx^\star)$ for SGD, ASGD, and Newton methods, respectively. Panels (a)-(d) correspond to different combinations of methods and stepsizes: (a) ASGD with $\alpha_t=t^{-0.505}$ versus the Newton method with $\beta_t=t^{-1}$; (b) the Newton method with $\beta_t=t^{-0.505}$ versus $\beta_t=t^{-1}$; (c) SGD with $\alpha_t=t^{-0.505}$ versus the Newton method with $\beta_t=t^{-0.505}$; and (d) SGD with $\alpha_t=t^{-1}$ versus the Newton method with $\beta_t=t^{-1}$. The red and green reference lines correspond to $y=x$ and $y=x/\sqrt{2}$, respectively. Overall, the Q-Q plots are consistent with the theoretical asymptotic normality of the considered online methods (cf. discussions in Section \ref{sec:4.1}).}}\label{fig:3}	
\end{figure}

In this section, we investigate how Hessian preconditioning in Newton methods affects the statistical efficiency of the estimation procedure. To this end, we compare the limiting covariance matrices of the last iterate and the averaged iterate for both SGD and Newton methods.
We consider the linear regression setting described in Section \ref{sec5:linear}, with an Equi-correlation design covariance $\Sigma_a$ and correlation parameter $r = 0.2$, and fix the dimension to $d = 5$. We implement SGD following \eqref{equ:sgd} with stepsize $\alpha_t = 1/t^{\alpha}$, and the Newton method following \eqref{nequ:7} with stepsize $\bar{\alpha}_t$ generated according to \eqref{nequ:6}.
In total, we compare five methods: SGD with $\alpha = 0.505$ and $\alpha = 1$; Averaged SGD (ASGD) with $\alpha = 0.505$; and the Newton method with $\beta = 0.505$ and $\beta = 1$, using $\chi = 2\beta$ and $c_{\beta} = c_{\chi} = 1$. All results are based on $200$ independent runs.

Figure \ref{fig:3} presents two-sample Q-Q plots of the scaled errors for estimating $\b1^T\tx$ using SGD, ASGD, and Newton methods under different stepsize regimes. From Figure \ref{C11}, we observe that the last iterate of the Newton method with $\beta_t=t^{-1}$ attains the same optimal limiting covariance $\Omega^\star$ as ASGD with $\alpha_t=t^{-0.505}$, indicating that iterate averaging is unnecessary for Newton methods to achieve asymptotic optimal efficiency. Figure \ref{C12} further shows that the quantile relationship between the scaled errors of the Newton method with $\beta_t=t^{-0.505}$ and $\beta_t=t^{-1}$ closely aligns with the line $y=x/\sqrt{2}$, which is consistent with the theoretical result that $\Xi^\star=\Omega^\star$ for $\beta_t=t^{-1}$ and $\Xi^\star=\Omega^\star/2$ for $\beta_t=t^{-0.505}$. This implies that the last iterate of the Newton method with $\beta\in(0.5,1)$ can be used to estimate $\Omega^\star$ up to a known scaling factor. 
In Figures \ref{C21} and \ref{C22}, we observe that the slopes in the quantile comparisons between SGD and Newton exceed one, indicating that under the same stepsize, SGD exhibits a larger asymptotic variance than the Newton method. Overall, these observations are in agreement with our theoretical analysis and highlight the improved statistical efficiency of Newton updates in online algorithms compared to SGD.

\subsection{Impact of sketching iteration number on covariance estimation}\label{sec:5.5}

\begin{figure}[!t]
\centering   
\subfigure[Toeplitz $\Sigma_a$ with $r=0.5$]{\label{D11}\includegraphics[width=0.40\textwidth]{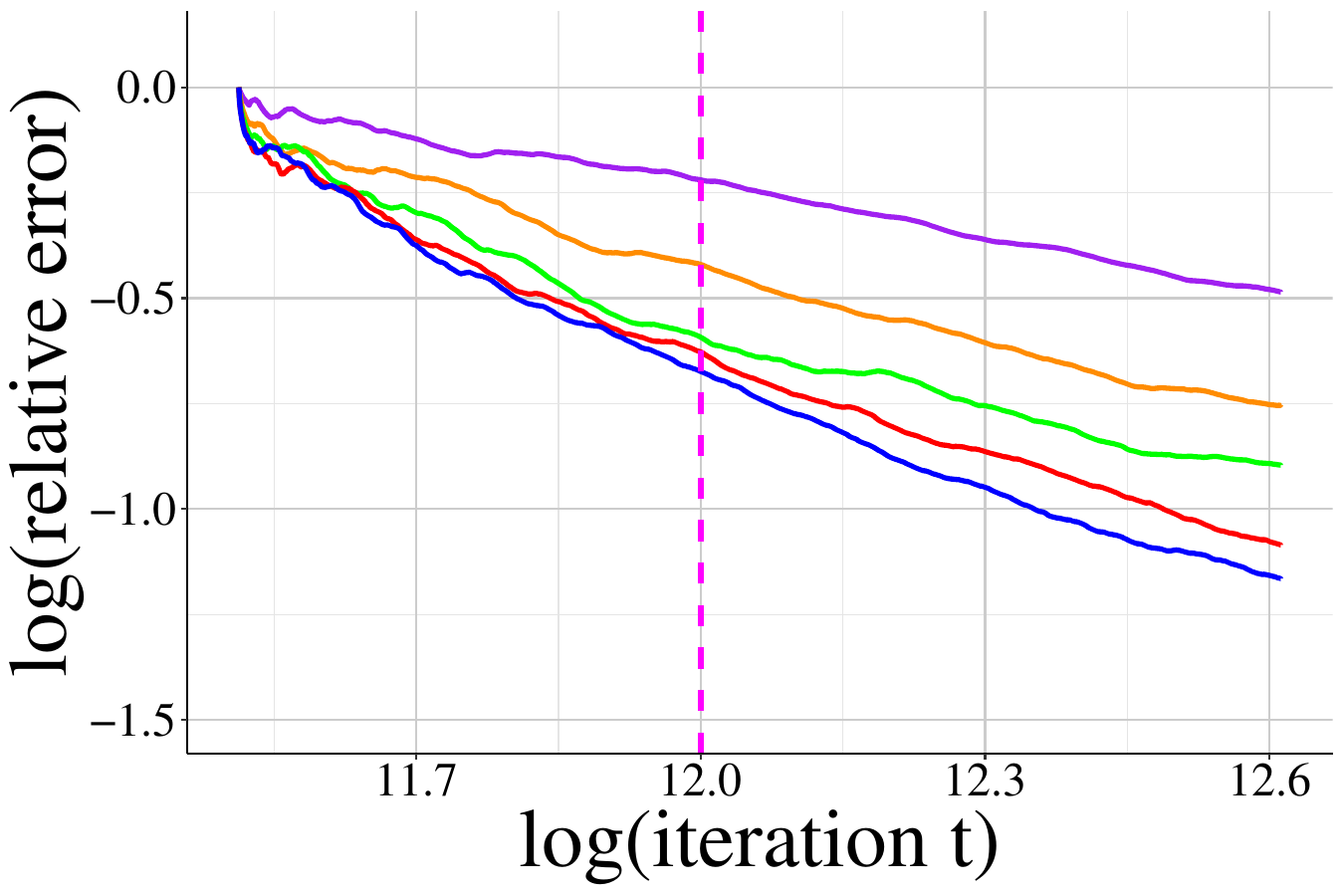}}
\subfigure[Equi-corr $\Sigma_a$ with $r=0.2$]{\label{D12}\includegraphics[width=0.40\textwidth]{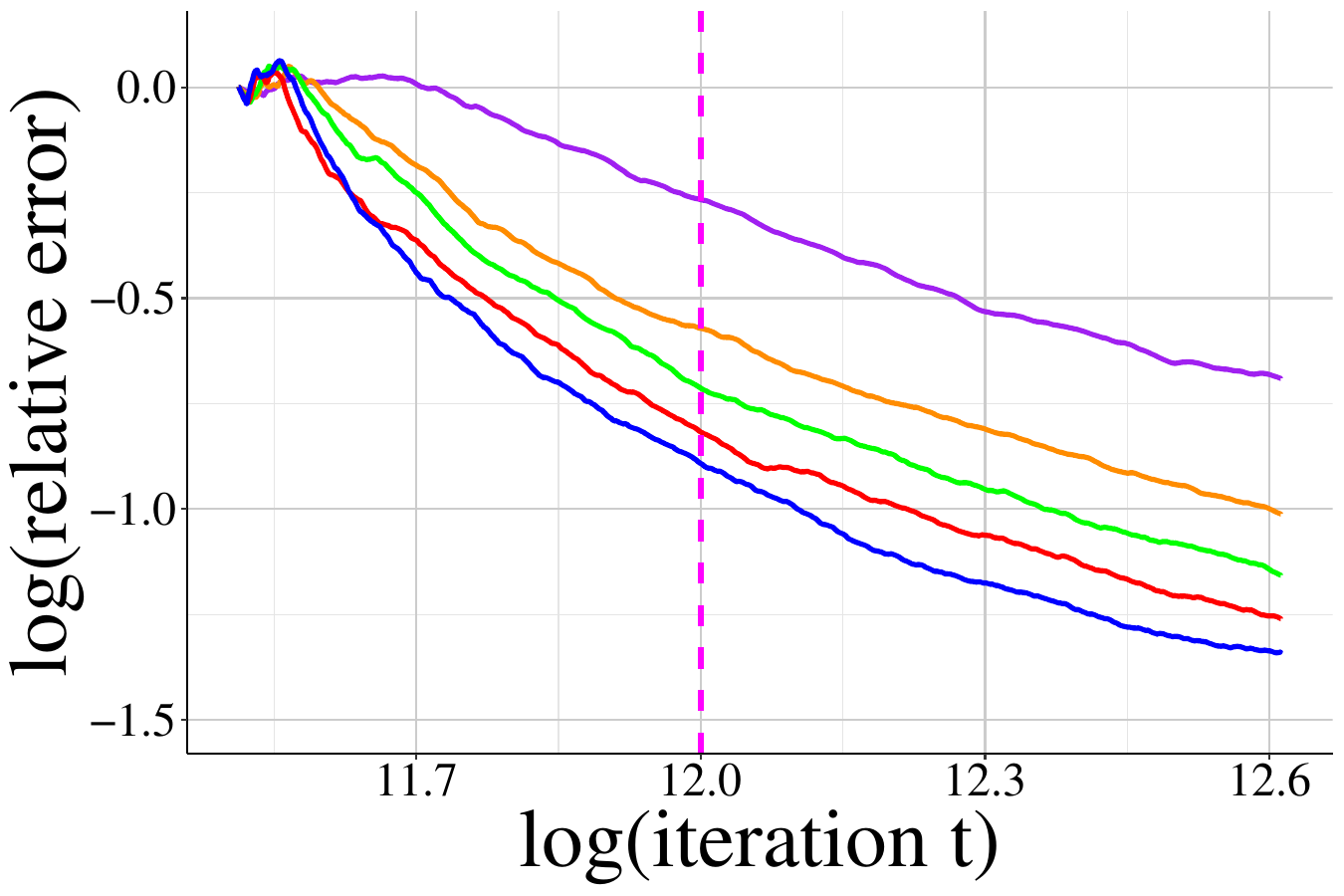}}
\vskip2pt
\centering{Linear regression}

\subfigure[Toeplitz $\Sigma_a$ with $r=0.5$]{\label{D21}\includegraphics[width=0.40\textwidth]{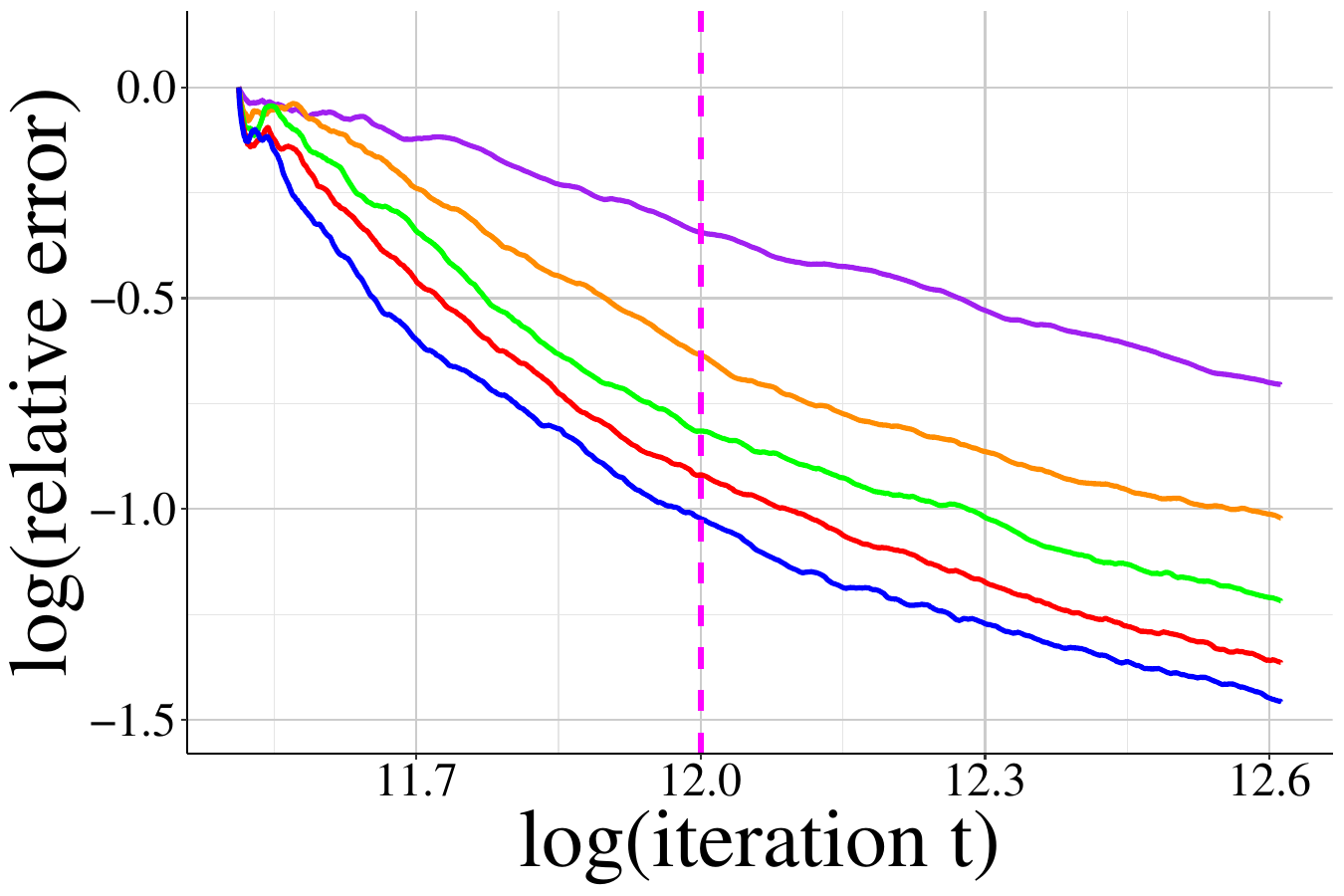}}
\subfigure[Equi-corr $\Sigma_a$ with $r=0.2$]{\label{D22}\includegraphics[width=0.40\textwidth]{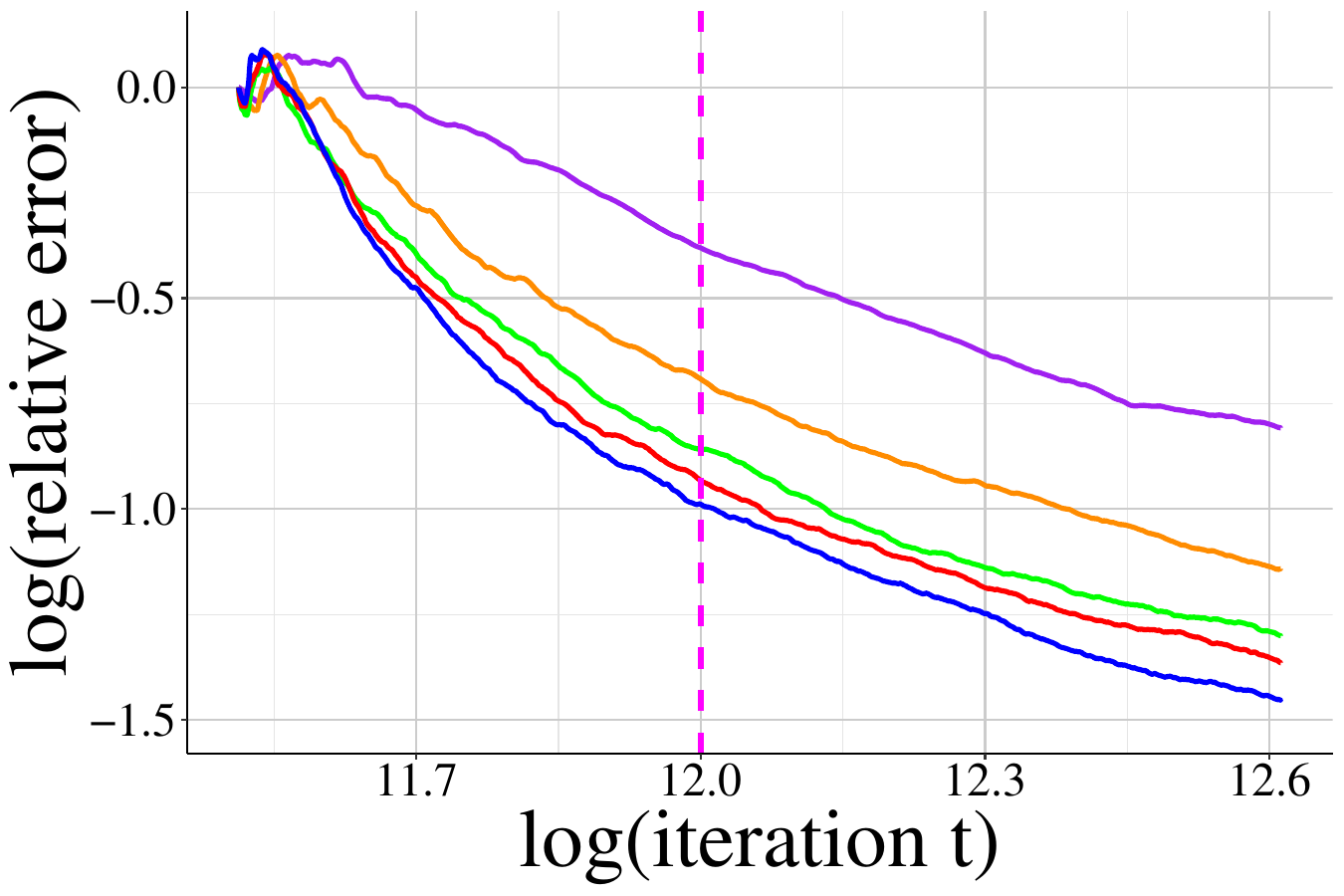}}
\vskip2pt
\centering{Logistic regression}
\vskip2pt
\includegraphics[width=0.6\textwidth]{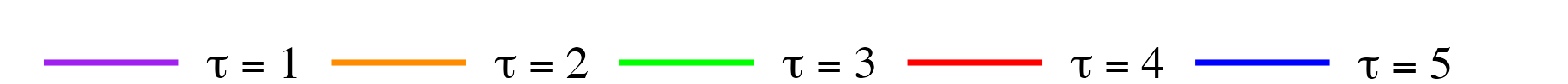}
\caption{\textit{The averaged trajectories for regression problems with $d=5$ under Toeplitz and Equi-correlation covariance matrices $\Sigma_a$. The top row corresponds to linear regression, while the bottom row corresponds to logistic regression. Panels (a) and (c) consider Toeplitz $\Sigma_a$ with $r=0.5$, and panels (b) and (d) consider Equi-correlation $\Sigma_a$ with $r=0.2$. The log relative covariance estimation error, $\log(\|\hat{\Xi}_t - \Xi^\star\|/\|\Xi^\star\|)$, is plotted against $\log t$ for the sketched Newton method with varying sketching iterations $\tau\in\{1,2,3,4,5\}$.}}
\label{fig:4}	
\end{figure}

In this section, we investigate how the sketching iteration number $\tau$ affects the convergence rate of the limiting covariance estimator $\hat{\Xi}_t$. 
We consider both linear regression and logistic regression problems and follow the experimental setups described in Sections \ref{sec5:linear} and \ref{sec5:logistic}, respectively. Two design covariance matrices are investigated: Toeplitz covariance $\Sigma_a$ with $r = 0.5$ and Equi-correlation covariance $\Sigma_a$ with $r = 0.2$. We fix the dimension to $d = 5$ and vary the sketching iteration number $\tau \in \{1,2,3,4,5\}$. All results are averaged over $200$ independent runs.

Figure \ref{fig:4} shows the averaged trajectories of the relative covariance estimation error. From Theorem \ref{sec4:thm1}, when $\beta = 0.505$ and $\chi = 2$, we have $\log (\mE[\|\hat{\Xi}_t-\Xi^\star\|]) \lesssim -1.5(1-\rho^\tau) - \frac{1-0.505}{2} \log t$. This bound indicates that the sketching iteration number $\tau$ affects the convergence rate through the constant factor, with larger values of $\tau$ leading to faster convergence. Across all four subfigures in Figure \ref{fig:4}, we observe that when $t$ is sufficiently large (e.g., after the vertical dashed line), the trajectories become approximately linear in $\log t$, and the ``intercept" decreases monotonically as $\tau$ increases from $1$ to $5$. This behavior is consistent with the theoretical ``intercept" $-1.5(1-\rho^\tau)$ predicted by Theorem \ref{sec4:thm1}. Note that the slopes also match those in Figures \ref{A13} and \ref{B13}. Moreover, the intercept difference between neighboring trajectories diminishes as $\tau$ increases, which aligns with the fact that the incremental change in $\rho^\tau$ becomes smaller for larger values of $\tau$. Overall, the empirical results in Figure \ref{fig:4} are in close agreement with our theoretical analysis.

\subsection{CUTEst benchmark problems}\label{sec5:subsec2}

In this section, we explore the empirical performance of $\hat{\Xi}_t$ in constrained optimization, as discussed in Section \ref{sec:4.3}. We perform four equality-constrained problems from the CUTEst test set: \texttt{MARATOS}, \texttt{HS7}, \texttt{BT9}, \texttt{HS39} \citep{Gould2014CUTEst}. 
For each problem and at each iteration, the CUTEst package provides true evaluations of the objective gradients and Hessians. With those quantities, we generate our estimates by letting  $\bar{g}_t\sim \mathcal{N}(\nabla F_t, \sigma^2(I+\boldsymbol{1}\boldsymbol{1}^T))$ and $[\bar{H}_t]_{i,j} =[\bar{H}_t]_{j,i} \sim \mathcal{N}([\nabla^2 F_t]_{i,j}, \sigma^2)$. We vary the sampling variance $\sigma^2$ from $\sigma^2\in\{10^{-4}, 10^{-2}, 10^{-1}, 1\}$ and set $\tau = 40$. The other parameters are set as in Section \ref{sec5:linear}, while the problem initialization is provided by the CUTEst package. The true solution $\tx$ is computed using the IPOPT solver \citep{Waechter2005implementation}. We construct 95\% confidence intervals for the averaged \textit{inactive} parameters $\sum_{i\in\I}\tx_i/|\I|$, where $\I\subseteq\{1,\ldots,d\}$ contains all the indices for which $\tx_i$ is not specified by the constraint (otherwise, $\tx_i$ has no randomness).

We evaluate the performance of $\hat{\Xi}_t$ on four CUTEst problems and summarize the results in Table \ref{table:3}. The table records the empirical coverage rate of the confidence intervals based on $\hat{\Xi}_t$ and the averaged relative estimation error of the variance of $\sum_{i\in\I}\tx_i/|\I|$.
From Table \ref{table:3}, we observe that variance estimation errors remain small across all settings. For \texttt{BT9} and \texttt{HS39}, the relative error in variance increases as the sampling variance $\sigma^2$ grows. This is expected since higher noise levels make the problem more challenging. In contrast, for \texttt{MARATOS} and \texttt{HS7}, the relative error in variance remains at the same magnitude across different values of $\sigma^2$, indicating that these problems are less sensitive to noise. Regarding statistical inference, we note that the coverage rates consistently center around the target 95\% confidence level. These results demonstrate the effectiveness of $\hat{\Xi}_t$ in constrained optimization and its robustness to varying noise levels.

\begin{table}[!t]
\centering
\resizebox{\linewidth}{!}{
\begin{tabular}{|c|cc|cc|cc|cc|}
\hline
\multirow{2}{*}{\diagbox{Prob}{$\sigma^2$}} &  \multicolumn{2}{c|}{$\sigma^2=10^{-4}$} &  \multicolumn{2}{c|}{$\sigma^2 = 10^{-2}$} &  \multicolumn{2}{c|}{$\sigma^2 = 10^{-1}$} &  \multicolumn{2}{c|}{$\sigma=1$} \\ 
& Cov (\%) & Var Err & Cov (\%) & Var Err & Cov (\%) & Var Err & Cov (\%) & Var Err \\ \hline
\texttt{MARATOS} & 97.50    & -0.0025  & 93.00    & -0.0079  & 92.50    & 0.0057   & 95.50    & 0.0124   \\ \hline
\texttt{HS7}     & 96.50    & -0.0053  & 96.50    & -0.0042  & 96.00    & -0.0021  & 94.50    & -0.0020  \\ \hline
\texttt{BT9}     & 94.50    & 0.0007   & 96.00    & 0.0067   & 94.00    & 0.0104   & 95.50    & 0.1668   \\ \hline
\texttt{HS39}    & 95.50    & -0.0030  & 94.50    & 0.0083   & 94.00    & 0.0192   & 96.50    & 0.1770   \\ \hline
\end{tabular}}
\caption{\textit{The empirical coverage rate of 95\% confidence intervals $(\textit{Cov})$ for four CUTEst problems under different sampling variance $\sigma^2\in \{10^{-4}, 10^{-2}, 10^{-1}, 1\}$; as well as the averaged relative estimation error of the variance $(\textit{Var Err})$ of $\sum_{i\in\I}\tx_i/|\I|$, given by $\sum_{i,j\in\I}([\hat{\Xi}_t]_{i,j}-[\Xi^\star]_{i,j})/\sum_{i,j\in\I}[\Xi^\star]_{i,j}$.}}
\label{table:3}
\end{table}

\section{Conclusion and Future Work}\label{sec:6}

In this paper, we designed a limiting covariance matrix estimator for sketched stochastic Newton methods. Our estimator is fully online and constructed entirely from the Newton iterates. We established the consistency and convergence rate of the estimator. Compared to plug-in estimators for second-order methods, our estimator is asymptotically consistent and more computationally efficient, requiring no matrix factorization. Compared to batch-means estimators for first-order methods, our estimator is batch-free and exhibits faster convergence. Based on our study, we can then construct asymptotically valid confidence intervals/regions for the model parameters using sketched Newton methods. We also discussed the generalization of our estimator to constrained stochastic problems. Extensive experiments on regression problems demonstrate the superior performance of our estimator.

For future research, it would be of interest to explore the lower bound for the online covariance estimation problem, as it provides insights into the statistical efficiency of our weighted sample covariance. Additionally, constructing different test statistics with different asymptotic distributions based on (sketched) Newton iterates could be promising. In particular, although the normality achieved by Newton methods is asymptotically minimax optimal \citep{Na2025Statistical}, recent studies have observed that confidence intervals constructed using other test statistics, such as $t$-statistics \citep{Zhu2024High} and their variants \citep{Lee2022Fast, Luo2022Covariance, Chen2024Online}, may exhibit better coverage rates in some problems due to the absence of further covariance estimation. A recent work \cite{Du2025Online} established asymptotic normality for the average of sketched Newton iterates, with a different limiting covariance matrix to $\Xi^\star$ in \eqref{equ:lyap}. Whether one can estimate that covariance using either batch-free or batch-means approaches, and demonstrate the advantages of leveraging Hessian information in statistical inference by establishing a faster convergence rate than those in \cite{Zhu2021Online, Chen2020Statistical}, remains largely open.
Lastly, performing inference based on Newton methods in non-asymptotic and high-dimensional settings, where the problem dimension grows with the sample size, would also be an interesting direction.

\section*{Acknowledgements}
This material was based upon work supported by the U.S. Department of Energy, Office of Science, Office of Advanced Scientific Computing Research (ASCR) under Contract DE-AC02-06CH11357.

\bibliographystyle{my-plainnat}
\bibliography{ref}

\appendix
\numberwithin{equation}{section}
\numberwithin{theorem}{section}

\mathtoolsset{showonlyrefs=true}

\section{Online Update of $\hat{\Xi}_t^{-1}$}\label{appendix:1}

We introduce how to online update $\hat{\Xi}_t^{-1}$ for constructing the confidence region in Corollary \ref{sec4:cor1}. By the definition of $\hat{\Xi}_t$ in \eqref{exp:Xihat}, we have
\begin{align*}
\hat{\Xi}_{t+1} & \stackrel{\mathclap{\eqref{exp:Xihat}}}{=} \frac{1}{t+1}\sum_{i=1}^{t+1}\frac{1}{\varphi_{i-1}}(\bx_i-\bar{\bx}_{t+1})(\bx_i-\bar{\bx}_{t+1})^T\\
& = \frac{1}{t+1}\sum_{i=1}^{t}\frac{1}{\varphi_{i-1}}(\bx_i-\bar{\bx}_{t+1})(\bx_i-\bar{\bx}_{t+1})^T + \frac{1/\varphi_t}{t+1}(\bx_{t+1}-\barx_{t+1})(\bx_{t+1}-\barx_{t+1})^T\\
& = \frac{t}{t+1}\bigg(\frac{1}{t}\sum_{i=1}^{t}\frac{1}{\varphi_{i-1}}(\bx_i-\bar{\bx}_{t})(\bx_i-\bar{\bx}_{t})^T + \frac{1}{t}\sum_{i=1}^{t}\frac{1}{\varphi_{i-1}}(\bx_i - \barx_{t})(\barx_{t} - \barx_{t+1})^T \\
&\quad\quad + \frac{1}{t}\sum_{i=1}^{t}\frac{1}{\varphi_{i-1}}(\barx_{t} - \barx_{t+1})(\bx_i - \barx_{t})^T + \frac{1}{t}\sum_{i=1}^{t}\frac{1}{\varphi_{i-1}}(\barx_{t} - \barx_{t+1})(\barx_{t} - \barx_{t+1})^T \bigg)\\
& \quad\quad + \frac{1/\varphi_{t}}{t+1}(\bx_{t+1}-\barx_{t+1})(\bx_{t+1}-\barx_{t+1})^T\\
& \stackrel{\mathclap{\eqref{nequ:9}}}{=} \frac{t}{t+1}\bigg(\hat{\Xi}_{t} + (\bv_{t} - a_{t}\barx_{t})(\barx_{t} - \barx_{t+1})^T + (\barx_{t} - \barx_{t+1})(\bv_{t} - a_{t}\barx_{t})^T + a_{t}(\barx_{t} - \barx_{t+1})(\barx_{t} - \barx_{t+1})^T\bigg) \\
&\quad\quad + \frac{1/\varphi_{t}}{t+1}(\bx_{t+1}-\barx_{t+1})(\bx_{t+1}-\barx_{t+1})^T.
\end{align*}
Let us define two matrices $R_t\in \mR^{d\times 3}$ and $\Lambda_t\in \mR^{3\times 3}$ as
\begin{equation*}
R_t = \rbr{\bv_t - a_t\barx_t; \barx_t - \barx_{t+1}; \bx_{t+1} - \barx_{t+1}}, \quad\quad\quad\quad \Lambda_t = \begin{pmatrix}
0 & 1 & 0\\
1 & a_t & 0\\
0 & 0 & 1/(t\varphi_{t})
\end{pmatrix}.
\end{equation*}
Then, we have
\begin{equation*}
\hat{\Xi}_{t+1} = \frac{t}{t+1}\rbr{\hat{\Xi}_{t} + R_t\Lambda_t R_t^T}. 
\end{equation*}
Thus, by Sherman–Morrison–Woodbury formula, we obtain
\begin{equation*}
\hat{\Xi}_{t+1}^{-1} = \frac{t+1}{t}\hat{\Xi}_{t}^{-1} - \frac{t+1}{t}\hat{\Xi}_{t}^{-1} R_t\rbr{\Lambda_t^{-1} + R_t^T\hat{\Xi}_{t}^{-1}R_t}^{-1} R_t^T\hat{\Xi}_{t}^{-1}.
\end{equation*}

\section{Preparation Lemmas}\label{appen:A1}

We introduce some preparation lemmas regarding the stepsize sequences and the update direction.

\begin{lemma}[\cite{Na2025Statistical}, Lemma B.1]\label{aux:lem3}
Suppose $\{\varphi_i\}_i$ is a positive sequence that satisfies $\lim\limits_{i\rightarrow\infty}i(1 - \varphi_{i-1}/\varphi_i) = \varphi$. Then, for any $p \geq 0$, we have $\lim\limits_{i\rightarrow\infty}i\rbr{1 - \varphi_{i-1}^p/\varphi_i^p} = p\cdot \varphi$.
\end{lemma}

\begin{lemma}[\cite{Na2025Statistical}, Lemma B.3(a)]\label{aux:lem1}
Let $\{\phi_i\}_i,\{\varphi_i\}_i,\{\sigma_i\}_i$ be three positive sequences. Suppose\footnote{In fact, $\phi<0$ is only required by Lemma B.3(b) in \cite{Na2025Statistical}, and the statements in Lemma B.3(a) hold for any constant $\phi$.}
\begin{equation}\label{appen:A1:cond1}
\lim\limits_{i\rightarrow \infty} i\rbr{1 - \phi_{i-1}/\phi_i} = \phi<0,\quad \quad\quad\lim\limits_{i\rightarrow\infty}\varphi_i = 0, \quad\quad\quad \lim\limits_{i\rightarrow \infty} i\varphi_i = \tvarphi
\end{equation}
for a constant $\phi$ and a (possibly infinite) constant $\tvarphi \in(0, \infty]$. For any $l\geq 1$, if we further have 
\begin{equation}\label{appen:A1:cond2}
\sum_{k=1}^{l}\sigma_k + \phi/\tvarphi>0,
\end{equation}
then the following results hold as $t\rightarrow \infty$:
\begin{align}
& \frac{1}{\phi_t}\sum_{i=0}^{t}\prod_{j = i+1}^t\prod_{k=1}^{l}(1-\varphi_j\sigma_k)\varphi_i\phi_i \longrightarrow \frac{1}{\sum_{k=1}^{l}\sigma_k + \phi/\tvarphi},\label{appen:A1:equ1a}\\
& \frac{1}{\phi_t}\bigg\{\sum_{i=0}^{t}\prod_{j = i+1}^t\prod_{k=1}^{l}(1-\varphi_j\sigma_k)\varphi_i\phi_i a_i + b\cdot \prod_{j = 0}^t\prod_{k=1}^{l}(1-\varphi_j\sigma_k) \bigg\}  \longrightarrow 0,\label{appen:A1:equ1b}
\end{align}
where the second result holds for any constant $b$ and any sequence $\{a_t\}_t$ such that $a_t\rightarrow 0$.
\end{lemma}

\begin{lemma}\label{aux:lem2}
Suppose $\{\phi_i, \sigma_i\}_i$ are two positive sequences, and $\{\phi_i\}_{i}$ satisfies $\lim_{i\rightarrow \infty} i(1 - \phi_{i-1}/\phi_i) = \phi<0$ for a constant $\phi$. Let $\varphi_i = c_{\varphi}/(i+1)^\varphi + o(1/(i+1)^{\varphi})$ for constants $c_{\varphi}>0$ and $\varphi\in (0,1)$. For any $l \geq 1$, we have
\begin{equation*}
\Big|\frac{1}{\phi_t}\sum_{i=0}^{t}\prod_{j = i+1}^t\prod_{k=1}^{l}(1-\varphi_j\sigma_k)\varphi_i\phi_i - \frac{1}{\sum_{k=1}^{l}\sigma_k}\Big| \lesssim 
\begin{dcases}
\varphi_t,  & \varphi \in(0,0.5),\\[2pt]
\rbr{0.5-\frac{\phi/c_{\varphi}^2}{(\sum_{k=1}^l\sigma_k)^2}}\varphi_t, & \varphi =0.5,\\[2pt]
-\frac{\phi}{(\sum_{k=1}^l\sigma_k)^2} \cdot \frac{1}{t\varphi_t},  & \varphi \in(0.5,1).
\end{dcases}
\end{equation*}
\end{lemma}

\begin{lemma}\label{aux:lem5}
Suppose $\{\phi_i\}_i,\{\varphi_i\}_i,\{\sigma_i\}_i$ be three positive sequences that satisfy the assumptions in Lemma \ref{aux:lem1}. Let $\{\eta_i\}_i$ be a positive sequence such that $\lim_{i\rightarrow\infty}\eta_i/\varphi_i =1$. For any $l\geq 1$, if $\sum_{k=1}^l\sigma_k/2 + \phi/\tilde{\varphi}>0$, then we have
\begin{equation*}
\prod_{i=0}^t \prod_{k=1}^l |1-\eta_i\sigma_k|  + \sum_{i=0}^t \prod_{j=i+1}^t \prod_{k=1}^l |1-\eta_j \sigma_k| \varphi_i \phi_i \lesssim \frac{1}{\sum_{k=1}^l\sigma_k/2 + \phi/\tilde{\varphi}}\cdot \phi_t.
\end{equation*}
\end{lemma}

\begin{lemma}\label{aux:lem4}
For the $t$-th iteration, let us define two sketching matrices
\begin{equation}\label{appen:A1:equ6}
\tilde{C}_{t,j} = I-\big(B_tS_{t,j}(S_{t,j}^TB_t^2S_{t,j})^{\dagger}S_{t,j}^TB_t\big)\quad\quad \text{ and }\quad\quad \tilde{C}_t = \prod_{j=1}^\tau \tilde{C}_{t,j},
\end{equation}
and we also let $C_t =\mE[\tilde{C}_t\mid \mF_{t-1}]$. Then, under Assumptions \ref{ass:2} and \ref{ass:4}, the following results hold (recall that $C^\star$ is defined in \eqref{exp:Cstar}):
\begin{enumerate}
\item We have $\bar{\Delta}\bx_t=(I-\tilde{C}_t)\Delta\bx_t=-(I-\tilde{C}_t)B_t^{-1}\bar{g}_t$ for any $ t\geq 0$.
\item We have $\mE[\bar{\Delta}\bx_t\mid\mF_{t-1}]=-(I-C_t)B_t^{-1}\nabla F_t$ for any $t\geq 0$.
\item We have $\|C_t\|\leq \rho^\tau$ for any $t\geq 0$ with $\rho=1-\gamma_S$. When $\bx_t\rightarrow\tx$, we also have $\|C^\star\|\leq\rho^\tau$.
\item When $\bx_t\rightarrow\tx$, we have $(1-\rho^\tau)I\preceq I-C^\star \preceq I$.
\end{enumerate}
\end{lemma}

\section{Proofs of Preparation Lemmas}

\subsection{Proof of Lemma \ref{aux:lem2}}

We note that \eqref{appen:A1:cond1} is satisfied with $\tilde{\varphi}=\infty$ and \eqref{appen:A1:cond2} holds as $\sum_{k=1}^l\sigma_k+\phi/\tilde{\varphi}=\sum_{k=1}^l\sigma_k>0$. Thus, Lemma \ref{aux:lem1} holds and its proof \cite[(C.1)]{Na2025Statistical} suggests the following decomposition 
\begin{multline}\label{appen:A1:equ2}
\frac{1}{\phi_t} \sum_{i=0}^t \prod_{j=i+1}^t \prod_{k=1}^l\left(1-\varphi_j \sigma_k\right) \varphi_i \phi_i-\frac{1}{\sum_{k=1}^l\sigma_k}=\frac{1}{\phi_t} \prod_{j=1}^t \prod_{k=1}^l (1-\varphi_j\sigma_k)\cdot \phi_0\bigg(\varphi_0-\frac{1}{\sum_{k=1}^{l}\sigma_k}\bigg)\\
+\frac{1}{\phi_t} \sum_{i=1}^t\prod_{j=i+1}^t \prod_{k=1}^l (1-\varphi_j\sigma_k)\phi_i  \bigg\{ \varphi_i-\frac{1}{\sum_{k=1}^{l}\sigma_k}\Big(1-\frac{\phi_{i-1}}{\phi_i}\prod_{k=1}^l(1-\varphi_i\sigma_k)\Big)\bigg\}=: \uppercase\expandafter{\romannumeral1} + \uppercase\expandafter{\romannumeral2}.
\end{multline}
We first calculate the rate of the term in the curly bracket in $\uppercase\expandafter{\romannumeral2}$. We note that
\begin{align*}
\prod_{k=1}^l(1-\varphi_i\sigma_k) & = 1-\sum_{k=1}^l\sigma_k\varphi_i + 0.5\Big\{\big(\sum_{k=1}^l \sigma_k\big)^2-\big(\sum_{k=1}^l \sigma_k^2\big)\Big\}\varphi_i^2 + o(\varphi_i^2),\\
\frac{\phi_{i-1}}{\phi_i} & = 1-\phi\cdot\frac{1}{i+1}+o\Big(\frac{1}{i+1}\Big).
\end{align*}
Thus, the rate of the multiplication of these two terms is
\begin{equation}\label{pequ:1}
{\footnotesize\frac{\phi_{i-1}}{\phi_i}\prod_{k=1}^l(1-\varphi_i\sigma_k)= 
\begin{dcases}
1 - \sum_{k=1}^l \sigma_k\varphi_i + 0.5\Big\{\big(\sum_{k=1}^l \sigma_k\big)^2-\big(\sum_{k=1}^l \sigma_k^2\big)\Big\}\varphi_i^2 + o(\varphi_i^2), &\varphi\in (0,0.5),\\
1 - \sum_{k=1}^l \sigma_k\varphi_i + \Big\{0.5\{\big(\sum_{k=1}^l \sigma_k\big)^2-\big(\sum_{k=1}^l \sigma_k^2\big)\}-\frac{\phi}{c_{\varphi}^2}\Big\}\varphi_i^2 + o(\varphi_i^2), &\varphi = 0.5,\\
1 -  \sum_{k=1}^l \sigma_k\varphi_i - \frac{\phi}{i+1} + o\Big(\frac{1}{i+1}\Big), &\varphi\in(0.5,1).
\end{dcases}}
\end{equation}
Let us first consider the case $\varphi\in(0.5,1)$. We plug the above display into $\uppercase\expandafter{\romannumeral2}$ in \eqref{appen:A1:equ2} and get
\begin{equation*}
\uppercase\expandafter{\romannumeral2} 
=-\frac{\phi}{\sum_{k=1}^{l}\sigma_k}\cdot \frac{1}{\phi_t} \sum_{i=1}^t\prod_{j=i+1}^t \prod_{k=1}^l (1-\varphi_j\sigma_k)\varphi_i\cdot\cbr{\frac{\phi_i}{(i+1)\varphi_i} + o\Big(\frac{\phi_i}{(i+1)\varphi_i}\Big)}.
\end{equation*}
We note that
\begin{equation*}
\lim\limits_{i\rightarrow \infty}i\bigg(1-\frac{\phi_{i-1}/\big(i\varphi_{i-1}\big)}{\phi_{i}/\big((i+1)\varphi_{i}\big)}\bigg) = \lim\limits_{i\rightarrow \infty}i\bigg(1-\frac{\phi_{i-1}}{\phi_{i}}+\frac{\phi_{i-1}}{\phi_{i}}\Big(1-\frac{1/\big(i\varphi_{i-1}\big)}{1/\big((i+1)\varphi_{i}\big)}\Big)\bigg) = \phi+\varphi-1<0,
\end{equation*}
so we can apply Lemma \ref{aux:lem1} and derive 
\begin{align*}
&\lim_{t\rightarrow \infty}\frac{1}{\phi_{t}/\big((t+1)\varphi_{t}\big)} \sum_{i=1}^t\prod_{j=i+1}^t \prod_{k=1}^l (1-\varphi_j\sigma_k)\varphi_i \cdot\frac{\phi_i}{(i+1)\varphi_i}\stackrel{\eqref{appen:A1:equ1a}}{=} \frac{1}{\sum_{k=1}^l \sigma_k},\\
&\lim_{t\rightarrow \infty}\frac{1}{\phi_{t}/\big((t+1)\varphi_{t}\big)} \sum_{i=1}^t\prod_{j=i+1}^t \prod_{k=1}^l (1-\varphi_j\sigma_k)\varphi_i \cdot o\Big(\frac{\phi_i}{(i+1)\varphi_i}\Big)\stackrel{\eqref{appen:A1:equ1b}}{=} 0.
\end{align*}
Combining the two displays, we have $|\uppercase\expandafter{\romannumeral2}| \lesssim -\frac{\phi}{(\sum_{k=1}^l \sigma_k)^2}\cdot\frac{1}{(t+1)\varphi_t}$. For the term $\uppercase\expandafter{\romannumeral1}$ in \eqref{appen:A1:equ2}, we have
\begin{equation*}
\lim_{t\rightarrow\infty}\frac{1}{\phi_{t}/\big((t+1)\varphi_{t}\big)} \prod_{j=1}^t \prod_{k=1}^l (1-\varphi_j\sigma_k)\cdot \phi_0\bigg(\varphi_0-\frac{1}{\sum_{k=1}^{l}\sigma_k}\bigg)\stackrel{\eqref{appen:A1:equ1b}}{=} 0.
\end{equation*}
This indicates $|\uppercase\expandafter{\romannumeral1}|=o(1/(t+1)\varphi_t)$. Combining the rates of $|\uppercase\expandafter{\romannumeral1}|$ and $|\uppercase\expandafter{\romannumeral2}|$ with \eqref{appen:A1:equ2}, we complete the proof for the case $\varphi\in(0.5,1)$. For the case $\varphi = 0.5$, we know from \eqref{pequ:1} and \eqref{appen:A1:equ2} that
\begin{equation*}
\uppercase\expandafter{\romannumeral2} = \frac{0.5\{\big(\sum_{k=1}^l \sigma_k\big)^2-\big(\sum_{k=1}^l \sigma_k^2\big)\}-\phi/c_{\varphi}^2}{\sum_{k=1}^{l}\sigma_k}\frac{1}{\phi_t} \sum_{i=1}^t\prod_{j=i+1}^t \prod_{k=1}^l (1-\varphi_j\sigma_k)\varphi_i\phi_i\cdot\cbr{\varphi_i+o(\varphi_i)}.
\end{equation*}
Following the same analysis as above and applying Lemma \ref{aux:lem1}, we obtain
\begin{equation*}
|\uppercase\expandafter{\romannumeral2}| \lesssim \frac{0.5\{\big(\sum_{k=1}^l \sigma_k\big)^2-\big(\sum_{k=1}^l \sigma_k^2\big)\}-\phi/c_{\varphi}^2}{(\sum_{k=1}^{l}\sigma_k)^2} \varphi_t\leq \rbr{0.5 - \frac{\phi/c_{\varphi}^2}{(\sum_{k=1}^l\sigma_k)^2}} \varphi_t.
\end{equation*}
We also have $|\uppercase\expandafter{\romannumeral1}| = o(\varphi_t)$ and, hence, complete the proof for the case $\varphi=0.5$. The proof for the case $\varphi\in(0,0.5)$ can be done similarly by noting that
\begin{equation*}
\uppercase\expandafter{\romannumeral2} = \frac{0.5\{\big(\sum_{k=1}^l \sigma_k\big)^2-\big(\sum_{k=1}^l \sigma_k^2\big)\}}{\sum_{k=1}^{l}\sigma_k}\frac{1}{\phi_t} \sum_{i=1}^t\prod_{j=i+1}^t \prod_{k=1}^l (1-\varphi_j\sigma_k)\varphi_i\phi_i\cdot\cbr{\varphi_i+o(\varphi_i)}.
\end{equation*}
We complete the proof.

\subsection{Proof of Lemma \ref{aux:lem5}}

Since $\lim_{t \rightarrow \infty}\eta_t/\varphi_t = 1$ and $\lim_{t \rightarrow \infty}\varphi_t = 0$, there exists a fixed integer $\tilde{t}$ such that for any $t\geq \tilde{t}$ and $1\leq k\leq l$, we have $\eta_t \geq \varphi_t/2$ and $0<1-\eta_t\sigma_k\leq 1-\varphi_t\sigma_k/2$. Define a sequence $\{\tilde{\phi}_t\}_{t=\tilde{t}-1}^{\infty}$ as follows:
\begin{equation*}
\tilde{\phi}_t = \begin{dcases}
\phi_{t} + \sum_{i=0}^{\tilde{t}-2}\prod_{j=i+1}^{\tilde{t}-1}\prod_{k=1}^l |1-\eta_j\sigma_k|\varphi_i \phi_i, &t = \tilde{t}-1,\\
\phi_t, &t\geq \tilde{t}.
\end{dcases}
\end{equation*}
With the above sequence, we use the techniques in \citep[(E.19)]{Na2025Statistical} and rewrite the following series as
\begin{align*}
&\sum_{i=0}^t \prod_{j=i+1}^t \prod_{k=1}^l |1-\eta_j \sigma_k| \varphi_i \phi_i
= \sum_{i = \tilde{t}-1}^t \prod_{j=i+1}^t \prod_{k=1}^l |1-\eta_j \sigma_k| \varphi_i \phi_i + \sum_{i = 0}^{\tilde{t}-2} \prod_{j=i+1}^t \prod_{k=1}^l |1-\eta_j \sigma_k| \varphi_i \phi_i\\
&=\sum_{i = \tilde{t}-1}^t \prod_{j=i+1}^t \prod_{k=1}^l |1-\eta_j \sigma_k| \varphi_i \phi_i + \prod_{j=\tilde{t}}^t \prod_{k=1}^l |1-\eta_j \sigma_k|\cdot \sum_{i = 0}^{\tilde{t}-2} \prod_{j=i+1}^{\tilde{t}-1} \prod_{k=1}^l |1-\eta_j \sigma_k| \varphi_i \phi_i\\
& = \sum_{i = \tilde{t}-1}^t \prod_{j=i+1}^t \prod_{k=1}^l |1-\eta_j \sigma_k| \varphi_i \tilde{\phi_i} \leq \sum_{i = \tilde{t}-1}^t \prod_{j=i+1}^t \prod_{k=1}^l (1-\varphi_j\sigma_k/2) \varphi_i \tilde{\phi_i}\\ 
&\stackrel{\mathclap{\eqref{appen:A1:equ1a}}}{\lesssim} \frac{1}{\sum_{k=1}^l\sigma_k/2 + \phi/\tilde{\varphi}}\cdot \phi_t.
\end{align*}
Additionally, we know
\begin{multline*}
\prod_{i=0}^t \prod_{k=1}^l |1-\eta_i\sigma_k| = \prod_{i=\tilde{t}}^t \prod_{k=1}^l |1-\eta_i\sigma_k| \cdot \prod_{i = 0}^{\tilde{t}-1}\prod_{k=1}^l |1-\eta_i\sigma_k|\\
\leq \prod_{i=\tilde{t}}^t \prod_{k=1}^l (1-\varphi_i\sigma_k/2) \cdot \prod_{i = 0}^{\tilde{t}-1}\prod_{k=1}^l |1-\eta_i\sigma_k| \stackrel{\eqref{appen:A1:equ1b}}{=} o(\phi_t).
\end{multline*}
We complete the proof.

\subsection{Proof of Lemma \ref{aux:lem4}}

Recalling $B_t\Delta\bx_t = -\bar{g}_t$, we subtract $\Delta\bx_t$ from both sides of \eqref{sec2:equ2} and obtain
\begin{equation*}
\Delta\bx_{t,j+1}-\Delta\bx_t =\big(I-B_tS_{t,j}(S_{t,j}^TB_t^2S_{t,j})^{\dagger}S_{t,j}^TB_t\big)(\Delta\bx_{t,j}-\Delta\bx_t) = \tilde{C}_{t,j}(\Delta\bx_{t,j}-\Delta\bx_t).
\end{equation*}
Since $\Delta\bx_{t,0}=\boldsymbol{0}$, we complete the proof of (a). By the independence between sketching and sampling and the unbiasedness of $\barg_t$ in Assumption \ref{ass:2}, we complete the proof of (b). (c) can be found in Lemma~4.4~and Corollary 5.4 in \cite{Na2025Statistical}. (d) is an immediate result from (c) by observing that $C^\star\succeq \boldsymbol{0}$.

\section{Proofs of Section \ref{sec3:subsec2}}\label{pf:sec3:subsec2}

Our proofs are adapted from \cite[Theorems 4.8 and 5.6]{Na2025Statistical} by restricting attention to the unconstrained strongly convex setting and refining some assumptions. In particular, we explicitly show how the proof of \cite[Theorem 4.8]{Na2025Statistical} is adapted to our proof of Theorem \ref{sec3:thm1} so that the results hold under a relaxed growth condition \eqref{ass:2:4th} on the gradient noise and a weaker condition on the parameter $\chi$ (from $\chi > 1$ in \cite[(4.4)]{Na2025Statistical}~to $\chi > 0.5(\beta+1)$). The proof of Theorem 4.2 follows from similar simplifications and modifications, and is therefore omitted for conciseness.
To ease the presentation, we assume throughout the proof and without loss of generality that all upper bound constants in the assumptions $\Upsilon_L,\Upsilon_{S}, \Upsilon_H, C_{g,1}, C_{g,2}, C_{H,1}, C_{H,2} \geq 1$, and the lower bound constant $0<\gamma_H\leq 1$. The range of these constants is not crucial to the analysis;~all results still hold by replacing $\gamma_H$ by $\gamma_H\wedge 1$ (similar for other constants).

\subsection{Proof of Theorem \ref{sec3:thm1}}\label{pf:sec3:thm1}

By Assumption \ref{ass:3}, $\|\nabla^2 F(\bx)\|\leq \Upsilon_H$. Applying Taylor expansion, we have
\begin{align}\label{appen:A2:equ8}
&\hskip-0.5cm  F_{t+1}-F^\star \nonumber\\
&\hskip-0.5cm \leq F_t-F^\star + \bar{\alpha}_t \nabla F_t^T\bar{\Delta} \bx_t + \frac{\Upsilon_{H}}{2}\bar{\alpha}_t^2\|\bar{\Delta}\bx_t\|^2 \nonumber\\
&\hskip-0.5cm = F_t-F^{\star}+\bar{\alpha}_t\mE[\nabla F_t^T\bar{\Delta} \bx_t\mid\mF_{t-1}]+\bar{\alpha}_t\Big(\nabla F_t^T\bar{\Delta}\bx_t-\mE[\nabla F_t^T\bar{\Delta} \bx_t\mid\mF_{t-1}]\Big)+\frac{\Upsilon_H}{2}\bar{\alpha}_t^2\|\bar{\Delta}\bx_t\|^2.
\end{align}
Then, we take expectation on both sides conditioning on $\mF_{t-1}$ and obtain
\begin{multline}\label{appen:A2:equ7}
\mE[F_{t+1}-F^\star\mid\mF_{t-1}] \leq F_t-F^\star
+ \mE\Big[\bar{\alpha}_t\mE[\nabla F_t^T\bar{\Delta} \bx_t\mid\mF_{t-1}]\mid \mF_{t-1}\Big]\\
+\mE\Big[\bar{\alpha}_t\Big\{\nabla F_t^T\bar{\Delta}\bx_t-\mE[\nabla F_t^T\bar{\Delta} \bx_t\mid\mF_{t-1}]\Big\}\mid\mF_{t-1}\Big]+\frac{\Upsilon_H}{2}\mE\big[\bar{\alpha}_t^2\|\bar{\Delta}\bx_t\|^2\mid\mF_{t-1}\big].
\end{multline}
For the second term on the right hand side, we apply Assumption \ref{ass:3}, Lemma \ref{aux:lem4}(b, c), and have
\begin{align}\label{appen:A2:equ4}
\mE\Big[\bar{\alpha}_t\mE[\nabla F_t^T\bar{\Delta} \bx_t\mid\mF_{t-1}]\mid \mF_{t-1}\Big]\; &=\; -\nabla F_t^T(I-C_t)B_t^{-1}\nabla F_t \cdot \mE\Big[\bar{\alpha}_t\mid \mF_{t-1}\Big]\nonumber\\
&\leq\Big(- \frac{1}{\Upsilon_H}\|\nabla F_t\|^2 + \frac{\rho^{\tau}}{\gamma_H}\|\nabla F_t\|^2\Big)\cdot \mE\Big[\bar{\alpha}_t\mid \mF_{t-1}\Big]\nonumber\\
&\leq -\frac{3}{4\Upsilon_H}\beta_t\|\nabla F_t\|^2\quad(\text{by }\rho^\tau\leq \gamma_H/4\Upsilon_H\text{ and }\beta_t\leq \bar{\alpha}_t).
\end{align}
For the third term in \eqref{appen:A2:equ7}, we note $\mE\big[\nabla F_t^T\bar{\Delta}\bx_t - \mE[\nabla F_t^T\bar{\Delta}\bx_t\mid\mF_{t-1}]\mid\mF_{t-1}\big]=0$. Thus, we have (recall $\varphi_t=\beta_t+\chi_t/2$)
\begin{multline}\label{appen:A2:equ1}
\mE\Big[\bar{\alpha}_t\Big\{\nabla F_t^T\bar{\Delta}\bx_t-\mE[\nabla F_t^T\bar{\Delta} \bx_t\mid\mF_{t-1}]\Big\}\mid\mF_{t-1}\Big]= \mE\big[(\bar{\alpha}_t-\varphi_t)\nabla F_t^T\big(\bar{\Delta}\bx_t-\mE[\bar{\Delta} \bx_t\mid\mF_{t-1}]\big)\mid\mF_{t-1}\big]\\
\leq \frac{\chi_t}{2}\|\nabla F_t\|\mE\Big[\big\|\bar{\Delta}\bx_t - \mE[\bar{\Delta}\bx_t\mid\mF_{t-1}]\big\|\mid\mF_{t-1}\Big]\quad (\text{by } |\bar{\alpha}_t-\varphi_t|\leq \chi_t/2).
\end{multline}
By Lemma \ref{aux:lem4}(a, b, c), we obtain
\begin{align*}
\big\|\bar{\Delta}\bx_t - \mE[\bar{\Delta}\bx_t\mid\mF_{t-1}]\big\| & = \big\|(I-\tilde{C}_t)B_t^{-1}\bar{g}_t - (I-C_t)B_t^{-1}\nabla F_t\big\|\\
& \leq \|C_t-\tilde{C}_t\|\|B_t^{-1}\|\|\nabla F_t\|+\|I-\tilde{C}_t\|\|B_t^{-1}\|\|\bar{g}_t-\nabla F_t\|\\
& \stackrel{\mathclap{\eqref{ass:3:c2}}}{\leq}\; \frac{{ 1+\rho^\tau}}{\gamma_H}\|\nabla F_t\|+\frac{2}{\gamma_H}\|\bar{g}_t-\nabla F_t\| \leq \frac{2}{\gamma_H}\|\nabla F_t\|+\frac{2}{\gamma_H}\|\bar{g}_t-\nabla F_t\|.
\end{align*}
Here, the third inequality also uses the bounds $\|C_t\|\le \rho^\tau$ (cf. Lemma \ref{aux:lem4}(c)) and $\|\tilde{C}_t\|\le 1$. In the constant factor $(1+\rho^\tau)/\gamma_H$, the term $1$ always dominates, and thus we can upper bound this factor by $2/\gamma_H$. This bounding argument is used repeatedly throughout the paper, and we omit further explanation. Applying Assumption \ref{ass:2}, we obtain
\begin{align}\label{appen:A2:equ2}
\mE\Big[\big\|\bar{\Delta}\bx_t - \mE[\bar{\Delta}\bx_t\mid\mF_{t-1}]\big\|\mid\mF_{t-1}\Big] &\leq \frac{2}{\gamma_H}\|\nabla F_t\| + \frac{2C_{g,1}^{1/4}}{\gamma_H}\|\bx_t-\bx^\star\|+\frac{2C_{g,2}^{1/4}}{\gamma_H}\nonumber\\
&\leq \frac{4C_{g,1}^{1/4}}{\gamma_H^2}\|\nabla F_t\| +\frac{2C_{g,2}^{1/4}}{\gamma_H} \quad (\text{by } C_{g,1}\geq 1),
\end{align}
where the last inequality also uses the property of strong convexity of $F(\bx)$ \citep{Nesterov2018Lectures} 
\begin{equation}\label{appen:A2:equ3}
\frac{\gamma_H}{2}\left\|\bx_t-\bx^{\star}\right\|^2 \leq F_t-F^{\star} \leq \frac{1}{2 \gamma_H}\left\|\nabla F_t\right\|^2.
\end{equation}
Combining \eqref{appen:A2:equ1} and \eqref{appen:A2:equ2}, we get
\begin{align}\label{appen:A2:equ5}
&\mE\Big[\bar{\alpha}_t\Big\{\nabla F_t^T\bar{\Delta}\bx_t-\mE[\nabla F_t^T\bar{\Delta} \bx_t\mid\mF_{t-1}]\Big\}\mid\mF_{t-1}\Big] \leq \frac{2C_{g,1}^{1/4}}{\gamma_H^2}\chi_t\|\nabla F_t\|^2 + \frac{C_{g,2}^{1/4}}{\gamma_H}\chi_t\|\nabla F_t\|\nonumber\\
&\leq \frac{2C_{g,1}^{1/4}}{\gamma_H^2}\chi_t\|\nabla F_t\|^2 +\frac{1}{4\Upsilon_H}\beta_t\|\nabla F_t\|^2 + \frac{\Upsilon_H C_{g,2}^{1/2}}{\gamma_H^2}\cdot \frac{\chi_t^2}{\beta_t}\quad   \text{(by Young's inequality)}.
\end{align}
Let $\eta_t=\beta_t+\chi_t$. We apply Lemma \ref{aux:lem4}(a) and bound the last term in \eqref{appen:A2:equ7} by
 \begin{align}\label{appen:A2:equ6}
\mE\big[\bar{\alpha}_t^2\|\bar{\Delta} \bx_t\|^2\mid \mF_{t-1}\big]&\leq \mE\big[\bar{\alpha}_t^2\|(I+\tC_t)\|^2\|B_t^{-1}\|^2\|\bar{g}_t\|^2\mid \mathcal{F}_{t-1}\big]\nonumber\\
&\leq \frac{8}{\gamma_H^2}\eta_t^2\Big(\|\nabla F_t\|^2 + \mE\big[\|\bar{g}_t-\nabla F_t\|^2\mid \mathcal{F}_{t-1}\big]\Big)\nonumber\\ 
&\leq \frac{16C_{g,1}^{1/2}}{\gamma_H^4}\eta_t^2\|\nabla F_t\|^2 + \frac{8C_{g,2}^{1/2}}{\gamma_H^2}\eta_t^2\quad \text{(by Assumption \ref{ass:2})}.
\end{align}
We plug \eqref{appen:A2:equ4}, \eqref{appen:A2:equ5}, and \eqref{appen:A2:equ6} into \eqref{appen:A2:equ7}, and obtain
\begin{multline*}
\mE[F_{t+1}-F^\star\mid\mF_{t-1}] \\ 
\leq F_t-F^\star - \Big(\frac{1}{2\Upsilon_H}\beta_t - \frac{2C_{g,1}^{1/4}}{\gamma_H^2}\chi_t-\frac{8\Upsilon_H C_{g,1}^{1/2}}{\gamma_H^4}\eta_t^2\Big)\|\nabla F_t\|^2 + \frac{4\Upsilon_H C_{g, 2}^{1/2}}{\gamma_H^2}\rbr{\frac{\chi_t^2}{\beta_t}+\eta_t^2}.
\end{multline*}
Since $\beta_t = c_{\beta}/(t+1)^{\beta}$ with $\beta\in(0.5, 1]$ and $\chi_t = c_{\chi}/(t+1)^{\chi}$ with $\chi>0.5(\beta+1)\geq \beta$, there exists a fixed integer $t_0$ such that $\frac{2C_{g,1}^{1/4}}{\gamma_H^2}\chi_t+\frac{8\Upsilon_H C_{g,1}^{1/2}}{\gamma_H^4}\eta_t^2\leq\frac{1}{4\Upsilon_H}\beta_t$ for all $t\geq t_0$. Thus, for $t\geq t_0$, we have
\begin{equation*}
\mE[F_{t+1}-F^\star\mid\mF_{t-1}]\leq 
F_t-F^\star - \frac{1}{4\Upsilon_H}\beta_t\|\nabla F_t\|^2 + \frac{4\Upsilon_H C_{g, 2}^{1/2}}{\gamma_H^2}\rbr{\frac{\chi_t^2}{\beta_t}+\eta_t^2}.
\end{equation*}
Note that $\sum_{t=t_0}^\infty \chi_t^2/\beta_t<\infty$ and $\sum_{t=t_0}^\infty \eta_t^2\lesssim \sum_{t=t_0}^\infty \beta_t^2 + \sum_{t=t_0}^\infty \chi_t^2<\infty$. Thus, we apply the Robbins-Siegmund Theorem \cite[Theorem 1.3.12]{Duflo2013Random} and conclude that $F_t-F^\star$ converges to a finite random variable, and $\sum_{t=t_0}^{\infty}\beta_t\left\|\nabla F_t\right\|^2 < \infty$ almost surely. Furthermore, we have $\liminf_{t\rightarrow\infty}\|\nabla F_t\|=0$ due to $\sum_{t=t_0}^\infty\beta_t = \infty$, which leads to $\liminf_{t\rightarrow\infty}(F_t-F^\star) =0$ according to \eqref{appen:A2:equ3}. Since $F_t-F^\star$ converges almost surely, the conclusion can be strengthened to $\lim_{t\rightarrow\infty}F_t-F^\star =0$. Again, we apply \eqref{appen:A2:equ3} and obtain $\lim_{t\rightarrow\infty}\bx_t = \bx^\star$ almost surely. This completes the proof.

\subsection{Proof of Theorem \ref{sec3:thm2}}\label{pf:sec3:thm2}

The proof of asymptotic normality is almost identical to the proof of Theorem 5.6 in \cite{Na2025Statistical}. Since $\chi>1.5\beta\Rightarrow \chi>0.5(\beta+1)$, we have $\bx_t\rightarrow \bx^\star$ almost surely, as proved in Theorem \ref{sec3:thm1}. Therefore, we only~have to note that our growth conditions in Assumptions \ref{ass:2} and \ref{ass:3} on gradients and Hessians do not affect the~proof of normality (though they affect the proof of convergence), since the term $\|\bx_t-\bx^\star\|$ in the growth conditions converges to $0$ almost surely.

\section{Proofs of Section \ref{sec4:subsec1}}

To clear up tedious constants, we assume $\eta_t=\beta_t+\chi_t\leq 1$, $\forall t\geq 0$, without loss of generality for the remainder of this paper. Note that this condition is non-essential, since $\eta_t \rightarrow 0$ and the condition will~always hold for sufficiently large, fixed threshold of $t$.

\subsection{Proof of Lemma \ref{sec4:lem1}}\label{pf:xtbound}

We separate the proof into two parts. 
\vskip4pt
\noindent \textbf{Part 1: Bound of $\mE[\|\bx_t-\bx^\star\|^4]$.} We take square on both sides of \eqref{appen:A2:equ8} and take expectation conditioning on $\mathcal{F}_{t-1}$, then we get
\begin{align}\label{appen:A3:equ1}
& \mE\big[(F_{t+1}-F^\star)^2\mid \mathcal{F}_{t-1}\big] \leq (F_t-F^\star)^2+\mE\big[2 \bar{\alpha}_t(F_t-F^\star) \nabla F_t^T \bar{\Delta} \bx_t\mid \mathcal{F}_{t-1}\big] + \mE\big[\bar{\alpha}_t^2 \Upsilon_{H}(F_t-F^\star)\|\bar{\Delta} \bx_t\|^2\mid \mathcal{F}_{t-1}\big] \nonumber \\
& +\mE\big[\bar{\alpha}_t^2(\nabla F_t^{T} \bar{\Delta} \bx_t)^2\mid \mathcal{F}_{t-1}\big] + \mE\big[\bar{\alpha}_t^3 \Upsilon_{H} \nabla F_t^{T} \bar{\Delta}\bx_t\|\bar{\Delta} \bx_t\|^2\mid \mathcal{F}_{t-1}\big] +\frac{1}{4}\mE\big[\bar{\alpha}_t^4 \Upsilon_{H}^2\|\bar{\Delta} \bx_t\|^4\mid \mathcal{F}_{t-1}\big].
\end{align}
We rearrange these terms by the order of $\bar{\alpha}_t$ and analyze them one by one.
\vskip4pt
\noindent $\bullet$ \textbf{Term 1:} $\mE[2 \bar{\alpha}_t(F_t-F^\star) \nabla F_t^T \bar{\Delta} \bx_t\mid \mF_{t-1}]$.

\noindent This term can be decomposed as
\begin{align*}
\mE[2 \bar{\alpha}_t(F_t-F^\star) \nabla F_t^T \bar{\Delta} \bx_t\mid \mF_{t-1}] & = 2(F_t-F^\star)\mE\Big[\bar{\alpha}_t\mE[\nabla F_t^T\bar{\Delta} \bx_t\mid\mF_{t-1}]\mid \mF_{t-1}\Big]\\
& \quad +2(F_t-F^\star)\mE\Big[\bar{\alpha}_t\Big\{\nabla F_t^T\bar{\Delta}\bx_t-\mE[\nabla F_t^T\bar{\Delta} \bx_t\mid\mF_{t-1}]\Big\}\mid\mF_{t-1}\Big].
\end{align*}
For the first term on the right hand side, by \eqref{appen:A2:equ4} and \eqref{appen:A2:equ3}, we have
\begin{equation}\label{appen:A3:equ3}
2(F_t-F^\star)\mE\Big[\bar{\alpha}_t\mE[\nabla F_t^T\bar{\Delta} \bx_t\mid\mF_{t-1}]\mid \mF_{t-1}\Big]\leq -\frac{3}{2\Upsilon_H}\beta_t(F_t-F^\star)\|\nabla F_t\|^2\leq -\frac{3\gamma_H}{\Upsilon_H}\beta_t(F_t-F^{\star})^2.
 \end{equation}
For the second term on the right hand side, \eqref{appen:A2:equ1} and \eqref{appen:A2:equ2} give us
\begin{align}\label{appen:A3:equ4}
& \hskip-0.6cm  2(F_t-F^\star)\mE\Big[\bar{\alpha}_t\Big\{\nabla F_t^T\bar{\Delta}\bx_t-\mE[\nabla F_t^T\bar{\Delta} \bx_t\mid\mF_{t-1}]\Big\}\mid\mF_{t-1}\Big] \nonumber\\
&\hskip-0.6cm \leq \chi_t(F_t-F^\star) \|\nabla F_t\|\Big(\frac{2}{\gamma_H}\|\nabla F_t\| +\frac{2C_{g,1}^{1/4}}{\gamma_H}\|\bx_t-\bx^\star\| + \frac{2C_{g,2}^{1/4}}{\gamma_H}\Big) \nonumber\\
&\hskip-0.6cm \leq \frac{4\Upsilon_H^{1/2}}{\gamma_H}\Big(\Upsilon_H^{1/2}+ \frac{C_{g,1}^{1/4}}{\gamma_H^{1/2}}\Big)\chi_t(F_t-F^\star)^2 + \frac{2\sqrt{2}\Upsilon_H^{1/2}C_{g,2}^{1/4}}{\gamma_H}\chi_t(F_t-F^\star)^{3/2} \nonumber\\
&\hskip-0.6cm \leq \rbr{\frac{4\Upsilon_H^{1/2}}{\gamma_H}\Big(\Upsilon_H^{1/2}+ \frac{C_{g,1}^{1/4}}{\gamma_H^{1/2}}\Big)\chi_t+ \frac{\gamma_H}{2\Upsilon_H}\beta_t}(F_t-F^\star)^2 + \frac{3^4\Upsilon_H^5C_{g,2}}{\gamma_H^7}\cdot\frac{\chi_t^4}{\beta_t^3}\quad \text{(Young's inequality)}.
\end{align}
Here, the second inequality is due to \eqref{appen:A2:equ3} and the following $\Upsilon_H$-Lipschitz continuity property of $\nabla F(\bx)$ \citep{Nesterov2018Lectures}:
\begin{equation}\label{appen:A3:equ2}
\frac{1}{2\Upsilon_{H}}\|\nabla F_t\|^2\leq F_t-F^{\star}\leq \frac{\Upsilon_{H}}{2}\|\bx_t-\bx^\star\|^2.
\end{equation}
\noindent$\bullet$ \textbf{Term 2:} $\mE[\bar{\alpha}_t^2 \Upsilon_{H}(F_t-F^\star)\|\bar{\Delta} \bx_t\|^2+\bar{\alpha}_t^2(\nabla F_t^{T} \bar{\Delta} \bx_t)^2\mid \mF_{t-1}]$.

\noindent Since $\bar{\alpha}_t\leq \eta_t$, we bound this term by
\begin{align}\label{appen:A3:equ5}
&\mE[\;\bar{\alpha}_t^2 \Upsilon_{H}(F_t - F^\star)\|\bar{\Delta} \bx_t\|^2 +\bar{\alpha}_t^2(\nabla F_t^{T} \bar{\Delta} \bx_t)^2\mid \mF_{t-1}] \leq \eta_t^2 \mE\big[\big(\Upsilon_{H}(F_t-F^\star) + \|\nabla F_t\|^2\big)\|\bar{\Delta} \bx_t\|^2\mid \mF_{t-1}\big]\nonumber\\
&\stackrel{\mathclap{\eqref{appen:A3:equ2}}}{\leq}\;\;3\Upsilon_{H}\eta_t^2\mE\big[(F_t-F^\star)\|\bar{\Delta} \bx_t\|^2\mid \mF_{t-1}\big] \leq \frac{\gamma_H}{2\Upsilon_H}(\beta_t+\chi_t) (F_t-F^\star)^2+\frac{9\Upsilon_H^3}{2\gamma_H}\eta_t^3 \mE\big[\|\bar{\Delta} \bx_t\|^4\mid \mF_{t-1}\big],
\end{align}
where the last inequality is by Young's inequality.

\vskip4pt
\noindent$\bullet$ \textbf{Term 3:} $\mE[\;\bar{\alpha}_t^3 \Upsilon_H \nabla F_t^{T} \bar{\Delta}\bx_t\|\bar{\Delta} \bx_t\|^2\mid \mF_{t-1}]$.

\noindent Similarly, we use Young's inequality, apply \eqref{appen:A3:equ2}, and have
\begin{align}\label{appen:A3:equ6}
&\hskip-0.7cm \mE[\;\bar{\alpha}_t^3 \Upsilon_{H} \nabla F_t^{T} \bar{\Delta}\bx_t\|\bar{\Delta} \bx_t\|^2\mid \mF_{t-1}]  \leq \eta_t^3\Upsilon_{H}\mE[\|\nabla F_t\| \|\bar{\Delta}\bx_t\|^3\mid \mF_{t-1}] \nonumber\\
&\hskip-0.7cm \leq \frac{1}{4}\eta_t^3\|\nabla F_t\|^4+\frac{3\Upsilon_{H}^{4/3}}{4}\eta_t^3\mE\big[\|\bar{\Delta}\bx_t\|^4\mid\mF_{t-1}\big]  \stackrel{\mathclap{\eqref{appen:A3:equ2}}}{\leq}\; \Upsilon_{H}^2\eta_t^3(F_t-F^\star)^2+\frac{3\Upsilon_{H}^{4/3}}{4}\eta_t^3\mE\big[\|\bar{\Delta}\bx_t\|^4\mid\mF_{t-1}\big].
\end{align}
Substituting \eqref{appen:A3:equ3}, \eqref{appen:A3:equ4}, \eqref{appen:A3:equ5}, and \eqref{appen:A3:equ6} into \eqref{appen:A3:equ1}, we obtain
\begin{multline*}
\mE[(F_{t+1}-F^\star)^2\mid\mF_{t-1}]\leq \Big(1-
\frac{2\gamma_H}{\Upsilon_H}\beta_t+\frac{\gamma_H}{2\Upsilon_H}\chi_t+\frac{4\Upsilon_H^{1/2}}{\gamma_H}\Big(\Upsilon_H^{1/2}+ \frac{C_{g,1}^{1/4}}{\gamma_H^{1/2}}\Big)\chi_t+\Upsilon_{H}^2\eta_t^3\Big)(F_t-F^\star)^2\\
+\frac{3^4\Upsilon_H^5C_{g,2}}{\gamma_H^7}\cdot\frac{\chi_t^4}{\beta_t^3} +\frac{6\Upsilon_H^3}{\gamma_H} \eta_t^3\mE\big[\|\bar{\Delta} \bx_t\|^4\mid \mF_{t-1}\big]\quad(\text{by }\eta_t\leq 1).
\end{multline*}
Following the analysis in \eqref{appen:A2:equ6}, we apply Assumption \ref{ass:2} and have
\begin{align*}
\mE\big[\|\bar{\Delta} \bx_t\|^4\mid \mF_{t-1}\big] & \leq \frac{2^4}{\gamma_H^4}\mE[\|\barg_t\|^4 \mid \mF_{t-1}] \leq \frac{2^7}{\gamma_H^4}\rbr{\mE[\|\barg_t - \nabla F_t\|^4 \mid \mF_{t-1}] + \|\nabla F_t\|^4}\\
& \stackrel{\mathclap{\eqref{appen:A3:equ2}}}{\leq}\; \frac{2^9\Upsilon_{H}^2}{\gamma_H^4}(F_t-F^\star)^2 + \frac{2^7C_{g,1}}{\gamma_H^4}\|\bx_t-\tx\|^4 + \frac{2^7C_{g,2}}{\gamma_H^4}\\
& \stackrel{\mathclap{\eqref{appen:A2:equ3}}}{\leq}\; \frac{2^9\Upsilon_{H}^2}{\gamma_H^4}(F_t-F^\star)^2 + \frac{2^9C_{g,1}}{\gamma_H^6}(F_t-F^\star)^2 + \frac{2^7C_{g,2}}{\gamma_H^4} \\
& = \frac{2^{9}(\Upsilon_H^2 + C_{g,1}/\gamma_H^2)}{\gamma_H^4}(F_t-F^\star)^2 + \frac{2^7C_{g,2}}{\gamma_H^4}.
\end{align*}
Combining the above two displays and taking full expectation, we obtain the recursion:
\begin{align*}
&\mE[(F_{t+1}-F^\star)^2] \\
& \leq \rbr{1-
\frac{2\gamma_H}{\Upsilon_H}\beta_t+\frac{\gamma_H}{2\Upsilon_H}\chi_t+ \frac{4\Upsilon_H^{1/2}}{\gamma_H}\Big(\Upsilon_H^{1/2}+ \frac{C_{g,1}^{1/4}}{\gamma_H^{1/2}}\Big)\chi_t + \frac{2^{12}\Upsilon_{H}^3(\Upsilon_{H}^2+C_{g,1}/\gamma_H^2)}{\gamma_H^5}\eta_t^3}\mE[(F_{t}-F^\star)^2]\\
& \quad +\frac{3^4\Upsilon_H^5C_{g,2}}{\gamma_H^7}\frac{\chi_t^4}{\beta_t^3}  + \frac{2^{10}\Upsilon_H^3C_{g,2}}{\gamma_H^5} \eta_t^3.
\end{align*}
We apply the above inequality recursively until $(F_0 - F^\star)^2$ and then apply Lemma \ref{aux:lem5} to compute~the rate of $\mE[(F_t-F^\star)^2]$. We first verify the assumptions. Since $\chi_t = o(\beta_t)$ by $\chi>\beta$, we know
\begin{equation*}
\lim_{i\rightarrow\infty}\frac{\beta_i - \frac{\Upsilon_H}{2\gamma_H}\rbr{\frac{\gamma_H}{2\Upsilon_H}\chi_i+\frac{4\Upsilon_H^{1/2}}{\gamma_H}\Big(\Upsilon_H^{1/2}+ \frac{C_{g,1}^{1/4}}{\gamma_H^{1/2}}\Big)\chi_i+ \frac{2^{12}\Upsilon_{H}^3(\Upsilon_{H}^2+C_{g,1}/\gamma_H^2)}{\gamma_H^5}\eta_i^3}}{\beta_i} = 1.
\end{equation*}
Since $\beta\in(0,1)$, we have $\lim_{i\rightarrow\infty}i\beta_i=\infty$ and \eqref{appen:A1:cond2} holds naturally. Furthermore, since $\lim_{i\rightarrow\infty}i(1-\beta_{i-1}/\beta_i)=-\beta$ and $\lim_{i\rightarrow\infty}i(1-\chi_{i-1}/\chi_i)=-\chi$, we obtain from Lemma \ref{aux:lem3} that $\lim_{i\rightarrow\infty}i(1-\beta_{i-1}^4/\beta_i^4)=-4\beta$ and $\lim_{i\rightarrow\infty}i(1-\chi_{i-1}^4/\chi_i^4)=-4\chi$. Thus, we have
\begin{equation*}
\begin{aligned}
&\lim_{i \rightarrow\infty} i\rbr{1-\frac{\chi_{i-1}^4/\beta_{i-1}^4}{\chi_i^4/\beta_i^4}}  = \lim_{i \rightarrow\infty} i\rbr{1-\frac{\chi_{i-1}^4}{\chi_{i}^4} + \frac{\chi_{i-1}^4}{\chi_{i}^4}\cbr{1-\frac{1/\beta_{i-1}^4}{1/\beta_i^4}}} = 4(\beta-\chi)<0, \\
&\lim_{i \rightarrow\infty} i\rbr{1-\frac{\eta_{i-1}^3/\beta_{i-1}}{\eta_i^3/\beta_i}}  = \lim_{i \rightarrow\infty} i\rbr{1-\frac{\eta_{i-1}^3}{\eta_{i}^3} + \frac{\eta_{i-1}^3}{\eta_{i}^3}\cbr{1-\frac{1/\beta_{i-1}}{1/\beta_i}}} = -2\beta<0.
\end{aligned}	
\end{equation*}
This suggests that \eqref{appen:A1:cond1} also holds. Now, we apply Lemma \ref{aux:lem5} and obtain
\begin{align}\label{xtrate}
\mE\big[\|\bx_t-\bx^\star\|^4\big]&\stackrel{\mathclap{\eqref{appen:A2:equ3}}}{\lesssim}\;\; \frac{1}{\gamma_H^2}\mE\big[(F_t-F^\star)^2\big]
\lesssim \frac{1}{\gamma_H^2}\cdot \frac{\Upsilon_{H}}{\gamma_H}  \rbr{\frac{\Upsilon_H^5C_{g,2}}{\gamma_H^7}\cdot\frac{\chi_t^4}{\beta_t^4}  + \frac{\Upsilon_H^3C_{g,2}}{\gamma_H^5}\cdot\frac{\eta_t^3}{\beta_t}} \nonumber\\
&\lesssim \frac{\Upsilon_H^4C_{g,2}}{\gamma_H^8} \beta_t^2 + \frac{\Upsilon_H^6C_{g,2}}{\gamma_H^{10}} \cdot\frac{\chi_t^4}{\beta_t^4}= O\rbr{\beta_t^2 + \frac{\chi_t^4}{\beta_t^4}}.
\end{align}
\noindent \textbf{Part 2: Bound of $\mE[\|B_t-B^\star\|^4]$.} By the construction of $B_t$ in \eqref{nequ:5}, we have
\begin{equation}\label{appen:A3:equ12}
\mE\big[\|B_t-B^\star\|^4\big] \lesssim\mE\bigg[\Big\|\frac{1}{t} \sum_{i=0}^{t-1}(\bar{H}_i-\nabla^2 F_i)\Big\|^4\bigg]
+\mE\bigg[\Big\|\frac{1}{t} \sum_{i=0}^{t-1}(\nabla^2 F_i-\nabla^2 F^\star)\Big\|^4\bigg].
\end{equation}
For the first term, we note that $\bar{H}_i-\nabla^2 F_i$ is a martingale difference sequence and \eqref{ass:3:4th} implies
\begin{equation*}
\mE\big[\left\|\bar{H}_i-\nabla^2 F_i\right\|_F^4\big] \lesssim \mE\big[\left\|\bar{H}_i-\nabla^2 F_i\right\|^4\big] \stackrel{\eqref{ass:3:4th}}{\lesssim} C_{H,1}\mE\big[\|\bx_t-\bx^\star\|^4\big] + C_{H,2}.
\end{equation*}
Therefore, same as in \cite[(63)]{Chen2020Statistical}, we apply \cite[Theorem 2.1]{Rio2008Moment} and obtain
\begin{align*}
\mE\bigg[\Big\|\frac{1}{t} \sum_{i=0}^{t-1}(\bar{H}_i-\nabla^2 F_i)\Big\|^4\bigg]&\leq  \mE\bigg[\Big\|\frac{1}{t} \sum_{i=0}^{t-1}(\bar{H}_i-\nabla^2 F_i)\Big\|_F^4\bigg]\lesssim \frac{1}{t^4}\bigg[\sum_{i=0}^{t-1}\Big(\mE\big[\|\bar{H}_i-\nabla^2 F_i\|_F^4\big]\Big)^{1 / 2}\bigg]^2\nonumber\\
&\lesssim \frac{1}{t^4}\Big(\sum_{i=0}^{t-1} C_{H,1}^{1/2}(\mE\big[\|\bx_t-\bx^\star\|^4\big])^{1/2}\Big)^2 + \frac{1}{t^4}\Big(\sum_{i=0}^{t-1} C_{H,2}^{1/2}\Big)^2\nonumber\\ 
&\stackrel{\mathclap{\eqref{xtrate}}}{\lesssim}\;\; \frac{1}{t^2}\bigg(\frac{\Upsilon_H^4 C_{g,2} C_{H,1}}{\gamma_H^8} \Big(\frac{1}{t}\sum_{i=0}^{t-1} \beta_i\Big)^2 + \frac{\Upsilon_H^6 C_{g,2} C_{H,1}}{\gamma_H^{10}}\Big(\frac{1}{t} \sum_{i=0}^{t-1} \frac{\chi_i^2}{\beta_i^2}\Big)^2 \bigg) + \frac{C_{H,2}}{t^2}.
\end{align*}
We only consider the case where $\chi\leq 1.5\beta$, otherwise $\chi_t^2/\beta_t^2 = o(\beta_t)$ and all $\chi_t^2/\beta_t^2$ terms in~the~following can be absorbed into $\beta_t$. We note that
\begin{equation}\label{appen:A3:equ11}
\frac{1}{t}\sum_{i=0}^{t-1}\beta_i = \frac{1}{t}\beta_0 + \sum_{i=1}^{t-1}\prod_{j=i+1}^{t-1}\Big(1-\frac{1}{j}\Big)\frac{1}{i}\beta_i\stackrel{\eqref{appen:A1:equ1a}}{\lesssim} \frac{1}{1-\beta}\beta_t\quad\text{and}\quad \frac{1}{t}\sum_{i=0}^{t-1}\frac{\chi_i^2}{\beta_i^2} \stackrel{\eqref{appen:A1:equ1a}}{\lesssim} \frac{1}{1-2(\chi-\beta)}\cdot\frac{\chi_t^2}{\beta_t^2},
\end{equation}
where we are able to apply Lemma \ref{aux:lem1} since the condition \eqref{appen:A1:cond2} is satisfied by $\beta<1$ and~$\chi\leq 1.5\beta\Rightarrow 2\chi-2\beta<1$. Thus, we combine the above two displays and have
\begin{equation}\label{appen:A3:equ13}
\footnotesize \frac{1}{t^2}\bigg(\frac{\Upsilon_H^4 C_{g,2} C_{H,1}}{\gamma_H^8} \Big(\frac{1}{t}\sum_{i=0}^{t-1} \beta_i\Big)^2 + \frac{\Upsilon_H^6 C_{g,2} C_{H,1}}{\gamma_H^{10}}\Big(\frac{1}{t} \sum_{i=0}^{t-1} \frac{\chi_i^2}{\beta_i^2}\Big)^2 \bigg) = o\rbr{\frac{1}{t^2}} \quad \text{and} \quad \mE\bigg[\Big\|\frac{1}{t} \sum_{i=0}^{t-1}(\bar{H}_i-\nabla^2 F_i)\Big\|^4\bigg]\lesssim \frac{C_{H,2}}{t^2}.
\end{equation}
For the second term on the right hand side in \eqref{appen:A3:equ12}, the $\Upsilon_L$-Lipschitz continuity of $\nabla^2 F(\bx)$ leads to
\begin{align}\label{appen:A3:equ14}
&\mE\bigg[\Big\|\frac{1}{t} \sum_{i=0}^{t-1}(\nabla^2 F_i-\nabla^2 F^\star)\Big\|^4\bigg] \leq  \mE\bigg[\Big(\frac{1}{t} \sum_{i=0}^{t-1}\|\nabla^2 F_i-\nabla^2 F^\star\|\Big)^4\bigg]
\leq\frac{\Upsilon_{L}^4}{t^4} \mE\bigg[\Big(\sum_{i=0}^{t-1}\|\bx_i-\bx^\star\|\Big)^4\bigg]\nonumber\\
&\leq \frac{\Upsilon_{L}^4}{t^4} \Big(\sum_{i=0}^{t-1}\big(\mE[\|\bx_i-\bx^\star\|^4]\big)^{1/4}\Big)^4 \quad (\text{by H\"older's inequality})\nonumber\\
&\stackrel{\mathclap{\eqref{xtrate}}}{\lesssim}\;\; \frac{\Upsilon_{L}^4\Upsilon_H^4C_{g,2}}{\gamma_H^8} \rbr{\frac{1}{t}\sum_{i=0}^{t-1}\beta_i^{1/2}}^4 + \frac{\Upsilon_{L}^4\Upsilon_H^6C_{g,2}}{\gamma_H^{10}}\rbr{\frac{1}{t}\sum_{i=0}^{t-1}\frac{\chi_i}{\beta_t}}^4\nonumber\\
&\stackrel{\mathclap{\eqref{appen:A3:equ11}}}{\lesssim}\;\; \frac{\Upsilon_{L}^4\Upsilon_H^4C_{g,2}}{\gamma_H^8} \beta_t^2 + \frac{\Upsilon_{L}^4\Upsilon_H^6C_{g,2}}{\gamma_H^{10}}\cdot \frac{\chi_t^4}{\beta_t^4}\quad (\text{by }1-\beta/2>1/2\quad\text{and}\quad 1-(\chi-\beta)>1/2).
\end{align}
Plugging \eqref{appen:A3:equ13} and \eqref{appen:A3:equ14} into \eqref{appen:A3:equ12}, we have
\begin{equation}\label{appen:A3:equ15}
\mE\big[\|B_t-B^\star\|^4\big] \lesssim \frac{\Upsilon_{L}^4\Upsilon_H^4C_{g,2}}{\gamma_H^8} \beta_t^2 + \frac{\Upsilon_{L}^4\Upsilon_H^6C_{g,2}}{\gamma_H^{10}}\cdot \frac{\chi_t^4}{\beta_t^4} = O\rbr{\beta_t^2 + \frac{\chi_t^4}{\beta_t^4}}.
\end{equation}
This completes the proof.

\subsection{Analysis of a dominant term in $\hat{\Xi}_t$}\label{appen:A4}

In this section, we focus on a dominant term of $\hat{\Xi}_t$ and show that the dominate term converges to the~limiting covariance matrix $\Xi^\star$. We first introduce a decomposition of the iterate $\bx_t$.

\begin{lemma}\label{appen:A4:lem1}\cite[Lemma 5.1]{Na2025Statistical}
The iterate sequence \eqref{nequ:7} can be decomposed as
\begin{equation}\label{appen:A4:equ1}
\bx_{t+1}-\bx^\star = \I_{1,t} + \I_{2,t} + \I_{3,t},
\end{equation}
where
\begin{align}
\I_{1,t} & = \sum_{i=0}^t\prod_{j=i+1}^{t}\cbr{I - \varphi_j(I-C^\star)}\varphi_i \btheta^i, \label{rec:a}\\
\I_{2,t} & = \sum_{i=0}^t\prod_{j=i+1}^{t} \cbr{I - \varphi_j(I-C^\star)}\rbr{\baralpha_i - \varphi_i}\bar{\Delta}\bx_i, \label{rec:b}\\
\I_{3,t} & = \prod_{i=0}^{t}\cbr{I - \varphi_i(I-C^\star)}(\bx_0 - \bx^\star) + \sum_{i=0}^t\prod_{j=i+1}^{t}\cbr{I - \varphi_j(I-C^\star)}\varphi_i\bdelta^i, 
\label{rec:c}
\end{align}
and
\begin{align}
C^\star &= (I - \mE[B^\star S(S^T(B^\star)^2S)^\dagger S^TB^\star])^\tau, \label{rec:def:a}\\ 
\btheta^i &= \bar{\Delta}\bx_i-\mE[\bar{\Delta}\bx_i\mid \mF_{i-1}] = -(I-\tilde{C}_i)B_i^{-1}\bar{g}_i + (I-C_i)B_i^{-1}\nabla F_i, \label{rec:def:b}\\ 
\bdelta^i &= -(I-C_i)\big\{(B^{\star})^{-1}\bpsi^i + \{B_i^{-1} - (B^\star)^{-1}\}\nabla F_i\big\}
+ \big(C_i-C^\star\big)(\bx_i - \bx^\star),\label{rec:def:c}\\ 
\bpsi^i &= \nabla F_i - B^\star(\bx_i-\bx^\star). \label{rec:def:d} 
\end{align}
\end{lemma}

Here, $\I_{1,t}$ includes the summation of martingale difference sequence; $\I_{2,t}$ characterizes the influence of the adaptive stepsize; and $\I_{3,t}$ encompasses all remaining errors. Based on \eqref{appen:A4:equ1}, we decompose~the following matrix as
\begin{equation}\label{appen:A4:equ20}
\frac{1}{t}\sum_{i=1}^t \frac{1}{\varphi_{i-1}}(\bx_i-\bx^\star)(\bx_i-\bx^\star)^T = \sum_{k=1}^3\sum_{l=1}^3\frac{1}{t}\sum_{i=0}^{t -1}\frac{1}{\varphi_i}\I_{k,i}\I_{l,i}.
\end{equation}
We study the dominant term $\frac{1}{t}\sum_{i=0}^{t-1}\frac{1}{\varphi_i}\I_{1,i}\I_{1,i}^T$ in this section and defer the analysis on the remaining~terms to Appendix \ref{appen:A5}. The next lemma shows consistency of the dominant term and establishes the convergence rate, with proof provided in Appendix \ref{pf:I1}.

\begin{lemma}\label{appen:A4:lem2}
Suppose Assumptions \ref{ass:1} -- \ref{ass:4} hold, the number of sketches satisfies $\tau\geq \log(\gamma_H/4\Upsilon_H)/\log \rho$ with $\rho = 1-\gamma_S$, and the stepsize parameters satisfy $\beta\in(0,1)$, $\chi> \beta$, and $c_{\beta}, c_{\chi}>0$. Then, we have
\begin{equation*}
\mE\bigg[\Big\|\frac{1}{t}\sum_{i=0}^{t-1}\frac{1}{\varphi_i}\I_{1,i}\I_{1,i}^T-\Xi^\star\Big\|\bigg]\lesssim \begin{dcases}
\frac{1}{{ 1-\rho^\tau} }\rbr{\sqrt{\beta_t} \;+\; \frac{\chi_t}{\beta_t}}, \quad & \beta\in(0,0.5),\\
\frac{1}{{ (1-\rho^\tau)^{1.5}}}\cdot \frac{1}{\sqrt{t\beta_t}} + \frac{1}{{ 1-\rho^\tau}}\cdot\frac{\chi_t}{\beta_t}, \quad & \beta\in[0.5,1),
\end{dcases}
\end{equation*}
where we explicitly track the dependency of the constant factor on $\rho = 1-\gamma_S$.
\end{lemma}

\subsubsection{Proof of Lemma \ref{appen:A4:lem2}}\label{pf:I1}

We define
\begin{equation}\label{Ctildestar}
\tilde{C}^\star_{k,j}=I-\big(B^\star S_{k,j}(S_{k,j}^T(B^\star)^2S_{k,j})^{\dagger}S_{k,j}^TB^\star\big)\quad \text{ and }\quad \tilde{C}^\star_k = \prod_{j=1}^\tau \tilde{C}^\star_{k,j},
\end{equation}
where $S_{k,j}$ is the same sketching matrix in $\tilde{C}_{k,j}$ at \eqref{appen:A1:equ6}. It is easy to verify $\mE\tilde{C}_k^\star = C^\star$ with $C^\star$ defined~in \eqref{rec:def:a}. We also define
\begin{equation}\label{appen:A4:equ3}
\tilde{\btheta}^k = -(I-\tilde{C}^\star_k)(B^{\star})^{-1}\nabla f(\bx^\star;\xi_k)\quad \text{ and }\quad \hat{\btheta}^k = \btheta^k-\tilde{\btheta}^k.
\end{equation}
Basically, $\tilde{\btheta}^k$ and $\btheta^k$ share the same randomness but $\tilde{\btheta}^k$ is constructed at $\bx^\star$ instead of $\bx_k$, which means we use the $iid$ copies $\{\tilde{\btheta}^k\}_k$ to approximate the martingale difference sequence $\{\btheta^k\}_k$. We decompose $\I_{1,i}$ as
\begin{equation}\label{appen:A4:equ2}
\I_{1,i} = \sum_{k=0}^i\prod_{l=k+1}^i\{I-\varphi_l(I-C^\star)\}\varphi_k \tilde{\btheta}^k + \sum_{k=0}^i\prod_{l=k+1}^i\{I-\varphi_l(I-C^\star)\}\varphi_k \hat{\btheta}^k =:  \tilde{\I}_{1,i} +\hat{\I}_{1,i}.
\end{equation}
Intuitively, as $\bx_i$ converges to $\bx^\star$, $\tilde{\I}_{1,i}$ should be a good approximation to $\I_{1,i}$ and $\hat{\I}_{1,i}$ should be negligible. The next two lemma provide bounds for $\tilde{\I}_{1,i}$ and $\hat{\I}_{1,i}$, respectively. The proofs are provided~in Appendices \ref{pf:tildeI1} and \ref{pf:hatI1}.

\begin{lemma}\label{appen:A4:lem3}
Under the assumptions of Lemma \ref{appen:A4:lem2}, we have
\begin{equation*}
\mE\bigg[\Big\|\frac{1}{t}\sum_{i=0}^{t-1}\frac{1}{\varphi_i}\tilde{\I}_{1,i}\tilde{\I}_{1,i}^T-\Xi^\star\Big\|\bigg] \lesssim \left\{\begin{aligned}
&\beta_t, \quad & \beta\in(0,1/3),\\
&\frac{1}{{(1-\rho^\tau)^{1.5}}}\cdot \frac{1}{\sqrt{t\beta_t}}, \quad & \beta\in [1/3,1).   
\end{aligned}\right.
\end{equation*}
\end{lemma}

\begin{lemma}\label{appen:A4:lem4}
Under the assumptions of Lemma \ref{appen:A4:lem2}, we have
\begin{equation*}
\mE\bigg[\Big\|\frac{1}{t}\sum_{i=0}^{t-1}\frac{1}{\varphi_i}\hat{\I}_{1,i}\hat{\I}_{1,i}^T\Big\|\bigg]\leq\frac{1}{t}\sum_{i=0}^{t-1} \frac{1}{\varphi_i} \mE\big[\|\hat{\I}_{1,i}\|^2\big] \lesssim \frac{1}{{1-\rho^\tau}}\rbr{\beta_t \;+\; \frac{\chi_t^2}{\beta_t^2}}.
\end{equation*}
\end{lemma}

By the decomposition \eqref{appen:A4:equ2}, we have
\begin{multline}\label{appen:A4:equ23}
\mE\bigg[\Big\|\frac{1}{t}\sum_{i=0}^{t-1}\frac{1}{\varphi_i}\I_{1,i}\I_{1,i}^T-\Xi^\star\Big\|\bigg] \leq \mE\bigg[\Big\|\frac{1}{t}\sum_{i=0}^{t-1}\frac{1}{\varphi_i}\tilde{\I}_{1,i}\tilde{\I}_{1,i}^T-\Xi^\star\Big\|\bigg]\\
+\mE\bigg[\Big\|\frac{1}{t}\sum_{i=0}^{t-1}\frac{1}{\varphi_i}\hat{\I}_{1,i}\hat{\I}_{1,i}^T\Big\|\bigg] + 2\mE\bigg[\Big\|\frac{1}{t}\sum_{i=0}^{t-1}\frac{1}{\varphi_i}\tilde{\I}_{1,i}\hat{\I}_{1,i}^T\Big\|\bigg].
\end{multline}
We apply H\"older's inequality twice to the last term and obtain
\begin{multline}\label{appen:A4:equ19}
\mE\bigg[\Big\|\frac{1}{t}\sum_{i=0}^{t-1}\frac{1}{\varphi_i}\tilde{\I}_{1,i}\hat{\I}_{1,i}^T\Big\|\bigg]
\leq \mE\bigg[\frac{1}{t}\sum_{i=0}^{t-1}\frac{1}{\varphi_i}\|\tilde{\I}_{1,i}\|\|\hat{\I}_{1,i}\|\bigg]\\
\leq \mE\Bigg[\sqrt{\frac{1}{t}\sum_{i=0}^{t-1}\frac{1}{\varphi_i}\|\tilde{\I}_{1,i}\|^2}\sqrt{\frac{1}{t}\sum_{i=0}^{t-1}\frac{1}{\varphi_i}\|\hat{\I}_{1,i}\|^2}\Bigg]\leq\sqrt{\frac{1}{t}\sum_{i=0}^{t-1}\frac{1}{\varphi_i}\mE\|\tilde{\I}_{1,i}\|^2} \sqrt{\frac{1}{t}\sum_{i=0}^{t-1}\frac{1}{\varphi_i}\mE\|\hat{\I}_{1,i}\|^2}.
\end{multline}
Given Lemmas \ref{appen:A4:lem3} and \ref{appen:A4:lem4}, it suffices to bound $\frac{1}{t}\sum_{i=0}^{t-1}\frac{1}{\varphi_i}\mE\|\tilde{\I}_{1,i}\|^2$. We have
\begin{align}\label{appen:A4:equ4}
&\mE\big[\|\tilde{\I}_{1,i}\|^2\big]\;\; \stackrel{\mathclap{\eqref{appen:A4:equ2}}}{=}\;\; \sum_{k_1,k_2=0}^i\varphi_{k_1}\varphi_{k_2}\mE\Big[(\tilde{\btheta}^{k_1})^T\Big(\prod_{l_1=k_1+1}^i\big\{I-\varphi_{l_1}(I-C^\star)\big\}\Big)^T \Big(\prod_{l_2=k_2+1}^i\big\{I-\varphi_{l_2}(I-C^\star)\big\}\Big) \tilde{\btheta}^{k_2}\Big]\nonumber\\
& = \sum_{k=0}^i\varphi_k^2 \mE\bigg[\Big\|\prod_{l=k+1}^i\big\{I-\varphi_{l}(I-C^\star)\big\}\tilde{\btheta}^k\Big\|^2\bigg] \leq \sum_{k=0}^i\varphi_k^2 \prod_{l=k+1}^i\big\|I-\varphi_{l}(I-C^\star)\big\|^2\mE\big[\|\tilde{\btheta}^k\|^2\big]\nonumber\\
&\leq \sum_{k=0}^i \prod_{l=k+1}^i\big(1-(1-\rho)^\tau\varphi_l\big)^2 \varphi_k^2 \mE\big[\|\tilde{\btheta}^k\|^2\big],
\end{align}
where the second equality uses the fact that $\{\tilde{\btheta}^k\}_k$ are mean zero and independent, and the last inequality uses the fact $\varphi_t\leq \eta_t\leq 1$ (cf. Appendix \ref{sec4:subsec1}) and Lemma \ref{aux:lem4}(d). Note that $\varphi_t\leq 1$ is not essential; given $\varphi_t\rightarrow 0$, we can apply Lemma \ref{aux:lem5} to derive the same results without this condition. Next, we bound the moment of $\|\tilde{\btheta}^k\|$. We note for $m=2,4$ and any $k\geq 0$,
\begin{align}\label{appen:A4:equ22}
&\mE\big[\|\nabla f(\bx^\star; \xi_k)\|^m\big] \lesssim \mE\big[\|\nabla f(\bx^\star;\xi_k)-\nabla f(\bx_k;\xi_k)\|^m\big] + \mE\big[\|\nabla f(\bx_k;\xi_k)-\nabla F_k\|^m\big] + \mE\big[\|\nabla F_k - \nabla F^\star\|^m\big]\nonumber\\
&\lesssim \Upsilon_H^m\mE\big[\|\bx_k-\bx^\star\|^m\big] + C_{g,1}^{m/4}\mE\big[\|\bx_k-\bx^\star\|^m\big] + C_{g,2}^{m/4} + \Upsilon_H^m\mE\big[\|\bx_k-\bx^\star\|^m\big]\;\;(\text{Assumptions \ref{ass:2}, \ref{ass:3}})\nonumber\\
&\lesssim C_{g,2}^{m/4}\quad\quad (\mE\big[\|\bx_k-\bx^\star\|^m\big]=o(1) \text{ by Lemma \ref{sec4:lem1}}).
\end{align}
Then, by \eqref{ass:3:c2} and Lemma \ref{aux:lem4}(c), we have for $m=2,4$
\begin{equation}\label{appen:A4:equ5}
\hskip-0.1cm \mE\big[\|\tilde{\btheta}^k\|^m\big] \stackrel{\eqref{appen:A4:equ3}}{\leq} \mE\big[\|I- \tilde{C}_k^\star\|^m\|(B^{\star})^{-1}\|^m\|\nabla f(\bx^\star;\xi_k)\|^m\big] \leq\frac{2^m}{\gamma_H^m}\mE\big[\|\nabla f(\bx^\star;\xi_k)\|^m\big] \stackrel{\eqref{appen:A4:equ22}}{\lesssim} \frac{C_{g,2}^{m/4}}{\gamma_H^m}.
\end{equation}
Plugging \eqref{appen:A4:equ5}  into \eqref{appen:A4:equ4}, we get
\begin{multline}\label{appen:A4:equ17}
\frac{1}{t}\sum_{i=0}^{t-1} \frac{1}{\varphi_i} \mE\big[\|\tilde{\I}_{1,i}\|^2\big] \leq \frac{1}{t}\sum_{i=0}^{t-1} \frac{1}{\varphi_i}\sum_{k=0}^i \prod_{l=k+1}^i (1-(1-\rho^{\tau}) \varphi_l)^2\varphi_k^2 \mE\big[\|\tilde{\btheta}^k\|^2\big]\\
\lesssim \frac{C_{g,2}^{1/2}}{\gamma_H^2}\cdot\frac{1}{t}\sum_{i=0}^{t-1}\underbrace{\frac{1}{\varphi_i}\sum_{k=0}^{i}\prod_{l=k+1}^i(1-(1-\rho^{\tau})\varphi_l)^2 \varphi_k^2}_{\longrightarrow 0.5/(1-\rho^\tau)\quad\text{by Lemma \ref{aux:lem1}}}\lesssim \frac{C_{g,2}^{1/2}}{\gamma_H^2(1-\rho^\tau)},
\end{multline}
where the last inequality uses the fact that $\lim_{t\rightarrow\infty}\sum_{i=0}^{t-1} a_i/t = a \text{ if } \lim_{t\rightarrow\infty}a_t=a$. Combining \eqref{appen:A4:equ19}, \eqref{appen:A4:equ17}, and Lemma \ref{appen:A4:lem4} (particularly \eqref{appen:A4:equ21} in the proof), we derive
\begin{equation*}
\mE\bigg[\Big\|\frac{1}{t}\sum_{i=0}^{t-1}\frac{1}{\varphi_i}\tilde{\I}_{1,i}\hat{\I}_{1,i}^T\Big\|\bigg] \lesssim \frac{C_{g,2}^{1/4}C_{\hat{\btheta}}^{1/2}}{\gamma_H(1-\rho^\tau)} \rbr{\frac{1}{(1-\beta)^{1/2}}\sqrt{\beta_t} + \frac{\Upsilon_H^{1/2}\mathbf{1}_{\{\chi\leq 1.5\beta\}}}{\gamma_H^{1/2}(1-2(\chi-\beta))^{1/2}}\cdot\frac{\chi_t}{\beta_t}}
\end{equation*}
with a constant $C_{\hat{\btheta}}>0$ defined later in \eqref{appen:A4:equ18}. Finally, combining Lemma \ref{appen:A4:lem3} (\eqref{appen:A4:equ24} in the proof), Lemma \ref{appen:A4:lem4} (\eqref{appen:A4:equ21} in the proof), and \eqref{appen:A4:equ23}, we have
\begin{footnotesize}
\begin{multline}\label{appen:A4:equ27}
\mE\bigg[\Big\|\frac{1}{t}\sum_{i=0}^{t-1}\frac{1}{\varphi_i}\I_{1,i}\I_{1,i}^T-\Xi^\star\Big\|\bigg]\\
\lesssim \begin{dcases}
\frac{C_{g,2}^{1/4}C_{\hat{\btheta}}^{1/2}}{\gamma_H(1-\rho^\tau)}\sqrt{\beta_t} + \frac{\Upsilon_H^{1/2}C_{g,2}^{1/4} C_{\hat{\btheta}}^{1/2} \mathbf{1}_{\{\chi\leq 1.5\beta\}}}{\gamma_H^{3/2}(1-\rho^\tau)}\cdot\frac{\chi_t}{\beta_t}= O\rbr{\frac{1}{{1-\rho^\tau}}\rbr{\sqrt{\beta_t}+ \frac{\chi_t}{\beta_t}}}, &\beta\in(0,0.5),\\
\max\bigg(\frac{C_{g,2}^{1/4}C_{\hat{\btheta}}^{1/2}}{\gamma_H(1-\rho^\tau)}, \frac{\max(\|\Lambda\|_F, C_{g,2}^{1/2}/\gamma_H^2)}{c_{\beta}(1-\rho^\tau)^{3/2}}\bigg)\sqrt{\beta_t} + \frac{\Upsilon_H^{1/2}C_{g,2}^{1/4} C_{\hat{\btheta}}^{1/2} \mathbf{1}_{\{\chi\leq 1.5\beta\}}}{\gamma_H^{3/2}(1-\rho^\tau)}\cdot\frac{\chi_t}{\beta_t}=O\rbr{\frac{\sqrt{\beta_t}}{{(1-\rho^\tau)^{1.5}}} + \frac{\chi_t}{{(1-\rho^\tau)}\beta_t}} , &\beta=0.5,\\
\frac{\max(\|\Lambda\|_F, C_{g,2}^{1/2}/\gamma_H^2)}{(1-\rho^\tau)^{3/2}}\cdot \frac{1}{\sqrt{t\beta_t}} + \frac{\Upsilon_H^{1/2}C_{g,2}^{1/4} C_{\hat{\btheta}}^{1/2} \mathbf{1}_{\{\chi\leq 1.5\beta\}}}{\gamma_H^{3/2}(1-\rho^\tau)(1-2(\chi-\beta))^{1/2}}\cdot\frac{\chi_t}{\beta_t} =O\rbr{\frac{1}{{(1-\rho^\tau)^{1.5}}\sqrt{t\beta_t}}+ \frac{\chi_t}{{(1-\rho^\tau)}\beta_t}}, &\beta\in(0.5,1),
\end{dcases}
\end{multline}
\end{footnotesize}
\hskip -3.5pt where $\Lambda = \mE[(I- \tilde{C}^\star)\Omega^\star(I-\tilde{C}^\star)^T]$. This completes the proof.

\subsubsection{Proof of Lemma \ref{appen:A4:lem3}}\label{pf:tildeI1}

By the eigenvalue decomposition $I-C^\star = U\Sigma U^T$ with $\Sigma=\text{diag}(\sigma_1,\dots,\sigma_d)$ in \eqref{eigendecomp}, we have
\begin{equation}\label{appen:A4:equ6}
\tilde{\I}_{1,i} = \sum_{k=0}^i\prod_{l=k+1}^i\{I-\varphi_l(I-C^\star)\}\varphi_k \tilde{\btheta}^k =U\sum_{k=0}^i\prod_{l=k+1}^i\{I-\varphi_l\Sigma\}\varphi_k U^T\tilde{\btheta}^k.
\end{equation}   
Let $\tilde{\Q}_i = U^T\tilde{\I}_{1,i}$ and $\Gamma = U^T\Lambda U$ with $\Lambda = \mE[(I- \tilde{C}^\star)\Omega^\star(I- \tilde{C}^\star)^T]$. Recalling the expression of $\Xi^\star$ in \eqref{exp:Xistar}, we get
\begin{multline*}
\mE\bigg[\Big\|\frac{1}{t}\sum_{i=0}^{t-1}\frac{1}{\varphi_i}\tilde{\I}_{1,i}\tilde{\I}_{1,i}^T-\Xi^\star\Big\|\bigg] = 
\mE\bigg[\Big\|\frac{1}{t}\sum_{i=0}^{t-1}\frac{1}{\varphi_i}\tilde{\Q}_i\tilde{\Q}_i^T-\Theta\circ\Gamma\Big\|\bigg] \leq \mE\bigg[\Big\|\frac{1}{t}\sum_{i=0}^{t-1}\frac{1}{\varphi_i}\tilde{\Q}_i\tilde{\Q}_i^T-\Theta\circ\Gamma\Big\|_F\bigg]\\
\leq \sqrt{\mE\bigg[\Big\|\frac{1}{t}\sum_{i=0}^{t-1}\frac{1}{\varphi_i}\tilde{\Q}_i\tilde{\Q}_i^T-\Theta\circ\Gamma\Big\|_F^2\bigg]} \quad\text{(by H\"older's inequality)}.
\end{multline*}
We perform bias-variance decomposition on this term:
\begin{equation}\label{appen:A4:equ12}
\mE\bigg[\Big\|\frac{1}{t}\sum_{i=0}^{t-1}\frac{1}{\varphi_i}\tilde{\Q}_i\tilde{\Q}_i^T-\Theta\circ\Gamma\Big\|_F^2\bigg] = \sum_{p,q=1}^d \mE\bigg[\Big(\frac{1}{t} \sum_{i=0}^{t-1} \frac{1}{\varphi_i} \tilde{\Q}_{i, p} \tilde{\Q}_{i, q}-\Theta_{p,q}\Gamma_{p,q}\Big)^2\bigg] =: \uppercase\expandafter{\romannumeral1} + \uppercase\expandafter{\romannumeral2},
\end{equation}
with
\begin{align*}
\uppercase\expandafter{\romannumeral1} &= \sum_{p,q=1}^d\bigg\{\mE\bigg[\Big(\frac{1}{t} \sum_{i=0}^{t-1} \frac{1}{\varphi_i} \tilde{\Q}_{i, p} \tilde{\Q}_{i, q}\Big)^2\bigg]-\bigg(\mE\Big[\frac{1}{t} \sum_{i=0}^{t-1} \frac{1}{\varphi_i} \tilde{\Q}_{i, p} \tilde{\Q}_{i, q}\Big]\bigg)^2\bigg\} \quad (\text{variance}),\nonumber  \\
\uppercase\expandafter{\romannumeral2} &= \sum_{p,q=1}^d\bigg(\mE\Big[\frac{1}{t} \sum_{i=0}^{t-1} \frac{1}{\varphi_i} \tilde{\Q}_{i, p}\tilde{\Q}_{i, q}\Big]-\Theta_{p,q} \Gamma_{p,q}\bigg)^2\quad (\text{bias}^2),
\end{align*}
and $\tilde{\Q}_{i, p}$ and $\tilde{\Q}_{i, q}$ represent the $p$-th and $q$-th elements in $\tilde{\Q}_i$. We first look at $\uppercase\expandafter{\romannumeral2}$. By \eqref{appen:A4:equ6}, we get
\begin{multline*}
\mE\Big[\frac{1}{t} \sum_{i=0}^{t-1} \frac{1}{\varphi_i} \tilde{\Q}_{i, p}\tilde{\Q}_{i, q}\Big]= \frac{1}{t} \sum_{i=0}^{t-1} \frac{1}{\varphi_i} \sum_{k_1=0}^i \sum_{k_2=0}^i\prod_{l_1=k_1+1}^i(1-\sigma_p\varphi_{l_1}) \cdot\\
\prod_{l_2=k_2+1}^i(1-\sigma_q\varphi_{l_2}) \varphi_{k_1}\varphi_{k_2} \mE\Big[\big(U^T \tilde{\btheta}^{k_1}\tilde{\btheta}^{k_2^T} U\big)_{p,q}\Big].
\end{multline*}
Given the definition of $\tilde{\btheta}^k$ in \eqref{appen:A4:equ3} and the independence among $\{\tilde{\btheta}^k\}_k$, it is observed that
\begin{equation*}
\mE[U^T\tilde{\btheta}^{k_1} \tilde{\btheta}^{k_2^T}U] = 0\text{ for }k_1\neq k_2 \quad\quad \text{ and } \quad\quad \mE[U^T\tilde{\btheta}^{k}\tilde{\btheta}^{kT}U] = \Gamma.
\end{equation*}
Thus, combining the above two displays leads to 
\begin{equation*}
\mE\Big[\frac{1}{t} \sum_{i=0}^{t-1} \frac{1}{\varphi_i} \tilde{\Q}_{i, p}\tilde{\Q}_{i, q}\Big]= \frac{1}{t} \sum_{i=0}^{t-1}\frac{1}{\varphi_i}\sum_{k=0}^i\prod_{l=k+1}^i (1-\sigma_p\varphi_l)(1-\sigma_q\varphi_l)\varphi_k^2\Gamma_{p,q}.
\end{equation*}
We plug the above display into the term $\uppercase\expandafter{\romannumeral2}$ and apply Lemma \ref{aux:lem2} to bound it. For $\beta\in(0.5,1)$, we have
\begin{align}\label{appen:A4:equ8}
|\uppercase\expandafter{\romannumeral2}| &\leq  \sum_{p,q=1}^d\bigg(\frac{1}{t} \sum_{i=0}^{t-1} \Big|\frac{1}{\varphi_i} \sum_{k=0}^i \prod_{l=k+1 }^i\left(1-\sigma_p\varphi_l\right)\left(1-\sigma_q\varphi_l\right) \varphi_k^2-\Theta_{p, q}\Big|\bigg)^2\Gamma_{p,q}^2\nonumber\\
&\lesssim \sum_{p,q=1}^d\Big(\frac{\beta}{(\sigma_p+\sigma_q)^2}\cdot \frac{1}{t}\sum_{i=0}^{t-1}\frac{1}{i\varphi_i}\Big)^2\Gamma_{p,q}^2 \stackrel{\eqref{appen:A3:equ11}}{\lesssim} \sum_{p,q=1}^d\Big(\frac{\beta}{(\sigma_p+\sigma_q)^2}\cdot \frac{1}{1-(1-\beta)}\cdot \frac{1}{t\varphi_t}\Big)^2\Gamma_{p,q}^2\nonumber\\
&\lesssim \frac{\|\Gamma\|_F^2}{(1-\rho^\tau)^4} \cdot\frac{1}{t^2\varphi_t^2}\lesssim \frac{\|\Lambda\|_F^2}{(1-\rho^\tau)^4} \cdot\frac{1}{t^2\beta_t^2}\quad (\text{by Lemma \ref{aux:lem4}(d) and }\chi_t=o(\beta_t)).
\end{align}
Applying Lemma \ref{aux:lem2} for $\beta\in(0,0.5)$ and $\beta=0.5$, we similarly obtain
\begin{equation}\label{appen:A4:equ8a}
|\uppercase\expandafter{\romannumeral2}|\lesssim \|\Lambda\|_F^2\beta_t^2 \text{ for } \beta\in(0,0.5)\quad\quad \text{and}\quad\quad|\uppercase\expandafter{\romannumeral2}|\lesssim \Big(1+\frac{\beta/c_{\beta}^2}{2(1-\rho^\tau)^2}\Big)^2 \|\Lambda\|_F^2\beta_t^2 \text{ for }\beta = 0.5.
\end{equation}
Now we deal with the term $\uppercase\expandafter{\romannumeral1}$. By \eqref{appen:A4:equ6}, we expand $\uppercase\expandafter{\romannumeral1}$ as
\begin{multline*}
\uppercase\expandafter{\romannumeral1}= \sum_{p,q=1}^d\frac{1}{t^2} \sum_{i_1, i_2=0}^{t-1} \frac{1}{\varphi_{i_1}} \frac{1}{\varphi_{i_2}} \sum_{k_1,k_1^{\prime}=0}^{i_1} \sum_{k_2,k_2^{\prime}=0}^{i_2} \prod_{l_1=k_1+1}^{i_1}(1- \sigma_p\varphi_{l_1}) \prod_{l_1^{\prime}=k_1^{\prime}+1}^{i_1} (1-\sigma_q\varphi_{l_1^{\prime}}) \prod_{l_2=k_2+1}^{i_2}(1-\sigma_p\varphi_{l_2} ) \prod_{l_2^{\prime}=k_2^{\prime}+1}^{i_2}(1- \sigma_q\varphi_{l_2^{\prime}})\\
\varphi_{k_1}\varphi_{k_1^{\prime}}
\varphi_{k_2}\varphi_{k_2^{\prime}} \bigg\{\mE\Big[\big(U^T \tilde{\btheta}^{k_1}\tilde{\btheta}^{k_1^{\prime T}}U\big)_{p,q}\big(U^T \tilde{\btheta}^{k_2}\tilde{\btheta}^{k_2^{\prime T}}U\big)_{p,q}\Big]-\mE\Big[\big(U^T \tilde{\btheta}^{k_1}\tilde{\btheta}^{k_1^{\prime T}}U\big)_{p,q}\Big]\mE\Big[\big(U^T \tilde{\btheta}^{k_2}\tilde{\btheta}^{k_2^{\prime T}}U\big)_{p,q}\Big]\bigg\}.
\end{multline*}
It is noteworthy that the term in the curly braces is nonzero only when the indices $k_1,k_1^\prime,k_2,k_2^\prime$ are pairwise identical. Thus, we decompose $\uppercase\expandafter{\romannumeral1}$ into four terms $\uppercase\expandafter{\romannumeral1}_1, \uppercase\expandafter{\romannumeral1}_2, \uppercase\expandafter{\romannumeral1}_3, \uppercase\expandafter{\romannumeral1}_4$ by classifying the indices.

\vskip0.3cm
\noindent$\bullet$ \textbf{Term 1:} $k_1=k_1^{\prime}=k_2=k_2^{\prime}$.

\noindent Summing over all the indices under this case, we get
\begin{align*}
|\uppercase\expandafter{\romannumeral1}_1| = \sum_{p,q}\frac{1}{t^2} \sum_{i_1=0}^{t-1} \sum_{i_2=0}^{t-1} &\frac{1}{\varphi_{i_1}} \frac{1}{\varphi_{i_2}} \sum_{k=0}^{i_1 \wedge i_2} \prod_{l_1=k+1}^{i_1}(1-\sigma_p\varphi_{l_1})(1- \sigma_q\varphi_{l_1})\cdot\\
&\prod_{l_2=k+1}^{i_2}(1- \sigma_p\varphi_{l_2})(1-\sigma_q\varphi_{l_2}) \varphi_k^4 \bigg\{\mE\Big[\big(U^T \tilde{\btheta}^{k}\tilde{\btheta}^{k^ T}U\big)_{p,q}^2\Big]-\bigg(\mE\Big[\big(U^T \tilde{\btheta}^{k}\tilde{\btheta}^{k^ T}U\big)_{p,q}\Big]\bigg)^2\bigg\}\\
\leq \frac{1}{t^2} \sum_{i_1=0}^{t-1} \sum_{i_2=0}^{t-1} \frac{1}{\varphi_{i_1}} &\frac{1}{\varphi_{i_2}} \sum_{k=0}^{i_1 \wedge i_2} \prod_{l_1=k+1}^{i_1}(1-(1-\rho^{\tau})\varphi_{l_1})^2\prod_{l_2=k+1}^{i_2}(1-(1-\rho^{\tau})\varphi_{l_2})^2 \varphi_k^4\mE\Big[\sum_{p,q}\big(U^T \tilde{\btheta}^{k}\tilde{\btheta}^{k^T}U\big)_{p,q}^2\Big].
\end{align*}
Here, the equality holds because $1-\sigma_k\varphi_t>0$ for any $1\leq k\leq d$ and $t\geq 0$ following the same discussion as in \eqref{appen:A4:equ4}. By \eqref{appen:A4:equ5}, we know
\begin{equation*}
\mE\Big[\sum_{p,q}\big(U^T \tilde{\btheta}^{k}\tilde{\btheta}^{k^T}U\big)_{p,q}^2\Big] = \mE\big[\|\tilde{\btheta}^k\|^4\big] \lesssim\frac{C_{g,2}}{\gamma_H^4}.
\end{equation*}
Due to the symmetry between the indices $i_1$ and $i_2$, $|\uppercase\expandafter{\romannumeral1}_1|$ can be further bounded by
\begin{multline*}       
|\uppercase\expandafter{\romannumeral1}_1| \leq \frac{2}{t^2} \sum_{i_1=0}^{t-1} \frac{1}{\varphi_{i_1}} \sum_{i_2=0}^{i_1} \frac{1}{\varphi_{i_2}} \prod_{l_2=i_2+1}^{i_1}(1-(1-\rho^{\tau})\varphi_{l_2})^2\underbrace{\sum_{k=0}^{i_2} \prod_{l_1=k+1}^{i_2} (1-(1-\rho^{\tau})\varphi_{l_1})^4 \varphi_k^4}_{\lesssim \varphi_{i_2}^3/(1-\rho^\tau) \quad\text{by Lemma \ref{aux:lem1}}}\mE\big[\|\tilde{\btheta}^k\|^4\big]\\
\lesssim \frac{C_{g,2}}{\gamma_H^4(1-\rho^\tau)}\cdot\frac{1}{t^2} \sum_{i_1=0}^{t-1} \underbrace{\frac{1}{\varphi_{i_1}} \sum_{i_2=0}^{i_1} \prod_{l_2=i_2+1}^{i_1}(1-(1-\rho^{\tau}))\varphi_{l_2})^2 \varphi_{i_2}^2}_{\longrightarrow 0.5/(1-\rho^{\tau})\quad\text{by Lemma \ref{aux:lem1}}}\lesssim \frac{C_{g,2}}{\gamma_H^4(1-\rho^\tau)^2}\cdot\frac{1}{t}.
\end{multline*}
\noindent$\bullet$ \textbf{Term 2:} $k_1=k_1^{\prime}, k_2=k_2^{\prime}, k_1 \neq k_2$.

\noindent We note that
\begin{equation*}
\mE\Big[\big(U^T \tilde{\btheta}^{k_1}\tilde{\btheta}^{k_1^{T}}U\big)_{p,q}\big(U^T \tilde{\btheta}^{k_2}\tilde{\btheta}^{k_2^{T}}U\big)_{p,q}\Big]=\mE\Big[\big(U^T \tilde{\btheta}^{k_1}\tilde{\btheta}^{k_1^{T}}U\big)_{p,q}\Big]\mE\Big[\big(U^T \tilde{\btheta}^{k_2}\tilde{\btheta}^{k_2^{T}}U\big)_{p,q}\Big]\;\;\text{for }k_1\neq k_2.
\end{equation*}
This indicates that $\uppercase\expandafter{\romannumeral1}_2 = 0$.
\vskip0.3cm
\noindent$\bullet$ \textbf{Term 3:} $k_1=k_2,  k_1^{\prime}=k_2^{\prime},  k_1 \neq k_1^{\prime}$.

\noindent In this case, it is observed that
\begin{equation*}
\mE\Big[\big(U^T \tilde{\btheta}^{k_1}\tilde{\btheta}^{k_1^{\prime T}}U\big)_{p,q}\Big]\mE\Big[\big(U^T \tilde{\btheta}^{k_1}\tilde{\btheta}^{k_1^{\prime T}}U\big)_{p,q}\Big] = 0.
\end{equation*}
Thus, we have
\begin{multline*}
|\uppercase\expandafter{\romannumeral1}_3|\leq \frac{1}{t^2} \sum_{i_1=0}^{t-1} \sum_{i_2=0}^{t-1} \frac{1}{\varphi_{i_1}}\frac{1}{\varphi_{i_2}} \sum_{k_1=0}^{i_1 \wedge i_2} \prod_{l_1=k_1+1}^{i_1}(1- (1-\rho^\tau)\varphi_{l_1}) \prod_{l_2=k_1+1}^{i_2}(1- (1-\rho^\tau)\varphi_{l_2}) \varphi_{k_1}^2\cdot\\
\sum_{k_1^{\prime}=0, k_1\neq k_1^{\prime}}^{i_1 \wedge i_2} \prod_{l_1^{\prime}=k_1^{\prime}+1}^{i_1}(1- (1-\rho^\tau)\varphi_{l_1^{\prime}}) \prod_{l_2^{\prime}=k_1^{\prime}+1}^{i_2}(1- (1-\rho^\tau)\varphi_{l_2^{\prime}}) \varphi_{k_1^{\prime}}^2\mE\Big[\sum_{p, q}\big(U^T \tilde{\btheta}^{k_1}\big)_p^2\big(U^T \tilde{\btheta}^{k_1^{\prime}}\big)_q^2\Big].
\end{multline*}
Since $k_1\neq k_1^{\prime}$, we have
\begin{equation*}
\mE\Big[\sum_{p, q}\big(U^T \tilde{\btheta}^{k_1}\big)_p^2\big(U^T \tilde{\btheta}^{k_1^{\prime}}\big)_q^2\Big]= \mE\big[\|\tilde{\btheta}^{k_1}\|^2\big] \mE\big[\|\tilde{\btheta}^{k_1^\prime}\|^2\big] \stackrel{\eqref{appen:A4:equ5}}{\lesssim}\frac{C_{g,2}}{\gamma_H^4}.
\end{equation*}
By the symmetry of the indices $i_1$ and $i_2$, we can further bound $|\uppercase\expandafter{\romannumeral1}_3|$ by
\begin{align}\label{appen:A4:equ26}
|\uppercase\expandafter{\romannumeral1}_3|
&\lesssim \frac{1}{t^2} \sum_{i_1=0}^{t-1} \frac{1}{\varphi_{i_1}} \sum_{i_2=0}^{i_1} \frac{1}{\varphi_{i_2}} \prod_{l=i_2+1}^{i_1}(1- (1-\rho^\tau)\varphi_l)^2\Big\{\underbrace{\sum_{k_1=0}^{i_2} \prod_{l_1=k_1+1}^{i_2}(1- (1-\rho^\tau)\varphi_{l_1})^2 \varphi_{k_1}^2}_{\lesssim \varphi_{i_2}/(1-\rho^\tau)\quad\text{by Lemma \ref{aux:lem1}}}\Big\}^2\cdot\frac{C_{g,2}}{\gamma_H^4}\nonumber\\
&\lesssim \frac{C_{g,2}}{\gamma_H^4(1-\rho^\tau)^2}\cdot\frac{1}{t^2} \sum_{i_1=0}^{t-1}\frac{1}{\varphi_{i_1}}  \underbrace{\sum_{i_2=0}^{i_1} \prod_{l=i_2+1}^{i_1}(1-(1-\rho^\tau)\varphi_l)^2\varphi_{i_2}}_{\longrightarrow 0.5/(1-\rho^{\tau})\quad\text{by Lemma \ref{aux:lem1}}}\nonumber\\
&\lesssim \frac{C_{g,2}}{\gamma_H^4(1-\rho^\tau)^3}\cdot\frac{1}{t^2} \sum_{i_1=0}^{t-1}\frac{1}{\varphi_{i_1}}\lesssim \frac{C_{g,2}}{c_{\beta}\gamma_H^4(1-\rho^\tau)^3}\cdot\frac{1}{t^2} \sum_{i_1=0}^{t-1}(i_1+1)^\beta\lesssim \frac{C_{g,2}}{\gamma_H^4(1-\rho^\tau)^3}\cdot\frac{1}{t\beta_t}.
\end{align}
\noindent$\bullet$ \textbf{Term 4:} $k_1=k_2^{\prime}, k_2=k_1^{\prime},  k_1 \neq k_2$.

\noindent In this case, we have
\begin{equation*}
\mE\Big[\big(U^T \tilde{\btheta}^{k_1}\tilde{\btheta}^{k_1^{\prime T}}U\big)_{p,q}\Big]\mE\Big[\big(U^T \tilde{\btheta}^{k_1^{\prime}}\tilde{\btheta}^{k_1^T}U\big)_{p,q}\Big] = 0.
\end{equation*}
The analysis of $\uppercase\expandafter{\romannumeral1}_4$ is almost identical to $\uppercase\expandafter{\romannumeral1}_3$, only with the expectation term being replaced by
\begin{equation*}
\sum_{p,q}\mE\Big[\big(U^T \tilde{\btheta}^{k_1} \tilde{\btheta}^{k_1^T} U\big)_{p,q}\big(U^T \tilde{\btheta}^{k_2} \tilde{\btheta}^{k_2^T }U\big)_{p,q}\Big] = \sum_{p,q} \Gamma_{p,q}^2 = \|\Gamma\|_F^2=\|\Lambda\|_F^2 \;\;\text{when } k_1\neq k_2.
\end{equation*}
Therefore, we conclude that 
\begin{equation*}\label{appen:A4:equ11}            
|\uppercase\expandafter{\romannumeral1}_4| \lesssim \frac{\|\Lambda\|_F^2}{(1-\rho^\tau)^3}\cdot\frac{1}{t\beta_t}.
\end{equation*}
Combining the analyses of four terms, we obtain
\begin{equation}\label{appen:A4:equ8b}
|\uppercase\expandafter{\romannumeral1}| \leq \sum_{i=1}^4 |\uppercase\expandafter{\romannumeral1}_i|\lesssim \frac{\max(\|\Lambda\|_F^2, C_{g,2}/\gamma_H^4)}{(1-\rho^\tau)^3}\cdot\frac{1}{t\beta_t}.
\end{equation}
Plugging \eqref{appen:A4:equ8}, \eqref{appen:A4:equ8a}, and \eqref{appen:A4:equ8b} into to \eqref{appen:A4:equ12}, we have
\begin{equation}\label{appen:A4:equ24}
\hskip-0.1cm {\footnotesize\mE\bigg[\Big\|\frac{1}{t}\sum_{i=0}^{t-1}\frac{1}{\varphi_i}\tilde{\I}_{1,i}\tilde{\I}_{1,i}^T-\Xi^\star\Big\|\bigg] \lesssim \begin{dcases}
\|\Lambda\|_F\beta_t= O(\beta_t), &\beta\in(0,1/3),\\
\max\bigg(\|\Lambda\|_F, \frac{\max(\|\Lambda\|_F, C_{g,2}^{1/2}/\gamma_H^2)}{c_{\beta}^{3/2}(1-\rho^\tau)^{3/2}}\bigg) \beta_t = O\rbr{\frac{\beta_t}{ (1-\rho^\tau)^{1.5}}}, &\beta = 1/3,\\
\frac{\max(\|\Lambda\|_F, C_{g,2}^{1/2}/\gamma_H^2)}{(1-\rho^\tau)^{3/2}}\cdot \frac{1}{\sqrt{t\beta_t}}=O\rbr{\frac{1}{{ (1-\rho^\tau)^{1.5}} \sqrt{t\beta_t}}}, &\beta\in(1/3,1).   
\end{dcases}}
\end{equation}
We complete the proof.

\subsubsection{Proof of Lemma \ref{appen:A4:lem4}}\label{pf:hatI1}

We present the following lemma to bound $\hat{\btheta}^k$ defined in \eqref{appen:A4:equ3}, with proof deferred to Appendix \ref{pf:hattheta}. 

\begin{lemma}\label{appen:A4:lem5}
Under the assumptions of Lemma \ref{appen:A4:lem2}, we have
\begin{equation*}
\mE\big[\|\hat{\btheta}^k\|^2\big] \lesssim \frac{\Upsilon_H^2}{\gamma_H^2} \mE\big[\|\bx_k-\bx^\star\|^2\big] + \frac{\tau^2\Upsilon_{S}C_{g,2}^{1/2}}{\gamma_H^4}\mE\big[\|B_k-B^\star\|^2\big].
\end{equation*}
\end{lemma}

This lemma indicates that the difference between the martingale difference $\btheta^k$ and its approximation $\tilde{\btheta}^k$ vanishes. Combining Lemma \ref{appen:A4:lem5} with \eqref{xtrate} and \eqref{appen:A3:equ15} in the proof of Lemma \ref{sec4:lem1}, we get $\hskip1.5cm$
\begin{equation}\label{appen:A4:equ18}
\mE\big[\|\hat{\btheta}^k\|^2\big]\lesssim C_{\hat{\btheta}}\rbr{ \beta_k + \frac{\Upsilon_H}{\gamma_H}\cdot \frac{\chi_k^2}{\beta_k^2}} \;\;\;\;\text{with}\;\;\;\; C_{\hat{\btheta}} = \frac{\Upsilon_H^2 C_{g,2}^{1/2}}{\gamma_H^6}\max\Big(\Upsilon_H^2, \frac{\tau^2\Upsilon_S\Upsilon_L^2C_{g,2}^{1/2}}{\gamma_H^2}\Big).
\end{equation}
Recall the expression of $\hat{\I}_{1,i}$ in \eqref{appen:A4:equ2}. Since $\{\hat{\btheta}^k\}_k$ is a martingale difference sequence, we follow the analysis in \eqref{appen:A4:equ4} and \eqref{appen:A4:equ17}, and obtain
\begin{align}\label{appen:A4:equ21}
&\frac{1}{t}\sum_{i=0}^{t-1} \frac{1}{\varphi_i} \mE\big[\|\hat{\I}_{1,i}\|^2\big] \leq \frac{1}{t}\sum_{i=0}^{t-1} \frac{1}{\varphi_i}\sum_{k=0}^i \prod_{l=k+1}^i (1-(1-\rho^{\tau}) \varphi_l)^2\varphi_k^2 \mE\big[\|\hat{\btheta}^k\|^2\big]\nonumber\\
&\lesssim C_{\hat{\btheta}}\cdot \frac{1}{t}\sum_{i=0}^{t-1}\underbrace{\frac{1}{\varphi_i}\sum_{k=0}^i\prod_{l=k+1}^i(1-(1-\rho^{\tau})\varphi_l)^2 \varphi_k^2\beta_k}_{\lesssim \beta_i/(1-\rho^\tau)\quad\text{by Lemma \ref{aux:lem1}}} + \frac{\Upsilon_H C_{\hat{\btheta}}}{\gamma_H}\cdot \frac{1}{t}\sum_{i=0}^{t-1}\underbrace{\frac{1}{\varphi_i}\sum_{k=0}^i\prod_{l=k+1}^i(1-(1-\rho^{\tau})\varphi_l)^2 \frac{\varphi_k^2\chi_k^2}{\beta_k^2}}_{\lesssim (\chi_i^2/\beta_i^2)/(1-\rho^\tau)\quad\text{by Lemma \ref{aux:lem1}}}\nonumber\\
&\stackrel{\mathclap{\eqref{appen:A3:equ11}}}{\lesssim}\;\; \frac{C_{\hat{\btheta}}}{(1-\rho^\tau)(1-\beta)}\beta_t + \frac{\Upsilon_H C_{\hat{\btheta}}\mathbf{1}_{\{\chi\leq 1.5\beta\}}}{\gamma_H(1-\rho^\tau)(1-2(\chi-\beta))}\cdot\frac{\chi_t^2}{\beta_t^2} = O\rbr{\frac{1}{{ 1-\rho^\tau}}\cbr{\beta_t + \frac{\chi_t^2}{\beta_t^2}}}.
\end{align}    
We complete the proof.

\subsubsection{Proof of Lemma \ref{appen:A4:lem5}}\label{pf:hattheta}

We expand $\hat{\btheta}^k$ based on its definition in \eqref{appen:A4:equ3} as
\begin{equation*}
\hat{\btheta}^k = \btheta^k - \tilde{\btheta}^k = (I-C_k) B_k^{-1} \nabla F_k -(I-\tilde{C}_k) B_k^{-1} \nabla f(\bx_k; \xi_k) + (I-\tilde{C}_k^\star) (B^{\star})^{-1} \nabla f(\bx^\star; \xi_k).
\end{equation*}
Then, we can bound $\|\hat{\btheta}^k\|^2$ as
\begin{multline*}
\|\hat{\btheta}^k\|^2\lesssim \|I-C_k\|^2 \|B_k^{-1}\|^2 \|\nabla F_k\|^2+\|I-\tilde{C}_k\|^2 \|B_k^{-1}\|^2\|\nabla f(\bx_k; \xi_k)-\nabla f(\bx^\star; \xi_k)\|^2\\
+\|(I-\tilde{C}_k) B_k^{-1}-(1-\tilde{C}_k^\star) (B^{\star})^{-1}\|^2\|\nabla f(\bx^\star; \xi_k)\|^2    =:\uppercase\expandafter{\romannumeral1} + \uppercase\expandafter{\romannumeral2} +  \uppercase\expandafter{\romannumeral3}.
\end{multline*}
For the first two terms, by Assumption \ref{ass:3} and Lemma \ref{aux:lem4}(c), we get
\begin{equation}\label{appen:A4:equ14}
\mE[\uppercase\expandafter{\romannumeral1}] \lesssim \frac{\Upsilon_H^2}{\gamma_H^2}\mE\big[\|\bx_k-\bx^\star\|^2\big]\quad\quad\text{and}\quad\quad\mE[\uppercase\expandafter{\romannumeral2}]  \lesssim \frac{\Upsilon_H^2}{\gamma_H^2}\mE\big[\|\bx_k-\bx^\star\|^2\big].
\end{equation}
Regarding the term $\uppercase\expandafter{\romannumeral3}$, we have
\begin{multline*}
\|(I-\tilde{C}_k) B_k^{-1}-(1-\tilde{C}_k^\star) (B^{\star})^{-1}\|^2\leq\|I-\tilde{C}_k\|^2 \|B_k^{-1}\|^2\|(B^{\star})^{-1}\|^2\|B_k-B^\star\|^2\\
+\|(B^{\star})^{-1}\|^2 \|\tilde{C}_k-\tilde{C}_k^\star\|^2 \lesssim \frac{1}{\gamma_H^4}\|B_k-B^\star\|^2+ \frac{1}{\gamma_H^2}\|\tilde{C}_k-\tilde{C}_k^\star\|^2.
\end{multline*}
Then, we apply the tower property of conditional expectation to bound $\mE[\uppercase\expandafter{\romannumeral3}]$ by first conditioning on $\mF_{k-1}$, and have
\begin{align}\label{appen:A4:equ13}
& \mE[\uppercase\expandafter{\romannumeral3}]
\lesssim \mE\bigg[ \mE\Big[\Big(\frac{1}{\gamma_H^4}\|B_k-B^\star\|^2+ \frac{1}{\gamma_H^2}\|\tilde{C}_k-\tilde{C}_k^\star\|^2\Big)\|\nabla f(\bx^\star;\xi_k)\|^2\mid\mF_{k-1}\Big] \bigg]\nonumber\\
&=\frac{1}{\gamma_H^4}\mE\Big[\|B_k-B^\star\|^2\mE\big[\|\nabla f(\bx^\star;\xi_k)\|^2\mid\mF_{k-1}\big]\Big] + \frac{1}{\gamma_H^2}\mE\Big[\mE\big[\|\tilde{C}_k-\tilde{C}_k^\star\|^2\mid \mF_{k-1}\big] \mE\big[\|\nabla f(\bx^\star;\xi_k)\|^2\mid\mF_{k-1}\big]\Big]\nonumber\\
&\stackrel{\mathclap{\eqref{appen:A4:equ22}}}{\lesssim}\;\;\frac{C_{g,2}^{1/2}}{\gamma_H^4}\mE\big[\|B_k-B^\star\|^2\big]+\frac{C_{g,2}^{1/2}}{\gamma_H^2}\mE\big[\|\tilde{C}_k-\tilde{C}_k^\star\|^2\big].
\end{align}
Here, the second equality is due to $\sigma(\|B_k-B^\star\|)\in\mF_{k-1}$ and the independence between $\xi_k$ and the~sketching matrices $\{S_{k,j}\}_{j=0}^\tau$. Plugging in the definition of $\tilde{C}_k$ \eqref{appen:A1:equ6} and $\tilde{C}_k^\star$ \eqref{Ctildestar}, we have
\begin{align*}
\|\tilde{C}_k-\tilde{C}_k^\star\| &=\Big\|\prod_{j=0}^{\tau-1} \tilde{C}_{k, j}-\prod_{j=0}^{\tau-1} \tilde{C}_{k, j}^\star\Big\|
\leq \Big\|\prod_{j=0}^{\tau-2} \tilde{C}_{k, j}-\prod_{j=0}^{\tau-2} \tilde{C}_{k, j}^\star\Big\|\cdot \|C_{k, \tau-1}^\star\|
+\Big\|\prod_{j=0}^{\tau-2} \tilde{C}_{k, j}\Big\|\cdot \|\tilde{C}_{k, \tau-1}-\tilde{C}_{k, \tau-1}^\star\|\\
&\leq\dots\leq \sum_{j=0}^{\tau-1}\left\|\tilde{C}_{k, j}-\tilde{C}_{k, j}^\star\right\|\quad(\text{by }\|C_{k, \tau-1}^\star\|\leq 1\text{ and }\|\tilde{C}_{k,j}\|\leq 1).
\end{align*}
Applying \cite[Lemma 5.2]{Na2025Statistical} and Assumption \ref{ass:4}, we obtain
\begin{equation*}
\mE\big[\|\tilde{C}_k-\tilde{C}_k^\star\|^2\mid\mF_{k-1}\big]\leq
\frac{4\|B_k-B^\star\|^2}{\gamma_H^2}\mE[(\sum_{j=0}^{\tau-1}\|S_{k, j}\|\|S_{k,j}^\dagger\|)^2]\lesssim \frac{\tau^2\Upsilon_{S}}{\gamma_H^2}\|B_k-B^\star\|^2.
\end{equation*}
Combining the above display to \eqref{appen:A4:equ13}, we get
\begin{equation}\label{appen:A4:equ16}
\mE[\uppercase\expandafter{\romannumeral3}]\lesssim \frac{\tau^2\Upsilon_{S}C_{g,2}^{1/2}}{\gamma_H^4}\mE\big[\|B_k-B^\star\|^2\big].
\end{equation}
Combining \eqref{appen:A4:equ14} and \eqref{appen:A4:equ16} completes the proof.

\subsection{Proof of Theorem \ref{sec4:thm1}}\label{appen:A5}

The weighted sample covariance matrix $\hat{\Xi}_t$ can be decomposed as
\begin{multline}\label{appen:A5:equ1}
\widehat{\Xi}_t = \frac{1}{t}\sum_{i=1}^t\frac{1}{\varphi_{i-1}}(\bx_i-\bx^\star)(\bx_i-\bx^\star)^T + \frac{1}{t}\sum_{i=1}^t\frac{1}{\varphi_{i-1}}(\bar{\bx}_t-\bx^\star)(\bar{\bx}_t-\bx^\star)^T \\
- \frac{1}{t}\sum_{i=1}^t\frac{1}{\varphi_{i-1}}(\bx_i-\bx^\star)(\bar{\bx}_t-\bx^\star)^T - \frac{1}{t}\sum_{i=1}^t\frac{1}{\varphi_{i-1}}(\bar{\bx}_t-\bx^\star)(\bx_i-\bx^\star)^T.
\end{multline}
The next two lemmas show that $\frac{1}{t}\sum_{i=1}^t\frac{1}{\varphi_{i-1}}(\bx_i-\bx^\star)(\bx_i-\bx^\star)^T$ converges to $\Xi^{\star}$, and the remaining terms are negligible as $\bar{\bx}_t$ converges to $\bx^\star$ fast. The proofs are in Appendices \ref{pf:samplecov} and \ref{pf:barx}.

\begin{lemma}\label{appen:A5:lem1}

Suppose Assumptions \ref{ass:1} -- \ref{ass:4} hold, the number of sketches satisfies $\tau\geq \log(\gamma_H/4\Upsilon_H)/\log \rho$ with $\rho = 1-\gamma_S$, and the stepsize parameters satisfy $\beta\in(0,1)$, $\chi>1.5\beta$, and $c_{\beta}, c_{\chi}>0$. Then, we have
\begin{equation}\label{appen:A5:equ10}
\mE\bigg[\Big\|\frac{1}{t}\sum_{i=1}^t\frac{1}{\varphi_{i-1}}(\bx_i-\bx^\star)(\bx_i-\bx^\star)^T-\Xi^\star\Big\|\bigg] \lesssim \begin{dcases}
\frac{1}{{ (1-\rho^\tau)^{1.5}}}\bigg(\sqrt{\beta_t} \; +\; \frac{\chi_t}{\beta_t^{1.5}}\bigg), &\beta\in(0,0.5],\\
\frac{1}{{ (1-\rho^\tau)^{1.5}}}\bigg(\frac{1}{\sqrt{t\beta_t}} \; +\; \frac{\chi_t}{\beta_t^{1.5}}\bigg), &\beta\in(0.5,1),
\end{dcases}
\end{equation}
and
\begin{equation}\label{appen:A5:equ19}
\frac{1}{t}\sum_{i=0}^{t-1}\frac{1}{\varphi_{i-1}}\mE\big[\|\bx_i-\bx^\star\|^2\big] \lesssim \frac{1}{{ 1-\rho^\tau}}= O(1),
\end{equation}
where we explicitly track the dependency of the constant factor on $\rho = 1-\gamma_S$.

\end{lemma}

\begin{lemma}\label{appen:A5:lem2}

Suppose Assumptions \ref{ass:1} -- \ref{ass:4} hold, the number of sketches satisfies $\tau\geq \log(\gamma_H/4\Upsilon_H)/\log \rho$ with $\rho = 1-\gamma_S$, and the stepsize parameters satisfy $\beta\in(0,1)$, $\chi>1.5\beta$, and $c_{\beta}, c_{\chi}>0$. Then, we have
\begin{multline*}
\mE\bigg[\Big\|\frac{1}{t}\sum_{i=1}^t\frac{1}{\varphi_{i-1}}(\bar{\bx}_t-\bx^\star)(\bar{\bx}_t-\bx^\star)^T\Big\|\bigg]\leq\frac{1}{t}\sum_{i=0}^{t-1}\frac{1}{\varphi_{i-1}}\mE\big[\|\bar{\bx}_t-\bx^\star\|^2\big]\\
\lesssim \begin{dcases}
\frac{1}{{ (1-\rho^\tau)^{2}}}\bigg(\beta_t \; +\; \frac{\chi_t^2}{\beta_t^{3}}\bigg), &\beta\in(0,0.5],\\
\frac{1}{{ (1-\rho^\tau)^{2}}}\bigg(\frac{1}{t\beta_t} \; +\; \frac{\chi_t^2}{\beta_t^{3}}\bigg), &\beta\in(0.5,1),
\end{dcases}
\end{multline*}
where we explicitly track the dependency of the constant factor on $\rho = 1-\gamma_S$.
\end{lemma}

By the decomposition \eqref{appen:A5:equ1}, we follow the derivations in \eqref{appen:A4:equ23} and \eqref{appen:A4:equ19} and obtain
\begin{align*}
\mE\big[\|\widehat{\Xi}_t-\Xi^\star\|\big]&\leq
\mE\bigg[\Big\|\frac{1}{t}\sum_{i=1}^t\frac{1}{\varphi_{i-1}}(\bx_i-\bx^\star)(\bx_i-\bx^\star)^T-\Xi^\star\Big\|\bigg]+\mE\bigg[\Big\|\frac{1}{t}\sum_{i=1}^t\frac{1}{\varphi_{i-1}}(\bar{\bx}_t-\bx^\star)(\bar{\bx}_t-\bx^\star)^T\Big\|\bigg]\\
&\quad +2\sqrt{\frac{1}{t}\sum_{i=1}^t\frac{1}{\varphi_{i-1}}\mE\big[\|\bx_i-\bx^\star\|^2\big]}\sqrt{\frac{1}{t}\sum_{i=1}^t\frac{1}{\varphi_{i-1}}\mE\big[\|\bar{\bx}_t-\bx^\star\|^2\big]}.
\end{align*}
Plugging \eqref{appen:A5:equ16} and \eqref{appen:A5:equ21} in the proof of Lemma \ref{appen:A5:lem1} and \eqref{appen:A5:equ22} in the proof of Lemma \ref{appen:A5:lem2} into the above display, we obtain
\begin{scriptsize}
\begin{align}\label{pf:res}
& \mE\big[\|\widehat{\Xi}_t-\Xi^\star\|\big] \nonumber\\
&\lesssim \begin{dcases}
\frac{C_{g,2}^{1/4}}{\gamma_H(1-\rho^\tau)}\max \rbr{C_{\hat{\btheta}}^{1/2}, \frac{C_{\bdelta}^{1/2}}{(1-\rho^\tau)^{1/2}}}\sqrt{\beta_t} + \frac{C_{g,2}^{1/2}\mathbf{1}_{\{\chi\leq 2\beta\}}}{\gamma_H^2(1-\rho^\tau)^{3/2}}\sqrt{\frac{\chi_t^2}{\beta_t^{3}}} = O\rbr{\frac{1}{ (1-\rho^\tau)^{1.5}}\cbr{\sqrt{\beta_t} + \frac{\chi_t}{\beta_t^{1.5}}}}, &\beta\in(0,0.5),\\
\max\rbr{\frac{C_{g,2}^{1/4}}{\gamma_H(1-\rho^\tau)}\max\rbr{C_{\hat{\btheta}}^{1/2}, \frac{C_{\bdelta}^{1/2}}{(1-\rho^\tau)^{1/2}}}, \frac{\max(\|\Lambda\|_F, C_{g,2}^{1/2}/\gamma_H^2)}{c_{\beta}(1-\rho^\tau)^{3/2}}}\sqrt{\beta_t} + \frac{C_{g,2}^{1/2}\mathbf{1}_{\{\chi\leq 2\beta\}}}{\gamma_H^2(1-\rho^\tau)^{3/2}}\sqrt{\frac{\chi_t^2}{\beta_t^{3}}} = O\rbr{\frac{\sqrt{\beta_t}+\chi_t/\beta_t^{1.5}}{ (1-\rho^\tau)^{1.5}}}, &\beta = 0.5,\\
\frac{\max(\|\Lambda\|_F, C_{g,2}^{1/2}/\gamma_H^2)}{(1-\rho^\tau)^{3/2}}\cdot\frac{1}{\sqrt{t\beta_t}} + \frac{C_{g,2}^{1/2}\mathbf{1}_{\{\chi\leq\beta+0.5\}}}{\gamma_H^2(1-\rho^\tau)^{3/2} }\sqrt{\frac{\chi_t^2}{\beta_t^{3}}}= O\rbr{\frac{1}{ (1-\rho^\tau)^{1.5}}\cbr{\frac{1}{\sqrt{t\beta_t}} + \frac{\chi_t}{\beta_t^{1.5}}}}, &\beta\in(0.5,1),
\end{dcases}
\end{align}
\end{scriptsize}
\hskip-2.5pt where $\Lambda = \mE[(I- \tilde{C}^\star)\Omega^\star(I-\tilde{C}^\star)^T]$, the constant $C_{\hat{\btheta}}>0$ is defined in \eqref{appen:A4:equ18}, and the constant $C_{\bdelta}>0$~is later defined in \eqref{appen:A5:equ6}. This completes the proof.

\subsubsection{Proof of Lemma \ref{appen:A5:lem1}}\label{pf:samplecov}

By the decomposition \eqref{appen:A4:equ20}, we have proved the consistency of the dominant term $\frac{1}{t}\sum_{i=0}^{t-1}\frac{1}{\varphi_i}\I_{1,i}\I_{1,i}^T$ in Appendix \ref{appen:A4}. The next two lemmas suggest that the terms involving $\{\I_{2,i}\}_i$ and $\{\I_{3,i}\}_i$ are higher order errors, the proofs of which are deferred to Appendices \ref{pf:I2} and \ref{pf:I3}.

\begin{lemma}\label{appen:A5:lem3}

Suppose the assumptions in Lemma \ref{appen:A5:lem1} hold, we have
\begin{equation*}
\mE\bigg[\Big\|\frac{1}{t}\sum_{i=0}^{t-1}\frac{1}{\varphi_i}\I_{2,i}\I_{2,i}^T\Big\|\bigg]\leq\frac{1}{t}\sum_{i=0}^{t-1}\frac{1}{\varphi_i}\mE\big[\|\I_{2,i}\|^2\big] =O\rbr{\frac{1}{{ (1-\rho^\tau)^2}}\frac{\chi_t^2}{\beta_t^3}}\cdot \mathbf{1}_{\{\chi<1.5\beta+0.5\}} + o(\beta_t)\cdot \mathbf{1}_{\{\chi\geq 1.5\beta+0.5\}},
\end{equation*}
where we explicitly track the dependency of the constant factor on $\rho = 1-\gamma_S$.
\end{lemma}

\begin{lemma}\label{appen:A5:lem4}

Suppose the assumptions in Lemma \ref{appen:A5:lem1} hold, we have
\begin{equation*}
\mE\bigg[\Big\|\frac{1}{t}\sum_{i=0}^{t-1}\frac{1}{\varphi_i}\I_{3,i}\I_{3,i}^T\Big\|\bigg]\leq\frac{1}{t} \sum_{i=0}^{t-1} \frac{1}{\varphi_i} \mE\big[\|\I_{3, i}\|^2\big] \lesssim \frac{\beta_t}{{ (1-\rho^\tau)^2}},
\end{equation*}
where we explicitly track the dependency of the constant factor on $\rho = 1-\gamma_S$.
\end{lemma}

With the above two lemmas, we separate the proof Lemma \ref{appen:A5:lem1} by two parts.

\vskip 0.3cm
\noindent \textbf{Part 1: Proof of \eqref{appen:A5:equ10}.} By the decomposition \eqref{appen:A4:equ20}, we follow \eqref{appen:A4:equ23} and \eqref{appen:A4:equ19} and have
\begin{align}\label{appen:A5:equ9}
&\mE\bigg[\Big\|\frac{1}{t}\sum_{i=1}^t\frac{1}{\varphi_{i-1}}(\bx_i-\bx^\star)(\bx_i-\bx^\star)^T-\Xi^\star\Big\|\bigg]\nonumber\\
&\leq \mE\bigg[\Big\|\frac{1}{t}\sum_{i=0}^{t-1}\frac{1}{\varphi_i}\I_{1,i}\I_{1,i}^T-\Xi^\star\Big\|\bigg]
+\mE\bigg[\Big\|\frac{1}{t}\sum_{i=0}^{t-1}\frac{1}{\varphi_i}\I_{2,i}\I_{2,i}^T\Big\|\bigg]+\mE\bigg[\Big\|\frac{1}{t}\sum_{i=0}^{t-1}\frac{1}{\varphi_i}\I_{3,i}\I_{3,i}^T\Big\|\bigg]\nonumber\\
&+2\sum_{1\leq r<s\leq 3}\sqrt{\frac{1}{t}\sum_{i=0}^{t-1}\frac{1}{\varphi_i}\mE\big[\|\I_{r,i}\|^2\big]}  \sqrt{\frac{1}{t}\sum_{i=0}^{t-1}\frac{1}{\varphi_i}\mE\big[\|\I_{s,i}\|^2\big]}.
\end{align}
Given Lemmas \ref{appen:A4:lem2}, \ref{appen:A5:lem3}, and \ref{appen:A5:lem4}, it is sufficient to establish the bound for $\frac{1}{t}\sum_{i=0}^{t-1} \frac{1}{\varphi_i} \mE\big[\|\I_{1,i}\|^2\big]$. We first~bound the moment for $\|\btheta^k\|$. Based on its definition \eqref{rec:def:b}, we have
\begin{equation*}
\btheta^k = -(I-\tilde{C}_k)B_k^{-1}(\bar{g}_k-\nabla F_k) + (\tilde{C}_k-C_k)B_k^{-1}\nabla F_k.
\end{equation*}
Furthermore, by $\|\tilde{C}_k\|\leq 1$, $\|C_k\|\leq 1$, and \eqref{ass:3:c2}, we get
\begin{align}\label{appen:A5:equ2}
\mE\big[\|\btheta^k\|^2\big]&\lesssim \frac{1}{\gamma_H^2}\mE\big[\|\bar{g}_k-\nabla F_k\|^2\big] + \frac{1}{\gamma_H^2}\mE\big[\|\nabla F_k\|^2\big]\nonumber\\
&\leq \frac{C_{g,1}^{1/2}}{\gamma_H^2}\mE\big[\|\bx_k-\bx^\star\|^2\big] + \frac{C_{g,2}^{1/2}}{\gamma_H^2} + \frac{\Upsilon_H^2}{\gamma_H^2}\mE\big[\|\bx_k-\bx^\star\|^2\big]\quad(\text{by Assumptions \ref{ass:2} and \ref{ass:3}})\nonumber\\
&\lesssim \frac{C_{g,2}^{1/2}}{\gamma_H^2}\quad (\mE\big[\|\bx_k-\bx^\star\|^2\big]=o(1) \text{ by Lemma \ref{sec4:lem1}}).
\end{align}
Since $\btheta^k$ is a martingale difference sequence, we follow \eqref{appen:A4:equ21} and get
\begin{multline}\label{appen:A5:equ20}
\frac{1}{t}\sum_{i=0}^{t-1} \frac{1}{\varphi_i} \mE\big[\|\I_{1,i}\|^2\big] \leq \frac{1}{t}\sum_{i=0}^{t-1} \frac{1}{\varphi_i}\sum_{k=0}^i \prod_{l=k+1}^i (1-(1-\rho^{\tau}) \varphi_l)^2\varphi_k^2 \mE\big[\|\btheta^k\|^2\big]\\
\lesssim \frac{C_{g,2}^{1/2}}{\gamma_H^2}\frac{1}{t}\sum_{i=0}^{t-1}\underbrace{\frac{1}{\varphi_i}\sum_{k=0}^{i}\prod_{l=k+1}^i(1-(1-\rho^{\tau})\varphi_l)^2 \varphi_k^2}_{\rightarrow 0.5/(1-\rho^\tau)\quad\text{by Lemma \ref{aux:lem1}}}\lesssim \frac{C_{g,2}^{1/2}}{\gamma_H^2(1-\rho^\tau)}.
\end{multline}
Combining the above display, Lemma \ref{appen:A4:lem2} (\eqref{appen:A4:equ27} in the proof), Lemma \ref{appen:A5:lem3} (\eqref{appen:A5:equ14} in the proof), and Lemma \ref{appen:A5:lem4} (\eqref{appen:A5:equ15} in the proof), and plugging them into \eqref{appen:A5:equ9}, we get
\begin{scriptsize}
\begin{multline}\label{appen:A5:equ16}
\mE\bigg[\Big\|\frac{1}{t}\sum_{i=1}^t\frac{1}{\varphi_{i-1}}(\bx_i-\bx^\star)(\bx_i-\bx^\star)^T-\Xi^\star\Big\|\bigg]\\
\lesssim \begin{dcases}
\frac{C_{g,2}^{1/4}}{\gamma_H(1-\rho^\tau)}\max \rbr{C_{\hat{\btheta}}^{1/2}, \frac{C_{\bdelta}^{1/2}}{(1-\rho^\tau)^{1/2}}}\sqrt{\beta_t} + \frac{C_{g,2}^{1/2}\mathbf{1}_{\{\chi\leq 2\beta\}}}{\gamma_H^2(1-\rho^\tau)^{3/2}}\sqrt{\frac{\chi_t^2}{\beta_t^{3}}} = O\rbr{\frac{1}{ (1-\rho^\tau)^{1.5}}\cbr{\sqrt{\beta_t} + \frac{\chi_t}{\beta_t^{1.5}}}}, &\beta\in(0,0.5),\\
\max\rbr{\frac{C_{g,2}^{1/4}}{\gamma_H(1-\rho^\tau)}\max\rbr{C_{\hat{\btheta}}^{1/2}, \frac{C_{\bdelta}^{1/2}}{(1-\rho^\tau)^{1/2}}}, \frac{\max(\|\Lambda\|_F, C_{g,2}^{1/2}/\gamma_H^2)}{c_{\beta}(1-\rho^\tau)^{3/2}}}\sqrt{\beta_t} + \frac{C_{g,2}^{1/2}\mathbf{1}_{\{\chi\leq 2\beta\}}}{\gamma_H^2(1-\rho^\tau)^{3/2}}\sqrt{\frac{\chi_t^2}{\beta_t^{3}}} = O\rbr{\frac{\sqrt{\beta_t}+\chi_t/\beta_t^{1.5}}{ (1-\rho^\tau)^{1.5}}}, &\beta = 0.5,\\
\frac{\max(\|\Lambda\|_F, C_{g,2}^{1/2}/\gamma_H^2)}{(1-\rho^\tau)^{3/2}}\cdot\frac{1}{\sqrt{t\beta_t}} + \frac{C_{g,2}^{1/2}\mathbf{1}_{\{\chi\leq\beta+0.5\}}}{\gamma_H^2(1-\rho^\tau)^{3/2}}\sqrt{\frac{\chi_t^2}{\beta_t^{3}}}= O\rbr{\frac{1}{ (1-\rho^\tau)^{1.5}}\cbr{\frac{1}{\sqrt{t\beta_t}} + \frac{\chi_t}{\beta_t^{1.5}}}}, &\beta\in(0.5,1),
\end{dcases}
\end{multline}
\end{scriptsize}
\hskip-1.5pt where constants $C_{\hat{\btheta}}>0$ is defined in \eqref{appen:A4:equ18} and $C_{\bdelta}>0$ will be later defined in \eqref{appen:A5:equ6}. Here, we also~use the observation that $\chi_t^2/\beta_t^3 = o(\beta_t)$ when $\chi>2\beta$ and $\beta\in(0,0.5]$, and $\chi_t^2/\beta_t^3=o(1/t\beta_t)$ when $\chi>\beta+0.5$ and $\beta\in(0.5,1)$.

\vskip 0.3cm
\noindent \textbf{Part 2: Proof of \eqref{appen:A5:equ19}.} By the decomposition \eqref{appen:A4:equ20}, we have
\begin{equation}\label{appen:A5:equ21}
\frac{1}{t}\sum_{i=0}^{t-1}\frac{1}{\varphi_{i-1}}\mE\big[\|\bx_i-\bx^\star\|^2\big] \lesssim \sum_{k=1}^3 \frac{1}{t}\sum_{i=0}^{t-1} \frac{1}{\varphi_i} \mE\big[\|\I_{k,i}\|^2\big] \lesssim \frac{C_{g,2}^{1/2}}{\gamma_H^2(1-\rho^\tau)},
\end{equation}
where the last inequality follows from \eqref{appen:A5:equ20}, and Lemmas \ref{appen:A5:lem3} and \ref{appen:A5:lem4}. We complete the proof.

\subsubsection{Proof of Lemma \ref{appen:A5:lem2}}\label{pf:barx}

By \eqref{appen:A4:equ1}, we decompose $\bar{\bx}_t-\bx^\star$ as
\begin{equation}\label{appen:A5:equ4}
\bar{\bx}_t-\bx^\star = \frac{1}{t}\sum_{i=0}^{t-1}\I_{1,i} + \frac{1}{t}\sum_{i=0}^{t-1}\I_{2,i}
+\frac{1}{t}\sum_{i=0}^{t-1}\I_{3,i}=: \bar{\I}_{1,t}
+\bar{\I}_{2,t}+\bar{\I}_{3,t}.
\end{equation}
We expand $\bar{\I}_{1,t}$ by plugging in \eqref{rec:a} and exchange the indices. Then, we obtain
\begin{equation*}
\bar{\I}_{1, t}=\frac{1}{t} \sum_{i=0}^{t-1} \sum_{k=0}^i \prod_{l=k+1}^i \{I-\varphi_l (I-C^\star)\} \varphi_k \btheta^k =\frac{1}{t} \sum_{k=0}^{t-1}\sum_{i=k}^{t-1} \prod_{l=k+1}^i \{I-\varphi_l (I-C^\star)\} \varphi_k \btheta^k.    
\end{equation*}
Since $\btheta^k$ is a martingale difference sequence, the interaction terms in $\mE\big[\|\bar{\I}_{1,t}\|^2\big]$ are vanished. Thus, we have
\begin{multline*}
\mE\big[\|\bar{\I}_{1,t}\|^2\big] 
= \frac{1}{t^2} \sum_{k=0}^{t-1}\varphi_k^2
\mE\bigg[\Big\|\sum_{i=k}^{t-1} \prod_{l=k+1}^i \{I-\varphi_l (I-C^\star)\} \btheta^k\Big\|^2\bigg]\\
\leq \frac{1}{t^2} \sum_{k=0}^{t-1}
\Big(\sum_{i=k}^{t-1} \prod_{l=k+1}^i (1- (1-\rho^\tau)\varphi_l)\Big)^2 \varphi_k^2 \mE\big[\|\btheta^k\|^2\big]=:(\#)\quad\text{(by Lemma \ref{aux:lem4}(d))}.
\end{multline*}
We rewrite the above display by exchanging the indices, and obtain
\begin{align*}
(\#) &= \frac{1}{t^2} \sum_{k=0}^{t-1}
\sum_{i_1=k}^{t-1}\sum_{i_2=k}^{t-1} \prod_{l_1=k_1+1}^{i_1} (1- (1-\rho^\tau)\varphi_{l_1}) \prod_{l_2=k_2+1}^{i_2} (1- (1-\rho^\tau)\varphi_{l_2})\varphi_k^2 \mE\big[\|\btheta^k\|^2\big]\\
&=\frac{1}{t^2} \sum_{i_1=0}^{t-1} \sum_{i_2=0}^{t-1} \sum_{k=0}^{i_1\wedge i_2} \prod_{l_1=k+1}^{i_1}(1- (1-\rho^\tau)\varphi_{l_1}) \prod_{l_2=k+1}^{i_2}(1- (1-\rho^\tau)\varphi_{l_2}) \varphi_k^2 \mE\big[\|\btheta^k\|^2\big]\\
&\leq \frac{2}{t^2}\sum_{i_1=0}^{t-1} \sum_{i_2=0}^{i_1}\prod_{l_1=i_2+1}^{i_1}(1- (1-\rho^\tau)\varphi_{l_1}) \sum_{k=0}^{i_2} \prod_{l_2=k+1}^{i_2}(1- (1-\rho^\tau)\varphi_{l_2})^2 \varphi_k^2 \mE\big[\|\btheta^k\|^2\big],
\end{align*}
where the last inequality comes from the symmetry between the indices $i_1$ and $i_2$. We plug in \eqref{appen:A5:equ2}~and get
\begin{align}\label{appen:A5:equ12}
\mE\big[\|\bar{\I}_{1,t}\|^2\big] & \lesssim \frac{C_{g,2}^{1/2}}{\gamma_H^2} \cdot \frac{1}{t^2}\sum_{i_1=0}^{t-1} \sum_{i_2=0}^{i_1}\prod_{l_1=i_2+1}^{i_1}(1- (1-\rho^\tau)\varphi_{l_1}) \underbrace{\sum_{k=0}^{i_2} \prod_{l_2=k+1}^{i_2}(1- (1-\rho^\tau)\varphi_{l_2})^2 \varphi_k^2}_{\lesssim \varphi_{i_2}/(1-\rho^\tau)\quad\text{ by Lemma \ref{aux:lem1}}} \nonumber\\
& \lesssim \frac{C_{g,2}^{1/2}}{\gamma_H^2(1-\rho^\tau)} \cdot \frac{1}{t^2}\sum_{i_1=0}^{t-1} \underbrace{\sum_{i_2=0}^{i_1}\prod_{l_1=i_2+1}^{i_1}(1- (1-\rho^\tau)\varphi_{l_1})\varphi_{i_2}}_{\longrightarrow 1/(1-\rho^{\tau})\quad\text{ by Lemma \ref{aux:lem1}}}\lesssim  \frac{C_{g,2}^{1/2}}{\gamma_H^2(1-\rho^\tau)^2} \cdot \frac{1}{t}.
\end{align}
For the term $\bar{\I}_{2,t}$, we plug in \eqref{rec:b} and get
\begin{equation*}
\bar{\I}_{2, t}=\frac{1}{t} \sum_{i=0}^{t-1} \sum_{k=0}^i \prod_{l=k+1}^i \{I-\varphi_l (I-C^\star)\}(\bar{\alpha}_k- \varphi_k) \bar{\Delta}\bx_k.
\end{equation*}
Furthermore, by Lemma \ref{aux:lem4}(d) and the fact that $|\bar{\alpha}_k-\varphi_k|\leq \chi_k/2$, we know
\begin{multline}\label{appen:A5:equ8}
\mE\big[\|\bar{\I}_{2, t}\|^2\big] \lesssim 
\mE\bigg[\Big(\frac{1}{t}\sum_{i=0}^{t-1} \sum_{k=0}^i \prod_{l=k+1}^i (1-(1-\rho^\tau)\varphi_l)\chi_k \|\bar{\Delta}\bx_k\|\Big)^2\bigg]\\
\leq \Big(\frac{1}{t}\sum_{i=0}^{t-1} \sum_{k=0}^i \prod_{l=k+1}^i (1-(1-\rho^\tau)\varphi_l) \chi_k\sqrt{\mathbb{E}\big[\|\bar{\Delta}\bx_k\|^2\big]}\Big)^2\quad \text{(by H\"older's inequality)}.
\end{multline}
Using $\|\tilde{C}_k\|\leq 1$ and \eqref{ass:3:c2}, we bound $\mathbb{E}\big[\|\bar{\Delta}\bx_k\|^2\big]$ as
\begin{equation}\label{appen:A5:equ5}
\mathbb{E}\big[\|\bar{\Delta}\bx_k\|^2\big]\leq \mE\big[\|I-\tilde{C}_k\|^2\|B_k^{-1}\|^2\|\bar{g}_k\|^2\big] \lesssim \frac{1}{\gamma_H^2}\big(\mE\big[\|\bar{g}_k-\nabla F_k\|^2\big]+\mE\big[\|\nabla F_k\|^2\big]\big)
\stackrel{\eqref{appen:A5:equ2}}{\lesssim} \frac{C_{g,2}^{1/2}}{\gamma_H^2}.
\end{equation}
Consequently, we obtain
\begin{align}\label{appen:A5:equ13}
\mE\big[\|\bar{\I}_{2, t}\|^2\big]\cdot\mathbf{1}_{\{\chi<\beta+1\}} &\lesssim  \frac{C_{g,2}^{1/2}\mathbf{1}_{\{\chi<\beta+1\}}}{\gamma_H^2}\Big(\frac{1}{t}\sum_{i=0}^{t-1} \underbrace{\sum_{k=0}^{i} \prod_{l=k+1}^i\left(1-\left(1-\rho^\tau\right) \varphi_l\right)\varphi_k\cdot \frac{\chi_k}{\varphi_k}}_{\lesssim (\chi_{i}/\beta_i)/(1-\rho^\tau)\quad\text{by Lemma \ref{aux:lem1}}}\Big)^2\nonumber\\
&\lesssim \frac{C_{g,2}^{1/2} \mathbf{1}_{\{\chi<\beta+1\}}}{\gamma_H^2(1-\rho^\tau)^2}\Big(\frac{1}{t}\sum_{i=0}^{t-1}\frac{\chi_i}{\beta_i}\Big)^2    \stackrel{\eqref{appen:A3:equ11}}{\lesssim} \frac{C_{g,2}^{1/2}\mathbf{1}_{\{\chi<\beta+1\}}}{\gamma_H^2(1-\rho^\tau)^2(1-(\chi-\beta))^2}\cdot \frac{\chi_t^2}{\beta_t^2},\\
\mE\big[\|\bar{\I}_{2, t}\|^2\big]\cdot\mathbf{1}_{\{\chi\geq \beta+ 1\}} &= \Big(\frac{1}{t}\sum_{i=0}^{t-1} o(\beta_i)\Big)^2\cdot \mathbf{1}_{\{\chi\geq \beta+ 1\}} \stackrel{\eqref{appen:A3:equ11}}{=} o(\beta_t^2)\cdot \mathbf{1}_{\{\chi\geq \beta+ 1\}}. \nonumber
\end{align}
Here, we use the fact that $\chi\geq \beta+1 > 2\beta \Rightarrow \chi_t/\beta_t = o(\beta_t)$.~For the term $\bar{\I}_{3,t}$, \eqref{rec:c} gives us the following expansion
\begin{equation*}
\bar{\I}_{3, t}=\frac{1}{t} \sum_{i=0}^{t-1} \prod_{k=0}^i \{I-\varphi_k (I-C^\star)\} (\bx_0-\bx^\star) +\frac{1}{t} \sum_{i=0}^{t-1} \sum_{k=0}^i \sum_{ l=k+1}^i \{I-\varphi_l (I-C^\star)\} \varphi_k \bdelta^k.
\end{equation*}
Similar to \eqref{appen:A5:equ8}, by H\"older's inequality, we have
\begin{multline}\label{appen:A5:equ11}
\mE\big[\|\bar{\I}_{3,t}\|^2\big]\lesssim \Big(\frac{1}{t}\sum_{i=0}^{t-1}\prod_{k=0}^i(1-(1-\rho^\tau)\varphi_k)\Big)^2\|\bx_0-\bx^\star\|^2\\
+ \Big(\frac{1}{t}\sum_{i=0}^{t-1}\sum_{k=0}^i\prod_{l=k+1}^i(1-(1-\rho^\tau)\varphi_l)\varphi_k\sqrt{\mE\big[\|\bdelta^k\|^2\big]}\Big)^2.
\end{multline}
Next, we bound the rate of $\mE\big[\|\bdelta^k\|^2\big]$. By the definition of $\bdelta^k$ in \eqref{rec:def:c} and $\bpsi^k$  in \eqref{rec:def:d}, we have
\begin{align}\label{appen:A5:equreview}
\|\bdelta^k\|^2 &\lesssim \|C_k-C^\star\|^2\|\bx_k-\bx^\star\|^2+\|(B^{\star})^{-1}\|^2\|\bpsi^k\|^2 + \|B_k^{-1}\|^2\|(B^{\star})^{-1}\|^2\|B_k-B^\star\|^2\|\nabla F_k\|^2\\
&\leq \frac{\tau^2\Upsilon_S}{\gamma_H^2}\|B_k-B^\star\|^2 \|\bx_k-\bx^\star\|^2 + \frac{1}{\gamma_H^2}\cdot \frac{\Upsilon_L^2}{4}\|\bx_k-\bx^\star\|^4 + \frac{\Upsilon_H^2}{\gamma_H^4}\|B_k-B^\star\|^2 \|\bx_k-\bx^\star\|^2.
\end{align}
The second inequality is due to $\|C_k-C^\star\|\leq \tau\Upsilon_{S}^{1/2}\|B_k-B^\star\|/\gamma_H$ \cite[Lemma 5.2]{Na2025Statistical}, the $\Upsilon_L$-Lipschitz~continuity of $\nabla^2 F(\bx)$, and \eqref{ass:3:c2}. Thus, we take expectation and obtain
\begin{align}\label{appen:A5:equ6}
\mE\big[\|\bdelta^k\|^2\big] &\lesssim \Big(\frac{\tau^2\Upsilon_{S}}{\gamma_H^2} +\frac{\Upsilon_H^2}{\gamma_H^4}\Big) \mE\big[\|B_k-B^\star\|^2\|\bx_k-\bx^\star\|^2\big] + \frac{\Upsilon_L^2}{\gamma_H^2}\mE\big[\|\bx_k-\bx^\star\|^4\big]\nonumber\\
&\lesssim \Big(\frac{\tau^2\Upsilon_{S}}{\gamma_H^2}+\frac{\Upsilon_H^2}{\gamma_H^4}\Big)\sqrt{\mE\big[\|B_k-B^\star\|^4\big]}\sqrt{\mE\big[\|\bx_k-\bx^\star\|^4\big]} +\frac{\Upsilon_L^2}{\gamma_H^2} 
\mE\big[\|\bx_k-\bx^\star\|^4\big]\nonumber\\
&\lesssim \Big(\frac{\tau^2\Upsilon_S}{\gamma_H^2}+\frac{\Upsilon_H^2}{\gamma_H^4}\Big)\cdot \frac{\Upsilon_L^2\Upsilon_H^4C_{g,2}}{\gamma_H^8}\beta_t^2 \eqqcolon C_{\bdelta}\beta_t^2,
\end{align}
where the last inequality follows from Lemma \ref{sec4:lem1} (particularly \eqref{xtrate} and \eqref{appen:A3:equ15} in the proof) and the observation $\chi>1.5\beta \Rightarrow \chi_t^4/\beta_t^4 = o(\beta_t^2)$. We plug \eqref{appen:A5:equ6} into \eqref{appen:A5:equ11}, apply Lemma \ref{aux:lem1}, and get
\begin{align}\label{appen:A5:equ18}
\mE\big[\|\bar{\I}_{3,t}\|^2\big]&\lesssim \Big(\frac{1}{t}\sum_{i=0}^{t-1}\underbrace{\prod_{k=0}^i(1-(1-\rho^\tau)\varphi_k)}_{=o(\chi_t^2/\beta_t^2+\beta_t)\quad\text{by \eqref{appen:A1:equ1b}}}\Big)^2\|\bx_0-\bx^\star\|^2 + C_{\bdelta}\Big(\frac{1}{t}\sum_{i=0}^{t-1}\underbrace{\sum_{k=0}^i\prod_{l=k+1}^i(1-(1-\rho^\tau)\varphi_l)\varphi_k\cdot \beta_k}_{\lesssim \beta_i/(1-\rho^\tau)\quad\text{by \eqref{appen:A1:equ1a}}}\Big)^2\nonumber\\
&\stackrel{\mathclap{\eqref{appen:A3:equ11}}}{\lesssim}\;\; \frac{C_{\bdelta}}{(1-\rho^\tau)^2(1-\beta)^2}\beta_t^2.
\end{align}
We recall the fact that $\frac{1}{t}\sum_{i=0}^{t-1}\frac{1}{\varphi_i}\lesssim 1/\beta_t$ in \eqref{appen:A4:equ26},  combine \eqref{appen:A5:equ4},  \eqref{appen:A5:equ12}, \eqref{appen:A5:equ13}, and \eqref{appen:A5:equ18} together, and obtain
\begin{footnotesize}
\begin{equation}\label{appen:A5:equ22}
\hskip-0.37cm\frac{1}{t}\sum_{i=0}^{t-1}\frac{1}{\varphi_i}\mE\big[\|\bar{\bx}_t-\bx^\star\|^2\big]
\lesssim \begin{dcases}
\frac{C_{\bdelta}}{(1-\rho^\tau)^2}\beta_t + \frac{C_{g,2}^{1/2}\mathbf{1}_{\{\chi\leq 2\beta\}}}{\gamma_H^2(1-\rho^\tau)^2 }\cdot \frac{\chi_t^2}{\beta_t^3} = O\rbr{\frac{1}{ (1-\rho^\tau)^{2}}\cbr{\beta_t + \frac{\chi_t^2}{\beta_t^3}}}, &\beta\in(0,0.5),\\
\max\rbr{\frac{C_{g,2}^{1/2}}{c_{\beta}^2 \gamma_H^2(1-\rho^\tau)^2}, \frac{ C_{\bdelta}}{(1-\rho^\tau)^2}}\beta_t + \frac{C_{g,2}^{1/2}\mathbf{1}_{\{\chi\leq 2\beta\}}}{\gamma_H^2(1-\rho^\tau)^2 }\cdot \frac{\chi_t^2}{\beta_t^3} = O\rbr{\frac{\beta_t+\chi_t^2/\beta_t^3}{ (1-\rho^\tau)^{2}}} &\beta=0.5,\\
\frac{C_{g,2}^{1/2}}{\gamma_H^2(1-\rho^\tau)^2}\cdot\frac{1}{t\beta_t} + \frac{C_{g,2}^{1/2}\mathbf{1}_{\{\chi\leq \beta+0.5\}}}{\gamma_H^2(1-\rho^\tau)^2}\cdot \frac{\chi_t^2}{\beta_t^3} = O\rbr{\frac{1}{ (1-\rho^\tau)^{2}}\cbr{\frac{1}{t\beta_t} + \frac{\chi_t^2}{\beta_t^3}}}, &\beta\in(0.5,1).
\end{dcases}
\end{equation}
\end{footnotesize}
\hskip-3.5pt Here follows the same discussion as in \eqref{appen:A5:equ16}. This completes the proof.

\subsubsection{Proof of Lemma \ref{appen:A5:lem3}}\label{pf:I2}

Based on the definition of $\I_{2,i}$ in \eqref{rec:b}, we apply Lemma \ref{aux:lem4}(d) and the fact that $|\bar{\alpha}_k-\varphi_k|\leq \chi_k/2$, then we have
\begin{equation*}
\begin{aligned}
\mE\|\I_{2,i}\|^2 &\leq \mE\sbr{\rbr{\sum_{k=0}^i \prod_{l=k+1}^i(1-(1-\rho^\tau)\varphi_l) \frac{\chi_k}{2}\|\bar{\Delta} \bx_k\|}^2}\\
& \lesssim \Big(\sum_{k=0}^i \prod_{l=k+1}^i(1-(1-\rho^\tau)\varphi_l) \chi_k\sqrt{\mE\big[\|\bar{\Delta} \bx_k\|^2\big]}\Big)^2\quad(\text{by H\"older's inequality})\\
&\stackrel{\mathclap{\eqref{appen:A5:equ5}}}{\lesssim}\;\; \frac{C_{g,2}^{1/2}}{\gamma_H^2}\Big(\sum_{k=0}^i \prod_{l=k+1}^i(1-(1-\rho^\tau)\varphi_l)\varphi_k\cdot\frac{\chi_k}{\varphi_k}\Big)^2 \lesssim \frac{C_{g,2}^{1/2}}{\gamma_H^2(1-\rho^\tau)^2}\cdot \frac{\chi_i^2}{\varphi_i^2}\quad(\text{by Lemma \ref{aux:lem1}}).    
\end{aligned}
\end{equation*}
With the above display, we obtain
\begin{align}\label{appen:A5:equ14}
&\frac{1}{t}\sum_{i=0}^{t-1}\frac{1}{\varphi_i}\mE\big[\|\I_{2,i}\|^2\big]\cdot \mathbf{1}_{\{\chi< 1.5\beta+0.5\}} \lesssim \frac{C_{g,2}^{1/2}\mathbf{1}_{\{\chi< 1.5\beta+0.5\}}}{\gamma_H^2(1-\rho^\tau)^2}\cdot \frac{1}{t}\sum_{i=0}^{t-1}\frac{\chi_i^2}{\varphi_i^3}\nonumber\\
&\quad\quad \stackrel{\mathclap{\eqref{appen:A3:equ11}}}{\lesssim}\;\; \frac{C_{g,2}^{1/2}\mathbf{1}_{\{\chi< 1.5\beta+0.5\}}}{\gamma_H^2(1-\rho^\tau)^2(1-(2\chi-3\beta))}\cdot\frac{\chi_t^2}{\beta_t^3} = O\rbr{\frac{\chi_t^2}{{ (1-\rho^\tau)^2}\beta_t^3}}\cdot \mathbf{1}_{\{\chi< 1.5\beta+0.5\}},\\
&\frac{1}{t}\sum_{i=0}^{t-1}\frac{1}{\varphi_i}\mE\big[\|\I_{2,i}\|^2\big]\cdot \mathbf{1}_{\{\chi\geq 1.5\beta+0.5\}} = \frac{1}{t}\sum_{i=0}^{t-1} o(\beta_t)\cdot \mathbf{1}_{\{\chi\geq 1.5\beta+0.5\}} \stackrel{\eqref{appen:A3:equ11}}{=} o(\beta_t)\cdot \mathbf{1}_{\{\chi\geq 1.5\beta+0.5\}}.    \nonumber
\end{align}
This completes the proof.

\subsubsection{Proof of Lemma \ref{appen:A5:lem4}}\label{pf:I3}

Given the expression of $\I_{3,i}$ in \eqref{rec:c}, we apply Lemma \ref{aux:lem4}(d) and have
\begin{align*}
\mE\big[\|\I_{3, i}\|^2\big] &\lesssim \prod_{k=0}^i(1-(1-\rho^{\tau})\varphi_k)^2\|\bx_0-\bx^\star\|^2 + \Big(\sum_{k=0}^i\prod_{l=k+1}^i(1-(1-\rho)^\tau)\varphi_l) \varphi_k\|\bdelta^k\|\Big)^2\\
&\leq \prod_{k=0}^i(1-(1-\rho^{\tau})\varphi_k)^2\|\bx_0-\bx^\star\|^2 + \Big(\sum_{k=0}^i\prod_{l=k+1}^i(1-(1-\rho^\tau)\varphi_l) \varphi_k \sqrt{\mE\big[\|\bdelta^k\|^2\big]}\Big)^2,
\end{align*}
where the last inequality is due to H\"older's inequality. We plugging in \eqref{appen:A5:equ6}, apply Lemma \ref{aux:lem1}, and get
\begin{align}\label{appen:A5:equ15}
\frac{1}{t} \sum_{i=0}^{t-1} \frac{1}{\varphi_i} \mE\big[\|\I_{3, i}\|^2\big] &\lesssim  \frac{1}{t} \sum_{i=0}^{t-1} \frac{1}{\varphi_i} \Big(\underbrace{\prod_{k=0}^i(1-(1-\rho^{\tau})\varphi_k)}_{=o(\beta_i)\quad\text{by \eqref{appen:A1:equ1a}}}\Big)^2\cdot \|\bx_0-\bx^\star\|^2\nonumber\\
&\quad\quad + \frac{C_{\bdelta}}{t} \sum_{i=0}^{t-1} \frac{1}{\varphi_i}\bigg(\underbrace{\sum_{k=0}^i \prod_{l=k+1}^i (1-(1-\rho^\tau)\varphi_l) \varphi_k \cdot\beta_k}_{\lesssim \beta_i/(1-\rho^\tau)\quad\text{by \eqref{appen:A1:equ1b}}} \bigg)^2\nonumber\\
&\lesssim \frac{C_{\bdelta}}{(1-\rho^\tau)^2}\cdot \frac{1}{t}\sum_{i=0}^{t-1}\beta_i \stackrel{\eqref{appen:A3:equ11}}{\lesssim} \frac{C_{\bdelta}}{(1-\rho^\tau)^2(1-\beta)}\beta_t = O\rbr{\beta_t/{ (1-\rho^\tau)^2}}.
\end{align}
This completes the proof.

\section{Additional Experiment Results}\label{appen:exp}

In this section, we complement Section \ref{sec:5} by presenting additional experimental~results~on regression problems. Specifically, we evaluate the performance of three online covariance estimators ($\bar{\Xi}_t, \tilde{\Xi}_t$,~and $\hat{\Xi}_t$) across different design covariance matrices $\Sigma_a$. Following the experimental setup in~Sections~\ref{sec5:linear} and \ref{sec5:logistic}, we construct 95\% confidence intervals for $\sum_{i=1}^d \bx^\star_i / d$. To assess performance, we vary $r \in \{0.4, 0.5, 0.6\}$ for Toeplitz $\Sigma_a$ and $r \in \{0.1, 0.2, 0.3\}$ for Equi-correlation $\Sigma_a$. Tables \ref{appen:table:1} -- \ref{appen:table:4}~summarize the empirical coverage rates of the confidence intervals and the averaged relative variance estimation error for $\sum_{i=1}^d (\bx_t)_i / d$ at the final iteration.

Overall, the results align with the analyses in Sections \ref{sec5:linear} and \ref{sec5:logistic}. These results further~demonstrate the superior performance of $\hat{\Xi}_t$ in statistical inference compared to $\bar{\Xi}_t$ and $\tilde{\Xi}_t$. Regarding the influence of $r$, a general trend is that increasing $r$ makes the problem more challenging. This is because a larger $r$ increases the condition number of $\Sigma_a$, which leads to harder problems. This can be observed in several ways. 
First, for Toeplitz $\Sigma_a$, $\hat{\Xi}_t$ performs well when $r = 0.4$ and $0.5$. However,~for $r = 0.6$ and $d = 100$, both $\tilde{\Xi}_t$ and $\bar{\Xi}_t$ fail to converge. Although their performance improves when $\tau$ increases from 10 to 40, neither achieves convergence for $r = 0.6$. This suggests that the iterate $\bx_t$ does not converge well, and $\tau = 40$ is insufficient to achieve desirable accuracy in approximating~the~Newton~direction.~Second, in Table \ref{appen:table:4}, we observe that the coverage rate corresponding to~$\bar{\Xi}_t$ decreases as $r$ increases from $0.1$ to $0.3$. Similarly, in Tables \ref{appen:table:3} and \ref{appen:table:4}, for the bold settings, the coverage rate corresponding to $\tilde{\Xi}_t$ decreases as $r$ increases. These results reinforce the impact of $r$ on problem difficulty and highlight the robustness of $\hat{\Xi}_t$ across different scenarios.

\begin{table}[!t]
\centering
\resizebox{\linewidth}{!}{
\begin{tabular}{|c|c|c|c|cccccccc|}
\hline
\multirow{3}{*}{Toeplitz $\Sigma_a$} &
\multirow{3}{*}{d} &
\multirow{3}{*}{Criterion} &
\multirow{2}{*}{SGD} &
\multicolumn{8}{c|}{Sketched Newton Method} \\ \cline{5-12} 
&
&
&
&
\multicolumn{2}{c|}{$\tau=\infty$} &
\multicolumn{2}{c|}{$\tau=10$} &
\multicolumn{2}{c|}{$\tau=20$} &
\multicolumn{2}{c|}{$\tau=40$} \\ \cline{4-12} 
&
&
&
{\footnotesize$\bar{\Xi}_t$} &
{\footnotesize$\tilde{\Xi}_t$} &
\multicolumn{1}{c|}{{\footnotesize$\hat{\Xi}_t$}} &
{\footnotesize$\tilde{\Xi}_t$} &
\multicolumn{1}{c|}{{\footnotesize$\hat{\Xi}_t$}} &
{\footnotesize$\tilde{\Xi}_t$} &
\multicolumn{1}{c|}{{\footnotesize$\hat{\Xi}_t$}} &
{\footnotesize$\tilde{\Xi}_t$} &
{\footnotesize$\hat{\Xi}_t$} \\ \hline
\multirow{8}{*}{r=0.4} &
\multirow{2}{*}{20} &
Cov (\%) &
92.00 &
93.50 &
\multicolumn{1}{c|}{93.50} &
92.00 &
\multicolumn{1}{c|}{97.00} &
94.50 &
\multicolumn{1}{c|}{97.00} &
91.00 &
93.00 \\
&
&
Var Err &
-0.166 &
0.025 &
\multicolumn{1}{c|}{0.017} &
-0.321 &
\multicolumn{1}{c|}{0.014} &
-0.266 &
\multicolumn{1}{c|}{0.015} &
-0.177 &
0.014 \\ \cline{2-12} 
&
\multirow{2}{*}{40} &
Cov (\%) &
92.50 &
92.50 &
\multicolumn{1}{c|}{92.00} &
87.50 &
\multicolumn{1}{c|}{96.50} &
84.50 &
\multicolumn{1}{c|}{93.00} &
90.50 &
97.50 \\
&
&
Var Err &
-0.095 &
0.049 &
\multicolumn{1}{c|}{0.051} &
-0.350 &
\multicolumn{1}{c|}{0.030} &
-0.319 &
\multicolumn{1}{c|}{0.029} &
-0.263 &
0.031 \\ \cline{2-12} 
&
\multirow{2}{*}{60} &
Cov (\%) &
90.50 &
95.00 &
\multicolumn{1}{c|}{95.00} &
91.00 &
\multicolumn{1}{c|}{97.50} &
84.00 &
\multicolumn{1}{c|}{93.50} &
89.00 &
95.00 \\
&
&
Var Err &
-0.112 &
0.072 &
\multicolumn{1}{c|}{0.066} &
-0.350 &
\multicolumn{1}{c|}{0.048} &
-0.332 &
\multicolumn{1}{c|}{0.041} &
-0.292 &
0.060 \\ \cline{2-12} 
&
\multirow{2}{*}{100} &
Cov (\%) &
90.50 &
100.0 &
\multicolumn{1}{c|}{100.0} &
87.50 &
\multicolumn{1}{c|}{95.50} &
92.00 &
\multicolumn{1}{c|}{98.00} &
90.00 &
97.00 \\
&
&
Var Err &
-0.100 &
$\infty$ &
\multicolumn{1}{c|}{$\infty$} &
-0.313 &
\multicolumn{1}{c|}{0.128} &
-0.327 &
\multicolumn{1}{c|}{0.088} &
-0.303 &
0.085 \\ \hline
\multirow{8}{*}{r=0.5} &
\multirow{2}{*}{20} &
Cov (\%) &
\textbf{87.00} &
94.50 &
\multicolumn{1}{c|}{94.50} &
\textbf{89.00} &
\multicolumn{1}{c|}{\textbf{94.00}} &
89.00 &
\multicolumn{1}{c|}{94.00} &
90.00 &
93.00 \\
&
&
Var Err &
\textbf{-0.104} &
0.025 &
\multicolumn{1}{c|}{0.026} &
\textbf{-0.339} &
\multicolumn{1}{c|}{\textbf{0.003}} &
-0.283 &
\multicolumn{1}{c|}{0.009} &
-0.208 &
0.018 \\ \cline{2-12} 
&
\multirow{2}{*}{40} &
Cov (\%) &
91.00 &
96.50 &
\multicolumn{1}{c|}{96.50} &
89.50 &
\multicolumn{1}{c|}{94.00} &
85.50 &
\multicolumn{1}{c|}{95.50} &
89.00 &
94.50 \\
&
&
Var Err &
-0.074 &
0.048 &
\multicolumn{1}{c|}{0.040} &
-0.376 &
\multicolumn{1}{c|}{0.016} &
-0.343 &
\multicolumn{1}{c|}{0.022} &
-0.285 &
0.029 \\ \cline{2-12} 
&
\multirow{2}{*}{60} &
Cov (\%) &
86.50 &
94.00 &
\multicolumn{1}{c|}{94.50} &
83.50 &
\multicolumn{1}{c|}{92.50} &
85.50 &
\multicolumn{1}{c|}{93.00} &
84.50 &
94.00 \\
&
&
Var Err &
-0.061 &
0.072 &
\multicolumn{1}{c|}{0.074} &
-0.383 &
\multicolumn{1}{c|}{0.044} &
-0.361 &
\multicolumn{1}{c|}{0.029} &
-0.317 &
0.046 \\ \cline{2-12} 
&
\multirow{2}{*}{100} &
Cov (\%) &
93.50 &
100.0 &
\multicolumn{1}{c|}{100.0} &
90.00 &
\multicolumn{1}{c|}{96.00} &
89.00 &
\multicolumn{1}{c|}{95.00} &
89.50 &
97.00 \\
&
&
Var Err &
-0.083 &
$\infty$ &
\multicolumn{1}{c|}{$\infty$} &
1.156 &
\multicolumn{1}{c|}{2.659} &
-0.069 &
\multicolumn{1}{c|}{0.582} &
-0.335 &
0.067 \\ \hline
\multirow{8}{*}{r=0.6} &
\multirow{2}{*}{20} &
Cov (\%) &
92.00 &
95.00 &
\multicolumn{1}{c|}{95.50} &
88.00 &
\multicolumn{1}{c|}{93.50} &
87.50 &
\multicolumn{1}{c|}{94.00} &
91.50 &
95.00 \\
&
&
Var Err &
-0.110 &
0.024 &
\multicolumn{1}{c|}{0.031} &
-0.338 &
\multicolumn{1}{c|}{0.003} &
-0.285 &
\multicolumn{1}{c|}{0.004} &
-0.225 &
0.004 \\ \cline{2-12} 
&
\multirow{2}{*}{40} &
Cov (\%) &
89.50 &
95.00 &
\multicolumn{1}{c|}{95.00} &
88.50 &
\multicolumn{1}{c|}{94.50} &
91.50 &
\multicolumn{1}{c|}{96.00} &
92.00 &
96.50 \\
&
&
Var Err &
-0.115 &
0.048 &
\multicolumn{1}{c|}{0.043} &
-0.381 &
\multicolumn{1}{c|}{0.023} &
-0.349 &
\multicolumn{1}{c|}{0.015} &
-0.294 &
0.017 \\ \cline{2-12} 
&
\multirow{2}{*}{60} &
Cov (\%) &
\textbf{89.50} &
97.00 &
\multicolumn{1}{c|}{98.00} &
86.50 &
\multicolumn{1}{c|}{98.00} &
84.00 &
\multicolumn{1}{c|}{94.50} &
\textbf{87.50} &
\textbf{95.50} \\
&
&
Var Err &
\textbf{-0.079} &
0.073 &
\multicolumn{1}{c|}{0.062} &
-0.290 &
\multicolumn{1}{c|}{0.232} &
-0.359 &
\multicolumn{1}{c|}{0.058} &
\textbf{-0.327} &
\textbf{0.037} \\ \cline{2-12} 
&
\multirow{2}{*}{100} &
Cov (\%) &
92.50 &
100.0 &
\multicolumn{1}{c|}{100.0} &
96.50 &
\multicolumn{1}{c|}{99.00} &
97.50 &
\multicolumn{1}{c|}{98.50} &
95.00 &
99.00 \\
&
&
Var Err &
-0.036 &
$\infty$ &
\multicolumn{1}{c|}{$\infty$} &
119.0 &
\multicolumn{1}{c|}{203.4} &
127.3 &
\multicolumn{1}{c|}{232.9} &
29.02 &
49.50 \\ \hline		
\end{tabular}}
\caption{\textit{Linear regression with Toeplitz $\Sigma_a$ across $r\in\{0.4, 0.5, 0.6\}$: the empirical coverage rate of 95\% confidence intervals $(\textit{Cov})$ and the averaged relative estimation error of the variance $(\textit{Var Err})$ of $\b1^T\bx_t/d$, given by~\mbox{$\b1^T(\hat{\Xi}_t-\Xi^\star)\b1/\b1^T\Xi^\star\b1$}. We bold entries to highlight scenarios~where~$\hat{\Xi}_t$ performs significantly better than others.}}
\label{appen:table:1}
\end{table}

\begin{table}[!h]
\centering
\resizebox{\linewidth}{!}{
\begin{tabular}{|c|c|c|c|cccccccc|}
\hline
\multirow{3}{*}{Equi-corr $\Sigma_a$} &
\multirow{3}{*}{d} &
\multirow{3}{*}{Criterion} &
\multirow{2}{*}{SGD} &
\multicolumn{8}{c|}{Sketched Newton Method} \\ \cline{5-12} 
&
&
&
&
\multicolumn{2}{c|}{$\tau=\infty$} &
\multicolumn{2}{c|}{$\tau=10$} &
\multicolumn{2}{c|}{$\tau=20$} &
\multicolumn{2}{c|}{$\tau=40$} \\ \cline{4-12} 
&
&
&
{\footnotesize$\bar{\Xi}_t$} &
{\footnotesize$\tilde{\Xi}_t$} &
\multicolumn{1}{c|}{{\footnotesize$\hat{\Xi}_t$}} &
{\footnotesize$\tilde{\Xi}_t$} &
\multicolumn{1}{c|}{{\footnotesize$\hat{\Xi}_t$}} &
{\footnotesize$\tilde{\Xi}_t$} &
\multicolumn{1}{c|}{{\footnotesize$\hat{\Xi}_t$}} &
{\footnotesize$\tilde{\Xi}_t$} &
{\footnotesize$\hat{\Xi}_t$} \\ \hline
\multirow{8}{*}{r=0.1} &
\multirow{2}{*}{20} &
Cov (\%) &
90.50 &
92.50 &
\multicolumn{1}{c|}{93.00} &
85.50 &
\multicolumn{1}{c|}{97.00} &
87.00 &
\multicolumn{1}{c|}{93.00} &
91.50 &
94.50 \\
&
&
Var Err &
-0.101 &
0.025 &
\multicolumn{1}{c|}{0.024} &
-0.459 &
\multicolumn{1}{c|}{0.017} &
-0.355 &
\multicolumn{1}{c|}{0.017} &
-0.183 &
0.017 \\ \cline{2-12} 
&
\multirow{2}{*}{40} &
Cov (\%) &
\textbf{91.50} &
93.00 &
\multicolumn{1}{c|}{93.00} &
\textbf{80.00} &
\multicolumn{1}{c|}{\textbf{94.50}} &
85.00 &
\multicolumn{1}{c|}{97.00} &
82.00 &
94.50 \\
&
&
Var Err &
\textbf{-0.118} &
0.048 &
\multicolumn{1}{c|}{0.046} &
\textbf{-0.612} &
\multicolumn{1}{c|}{\textbf{0.027}} &
-0.565 &
\multicolumn{1}{c|}{0.034} &
-0.467 &
0.037 \\ \cline{2-12} 
&
\multirow{2}{*}{60} &
Cov (\%) &
93.50 &
94.50 &
\multicolumn{1}{c|}{92.50} &
71.00 &
\multicolumn{1}{c|}{94.00} &
75.00 &
\multicolumn{1}{c|}{95.50} &
77.00 &
93.00 \\
&
&
Var Err &
-0.059 &
0.072 &
\multicolumn{1}{c|}{0.070} &
-0.681 &
\multicolumn{1}{c|}{0.048} &
-0.655 &
\multicolumn{1}{c|}{0.046} &
-0.600 &
0.047 \\ \cline{2-12} 
&
\multirow{2}{*}{100} &
Cov (\%) &
92.50 &
100.0 &
\multicolumn{1}{c|}{100.0} &
68.00 &
\multicolumn{1}{c|}{96.50} &
64.50 &
\multicolumn{1}{c|}{93.50} &
70.50 &
98.00 \\
&
&
Var Err &
-0.062 &
$\infty$ &
\multicolumn{1}{c|}{$\infty$} &
-0.748 &
\multicolumn{1}{c|}{0.072} &
-0.737 &
\multicolumn{1}{c|}{0.070} &
-0.712 &
0.075 \\ \hline
\multirow{8}{*}{r=0.2} &
\multirow{2}{*}{20} &
Cov (\%) &
92.00 &
93.00 &
\multicolumn{1}{c|}{92.50} &
79.00 &
\multicolumn{1}{c|}{94.00} &
83.00 &
\multicolumn{1}{c|}{94.00} &
91.50 &
95.50 \\
&
&
Var Err &
-0.063 &
0.024 &
\multicolumn{1}{c|}{0.023} &
-0.538 &
\multicolumn{1}{c|}{0.013} &
-0.468 &
\multicolumn{1}{c|}{0.016} &
-0.334 &
0.012 \\ \cline{2-12} 
&
\multirow{2}{*}{40} &
Cov (\%) &
90.50 &
95.50 &
\multicolumn{1}{c|}{94.50} &
75.00 &
\multicolumn{1}{c|}{96.50} &
82.50 &
\multicolumn{1}{c|}{96.50} &
80.50 &
94.50 \\
&
&
Var Err &
-0.139 &
0.048 &
\multicolumn{1}{c|}{0.040} &
-0.654 &
\multicolumn{1}{c|}{0.022} &
-0.630 &
\multicolumn{1}{c|}{0.018} &
-0.580 &
0.024 \\ \cline{2-12} 
&
\multirow{2}{*}{60} &
Cov (\%) &
91.00 &
95.50 &
\multicolumn{1}{c|}{95.50} &
72.00 &
\multicolumn{1}{c|}{91.50} &
68.00 &
\multicolumn{1}{c|}{94.50} &
81.50 &
96.50 \\
&
&
Var Err &
-0.015 &
0.072 &
\multicolumn{1}{c|}{0.067} &
-0.697 &
\multicolumn{1}{c|}{0.019} &
-0.685 &
\multicolumn{1}{c|}{0.027} &
-0.660 &
0.029 \\ \cline{2-12} 
&
\multirow{2}{*}{100} &
Cov (\%) &
93.50 &
100.0 &
\multicolumn{1}{c|}{100.0} &
69.50 &
\multicolumn{1}{c|}{96.50} &
68.00 &
\multicolumn{1}{c|}{97.50} &
73.00 &
97.50 \\
&
&
Var Err &
-0.022 &
$\infty$ &
\multicolumn{1}{c|}{$\infty$} &
-0.732 &
\multicolumn{1}{c|}{0.030} &
-0.727 &
\multicolumn{1}{c|}{0.028} &
-0.718 &
0.035 \\ \hline
\multirow{8}{*}{r=0.3} &
\multirow{2}{*}{20} &
Cov (\%) &
94.00 &
95.00 &
\multicolumn{1}{c|}{96.50} &
83.50 &
\multicolumn{1}{c|}{98.50} &
85.50 &
\multicolumn{1}{c|}{93.50} &
89.50 &
94.00 \\
&
&
Var Err &
-0.057 &
0.025 &
\multicolumn{1}{c|}{0.026} &
-0.543 &
\multicolumn{1}{c|}{0.010} &
-0.504 &
\multicolumn{1}{c|}{0.010} &
-0.424 &
0.008 \\ \cline{2-12} 
&
\multirow{2}{*}{40} &
Cov (\%) &
\textbf{90.50} &
97.50 &
\multicolumn{1}{c|}{97.50} &
\textbf{73.00} &
\multicolumn{1}{c|}{\textbf{94.50}} &
76.00 &
\multicolumn{1}{c|}{96.50} &
85.50 &
95.50 \\
&
&
Var Err &
\textbf{-0.106} &
0.048 &
\multicolumn{1}{c|}{0.048} &
\textbf{-0.617} &
\multicolumn{1}{c|}{\textbf{0.014}} &
-0.605 &
\multicolumn{1}{c|}{0.017} &
-0.581 &
0.007 \\ \cline{2-12} 
&
\multirow{2}{*}{60} &
Cov (\%) &
92.50 &
96.00 &
\multicolumn{1}{c|}{95.50} &
73.00 &
\multicolumn{1}{c|}{93.00} &
72.00 &
\multicolumn{1}{c|}{94.00} &
72.50 &
94.00 \\
&
&
Var Err &
-0.035 &
0.073 &
\multicolumn{1}{c|}{0.065} &
-0.643 &
\multicolumn{1}{c|}{0.010} &
-0.637 &
\multicolumn{1}{c|}{0.013} &
-0.625 &
0.005 \\ \cline{2-12} 
&
\multirow{2}{*}{100} &
Cov (\%) &
91.00 &
100.0 &
\multicolumn{1}{c|}{100.0} &
78.00 &
\multicolumn{1}{c|}{97.00} &
72.50 &
\multicolumn{1}{c|}{95.00} &
73.50 &
94.50 \\
&
&
Var Err &
0.002 &
$\infty$ &
\multicolumn{1}{c|}{$\infty$} &
-0.416 &
\multicolumn{1}{c|}{0.805} &
-0.658 &
\multicolumn{1}{c|}{0.018} &
-0.656 &
0.014 \\ 
\hline		
\end{tabular}}
\caption{\textit{Linear regression with Equi-correlation $\Sigma_a$ across $r\in\{0.1, 0.2, 0.3\}$. See Table \ref{appen:table:1} for interpretation.}}
\label{appen:table:2}
\end{table}

\begin{table}[!h]
\centering
\resizebox{\linewidth}{!}{
\begin{tabular}{|c|c|c|c|cccccccc|}
\hline
\multirow{3}{*}{Toeplitz $\Sigma_a$} &
\multirow{3}{*}{d} &
\multirow{3}{*}{Criterion} &
\multirow{2}{*}{SGD} &
\multicolumn{8}{c|}{Sketched Newton Method} \\ \cline{5-12} 
&
&
&
&
\multicolumn{2}{c|}{$\tau=\infty$} &
\multicolumn{2}{c|}{$\tau=10$} &
\multicolumn{2}{c|}{$\tau=20$} &
\multicolumn{2}{c|}{$\tau=40$} \\ \cline{4-12} 
&
&
&
{\footnotesize$\bar{\Xi}_t$} &
{\footnotesize$\tilde{\Xi}_t$} &
\multicolumn{1}{c|}{{\footnotesize$\hat{\Xi}_t$}} &
{\footnotesize$\tilde{\Xi}_t$} &
\multicolumn{1}{c|}{{\footnotesize$\hat{\Xi}_t$}} &
{\footnotesize$\tilde{\Xi}_t$} &
\multicolumn{1}{c|}{{\footnotesize$\hat{\Xi}_t$}} &
{\footnotesize$\tilde{\Xi}_t$} &
{\footnotesize$\hat{\Xi}_t$} \\ \hline
\multirow{8}{*}{r=0.4} &
\multirow{2}{*}{20} &
Cov (\%) &
87.50 &
94.50 &
\multicolumn{1}{c|}{94.50} &
93.50 &
\multicolumn{1}{c|}{97.50} &
92.00 &
\multicolumn{1}{c|}{96.00} &
92.50 &
95.00 \\
&
&
Var Err &
-0.226 &
0.040 &
\multicolumn{1}{c|}{0.035} &
-0.197 &
\multicolumn{1}{c|}{0.026} &
-0.159 &
\multicolumn{1}{c|}{0.032} &
-0.079 &
0.033 \\ \cline{2-12} 
&
\multirow{2}{*}{40} &
Cov (\%) &
85.00 &
96.00 &
\multicolumn{1}{c|}{96.00} &
94.50 &
\multicolumn{1}{c|}{96.00} &
88.00 &
\multicolumn{1}{c|}{93.00} &
91.50 &
94.50 \\
&
&
Var Err &
-0.227 &
0.085 &
\multicolumn{1}{c|}{0.074} &
-0.190 &
\multicolumn{1}{c|}{0.077} &
-0.180 &
\multicolumn{1}{c|}{0.079} &
-0.141 &
0.073 \\ \cline{2-12} 
&
\multirow{2}{*}{60} &
Cov (\%) &
\textbf{86.00} &
93.50 &
\multicolumn{1}{c|}{93.00} &
\textbf{93.00} &
\multicolumn{1}{c|}{\textbf{96.50}} &
93.00 &
\multicolumn{1}{c|}{96.50} &
93.00 &
96.00 \\
&
&
Var Err &
\textbf{-0.249} &
0.130 &
\multicolumn{1}{c|}{0.122} &
\textbf{-0.164} &
\multicolumn{1}{c|}{\textbf{0.102}} &
-0.163 &
\multicolumn{1}{c|}{0.100} &
-0.140 &
0.113 \\ \cline{2-12} 
&
\multirow{2}{*}{100} &
Cov (\%) &
87.50 &
94.50 &
\multicolumn{1}{c|}{94.50} &
87.50 &
\multicolumn{1}{c|}{92.00} &
92.50 &
\multicolumn{1}{c|}{95.00} &
95.50 &
98.00 \\
&
&
Var Err &
-0.138 &
0.232 &
\multicolumn{1}{c|}{0.220} &
-0.093 &
\multicolumn{1}{c|}{0.184} &
-0.098 &
\multicolumn{1}{c|}{0.192} &
-0.093 &
0.190 \\ \hline
\multirow{8}{*}{r=0.5} &
\multirow{2}{*}{20} &
Cov (\%) &
88.00 &
98.50 &
\multicolumn{1}{c|}{98.00} &
93.50 &
\multicolumn{1}{c|}{96.00} &
94.00 &
\multicolumn{1}{c|}{95.50} &
93.50 &
94.50 \\
&
&
Var Err &
-0.199 &
0.038 &
\multicolumn{1}{c|}{0.032} &
-0.226 &
\multicolumn{1}{c|}{0.028} &
-0.179 &
\multicolumn{1}{c|}{0.028} &
-0.097 &
0.020 \\ \cline{2-12} 
&
\multirow{2}{*}{40} &
Cov (\%) &
85.50 &
96.00 &
\multicolumn{1}{c|}{96.50} &
91.50 &
\multicolumn{1}{c|}{95.00} &
90.00 &
\multicolumn{1}{c|}{93.50} &
95.00 &
97.50 \\
&
&
Var Err &
-0.190 &
0.079 &
\multicolumn{1}{c|}{0.077} &
-0.233 &
\multicolumn{1}{c|}{0.063} &
-0.217 &
\multicolumn{1}{c|}{0.066} &
-0.160 &
0.064 \\ \cline{2-12} 
&
\multirow{2}{*}{60} &
Cov (\%) &
92.00 &
95.50 &
\multicolumn{1}{c|}{94.50} &
\textbf{92.00} &
\multicolumn{1}{c|}{\textbf{94.50}} &
89.50 &
\multicolumn{1}{c|}{94.00} &
92.50 &
97.00 \\
&
&
Var Err &
-0.170 &
0.122 &
\multicolumn{1}{c|}{0.115} &
\textbf{-0.215} &
\multicolumn{1}{c|}{\textbf{0.081}} &
-0.208 &
\multicolumn{1}{c|}{0.100} &
-0.174 &
0.094 \\ \cline{2-12} 
&
\multirow{2}{*}{100} &
Cov (\%) &
87.50 &
97.00 &
\multicolumn{1}{c|}{96.00} &
90.50 &
\multicolumn{1}{c|}{93.50} &
92.00 &
\multicolumn{1}{c|}{96.50} &
90.00 &
93.50 \\
&
&
Var Err &
-0.159 &
0.221 &
\multicolumn{1}{c|}{0.215} &
-0.158 &
\multicolumn{1}{c|}{0.163} &
-0.161 &
\multicolumn{1}{c|}{0.166} &
-0.146 &
0.164 \\ \hline
\multirow{8}{*}{r=0.6} &
\multirow{2}{*}{20} &
Cov (\%) &
86.00 &
91.00 &
\multicolumn{1}{c|}{90.50} &
91.00 &
\multicolumn{1}{c|}{94.50} &
93.00 &
\multicolumn{1}{c|}{94.50} &
92.50 &
94.00 \\
&
&
Var Err &
-0.188 &
0.036 &
\multicolumn{1}{c|}{0.023} &
-0.241 &
\multicolumn{1}{c|}{0.032} &
-0.185 &
\multicolumn{1}{c|}{0.031} &
-0.113 &
0.028 \\ \cline{2-12} 
&
\multirow{2}{*}{40} &
Cov (\%) &
84.50 &
95.50 &
\multicolumn{1}{c|}{95.00} &
88.00 &
\multicolumn{1}{c|}{94.50} &
87.50 &
\multicolumn{1}{c|}{96.00} &
91.00 &
94.50 \\
&
&
Var Err &
-0.203 &
0.069 &
\multicolumn{1}{c|}{0.062} &
-0.266 &
\multicolumn{1}{c|}{0.065} &
-0.236 &
\multicolumn{1}{c|}{0.064} &
-0.176 &
0.051 \\ \cline{2-12} 
&
\multirow{2}{*}{60} &
Cov (\%) &
86.00 &
94.50 &
\multicolumn{1}{c|}{95.00} &
\textbf{90.50} &
\multicolumn{1}{c|}{\textbf{94.50}} &
91.00 &
\multicolumn{1}{c|}{93.50} &
93.00 &
98.00 \\
&
&
Var Err &
-0.108 &
0.115 &
\multicolumn{1}{c|}{0.104} &
\textbf{-0.257} &
\multicolumn{1}{c|}{\textbf{0.088}} &
-0.242 &
\multicolumn{1}{c|}{0.083} &
-0.200 &
0.091 \\ \cline{2-12} 
&
\multirow{2}{*}{100} &
Cov (\%) &
86.50 &
95.00 &
\multicolumn{1}{c|}{94.50} &
90.00 &
\multicolumn{1}{c|}{96.00} &
89.00 &
\multicolumn{1}{c|}{94.00} &
92.00 &
95.00 \\
&
&
Var Err &
-0.119 &
0.202 &
\multicolumn{1}{c|}{0.184} &
-0.219 &
\multicolumn{1}{c|}{0.163} &
-0.213 &
\multicolumn{1}{c|}{0.149} &
-0.189 &
0.153 \\ \hline		
\end{tabular}}
\caption{\textit{Logistic regression with Toeplitz $\Sigma_a$ across $r\in\{0.4, 0.5, 0.6\}$. See Table \ref{appen:table:1} for interpretation.}}
\label{appen:table:3}
\end{table}

\clearpage

\begin{table}[!h]
\centering
\resizebox{\linewidth}{!}{
\begin{tabular}{|c|c|c|c|cccccccc|}
\hline
\multirow{3}{*}{Equi-corr $\Sigma_a$} &
\multirow{3}{*}{d} &
\multirow{3}{*}{Criterion} &
\multirow{2}{*}{SGD} &
\multicolumn{8}{c|}{Sketched Newton Method} \\ \cline{5-12} 
&
&
&
&
\multicolumn{2}{c|}{$\tau=\infty$} &
\multicolumn{2}{c|}{$\tau=10$} &
\multicolumn{2}{c|}{$\tau=20$} &
\multicolumn{2}{c|}{$\tau=40$} \\ \cline{4-12} 
&
&
&
{\footnotesize$\bar{\Xi}_t$} &
{\footnotesize$\tilde{\Xi}_t$} &
\multicolumn{1}{c|}{{\footnotesize$\hat{\Xi}_t$}} &
{\footnotesize$\tilde{\Xi}_t$} &
\multicolumn{1}{c|}{{\footnotesize$\hat{\Xi}_t$}} &
{\footnotesize$\tilde{\Xi}_t$} &
\multicolumn{1}{c|}{{\footnotesize$\hat{\Xi}_t$}} &
{\footnotesize$\tilde{\Xi}_t$} &
{\footnotesize$\hat{\Xi}_t$} \\ \hline
\multirow{8}{*}{r=0.1} &
\multirow{2}{*}{20} &
Cov (\%) &
89.50 &
94.50 &
\multicolumn{1}{c|}{95.00} &
91.50 &
\multicolumn{1}{c|}{96.00} &
89.00 &
\multicolumn{1}{c|}{93.00} &
94.50 &
96.00 \\
&
&
Var Err &
-0.185 &
0.043 &
\multicolumn{1}{c|}{0.045} &
-0.324 &
\multicolumn{1}{c|}{0.041} &
-0.230 &
\multicolumn{1}{c|}{0.039} &
-0.088 &
0.032 \\ \cline{2-12} 
&
\multirow{2}{*}{40} &
Cov (\%) &
90.00 &
96.50 &
\multicolumn{1}{c|}{95.50} &
82.50 &
\multicolumn{1}{c|}{96.00} &
87.50 &
\multicolumn{1}{c|}{94.50} &
88.00 &
94.00 \\
&
&
Var Err &
-0.171 &
0.094 &
\multicolumn{1}{c|}{0.086} &
-0.458 &
\multicolumn{1}{c|}{0.073} &
-0.399 &
\multicolumn{1}{c|}{0.065} &
-0.288 &
0.069 \\ \cline{2-12} 
&
\multirow{2}{*}{60} &
Cov (\%) &
89.00 &
97.00 &
\multicolumn{1}{c|}{97.00} &
72.00 &
\multicolumn{1}{c|}{92.50} &
83.00 &
\multicolumn{1}{c|}{94.50} &
88.50 &
95.50 \\
&
&
Var Err &
-0.115 &
0.152 &
\multicolumn{1}{c|}{0.146} &
-0.527 &
\multicolumn{1}{c|}{0.106} &
-0.485 &
\multicolumn{1}{c|}{0.107} &
-0.411 &
0.108 \\ \cline{2-12} 
&
\multirow{2}{*}{100} &
Cov (\%) &
\textbf{79.00} &
98.00 &
\multicolumn{1}{c|}{98.00} &
78.00 &
\multicolumn{1}{c|}{92.50} &
\textbf{77.00} &
\multicolumn{1}{c|}{\textbf{96.00}} &
\textbf{82.00} &
\textbf{96.00} \\
&
&
Var Err &
\textbf{-0.161} &
0.262 &
\multicolumn{1}{c|}{0.251} &
-0.595 &
\multicolumn{1}{c|}{0.182} &
\textbf{-0.575} &
\multicolumn{1}{c|}{\textbf{0.177}} &
\textbf{-0.533} &
\textbf{0.177} \\ \hline
\multirow{8}{*}{r=0.2} &
\multirow{2}{*}{20} &
Cov (\%) &
90.00 &
96.00 &
\multicolumn{1}{c|}{96.00} &
88.50 &
\multicolumn{1}{c|}{96.50} &
89.00 &
\multicolumn{1}{c|}{96.50} &
92.50 &
96.00 \\
&
&
Var Err &
-0.172 &
0.041 &
\multicolumn{1}{c|}{0.037} &
-0.394 &
\multicolumn{1}{c|}{0.028} &
-0.302 &
\multicolumn{1}{c|}{0.037} &
-0.153 &
0.039 \\ \cline{2-12} 
&
\multirow{2}{*}{40} &
Cov (\%) &
86.00 &
95.00 &
\multicolumn{1}{c|}{95.00} &
78.00 &
\multicolumn{1}{c|}{95.00} &
81.00 &
\multicolumn{1}{c|}{94.50} &
88.00 &
96.50 \\
&
&
Var Err &
-0.111 &
0.083 &
\multicolumn{1}{c|}{0.084} &
-0.530 &
\multicolumn{1}{c|}{0.062} &
-0.490 &
\multicolumn{1}{c|}{0.046} &
-0.402 &
0.050 \\ \cline{2-12} 
&
\multirow{2}{*}{60} &
Cov (\%) &
80.00 &
94.00 &
\multicolumn{1}{c|}{93.50} &
78.50 &
\multicolumn{1}{c|}{94.00} &
80.00 &
\multicolumn{1}{c|}{97.00} &
82.50 &
96.00 \\
&
&
Var Err &
-0.144 &
0.130 &
\multicolumn{1}{c|}{0.110} &
-0.592 &
\multicolumn{1}{c|}{0.076} &
-0.569 &
\multicolumn{1}{c|}{0.068} &
-0.518 &
0.072 \\ \cline{2-12} 
&
\multirow{2}{*}{100} &
Cov (\%) &
\textbf{66.50} &
97.50 &
\multicolumn{1}{c|}{96.00} &
73.50 &
\multicolumn{1}{c|}{96.00} &
\textbf{73.00} &
\multicolumn{1}{c|}{\textbf{96.00}} &
\textbf{80.00} &
\textbf{97.00} \\
&
&
Var Err &
\textbf{-0.108} &
0.234 &
\multicolumn{1}{c|}{0.227} &
-0.647 &
\multicolumn{1}{c|}{0.116} &
\textbf{-0.636} &
\multicolumn{1}{c|}{\textbf{0.115}} &
\textbf{-0.615} &
\textbf{0.109} \\ \hline
\multirow{8}{*}{r=0.3} &
\multirow{2}{*}{20} &
Cov (\%) &
86.00 &
93.50 &
\multicolumn{1}{c|}{93.50} &
83.00 &
\multicolumn{1}{c|}{94.00} &
85.50 &
\multicolumn{1}{c|}{92.00} &
89.50 &
95.50 \\
&
&
Var Err &
-0.139 &
0.038 &
\multicolumn{1}{c|}{0.024} &
-0.422 &
\multicolumn{1}{c|}{0.027} &
-0.347 &
\multicolumn{1}{c|}{0.011} &
-0.220 &
0.028 \\ \cline{2-12} 
&
\multirow{2}{*}{40} &
Cov (\%) &
81.00 &
93.50 &
\multicolumn{1}{c|}{93.50} &
80.50 &
\multicolumn{1}{c|}{96.50} &
85.50 &
\multicolumn{1}{c|}{95.00} &
79.50 &
95.50 \\
&
&
Var Err &
-0.124 &
0.078 &
\multicolumn{1}{c|}{0.071} &
-0.536 &
\multicolumn{1}{c|}{0.045} &
-0.510 &
\multicolumn{1}{c|}{0.026} &
-0.450 &
0.044 \\ \cline{2-12} 
&
\multirow{2}{*}{60} &
Cov (\%) &
74.00 &
92.00 &
\multicolumn{1}{c|}{91.50} &
82.00 &
\multicolumn{1}{c|}{94.50} &
76.00 &
\multicolumn{1}{c|}{93.50} &
82.50 &
95.00 \\
&
&
Var Err &
-0.137 &
0.111 &
\multicolumn{1}{c|}{0.096} &
-0.584 &
\multicolumn{1}{c|}{0.051} &
-0.567 &
\multicolumn{1}{c|}{0.055} &
-0.539 &
0.046 \\ \cline{2-12} 
&
\multirow{2}{*}{100} &
Cov (\%) &
\textbf{54.00} &
97.00 &
\multicolumn{1}{c|}{96.50} &
78.50 &
\multicolumn{1}{c|}{96.00} &
\textbf{68.50} &
\multicolumn{1}{c|}{\textbf{94.00}} &
\textbf{76.50} &
\textbf{94.50} \\
&
&
Var Err &
\textbf{-0.115} &
0.203 &
\multicolumn{1}{c|}{0.185} &
-0.621 &
\multicolumn{1}{c|}{0.067} &
\textbf{-0.619} &
\multicolumn{1}{c|}{\textbf{0.064}} &
\textbf{-0.604} &
\textbf{0.069} \\ \hline		
\end{tabular}}
\caption{\textit{Logistic regression with Equi-correlation $\Sigma_a$ across $r\in\{0.1, 0.2, 0.3\}$. See Table \ref{appen:table:1} for interpretation.}}
\label{appen:table:4}
\end{table}

\begin{flushright}
\scriptsize \framebox{\parbox{2.5in}{Government License: The submitted manuscript has been created by UChicago Argonne, LLC, Operator of Argonne National Laboratory (``Argonne"). Argonne, a U.S. Department of Energy Office of Science laboratory, is operated under Contract No. DE-AC02-06CH11357.  The U.S. Government retains for itself, and others acting on its behalf, a paid-up nonexclusive, irrevocable worldwide license in said article to reproduce, prepare derivative works, distribute copies to the public, and perform publicly and display publicly, by or on behalf of the Government. The Department of Energy will provide public access to these results of federally sponsored research in accordance with the DOE Public Access Plan. http://energy.gov/downloads/doe-public-access-plan.}}
\normalsize
\end{flushright}

\end{document}